\documentclass[journal]{IEEEtran}
\ifCLASSINFOpdf
\else
\fi

\usepackage{cite}
\usepackage{graphicx}
\usepackage{amsmath}
\usepackage{amssymb}
\usepackage{multirow}
\usepackage{arydshln}
\usepackage{adjustbox}
\usepackage{multirow}
\usepackage{url}
\usepackage{rotating}
\usepackage{makecell}
\usepackage{colortbl}
\usepackage{collcell,array}
\usepackage{arydshln}
\usepackage{xcolor,colortbl}
\usepackage{pgfplots}
\usepackage{bm}

\newcommand{\BalancedScore}{\mathcal{B}}

\begin{document}
%


\title{Multitemporal and multispectral data fusion for super-resolution of Sentinel-2 images}

%
%
%

\author{Tomasz~Tarasiewicz,~\IEEEmembership{Student Member,~IEEE,}
        Jakub~Nalepa,~\IEEEmembership{Senior Member,~IEEE,}
        Reuben~A.~Farrugia,~\IEEEmembership{Senior Member,~IEEE,}
        Gianluca~Valentino,~\IEEEmembership{Senior Member,~IEEE,}
        Mang~Chen,
        Johann~A.~Briffa,
        and~Michal~Kawulok,~\IEEEmembership{Member,~IEEE}
\thanks{T. Tarasiewicz, J. Nalepa, and M. Kawulok are with Silesian University of Technology, Gliwice, Poland, e-mail: \{tomasz.tarasiewicz, jakub.nalepa, michal.kawulok\}@polsl.pl}
\thanks{J. Nalepa and M. Kawulok are also with KP Labs, Gliwice, Poland}
\thanks{R.A. Farrugia, G. Valentino, M. Chen, and J.A. Briffa are with the Department of Communications and Computer Engineering, University of Malta, Msida, Malta.}
\thanks{Manuscript received April 19, 2005; revised August 26, 2015.}}

%
%

\markboth{SUBMITTED TO IEEE TRANSACTIONS ON GEOSCIENCE AND REMOTE SENSING}%
{Shell \MakeLowercase{\textit{et al.}}: Bare Demo of IEEEtran.cls for IEEE Journals}
%



\maketitle

\begin{abstract}
Multispectral Sentinel-2 images are a valuable source of Earth observation data, however spatial resolution of their spectral bands limited to 10\,m, 20\,m, and 60\,m ground sampling distance remains insufficient in many cases. This problem can be addressed with super-resolution, aimed at reconstructing a high-resolution image from a low-resolution observation. For Sentinel-2, spectral information fusion allows for enhancing the 20\,m and 60\,m bands to the 10\,m resolution. Also, there were attempts to combine multitemporal stacks of individual Sentinel-2 bands, however these two approaches have not been combined so far.
In this paper, we introduce DeepSent---a new deep network for super-resolving multitemporal series of multispectral Sentinel-2 images. It is underpinned with information fusion performed simultaneously in the spectral and temporal dimensions to generate an enlarged multispectral image. 
In our extensive experimental study, we demonstrate that our solution outperforms other state-of-the-art techniques that realize either multitemporal or multispectral data fusion. Furthermore, we show that the advantage of DeepSent results from how these two fusion types are combined in a single architecture, which is superior to performing such fusion in a sequential manner. Importantly, we have applied our method to super-resolve real-world Sentinel-2 images, 
enhancing the spatial resolution of all the spectral bands to 3.3\,m nominal ground sampling distance, and we compare the outcome with very high-resolution WorldView-2 images. We will publish our implementation upon paper acceptance, and we expect it will increase the possibilities of exploiting super-resolved Sentinel-2 images in real-life applications.
\end{abstract}

\begin{IEEEkeywords}
super-resolution, deep learning, Sentinel-2, multispectral images, information fusion, multi-image super-resolution.
\end{IEEEkeywords}

\newcommand{\mymethod}{DeepSent}
\newcommand{\DSenR}{DSen2\textsubscript{R}}
\newcommand{\DSenHRn}{DSen2\textsubscript{HRn}}

%

\section{Introduction}
%
%
%
%
\IEEEPARstart{T}{he} Sentinel-2 (S-2) mission is based on two Earth observation satellites that acquire 13-band multispectral images (MSIs) at a revisit time of around 5~days~\cite{Drusch2012}. Free and open data policy adopted by the European Space Agency (ESA) for the S-2 products has attracted interest of the remote sensing community, leading to numerous interesting applications including land-cover mapping~\cite{Phiri2020}, precision agriculture~\cite{Segarra2020}, water resources monitoring~\cite{Sekertekin2018,Kaplan2017}, and more~\cite{Gibson2020}.
Spatial resolution of the individual S-2 bands ranges between 10\,m and 60\,m ground sampling distance (GSD). The blue, green, red, and near infrared (NIR) bands (B02, B03, B04, and B08, respectively) are of 10\,m GSD, the vegetation red edge bands (B05, B06, and B07), narrow NIR (B08a), and the short-wave infrared (SWIR) bands (B11 and B12) are of 20\,m GSD, while the remaining coastal aerosol (B01), water vapour (B09), and cirrus clouds estimation (B10) bands are of 60\,m GSD. While such resolution is sufficient in many cases, it also constitutes a serious limitation for the tasks that require higher accuracy in the spatial domain, like precision farming~\cite{DelFrate2014} or object delineation~\cite{LiuTang2022,Razzak2021}.


The aforementioned limitations can potentially be mitigated with effective super-resolution (SR) techniques, and therefore the latter are attracting considerable attention from the researchers and practitioners, including many attempts to enhance spatial resolution of S-2 spectral bands~\cite{Liebel2016,Lanaras2018,Paris2018,Gargiulo2019,Ulfarsson2019,TaoXiong2021,Razzak2021,QianJiang2023}. SR is a common term for a variety of techniques aimed at generating a high-resolution (HR) image from a low-resolution (LR) input, being either a single image, an image composed of multiple spectral bands, or a multitemporal series of images presenting the same scene captured at different times. 

In the last decade, we have witnessed unprecedented advancements in single-image SR (SISR) attributed to the use of deep learning~\cite{WangChen2021}, and such techniques have been applied in a band-wise manner for enhancing S-2 images~\cite{Romero2020,Galar2020}. However, SISR is an ill-posed problem---given a certain LR image, multiple possible HR solutions exist, and the larger the magnification ratio required, the more severe this problem becomes. Overall, while advanced SISR techniques lead to generating plausible images of high perceptual quality, their capabilities of reconstructing the real underlying HR information are still limited.

This problem can be addressed by using more informative input data that can be effectively fused to reconstruct the real HR information. There are two general categories of such approaches. First, SR can be based on multiple images that present the same area of interest, each of which carries a different portion of HR information. For satellite imagery, such multi-image SR (MISR)~\cite{Yue2016} is performed from images captured at different revisits of the imaging sensor, so the information fusion operates in the \emph{temporal} domain. For that reason, this process is also termed as \emph{multitemporal} SR~\cite{Molini2020}, and it has been applied for enhancing the individual S-2 bands~\cite{Kawulok2021IGARSS,Vaqueiro2021,Razzak2021}. The second group of approaches, aimed at MSIs or hyperspectral images (HSIs), relies on information fusion in the \emph{spectral} domain. It consists in exploiting spectral bands of higher spatial resolution to magnify the lower-resolution bands~\cite{HuHuang2021}. Such techniques have been also proposed for enhancing S-2 images, and they increase resolution of the 20\,m and 60\,m GSD bands up to 10\,m GSD~\cite{Lanaras2018,Lin2020,Gargiulo2019,Armannsson2021,Ulfarsson2019,Paris2018}.

Overall, there are two actively explored research directions for super-resolving S-2 images, namely (\textit{i})~multitemporal SR and (\textit{ii})~information fusion in the spectral domain. However, there were no reported attempts to combine the benefits of these two approaches, neither in a sequential manner, nor simultaneously.
In this paper, we address this research gap with DeepSent---a new \underline{Deep} network architecture for \underline{Sent}inel-2 SR that fuses information from multiple spectral bands in multitemporal S-2 images.  Our contribution can be summarized as follows:
\begin{itemize}
    \item We introduce the first SR network that benefits from information fusion performed both in the spectral and temporal domains. It combines multiple S-2 spectral bands in a multitemporal image series, enhancing the spatial resolution of the 10\,m, 20\,m, and 60\,m GSD bands by a factor of $3\times$, $6\times$, and $18\times$, respectively. Importantly, \mymethod\, has to be trained only once to reconstruct all the bands (they are processed using the same model). The method is described in detail in Section~\ref{sec:method}.
    \item We report the results of our extensive experimental validation (Section~\ref{sec:experiments}) which indicate that the proposed simultaneous fusion in the spectral and temporal domains outperforms the state-of-the-art methods underpinned with the multitemporal image fusion~\cite{deudon2020highresnet,Salvetti2020} or the spectral fusion~\cite{Lanaras2018}. Our ablation study clearly shows that fusion in both domains significantly improves the reconstruction accuracy for all the bands, and our simultaneous approach is more effective than combining these two fusion types in a sequential manner.
    \item We demonstrate that \mymethod~effectively super-resolves original S-2 data (i.e., not simulated ones like in many other approaches, as discussed in Section~\ref{sec:realworld}), decreasing the nominal GSD\footnote{Nominal (or geometric) GSD indicates the pixel size, and it is not to be confused with effective (or true) GSD~\cite{Topan2009} that corresponds to the information content, also referred to as ground resolved distance~\cite{Meissner2020}.} for all 12 bands down to 3.3\,m (as in other works~\cite{Lanaras2018,Lin2020}, we do not include the B10 band, because it does not present any HR details). The benefits of \mymethod\, over the state-of-the-art techniques are confirmed with the quantitative scores obtained for the MuS2 benchmark~\cite{Kowaleczko2022} with real-world HR reference images.
    \item We thoroughly investigate the reason why the artifacts are often introduced by the models trained with simulated LR data. Furthermore, we explain their origin, making it possible to identify the areas that are likely to be affected, and we present the measures that reduce the intensiveness of such artifacts.
    \item Upon acceptance, we will make the \mymethod~implementation publicly available, so that other research groups can easily reproduce the reported experiments and benefit from our 
    approaches.
\end{itemize}


\section{Related Work} \label{sec:related}

We outline the state of the art in SR taking into account the category of involved information fusion, including 
the methods that do not exploit any fusion (Section~\ref{sec:sisr}), followed by the techniques that employ fusion in the temporal domain (Section~\ref{sec:temporal}) and spectral domain (Section~\ref{sec:spectral}). Afterwards, we present the recent advancements in real-world SR (Section~\ref{sec:realworld}) to better contextualize the reported research within the efforts toward exploiting SR in practical remote sensing applications, and in Section~\ref{sec:s2}, we focus on the techniques proposed to super-resolve S-2 images.

\subsection{Fusion-free super-resolution} \label{sec:sisr}

The relation between LR and HR images can be effectively modeled using CNNs, and the advancements in feature representation and nonlinear mapping based on deep learning are often directly exploited to improve SISR solutions~\cite{HuangLi2021}. They remain the most actively explored category of SR~\cite{WangChen2021}
and were used for enhancing the remotely sensed images, including MSIs and HSIs~\cite{YuanZheng2017}. The first CNN proposed for SR (SRCNN)~\cite{Dong2014} was composed of just three convolutional layers, and it was shown that more complex models of much larger capacities, like enhanced deep SR network (EDSR)~\cite{LimSon2017}, are more effective in modeling the LR--HR relation. Recently, it was demonstrated that SISR can benefit from vision transformers~\cite{LuLi2022} which dynamically adjust the size of the feature maps, thus reducing the model complexity.


An important direction in SISR is concerned with the use of generative adversarial networks (GANs)~\cite{Ledig2017}, composed of a generator and a discriminator that compete between each other during training. 
Even though GANs are particularly effective in reconstructing images of high perceptual quality, they do not necessarily recover the actual HR information. Despite of that, they were explored for remote sensing~\cite{WangJiang2020}, and in~\cite{KimChung2019} it was demonstrated that imposing certain constraints on the adversarial loss may help increase the reliability of the reconstruction outcome obtained using GANs.

\subsection{Multitemporal image fusion} \label{sec:temporal}

Temporal fusion allows for higher reconstruction accuracy than SISR~\cite{Yue2016,Nasrollahi2014}, but it poses significant challenges including image co-registration and managing temporal variability among images. Thus, existing CNNs cannot be straightforwardly applied to MISR and elaborating dedicated architectures has been necessary---as a result, deep learning is not that commonly exploited here as for SISR, and the techniques underpinned with conventional image processing pipelines~\cite{Farsiu2004} are still used for enhancing satellite images~\cite{Zhu2016}.

The first reported approach to exploit deep learning for MISR was the EvoNet framework~\cite{Kawulok2020GRSL} which employs CNNs for SISR prior to evolutionary multi-image fusion~\cite{Kawulok2018Gecco}. This was followed by DeepSUM---the first end-to-end deep network for MISR~\cite{Molini2020}, later enhanced with non-local operations~\cite{Molini2020IGARSS}. 
DeepSUM~\cite{Martens2019} assumes a fixed number of LR inputs and requires a long training, being the result of fusing the upsampled LR images. These downsides were addressed in other MISR solutions, including HighRes-net~\cite{deudon2020highresnet} that combines the latent LR representations in a recursive manner to obtain the global representation which is upsampled to obtain the super-resolved image. Also, the attention mechanism was found useful for selecting the most valuable features extracted from LR inputs in the residual attention multi-image SR (RAMS) network~\cite{Salvetti2020}, and a recurrent network with gated recurrent units (MISR-GRU) was proposed in~\cite{Arefin2020}. An et al. focused on simplifying the training with the use of transformers~\cite{AnZhang2022} and reported competitive results for the PROBA-V dataset~\cite{Martens2019}. Another approach is to represent an input set of LR images as a graph which is processed with a graph neural network that produces the super-resolved image~\cite{Tarasiewicz2021}. Multitemporal fusion may also be performed to combine images captured by different satellites, including S-2 and Landsat-8~\cite{WuLin2022,Saunier2022}.


The existing MISR techniques can be applied to MSIs in a band-wise manner~\cite{Kawulok2021IGARSS}. Multiple images of each band are processed in isolation to synthesize a super-resolved image of that band, but there are no reports of exploiting the spectral correlation between the bands during such temporal fusion.


\subsection{Fusion in the spectral domain} \label{sec:spectral}

Fusion in the spectral domain is primarily explored for HSIs, but the developed techniques have also been applied to MSIs, including S-2 images (discussed later in Section~\ref{sec:s2}). HSIs are often accompanied with a panchromatic channel of higher spatial resolution which can be exploited for enhancing the remaining spectral bands. This process is known as pansharpening and it can be effectively performed with deep learning~\cite{HuangXiao2015}. 
Another possibility is to combine an LR HSI with an MSI of higher spatial resolution to obtain an HR HSI~\cite{HuHuang2021,Sara2021}. Recently, it has been proposed to exploit the deep prior for regularizing the optimization unfolded into a deep network~\cite{YangXiao2022} that performs the MSI--HSI fusion.

When no HR channel is available, the spectral correlation between the bands can also be exploited to generate an HR image---this is commonly performed with 3D convolutions~\cite{LiWang2021} or with tensor decomposition techniques~\cite{XueZhao2022TC}. The latter may be underpinned with low-rank tensor decomposition for tensor completion~\cite{XueZhao2022TNNLS} that is employed for recovering missing information. Such techniques were applied for HSI denoising~\cite{XueZhao2019}, super-resolution~\cite{DianLi2019}, and also for fusion of MSIs and HSIs~\cite{BuZhao2021}.
HSI SR based on 3D CNNs can also benefit from the attention modules that help capture local contextual features, as well as non-local interdependencies~\cite{YangXiao2021}.

To ensure that the spectral information is not disturbed when an HSI is being enhanced, the spectral angle mapper (SAM) is often monitored between the reconstructed image and the ground-truth~\cite{KwanChoi2018}. Also, SAM can be embedded into the loss function during training~\cite{LiZhang2018} for better spectral consistency.


\subsection{Real-world super-resolution} \label{sec:realworld}

Commonly, the SR algorithms are trained and validated following an artificial scenario, in which LR images are simulated by downsampling the original image, treated as an HR reference. Although the use of simulated LR images is widely adopted for SR~\cite{Kawulok2018ACIIDS}, there may be a substantial performance gap between such data and real images~\cite{Kohler2019,Lugmayr2020}.
Therefore, a lot of effort is invested nowadays into creating real-life sets encompassing pairs of original LR and HR images~\cite{ChenHe2022} for both SISR and MISR. The PROBA-V benchmark was the first large-scale satellite image dataset~\cite{Martens2019}, and recently two datasets were published that match S-2 images with HR data. The WorldStrat dataset~\cite{Cornebise2022} matches S-2 with SPOT images of higher resolution, whereas the MuS2 benchmark~\cite{Kowaleczko2022} couples the 10\,m S-2 bands with HR WorldView-2 images. The use of real-world datasets for evaluating SR is not straightforward, as they rely on LR and HR images captured in different conditions. Recently, Nguyen et al. proposed to adapt the PROBA-V dataset for reference-aware SR~\cite{NguyenAnger2021Proba} which addresses the problem of evaluation bias resulting from temporal changes among the input LR images.

The reconstruction accuracy can also be assessed in a blind way without using any reference image, but the scores may not be sensitive to the presence of artifacts~\cite{TaoXiong2021}. 
Another approach was adopted for the OpTiGAN system~\cite{TaoMuller2021} which was validated using images captured at Baotou Geocal site that presents different patterns at varying scale. This makes it possible to measure the effective GSD from a super-resolved image. 
Another possibility is to evaluate SR in the context of a specific image analysis task. Xu et al. demonstrated the benefits of employing SR for S-2 images to improve the quality of their semantic segmentation~\cite{XuTang2021}, and an MISR based on HighRes-net was validated for the delineation of buildings in~\cite{Razzak2021}.






\subsection{Super-resolving S-2 images} \label{sec:s2}

S-2 SR has already attracted considerable attention. In~\cite{Liebel2016}, SRCNN~\cite{Dong2014} was exploited for super-resolving the RGB S-2 bands (i.e., B02, B03, and B04). These bands were processed independently and more advanced SISR techniques, including EDSR~\cite{LimSon2017}, were also applied in a band-wise manner to enhance original multispectral S-2 images~\cite{Galar2020} (this method was evaluated quantitatively using 2.5\,m PlanetScope images).
GANs were also used for super-resolving S-2 images in a band-wise manner~\cite{Latif2022}. Beaulieu et al. employed them for upsampling the 10\,m bands to 2.5\,m~\cite{Beaulieu2018}, and Romero et al. used ESRGAN~\cite{WangYu2018} for enhancing the original S-2 images by a factor of $5\times$~\cite{Romero2020}. Importantly, it was noticed in~\cite{Romero2020} that the visual quality is often obtained by hallucinating image details, hence subject to a trade-off with the quantitative metrics which indicate the accuracy of reconstructing the actual HR information. In~\cite{Beaulieu2018} and~\cite{Romero2020}, the HR WorldView images were exploited for quantitative evaluation. Recently, Zabalza and Bernardini trained a residual network with spectral attention~\cite{Zabalza2022} using real-world pairs of S-2 10\,m bands coupled with HR PlanetScope data.

Also, there were a few attempts to enhance multitemporal S-2 images in a band-wise manner. Vaqueiro et al. proposed to exploit the georeferencing error when fusing S-2 images using the $k$-nearest pixels technique~\cite{Vaqueiro2021}. In~\cite{Valsesia2022permutation},
enforcing permutation invariance within a set of LR inputs was shown to improve the results for the PROBA-V images and S-2 B03 band with simulated LR images. Also, RAMS~\cite{Salvetti2020} and HighRes-net~\cite{deudon2020highresnet} trained with real-world PROBA-V images were used for enhancing multitemporal series of individual S-2 bands, and Razzak et al. enriched HighRes-net with a radiometric consistency loss before applying it for MISR of S-2 images~\cite{Razzak2021}.

A widely explored direction is concerned with exploiting the 10\,m S-2 bands to super-resolve the 20\,m and 60\,m bands relying on spectral fusion~\cite{WangShi2016}.
Brodu adapted existing pansharpening techniques to super-resolve the 20\,m and 60\,m S-2 bands using a simulated panchromatic channel obtained from the 10\,m RGB bands~\cite{Brodu2017}. Lanaras et al. analyzed correlations between the spectral bands and exploited the textural information from higher-resolution bands to magnify the remaining ones~\cite{Lanaras2017}. Their solution was followed with DSen2---a deep CNN architecture elaborated to magnify all the bands to 10\,m GSD~\cite{Lanaras2018}. As in other works~\cite{Lin2020}, the B10 band was not included. A similar fusion for super-resolving the B11 SWIR band was reported by Gargiulo et al.~\cite{Gargiulo2018} and it was extended to process all of the 20\,m bands~\cite{Gargiulo2019}. The SR outcome was inspected qualitatively against the artifacts, and quantitatively against spectral distortions in a reference-free manner. Paris et al. applied their S-2 sharpening method based on 3D filtering for real-world images~\cite{Paris2018} and inspected the results qualitatively. In~\cite{Ulfarsson2019}, the reduced-rank approximation was employed for S-2 sharpening and in~\cite{QianJiang2023}, the deep image prior was employed. Armannsson et al. compared several S-2 sharpening approaches and applied the Bayesian optimization to tune their hyperparameters~\cite{Armannsson2021}. All of these methods are limited to enhancing the resolution of low-resolution bands to 10\,m nominal GSD, but S-2 images can also be fused with HR imagery. In~\cite{Latte2020}, PlanetScope images were used to sharpen the 10\,m and 20\,m S-2 bands to 2.5\,m.

Lloyd et al. proposed to extract HR information from S-2 optical bands and fuse it with Sentinel-3 thermal bands for measuring sea surface temperature~\cite{Lloyd2021}. Lin and Bioucas-Dias introduced a self-similarity prior that is used as a regularizer when super-resolving the lower-resolution S-2 bands~\cite{Lin2020}. Their method, in contrast to the techniques mentioned earlier in this section, was fed with both simulated and original S-2 data, and the obtained results were inspected qualitatively.


Overall, there are many S-2 SR techniques underpinned with information fusion realized in the temporal or spectral dimension, but there were no attempts reported yet to combine multitemporal and multispectral information at once. Also, just a few methods were validated quantitatively using real-world HR images. Some researchers observed artifacts when the models trained using simulated data were fed with real-world images~\cite{Gargiulo2019,Kawulok2021IGARSS,Romero2020}, but the attempts to understand their origin and reduce their intensiveness were rather limited, and this problem is also tackled in the study reported here.

\newcommand{\BandGroup}{\bm{B}}
\newcommand{\Band}{b}
\newcommand{\MedianBand}{m}
\newcommand{\EmbeddingBlock}{\rm EB}
\newcommand{\FusionBlock}{\rm FB}
\newcommand{\EmbeddingBlockOutputVector}{\bf{E}}
\newcommand{\EmbeddingBlockOutputVectorPaddedWithZeroes}{\bf{E}'}
\newcommand{\EmbeddingBlockSingle}{\bf{e}}
\newcommand{\NumberOfFeatureMapsEB}{64}
\newcommand{\LR}{\mathcal{I}}
\newcommand{\NumberOfLRs}{N_\mathcal{I}}
\newcommand{\NumberOfLRsPaddedWithZeroes}{N_\mathcal{I}'}
\newcommand{\ImageHeight}{H}
\newcommand{\ImageWidth}{W}
\newcommand{\Resolution}{r}
\newcommand{\FusingIterations}{K}
\newcommand{\FusingIteration}{k}
\newcommand{\ScalingFactor}{s}
\newcommand{\TensorWidth}{W_T}
\newcommand{\TensorHeight}{H_T}
\newcommand{\NumberOfChannels}{C}
\newcommand{\IterationImageCount}{N}
\newcommand{\IterationTensor}{X}
\newcommand{\BandFusionFunction}{f_b}
\newcommand{\FirstHalf}{s^k_1}
\newcommand{\SecondHalf}{s^k_2}

\section{Method} \label{sec:method}

In this section, we present \mymethod\footnote{Implementation of \mymethod~will be published at \url{https://gitlab.com/tarasiewicztomasz/deepsent} upon acceptance}---a fully convolutional deep neural network for super-resolving S-2 images by means of spectral (multiple bands) and temporal (multiple images per band) information fusion. Such an approach allows us to upsample all the bands up to 3.3\,m GSD resolution, hence the magnification factors are of $3\times$, $6\times$ and, $18\times$ for 10\,m, 20\,m and 60\,m GSD bands, respectively. We can distinguish three main parts of the architecture: the band fusion, cross-resolution fusion, and super-resolution modules---they are discussed in detail in the following subsections.

\subsection{Overview of the architecture}
\begin{figure*}[ht!]
\centering
\includegraphics[width=\textwidth]{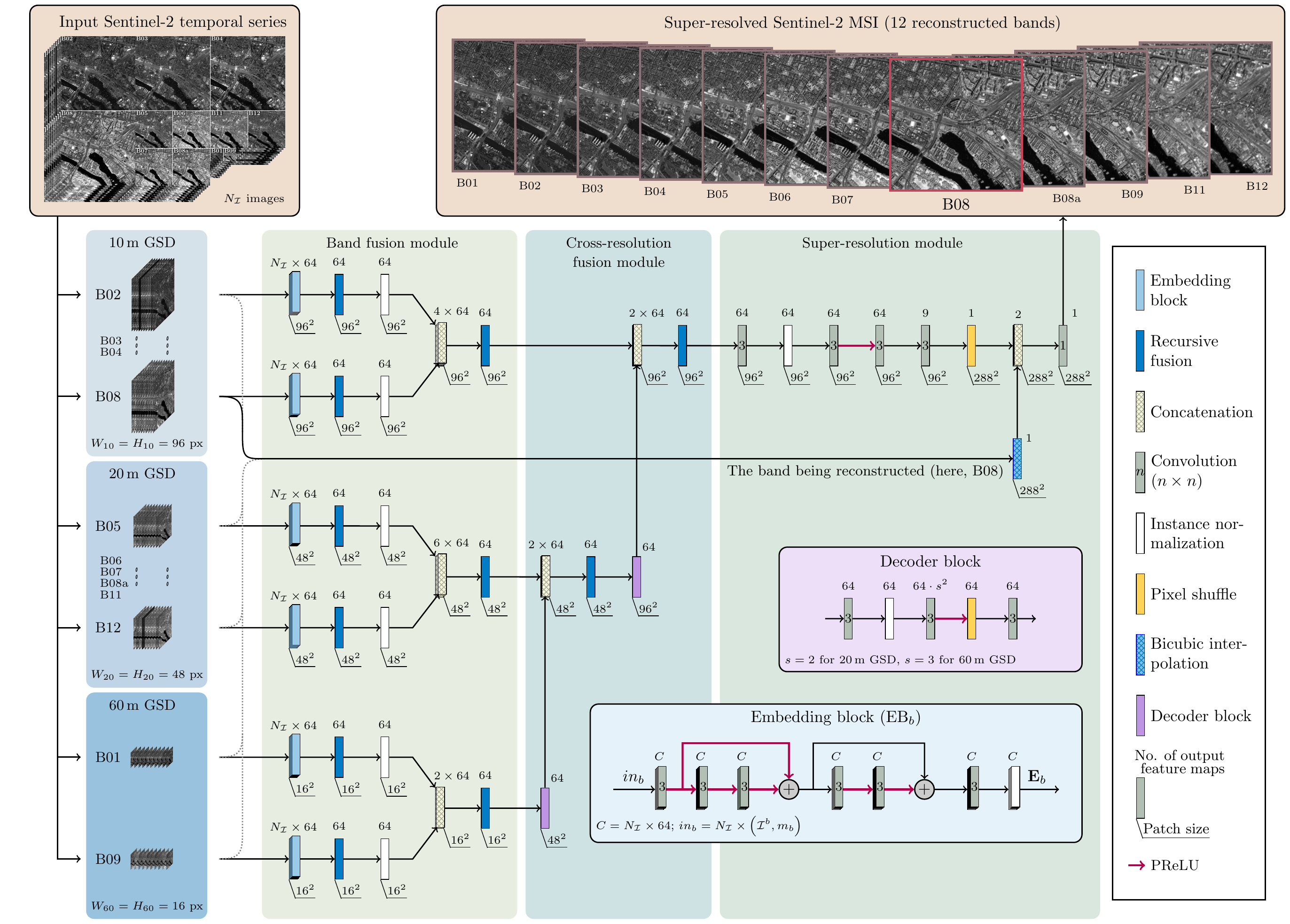}

\caption{The \mymethod\, architecture. A stack of input S-2 MSIs is processed, and every band of the output super-resolved image is reconstructed by passing the appropriate stack of the input bands---in the figure, this is presented for the B08 band, and the remaining alternative connections are shown with gray dotted lines.}
\label{fig:architecture}
\end{figure*}

Our \mymethod~architecture that benefits from the spectro-temporal information fusion is presented in Fig.~\ref{fig:architecture}. A temporal series of $\NumberOfLRs$ multispectral images ($\LR$) is organized into $12$ stacks, each of which contains $\NumberOfLRs$ images ($\LR^{\Band}$) of the same band $\Band$. In the \emph{band fusion module} (Section~\ref{sec:band_fusion_module}), the input bands ($\BandGroup$) are grouped by their spatial resolution, hence we obtain three mutually exclusive S-2 band subsets: $\BandGroup_{10}$, $\BandGroup_{20}$, and $\BandGroup_{60}$ with 10\,m, 20\,m, and 60\,m bands, respectively. We utilize the S-2 Level-2A products with B10 excluded, as it is an uncalibrated band exploited for detecting thin cirrus clouds (it is referred to as the ``cirrus'' band)~\cite{Raiyani2021}.

Having the bands grouped, we encode LR images and fuse the temporal information for each band independently. After that, we combine the spectral information between the bands assigned to the same spatial resolution group $\BandGroup_{\Resolution}$, where $\Resolution \in \{10, 20, 60\}$, resulting in three stacks of feature maps, each of different spatial resolution. Afterwards, we upsample and merge them sequentially to reach the final latent representation of a scene at 10\,m GSD using the \emph{cross-resolution fusion module} (Section~\ref{sec:cross_resolution_fusion}). Therefore, 60\,m feature maps are upsampled $3\times$ to match the resolution of 20\,m GSD, and then we recursively combine them with the $\BandGroup_{20}$ bands. We apply the same procedure to match the result of this operation with 10\,m bands through upsampling it $2\times$. Such representation of a scene is decoded and upsampled $3\times$ to reach the target resolution of 3.3\,m GSD using the \emph{super-resolution module} (Section~\ref{sec:super_resolution_module}). 
The intrinsic characteristics of the band being reconstructed are captured and propagated to the output using a skip connection that transfers the averaged stack of bicubically upsampled input images for that band. It is ultimately concatenated with the decoded scene and merged using a single convolutional layer. This is repeated for every band to reconstruct the entire MSI.

\subsection{Band fusion module}\label{sec:band_fusion_module}

We exploit a recursive fusion block inspired by the HighRes-net architecture~\cite{deudon2020highresnet} to perform fusion in the temporal dimension, followed by fusion in the spectral dimension. Recursive fusion was originally applied for merging a single stack of multitemporal LR images---here, we exploit it for this purpose (the first recursive fusion blocks at each band-wise processing path), as well as to fuse the features extracted from the individual spectral bands (the recursive fusion blocks later on the processing paths). Here, we extend this idea and integrate feature maps obtained for different input bands of the same spatial resolution. Since the characteristics of the available S-2 bands can significantly differ, all input images undergo $z$-score normalization, performed for each band independently. Afterwards, we exploit the embedding blocks ${\EmbeddingBlock}_\Band$, and apply them to each band of an input LR image. The $\EmbeddingBlock$s consist of the convolutional layers and residual blocks followed by the parametric rectified linear unit (PReLU) activation, and they extract $\NumberOfFeatureMapsEB$ feature maps for each $\LR^{\Band}$. Additionally, an $\EmbeddingBlock$ implicitly co-registers the input images (for each band separately), as each one is collated with the median band $\MedianBand$, in which each pixel at the $(x, y)$ position is the median pixel value from the entire LR stack. Thus, the embedded outputs become:
\begin{equation}
   {\EmbeddingBlockOutputVector}_{\Band} = \left[{\EmbeddingBlockSingle}_1^{\Band}, {\EmbeddingBlockSingle}_2^{\Band}, \dots, {\EmbeddingBlockSingle}_{\NumberOfLRs}^{\Band}\right],
\end{equation}
where
\begin{equation}
    {\EmbeddingBlockSingle}_i^{\Band} = {\EmbeddingBlock}_{\Band}({\LR}_i^{\Band}, {\MedianBand}_{\Band}) \in \mathbb{R}^{\NumberOfFeatureMapsEB\times \ImageHeight_{\Resolution}\times \ImageWidth_{\Resolution}},
\end{equation}

\noindent and ${\LR}_i^{\Band}$ denotes the $\Band$-\textit{th} band of the $i$-\textit{th} LR image, ${\MedianBand}_{\Band}$ is the median $\Band$-\textit{th} band, and $\ImageHeight_{\Resolution}$ and $\ImageWidth_{\Resolution}$ are the height and width of the LR image of spatial resolution $\Resolution$, respectively.

The temporal information fusion is achieved using the recursive approach which extracts the mutual features from the stack of input images, and halves their number with each iteration until all the images are fused into one. If $\NumberOfLRs$ is not the power of $2$, we pad the embedded tensor ${\EmbeddingBlockOutputVector}_{\Band}$ with zeros and reshape it to $(\NumberOfLRsPaddedWithZeroes, \NumberOfFeatureMapsEB, \ImageHeight_{\Resolution}, \ImageWidth_{\Resolution})$, where $\NumberOfLRsPaddedWithZeroes \in \{1, 2, 4, 8, \dots\}$, and such conditionally padded tensor is ${\EmbeddingBlockOutputVectorPaddedWithZeroes}_{\Band}$. In addition, we create a vector $\alpha^0_b \in \mathbb{B}^{\NumberOfLRsPaddedWithZeroes}$ indicating whether a specific temporal image in ${\EmbeddingBlockOutputVectorPaddedWithZeroes}$ was padded (zero) or not (one). The recursive fusion block $\FusionBlock$ for each band $b \in B$ operates on the corresponding ${\EmbeddingBlockOutputVectorPaddedWithZeroes}_{\Band}$, and it iteratively halves its temporal dimension until it reaches $1$. We merge the information transferred directly from $\EmbeddingBlock$s---in each fusing iteration $\FusingIteration \in \{0, 1, \dots, \FusingIterations-1\}$, where $\FusingIterations = \log_2(\NumberOfLRsPaddedWithZeroes)$, we generate the fusion map:

\begin{equation}\label{eq_band_fuse_it}
    {\IterationTensor}^{\FusingIteration+1}_{\Band} = {\FirstHalf} + \alpha^{\FusingIteration}_2 {\BandFusionFunction}({\FirstHalf}, {\SecondHalf}) \in \mathbb{R}^{{\IterationImageCount}^{\FusingIteration+1}\times \NumberOfFeatureMapsEB \times \ImageHeight_{\Resolution}\times \ImageWidth_{\Resolution}},
\end{equation}

\noindent where ${\IterationImageCount}^{\FusingIteration+1} = {\IterationImageCount}^{\FusingIteration}/2$, $\alpha_{\Band}^{{\FusingIteration}+1} = \alpha_1^{\FusingIteration} \in \mathbb{B}^{{\IterationImageCount}^{{\FusingIteration}+1}}$, ${\BandFusionFunction}$ is a band-specific residual block built with a skip connection and two convolutional layers, ${\FirstHalf}$ and ${\SecondHalf}$ represent two halves of ${\IterationTensor}^{\FusingIteration}_{\Band}$ split along its temporal dimension. Similarly, we divide the $\alpha_{\Band}^{\FusingIteration}$ in two equisized vectors, where $\alpha_1^{\FusingIteration}$ denotes its first half and $\alpha_2^{\FusingIteration}$ becomes the second one. Also, to initiate the fusion loop, we set ${\IterationTensor}^0_{\Band} = {\EmbeddingBlockOutputVectorPaddedWithZeroes}_{\Band}$ and ${\IterationImageCount}^0 = {\NumberOfLRsPaddedWithZeroes}$. Therefore, the output of the fusion block ${\FusionBlock}_{\Band}$ can be defined as:
\begin{equation}\label{eq_band_fuse}
    {\IterationTensor}^{\FusingIterations}_{\Band} = {\rm \FusionBlock}_{\Band}({\EmbeddingBlockOutputVectorPaddedWithZeroes}_{\Band}, \alpha_{\Band}^0) \in \mathbb{R}^{\NumberOfFeatureMapsEB \times \ImageHeight_{\Resolution}\times \ImageWidth_{\Resolution}}.
\end{equation}



The spectral fusion uses an analogous recursive method as the temporal one, but instead of combining a stack of LR images from a single band, we concatenate and merge fused features of bands having the same spatial resolution. It allows for an additional flow of information and enhancing the scene representation in the latent space. We apply the instance normalization layer for every fused feature map ${\IterationTensor}_{\Band}^{\FusingIteration}$ to minimize the luminance differences between the bands, and focus more on their textural features. Then, we concatenate the feature maps of bands belonging to the same subset $\BandGroup_{\Resolution}$, and pass them to the recursive fusion layers, defined exclusively for each $\BandGroup_{\Resolution}$. The only conceptual difference is that the first dimension represents the spectral entries and not the temporal ones. Similarly to the temporal fusion, we also pad the input tensors to match their size to a power of $2$, and provide the vectors indicating which entry was padded. Such recursion performed on multiple levels ensures flexibility of our model in terms of the input data, as it does not require passing the same number of LR images for every band. Moreover, it can super-resolve a scene even if we provide LR images only for some selected bands---it will be experimentally shown in Section~\ref{sec:experiments}.



\subsection{Cross-resolution fusion module}\label{sec:cross_resolution_fusion}

To combine information extracted from bands of different resolutions, we sequentially upsample the feature maps to the 10\,m GSD resolution. We perform the upsampling in multiple parts of the network using pixel shuffling~\cite{Salvetti2020}. It reorganizes elements in the input tensor with the dimensionality of $(\NumberOfChannels{\ScalingFactor^2} \times \TensorHeight \times \TensorWidth)$ to $(\NumberOfChannels \times \ScalingFactor\TensorHeight \times \ScalingFactor\TensorWidth)$, where $\NumberOfChannels$ represents the number of input channels, and $\ScalingFactor$ denotes the upsampling factor. In the upsampling block, we utilize a 2D convolution (with the instance normalization), followed by another 2D convolution which becomes the input to the pixel shuffle operation. Finally, its output is passed to the last convolutional layer which prunes possible artifacts from the upsampled image. Importantly, the second convolutional layer multiplies the number of feature maps by a square of the block's upsampling factor. The reason for that is to ensure that the upsampled image consists of the same number of feature maps as the input one.

Afterwards, the 60\,m latent scene representation is upsampled by a factor of three, fused with the 20\,m feature maps using concatenation followed by a recursive fusion layer, and then fed into the upsampling layer of the 20\,m processing path. This is later merged with the 10\,m bands in the same way as the 60-to-20\,m fusion is performed. Here, we use the recursive fusion layer that was previously exploited to combine the temporal and spectral features. It allows us to make \mymethod~functional over incomplete input data---in this case, it can perform a successful forward pass even if the bands of a given resolution are missing in the input LR image stack.

\subsection{Super-resolution module}\label{sec:super_resolution_module}

At this point, the model stores multiple embedded feature maps at 10\,m GSD resulted from the fusion performed on multiple levels of the \mymethod~architecture. To upsample the images even further (to the target 3.3\,m GSD), we transform the embedded scene representation using pixel shuffling. This operation requires the number of channels in an input image to be divisible by a square of the upsampling factor. Thus, we pass it through a convolutional layer which generates nine feature maps with subpixel shifts between them, and then the pixel shuffle transforms these maps into a single-channel tensor with the tripled spatial dimensions. We utilize 12 such convolutional layers (one for each S-2 band), and exploit them as a reconstruction mechanism which recreates the band-specific characteristics based on the latent representation.

Since the original information captured by each S-2 band is entangled in the latent space, it may not be straightforward to  reconstruct its specific spectral characteristics. To address this issue, we benefit from the skip connections which bypass the input bands of interest directly to the super-resolution module, averaged across the temporal dimension and bicubically upsampled to match the output resolution of the model. Finally, as \mymethod~operates on standardized images, the resulting super-resolved images are standardized as well.

\section{Experimental Validation}\label{sec:experiments}

In this section, we present and discuss the results of our experimental study performed over both simulated and real-life image data, including the recently published MuS2 benchmark~\cite{Kowaleczko2022}. The objectives of the experiments are two-fold: (\textit{i})~to investigate the reconstruction capabilities of \mymethod\, and to confront it with the existing SR methods benefiting from the temporal or spectral information fusion for both simulated and real-world images, and (\textit{ii})~to thoroughly understand the benefits of spectral and temporal fusion and to justify the proposed processing flow.


\subsection{Experimental setup}
\newcommand{\correctionBias}{b}
\newcommand{\hr}{HR}
\newcommand{\sr}{SR}
\newcommand{\sims}{Sim-S2}
\newcommand{\simhsi}{Sim-HSI}
\newcommand{\reals}{Org-S2}

In the reported study, we have used the following datasets: (\textit{i})~simulated data obtained from original S-2 images (termed as \sims), (\textit{ii})~data simulated from HR HSIs (termed as \simhsi), (\textit{iii})~real-world S-2 images without HR reference (termed as \reals), and (\textit{iv})~the MuS2 benchmark~\cite{Kowaleczko2022} with real-world S-2 images, whose 10\,m bands are matched with corresponding HR WorldView-2 bands (downsampled to 3.3\,m GSD). When HR reference images are available, we assess the reconstruction quality using peak signal-to-noise ratio (PSNR), structural similarity index (SSIM), learned perceptual image patch similarity (LPIPS)~\cite{ZhangIsola2018}, and SAM (the better reconstruction is indicated with higher PSNR and SSIM, and lower LPIPS and SAM scores). For assessing MISR, the PSNR and SSIM metrics require compensation with regards to the average brightness and full-pixel displacements between SR output and its corresponding HR reference. Therefore, as proposed in ~\cite{Martens2019} and followed in many works on MISR~\cite{Molini2020,deudon2020highresnet,Salvetti2020}, we report the scores after applying such compensation, and these adjusted metrics are termed as cPSNR and cSSIM. For the \reals\, images which are not coupled with any HR reference, we compute SAM after downsampling the super-resolved image back to the LR size to verify the radiometric consistency, and the quality of the super-resolved image is assessed using the reference-less naturalness image quality evaluator (NIQE)~\cite{Mittal2013}. To verify whether the differences between the scores obtained using \mymethod\, and other methods are statistically significant, we employed the two-tailed Wilcoxon signed-rank test (at $p=0.05$) with the null hypothesis saying that two different methods allow for the same reconstruction quality. The differences between two techniques are considered significant, if the null hypothesis can be rejected at $p<0.05$.

In order to create the \sims\, and \reals\, datasets, we collected eight different scenes with total coverage of around 100\,000\,km$^2$, each composed of 15 geographically co-registered S-2 Level-2A tiles captured at different times. Each tile comprises cloud masks and twelve square-shaped images ($10980\times 10980$, $5490\times 5490$ and $1830\times 1830$ pixels for 10\,m, 20\,m, and 60\,m GSD bands, respectively). To generate the \sims\, dataset, we first downsampled the 10\,m bands in the original S-2 images $6\times$ and 20\,m bands $3\times$, so that the resolution of all bands is uniform and equal to the original resolution of the 60\,m bands. This was performed to prepare an HR reference for \mymethod, as it super-resolves all the bands to the same resolution. Next, the tiles were split into $288\times288$ patches (each composed of 12 bands) and we simulated nine LR MSIs from each patch (later treated as our target HR). The simulation consisted of the following steps: (\textit{i})~translating a patch with a random sub-pixel shift, (\textit{ii})~manipulating its contrast and brightness, (\textit{iii})~contaminating it with Gaussian blur and additive Gaussian noise. These operations were performed independently for each band within an image, except for the translations which were shared across the bands---this was to simulate different spectral angles between the input LR images, because they are also observed for original S-2 data, as explained later in Section~\ref{sec:realworld}. 
Finally, to convert such patches into LR input data, we downsampled them bicubically by the factors of $3\times$ (to simulate the 10\,m bands), $6\times$ (to get the 20\,m bands) and $18\times$ (to get the 60\,m bands). Therefore, we obtained the square images of $96\times 96$, $48\times 48$, and $16\times 16$ pixels for 10\,m, 20\,m, and 60\,m bands, respectively. Overall, we obtained 4320 multispectral patches that have been split into training (80\% of patches), validation (10\%) and test sets (10\%), without any overlaps in the spatial domain. The training set retrieved from the \sims\, images was used to train \mymethod\, and other techniques considered in this study.

To create the \reals\, dataset (without HR references), the original Level-2A S-2 tiles were partitioned into $288\times288$ patches for the 10\,m, $144\times144$ for 20\,m and $48\times48$ for 60\,m bands, and we picked the patches located inside the patches selected earlier to the \sims\, test set. A single patch in \sims\, covers the same region (at a different scale) as 36 patches from the \reals\, set. For each of these patches, we have 15 MSIs acquired at different times, out of which we selected nine with the smallest cloud coverage (based on the cloud masks). Therefore, our \reals\, set is composed of 1044 MSI stacks, each one encompassing nine observations acquired at different times (a single observation being a cropped Level-2A S-2 MSI at its original resolution).

The \simhsi\, dataset was generated using the hyperspectral data acquired using Airborne Visible/Infrared Imaging Spectrometer (AVIRIS)\footnote{The AVIRIS data are available at \url{https://aviris.jpl.nasa.gov/dataportal}} at 3.4\,m resolution. We extracted 42 patches of $486 \times 486$ pixels each from the f130818t01p00r08 flight presenting urban area and different types of terrain. At first, we simulated the multispectral bands of the same characteristic as that of S-2 MSIs~\cite{Marcinkiewicz2019,Blonski2003}. They were treated as the HR reference and the LR data were simulated following the same protocol as for the \sims\, dataset, resulting in nine LR input images sized $162 \times 162$, $81 \times 81$, and $27 \times 27$ pixels for the simulated 10\,m, 20\,m, and 60\,m bands, respectively, for each HR patch.
While \sims\, has been obtained from real S-2 images (preserving the original spectral properties of S-2 images), the \simhsi\, is supposed to better reflect the scale, as the spatial resolution of the input HSI equals 3.4\,m GSD, thus the simulated LR images are at the same scale as original S-2 images.

\mymethod\, is confronted with two state-of-the-art MISR architectures, RAMS~\cite{Salvetti2020} and HighRes-net~\cite{deudon2020highresnet} that enhance the spatial resolution $3\times$, as well as with the DSen2~\cite{Lanaras2018} network that operates from a single MSI and enhances the resolution of 20\,m and 60\,m bands up to 10\,m GSD. To make our experiments more comprehensive, we have combined DSen2 with RAMS and HighRes-net in a sequential manner, so that the output of the band-wise multitemporal SR (performed with the latter models) is directly transferred to DSen2 (thus it operates on 60\,m and 20\,m bands already enlarged $3\times$). We denote such conjunctions as \DSenR~and \DSenHRn~for RAMS and HighRes-net, respectively. Such combinations have the same upsampling capabilities as \mymethod~and, similarly to our method, they exploit fusion of both temporal and spectral information. As in some cases the images produced by other methods are of lower spatial resolution than the size of the HR reference that matches the upsampling capabilities of \mymethod\, ($3\times$ larger than the input 10\,m bands), we bicubically upsample these images to compare them with the reference ones. The images (for all the bands) generated by \mymethod, \DSenHRn, \DSenR, as well as the super-resolved images of 10\,m bands for RAMS and HighRes-net are of the same spatial resolution and they do not require any additional adjustments.

\begin{table*}[h!]
\centering
\caption{Reconstruction accuracy scores obtained for the simulated data (\sims\, and \simhsi), calculated for every band, and averaged within each spatial resolution group for the images in the test set. The best results in each group are boldfaced and the second best are underlined.}
\renewcommand{\tabcolsep}{1.6mm}
\begin{tabular}{clrrrcrrrcrrrcr}
    \Xhline{2\arrayrulewidth}
     Simulated &
    \multicolumn{1}{r}{Metric$\rightarrow$} & \multicolumn{3}{c}{cPSNR} &~& \multicolumn{3}{c}{cSSIM} &~& \multicolumn{3}{c}{LPIPS} &~& SAM\\
    \cline{3-5} \cline{7-9} \cline{11-13}
    dataset &
    \multicolumn{1}{r}{Method~~~~Bands $\rightarrow$} &
      \multicolumn{1}{c}{60\,m} &
      \multicolumn{1}{c}{20\,m} &
      \multicolumn{1}{c}{10\,m} & &
      \multicolumn{1}{c}{60\,m} &
      \multicolumn{1}{c}{20\,m} &
      \multicolumn{1}{c}{10\,m} & &
      \multicolumn{1}{c}{60\,m} &
      \multicolumn{1}{c}{20\,m} &
  \multicolumn{1}{c}{10\,m} &~&
   \\ \hline
      \multirow{7}{*}{\begin{tabular}{c} \sims \\ (test set) \end{tabular}} &
    \multicolumn{1}{l}{Bicubic} & 38.66  & 33.07  & 39.46 & & 0.9040   & 0.7697   & 0.9066        & & 0.4564 & 0.6228 & 0.3547 & & 0.1155\\
&    \multicolumn{1}{l}{RAMS}        & 38.26  & 33.85  & 41.50 & & 0.9036   & 0.7986   & 0.9457 & & 0.4447 & 0.4978 & 0.1571 & & 0.1130       \\
&    \multicolumn{1}{l}{HighRes-net}  & 38.11  & 34.25  & \underline{41.55} & & 0.9035   & 0.8180   & \underline{0.9487} & & 0.4465 & 0.4706 & \underline{0.1425} & & 0.1171       \\
&    \multicolumn{1}{l}{DSen2}       & 42.74  & 35.07  & ---      & & 0.9602   & 0.8485   & --- & & 0.2034 & 0.3617 & --- & & 0.0915           \\
&    \multicolumn{1}{l}{\DSenR} & \underline{43.68} & 36.16 & --- & & 0.9702 & 0.8820 & ---
& & 0.0868 & 0.1831 & --- & & 0.0909\\
&    \multicolumn{1}{l}{\DSenHRn} & 43.60 & \underline{36.34} & --- & & \underline{0.9720} & \underline{0.8912} & --- & & \underline{0.0728} & \underline{0.1550} & --- & & \underline{0.0903} \\
&    \multicolumn{1}{l}{DeepSent} &
      \textbf{49.06} &
      \textbf{39.26} &
      \textbf{43.68} & &
      \textbf{0.9872} &
      \textbf{0.9348} &
      \textbf{0.9629} & &
      \textbf{0.0450} &
      \textbf{0.1219} &
      \textbf{0.0735} & &
      \textbf{0.0553}\\ \cline{1-15}

     \multirow{7}{*}{ \simhsi } &
    \multicolumn{1}{l}{Bicubic} &  32.29 & 30.78 & 30.70 &&
    0.8077 & 0.7565 & 0.8253  &&
    0.6162 & 0.4682 & 0.3183  && 0.1293\\
&    \multicolumn{1}{l}{RAMS} & 32.18 & 31.21 & \underline{32.08} &&
    0.8399 & 0.7717 & \underline{0.8844} &&
    0.5746 & 0.4415 & 0.2488 && 0.1677    \\
&    \multicolumn{1}{l}{HighRes-net}  & 31.70 & 30.92 & 31.53 &&
    0.8338 & 0.7684 & 0.8817 &&
    0.5737 & 0.4249 & \underline{0.2215} && 0.1815\\
&    \multicolumn{1}{l}{DSen2} & \underline{37.19} & \underline{33.57} & --- &&
    \underline{0.9327} & 0.8717 & --- &&
    0.2177 & 0.2936 & --- && \underline{0.1031} \\
&    \multicolumn{1}{l}{\DSenR} & 36.63 & 33.05 & --- &&
    0.9294 & \underline{0.8744} & --- &&
    0.1745 & 0.1936 & --- && 0.1205\\
&    \multicolumn{1}{l}{\DSenHRn} & 35.94 & 32.57 & --- &&
    0.9244 & 0.8672 & --- &&
    \underline{0.1538} & \underline{0.1855} & --- && 0.1282\\
&    \multicolumn{1}{l}{DeepSent} & \textbf{38.59} & \textbf{34.73} & \textbf{32.28} &&
    \textbf{0.9566} & \textbf{0.9186} & \textbf{0.9124} &&
    \textbf{0.1065} & \textbf{0.1260} & \textbf{0.1322} && \textbf{0.0793}\\

\Xhline{2\arrayrulewidth}

\end{tabular}
\label{table:sim_results_table}
\end{table*}

\begin{figure}[h!]
\centering
\renewcommand{\tabcolsep}{0mm}
\begin{tabular}{c}
\includegraphics[width=0.48\textwidth]{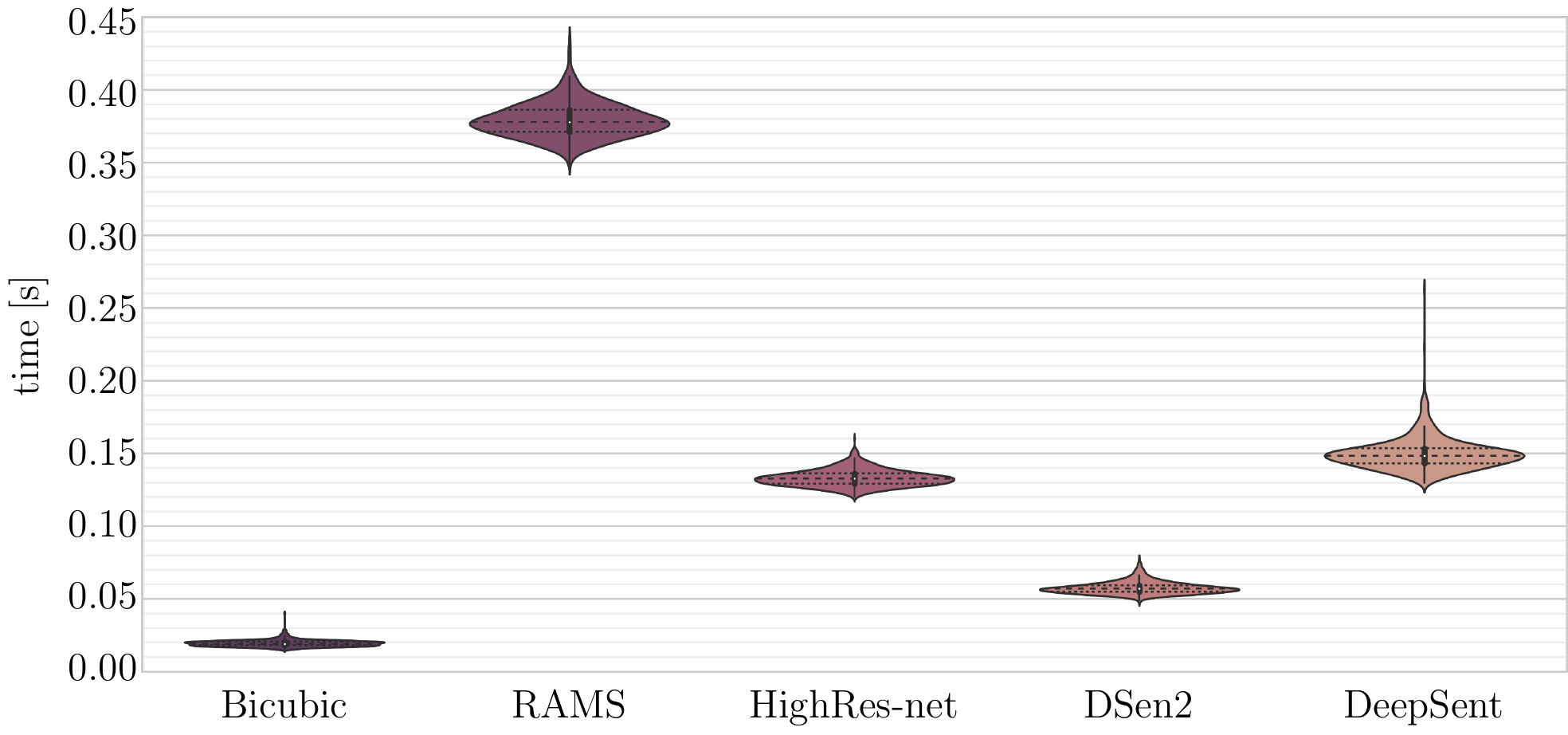} \\
\end{tabular}

\caption{Distribution of the processing times necessary to reconstruct a multispectral image of $288 \times 288$ pixels. The dashed lines indicate the median and quartile values. }
\label{fig:times}
\end{figure}

We have trained RAMS, HighRes-net and DSen2 using the LR images from the \sims\, set, and we bicubically readjusted the size of all HR target images to match the shape of their output. Thus, HR images used to train HighRes-net and RAMS networks are $3\times$ larger than their LR counterparts, while for DSen2 we maintain the upsampling factor of $6\times$ and $2\times$ for 60\,m and 20\,m bands. Moreover, we ensured that each model was provided with the training samples in the very same order during training. The batch for \mymethod\, includes four patches, and we utilized the Adam optimizer with the learning rate of $2\cdot 10^{-4}$, maximum of 200 epochs, and the early stopping condition was activated, if there was no improvement in the loss value in 15 consecutive epochs. As the loss function, we utilized the corrected mean squared error (cMSE) which is widely applied for MISR problems~\cite{Arefin2020, Dorr2020} and tightly related to the cPSNR metric. The training hyperparameters (excluding the batch size, as discussed below) were kept consistent across all investigated architectures. 
For HighRes-net and RAMS, we used the batches of 32, 64 and 196 patches for 10\,m, 20\,m and 60\,m S-2 bands, and we trained the DSen2 models for reconstructing 20\,m and 60\,m bands with the batch sizes of 24 and 16 patches. This is strictly related to the available memory of our GPU device, which is NVIDIA RTX 3090 equipped with 24 GB VRAM. Our techniques were implemented in Python 3.7 and Pytorch 1.10 deep learning library. The distribution of processing times needed to reconstruct a 12-band multispectral patch of $288\times 288$ pixels are presented in Fig.~\ref{fig:times} (DSen2 reconstructs 8-bands, as it cannot be applied to the 10\,m bands). DeepSent is slightly (though significantly in the statistical sense) slower than HighRes-net, but over $2\times$ faster than RAMS.

\subsection{Validation results for the simulated data}

\begin{figure}[ht!]
\centering
\renewcommand{\tabcolsep}{0mm}
\begin{tabular}{c}
\includegraphics[width=0.48\textwidth]{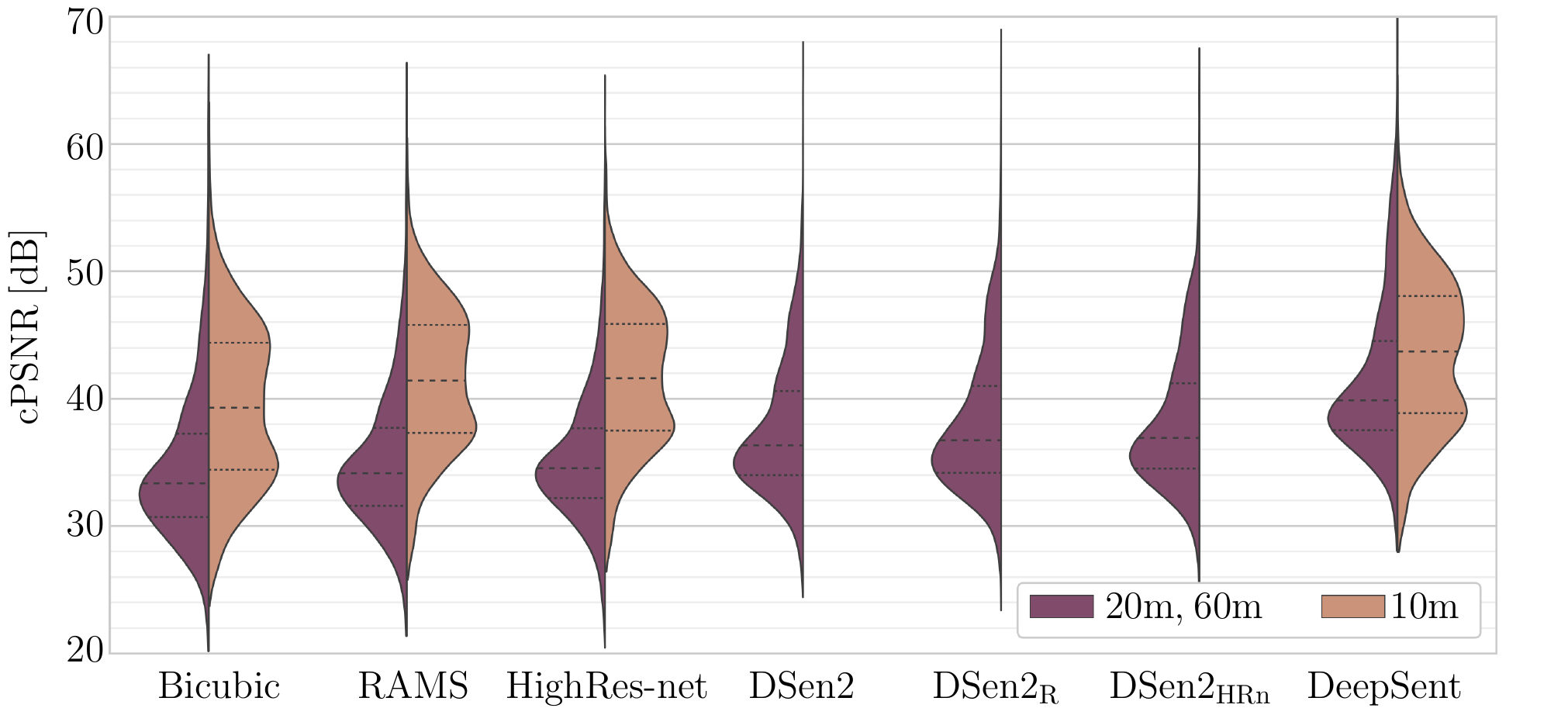} \\
\rule{0pt}{5mm}
\includegraphics[width=0.48\textwidth]{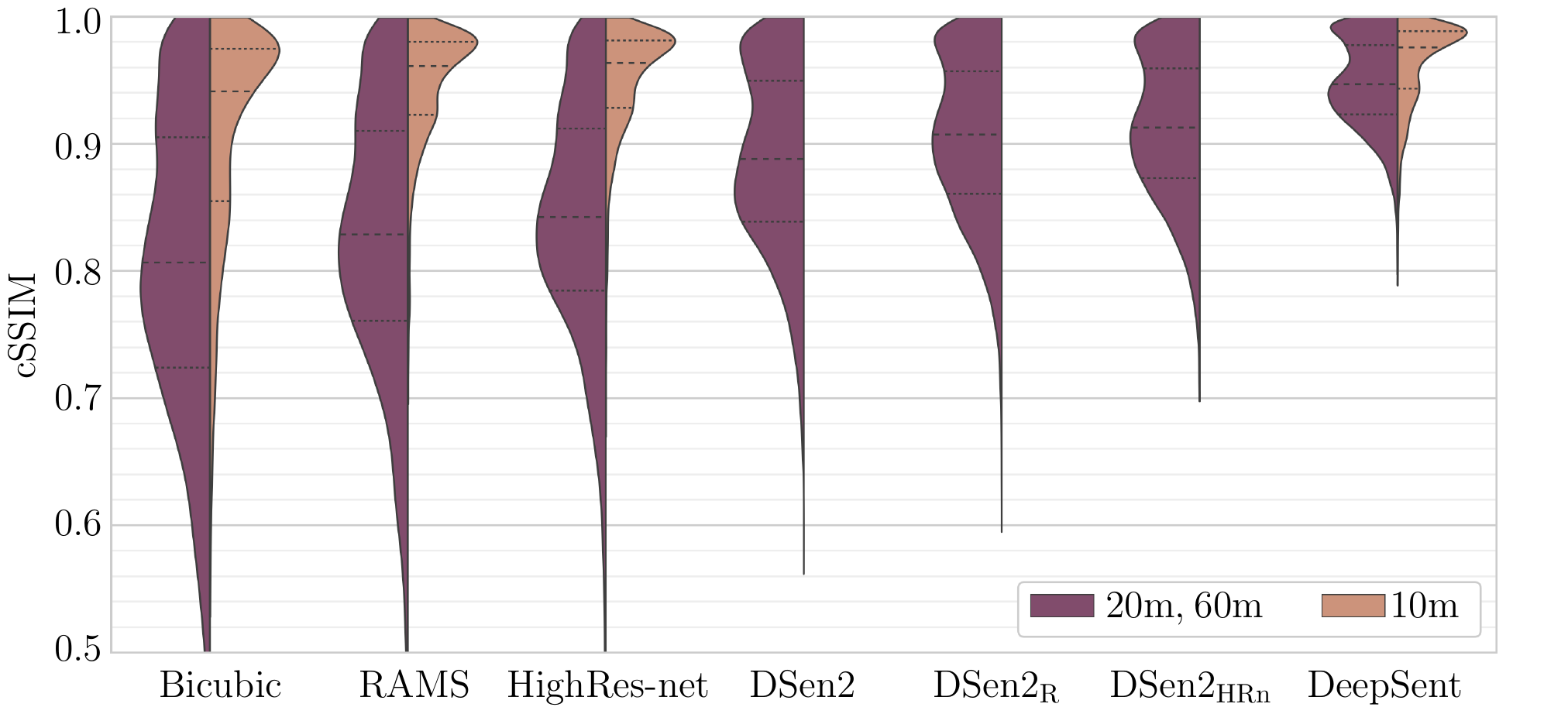} \\
\includegraphics[width=0.48\textwidth]{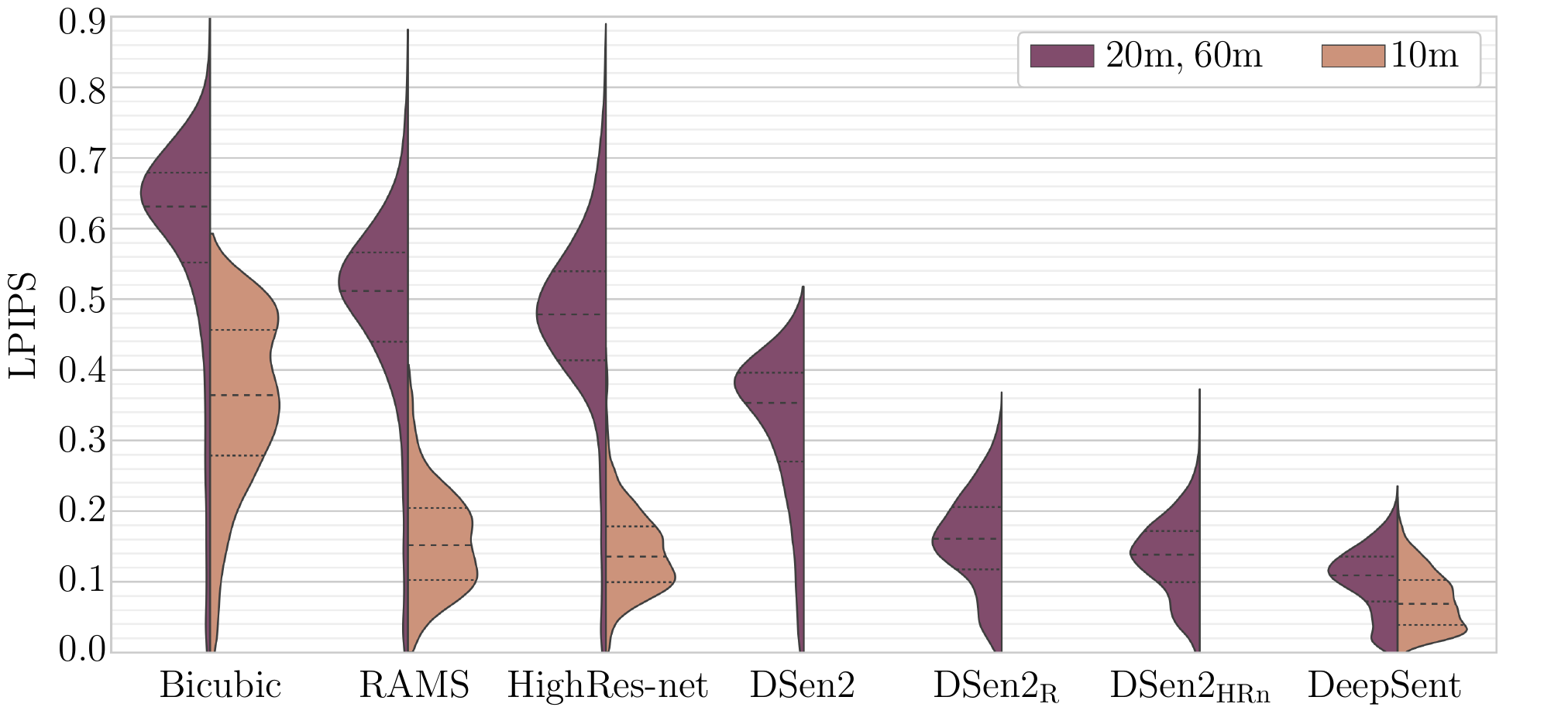} \\
\includegraphics[width=0.48\textwidth]{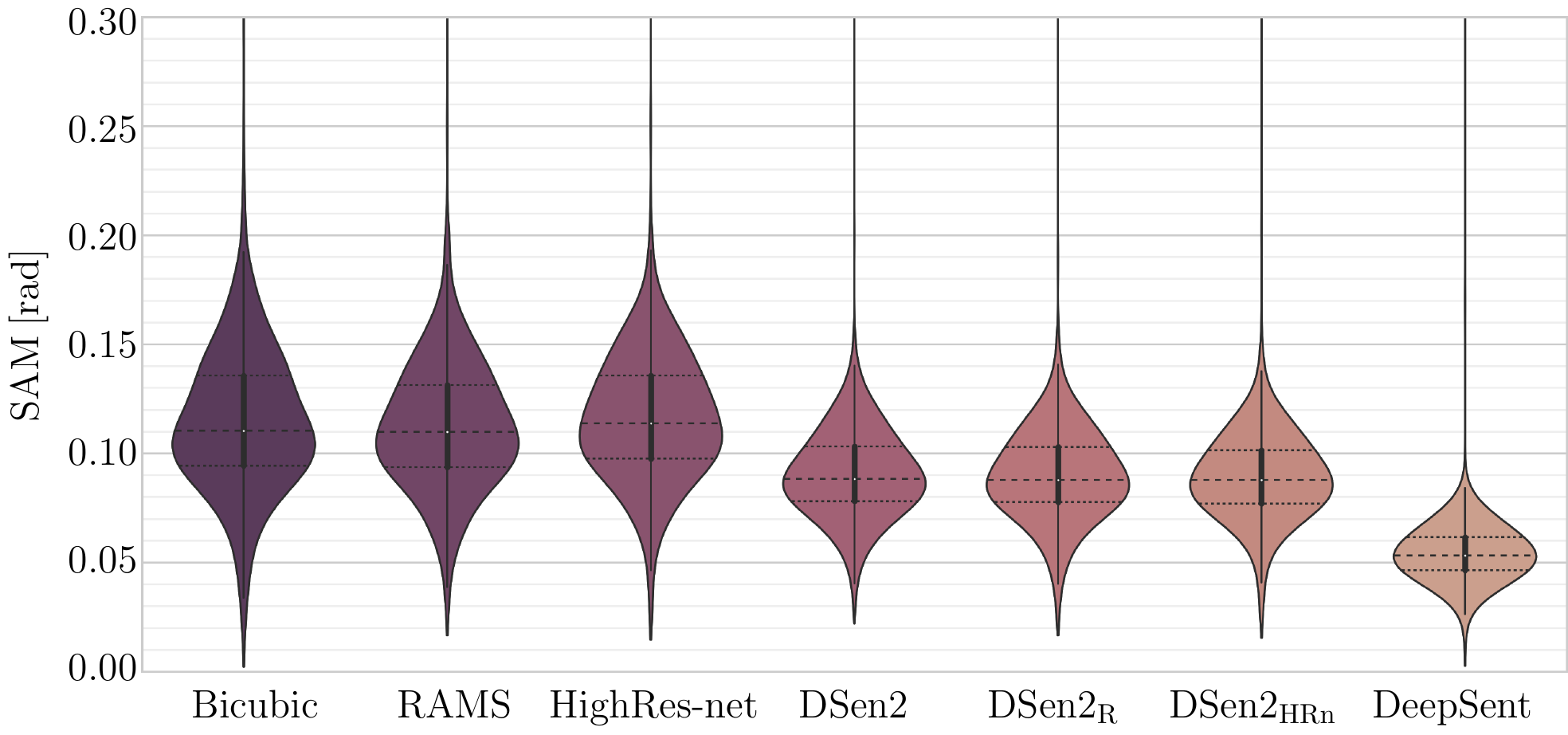}
\end{tabular}

\caption{Distribution of the reconstruction accuracy scores obtained for \sims\, dataset (its test set part), presented for the bands grouped according to their resolution. The dashed lines indicate the median and quartile values. }
\label{fig:sim_results_plot}
\end{figure}

\begin{figure}[ht!]
\centering
\renewcommand{\tabcolsep}{0mm}
\begin{tabular}{c}
\includegraphics[width=0.48\textwidth]{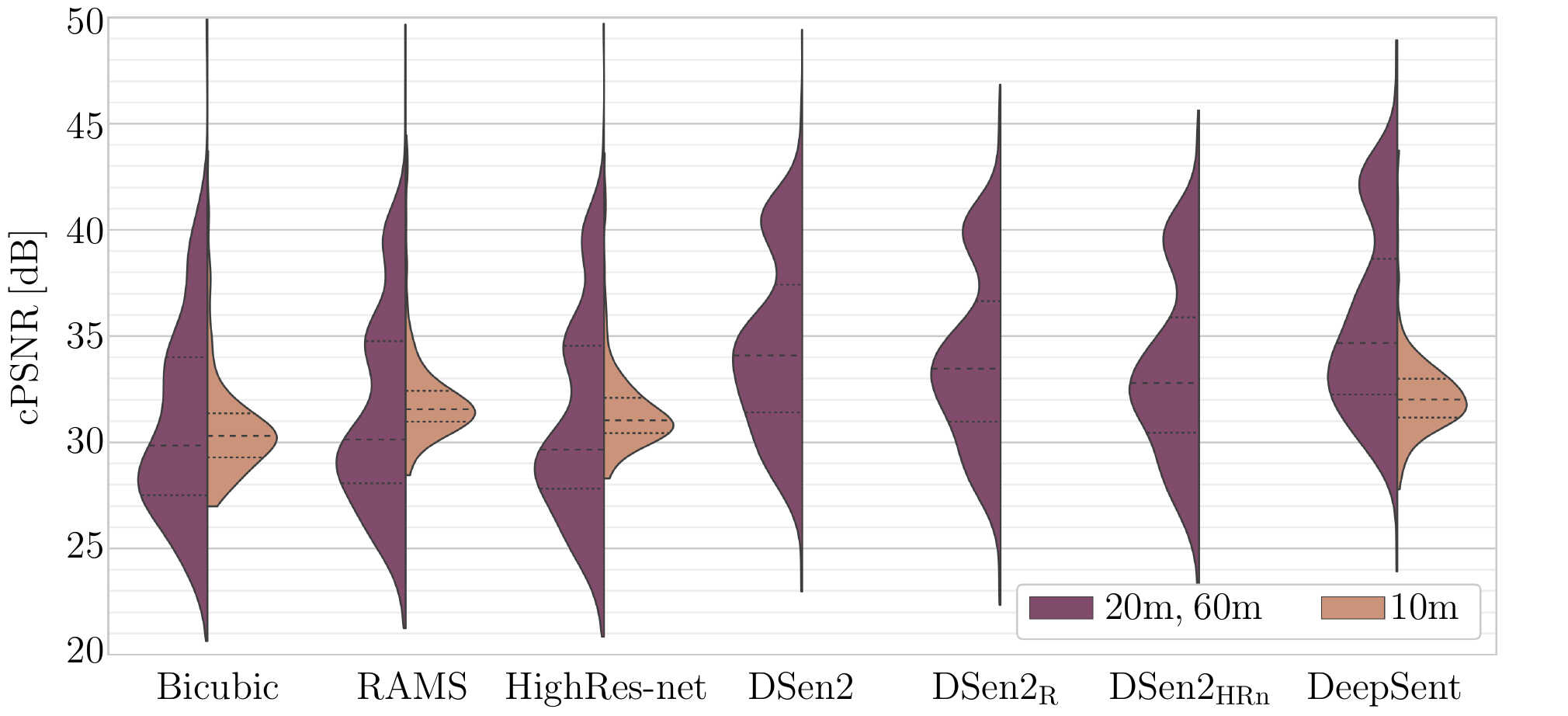} \\
\rule{0pt}{5mm}
\includegraphics[width=0.48\textwidth]{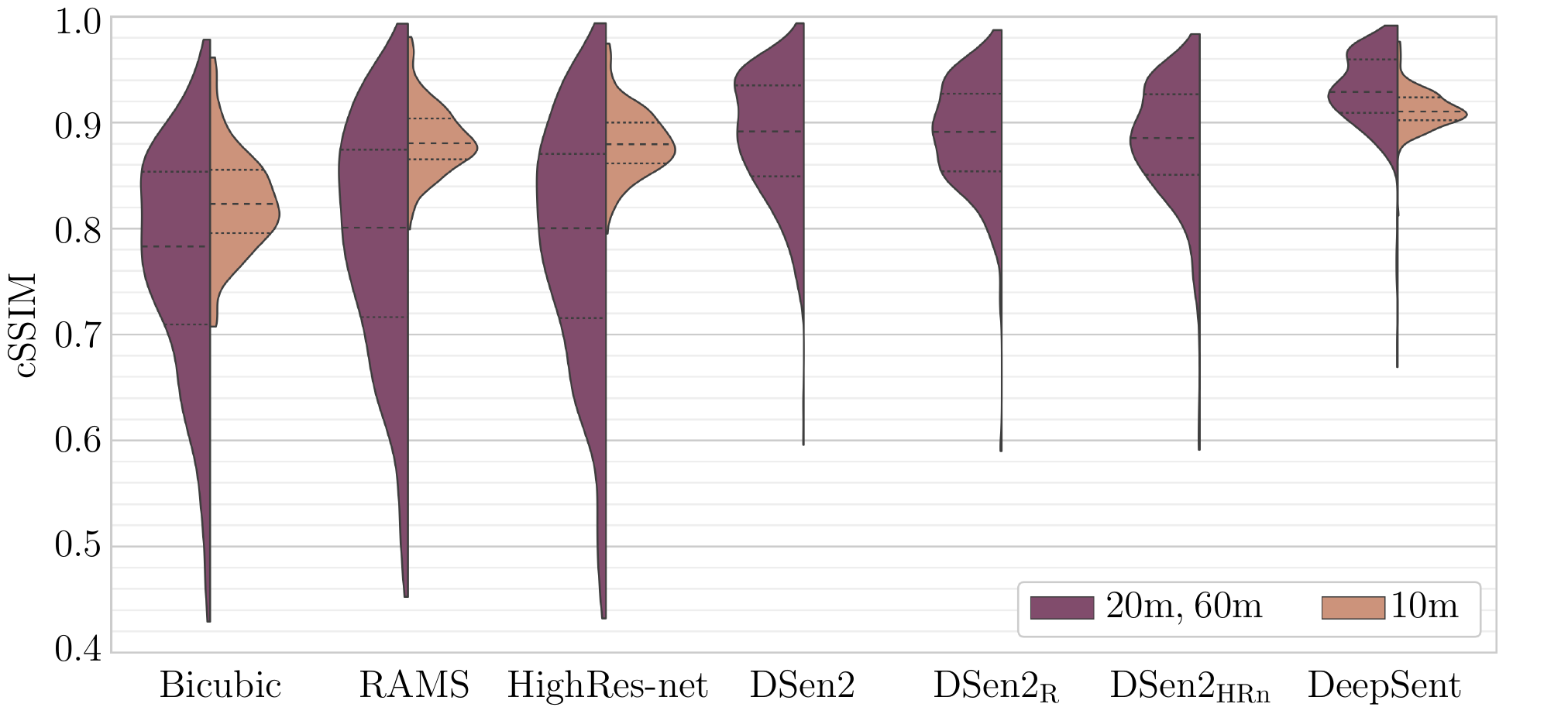} \\
\includegraphics[width=0.48\textwidth]{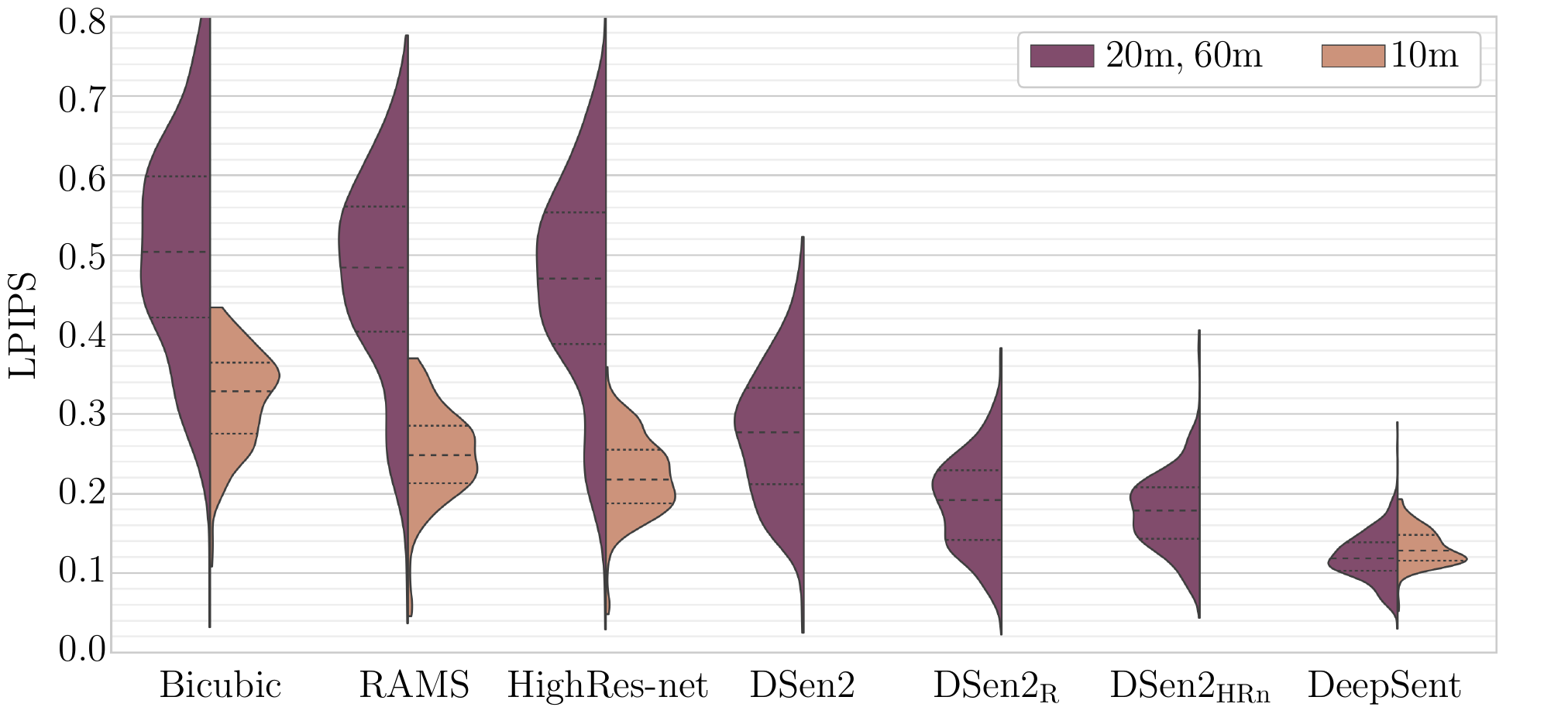} \\
\includegraphics[width=0.48\textwidth]{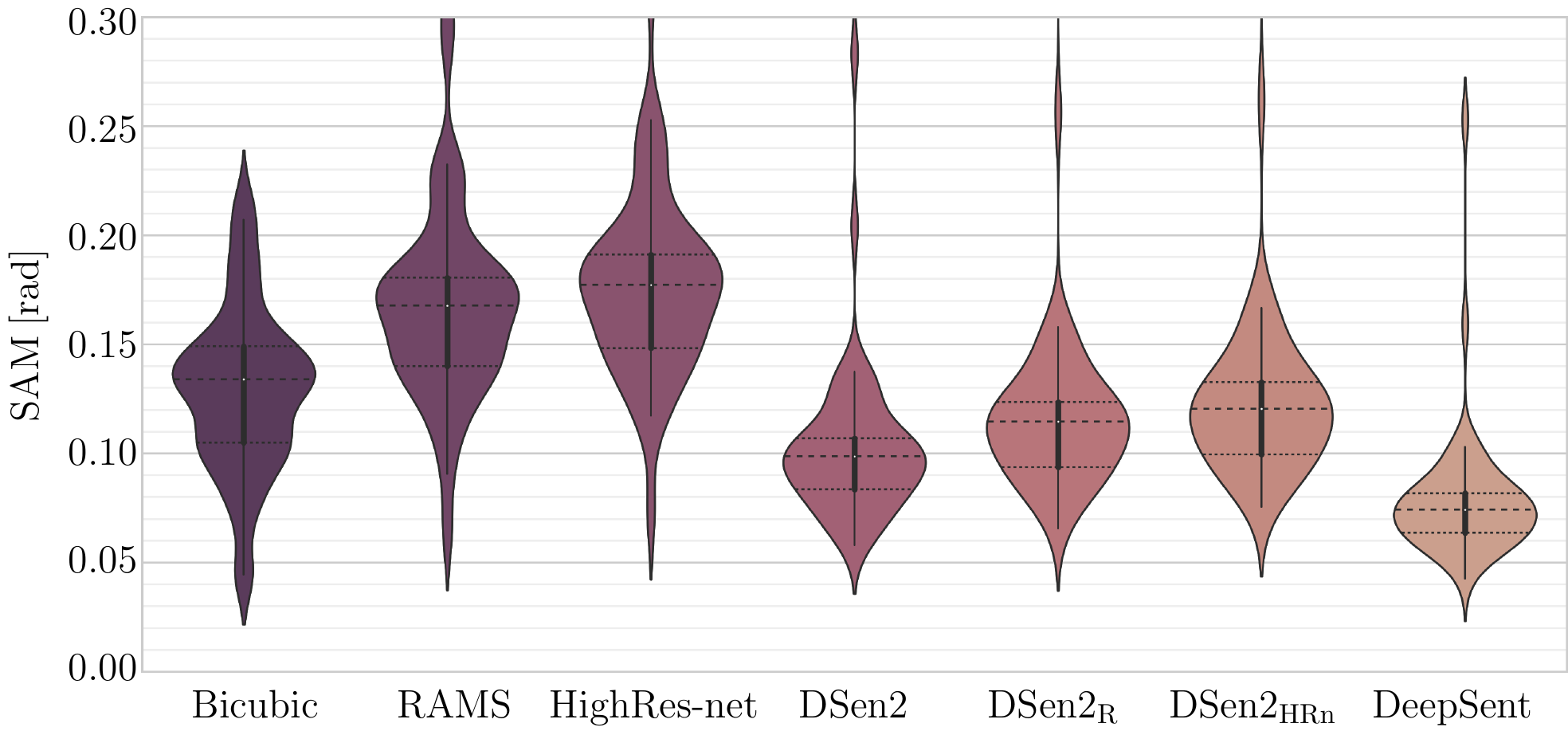} \\
\end{tabular}

\caption{Distribution of the reconstruction accuracy scores obtained for the \simhsi\, images, presented for the bands grouped according to their resolution. The dashed lines indicate the median and quartile values. }
\label{fig:sim_results_plot_hsi}
\end{figure}

In Table~\ref{table:sim_results_table}, we gather the metrics for every model over the simulated data (test set for \sims\, and \simhsi), and in Figs.~\ref{fig:sim_results_plot} and~\ref{fig:sim_results_plot_hsi} we show the distribution of the scores for the bands, grouped by their GSD resolution\footnote{The detailed quantitative scores for every band are presented in the Supplementary Material available at \url{https://gitlab.com/tarasiewicztomasz/deepsent}.}. 
DSen2 effectively benefits from the information contained in the higher-resolution spectral bands when reconstructing the 20\,m and 60\,m bands, and it surpasses both RAMS and HighRes-net.
This includes the 20\,m bands, even though DSen2 upsamples them $2\times$, compared with $3\times$ for RAMS and HighRes-net---this shows the benefits of exploiting information from the 10\,m bands.
Although multitemporal data fusion is affected by the natural variability of the Earth surface appearance, primarily caused by the vegetation characteristics, RAMS and HighRes-net offer substantial gain over the bicubic interpolation for 20\,m and 10\,m bands, similar to that reported for the PROBA-V dataset~\cite{Salvetti2020, deudon2020highresnet}.
For the \sims\, dataset, the \DSenR~and \DSenHRn\, variants that combine the multispectral and multitemporal data fusion in a sequential manner render higher scores (with all metrics) than any of their components standalone, and for \simhsi\, some of the metrics, including SAM, indicate DSen2 as a more accurate technique. However, in all the cases it is \mymethod\,  
which delivers significantly better results than all other techniques (in the statistical sense, verified for all the metrics, and in all bands separately). 

\begin{figure*}[ht!]
\centering
\footnotesize
\renewcommand{\tabcolsep}{0.3mm}
\renewcommand{\arraystretch}{0.66}
\newcommand{\mywidth}{0.16}
\begin{tabular}{ccccccc}
   ~~ & LR image & Bicubic interpolation & RAMS & HighRes-net  & DeepSent & HR image\\

    \raisebox{13.5mm}{a)} &
    \includegraphics[width=\mywidth\textwidth]{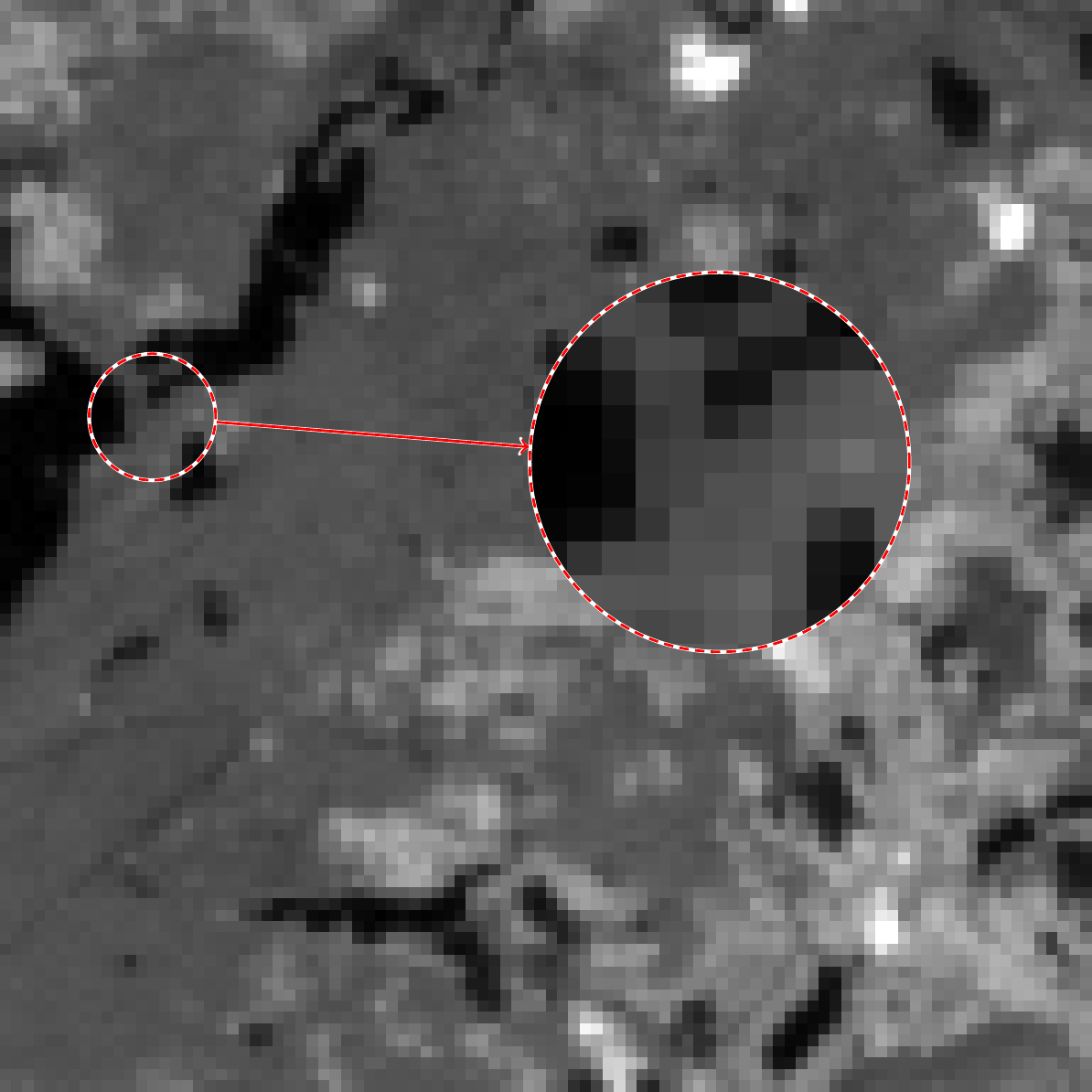} &
    \includegraphics[width=\mywidth\textwidth]{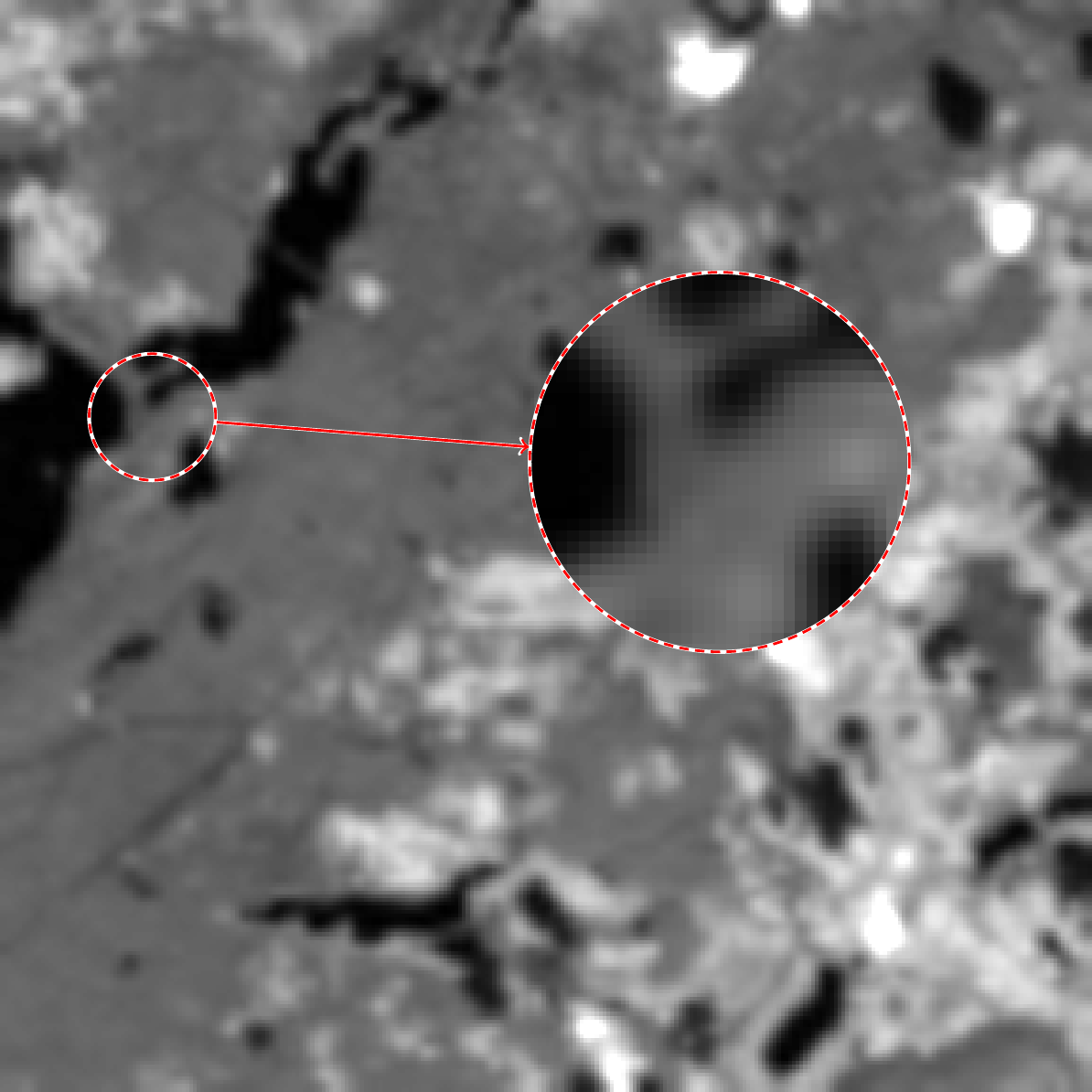} &
    \includegraphics[width=\mywidth\textwidth]{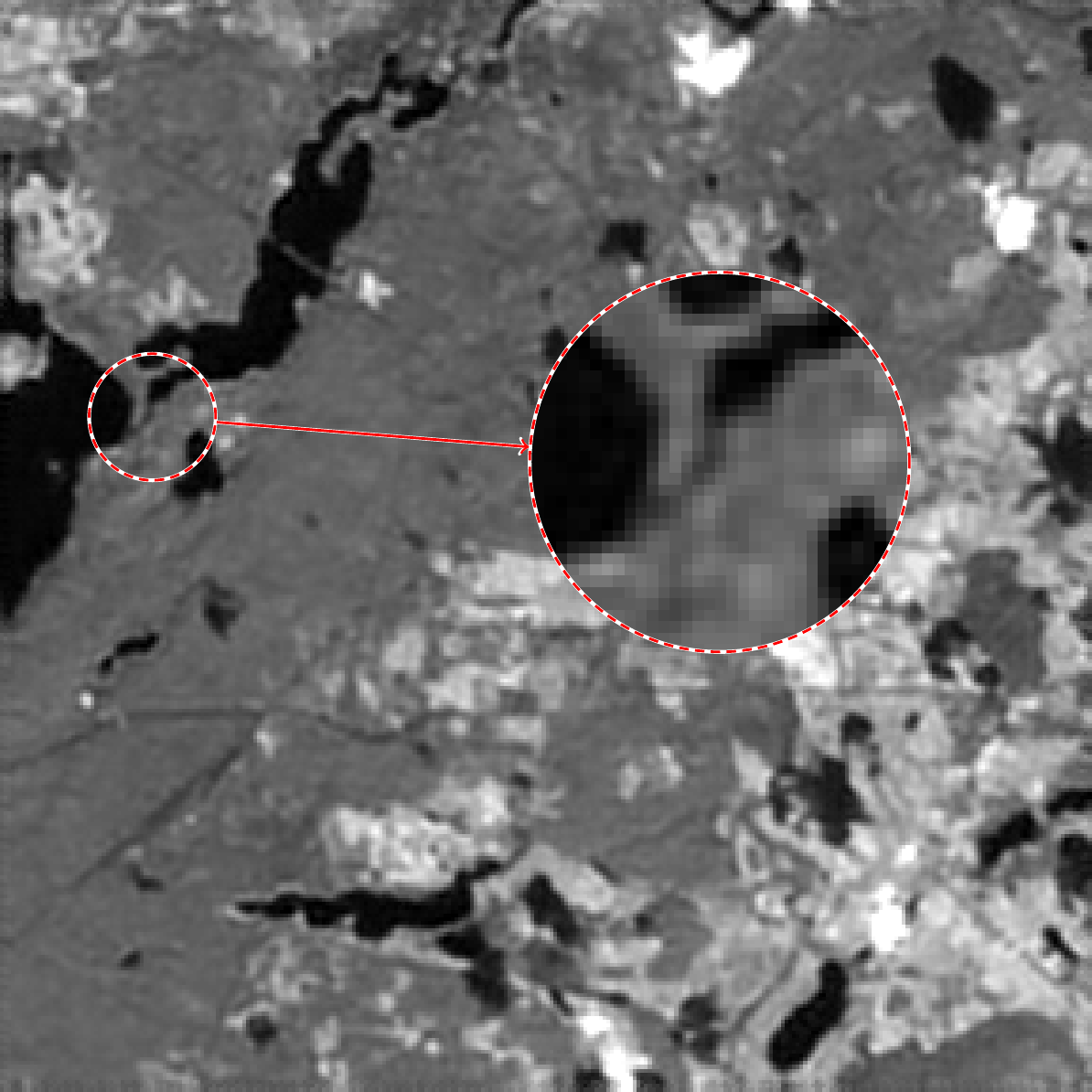} &
    \includegraphics[width=\mywidth\textwidth]{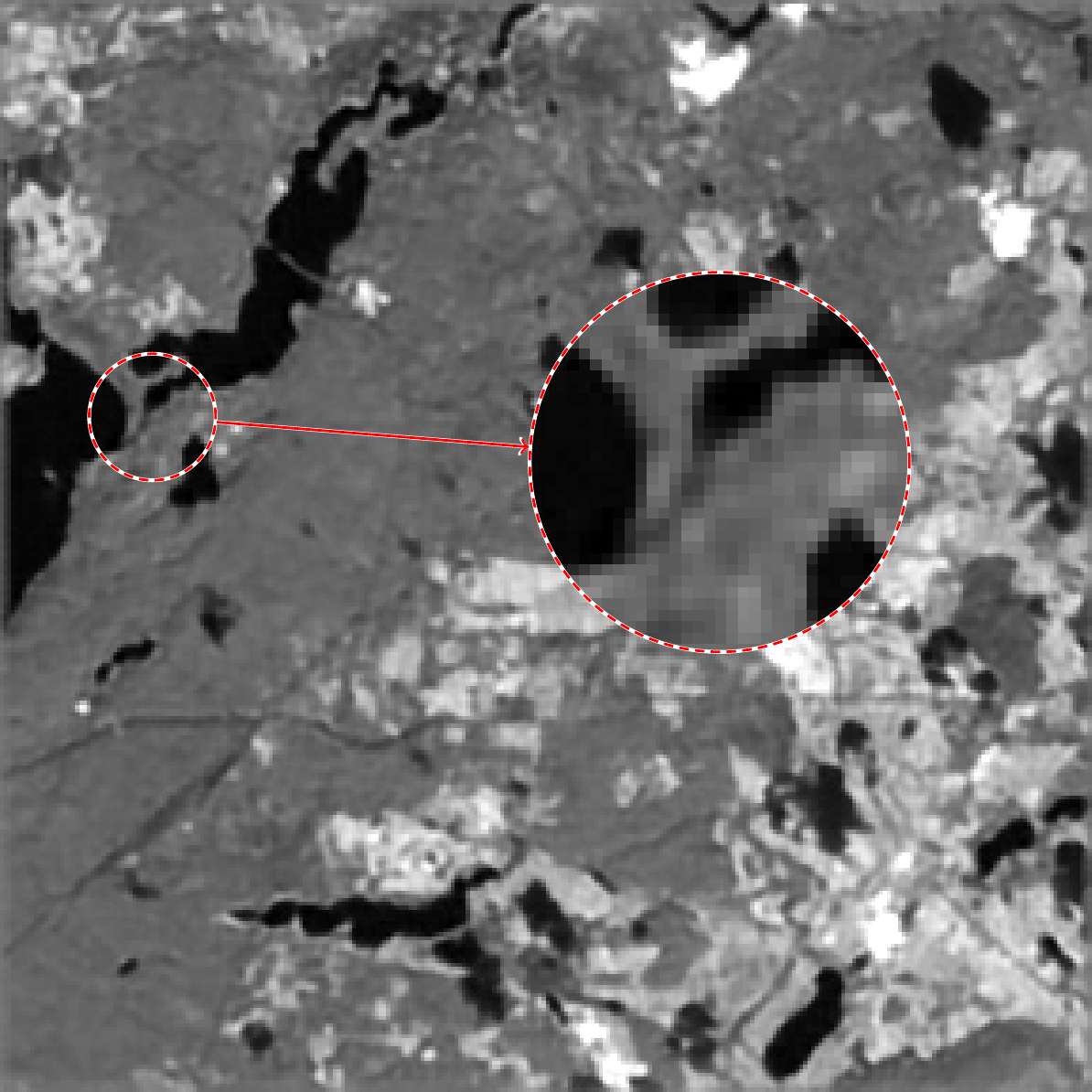} &
    \includegraphics[width=\mywidth\textwidth]{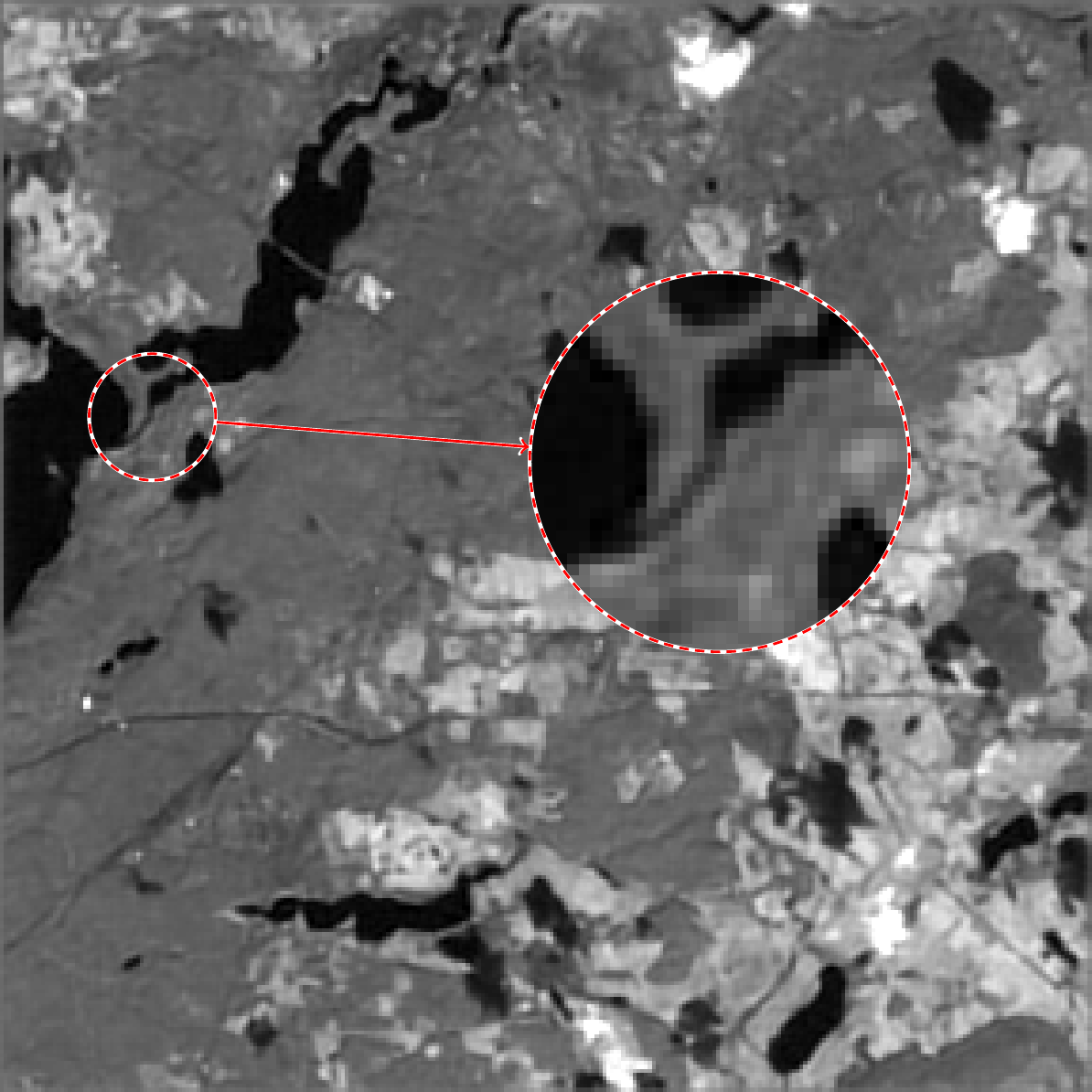} &
    \includegraphics[width=\mywidth\textwidth]{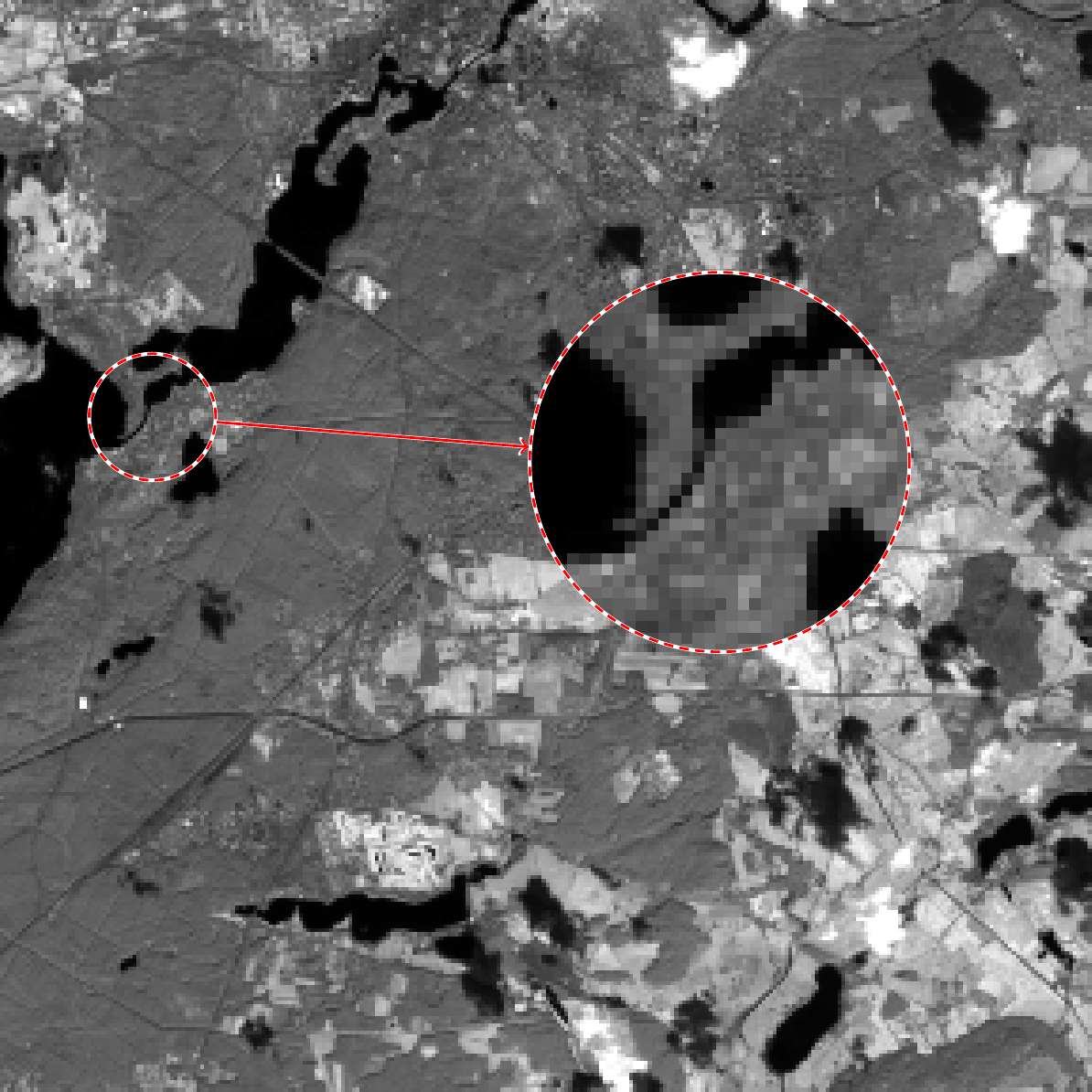}  \\

    \raisebox{13.5mm}{b)} &
    \includegraphics[width=\mywidth\textwidth]{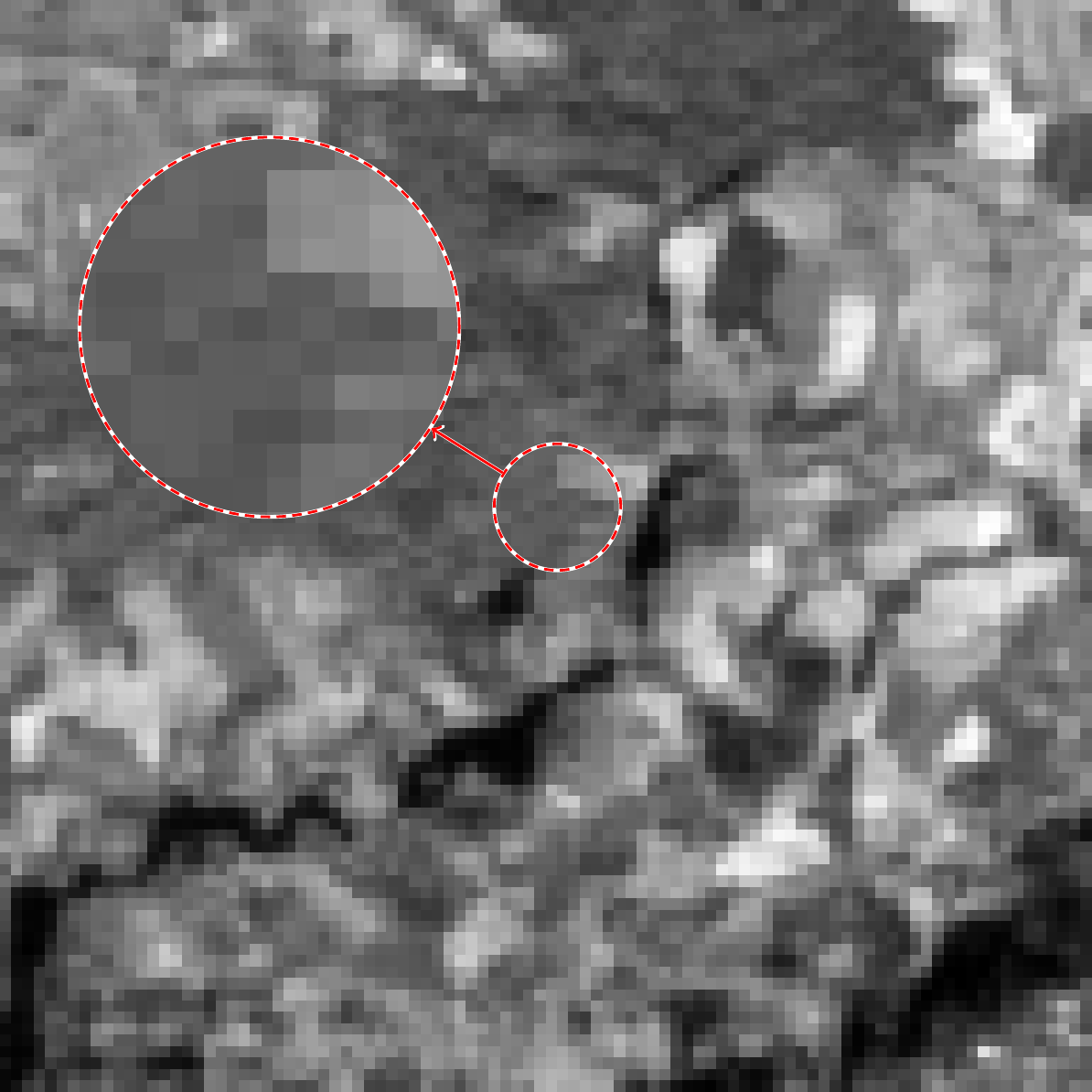} &
    \includegraphics[width=\mywidth\textwidth]{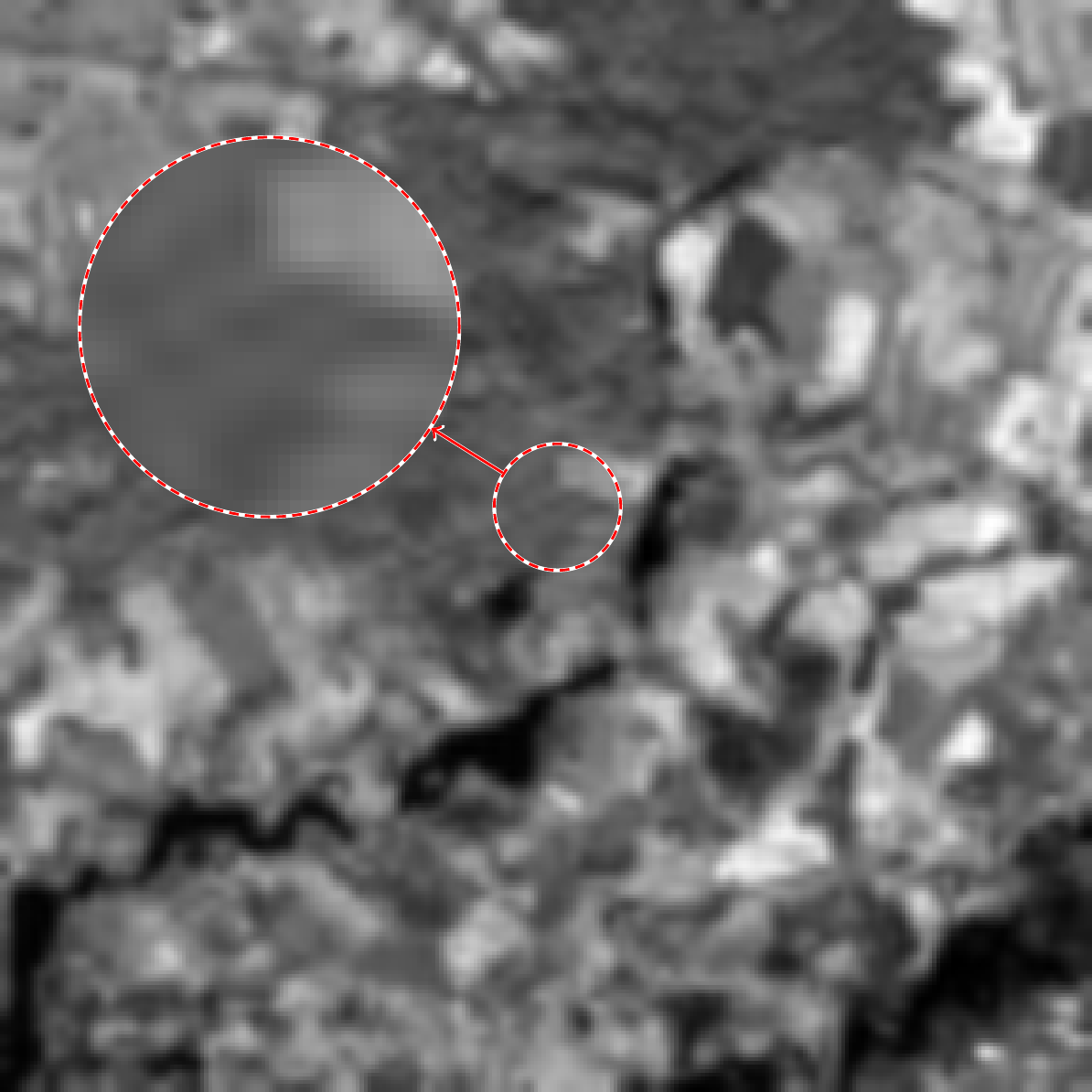} &
    \includegraphics[width=\mywidth\textwidth]{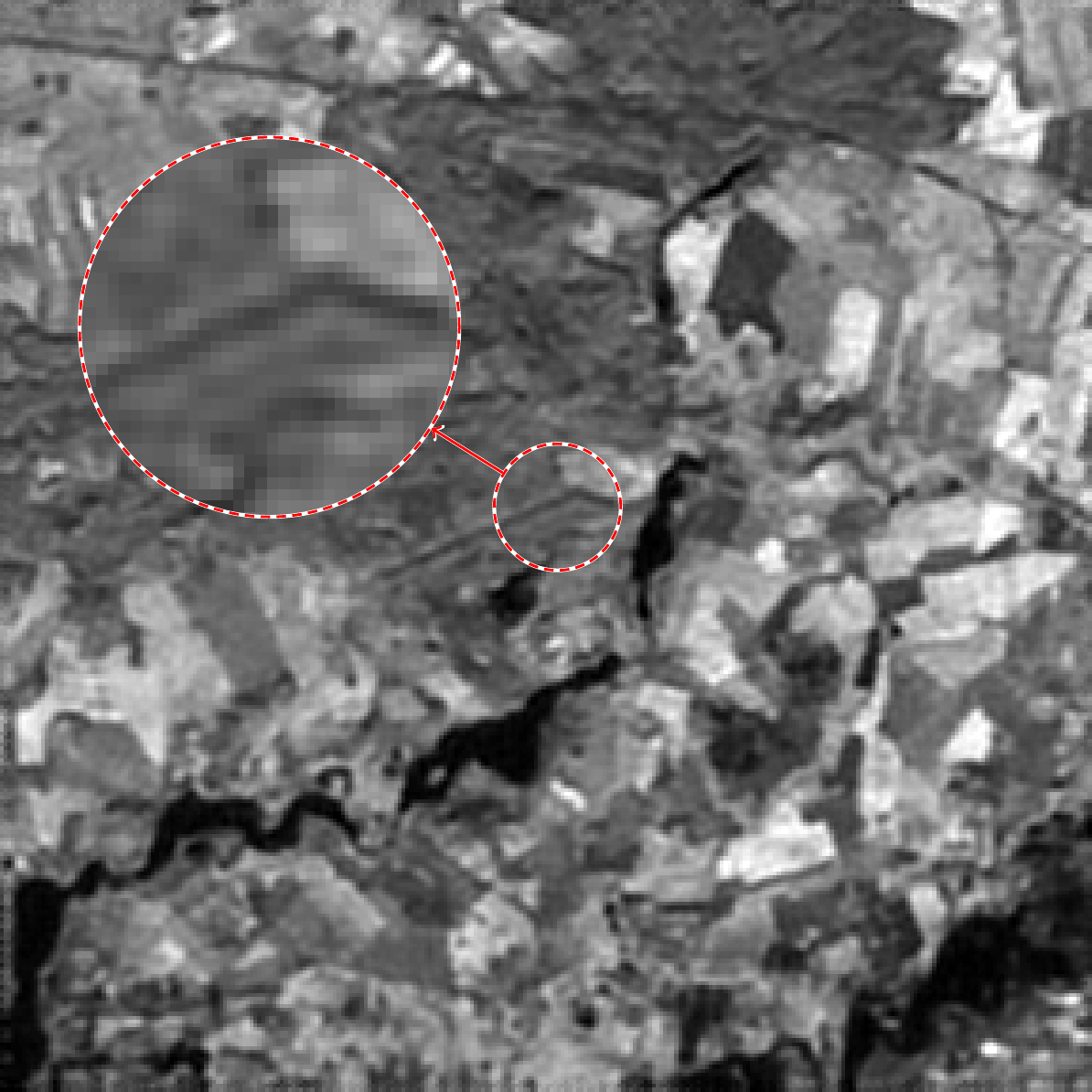} &
    \includegraphics[width=\mywidth\textwidth]{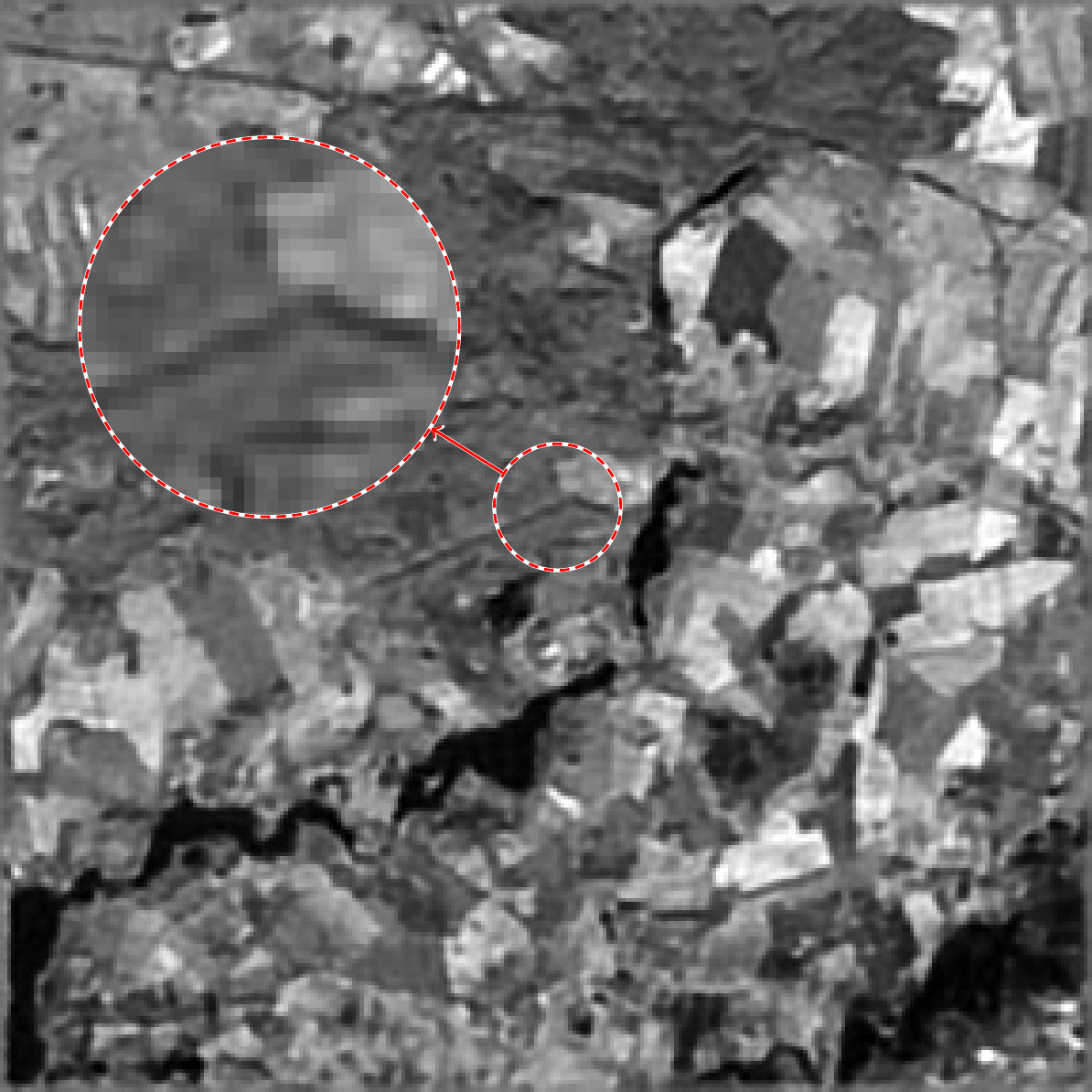} &
    \includegraphics[width=\mywidth\textwidth]{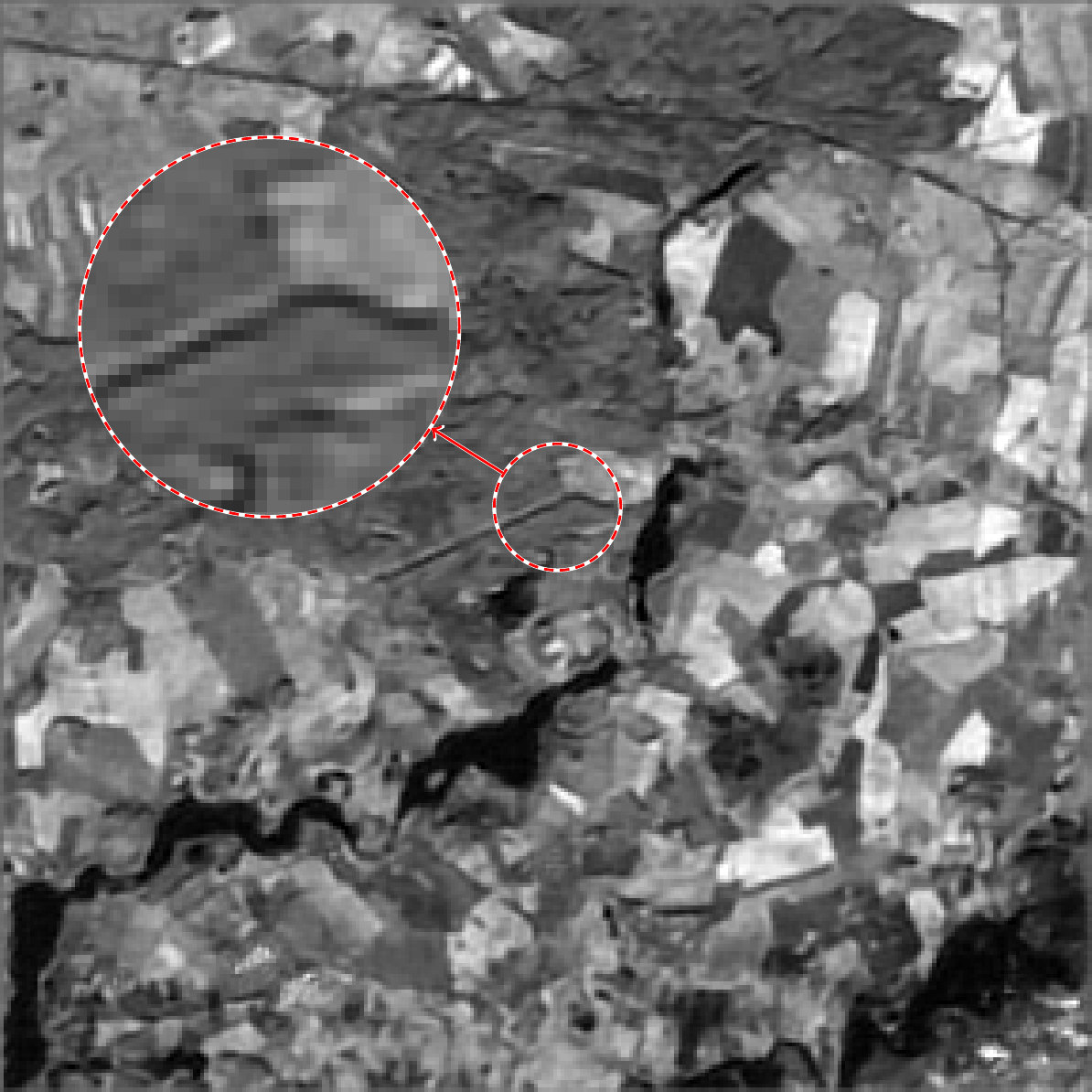} &
    \includegraphics[width=\mywidth\textwidth]{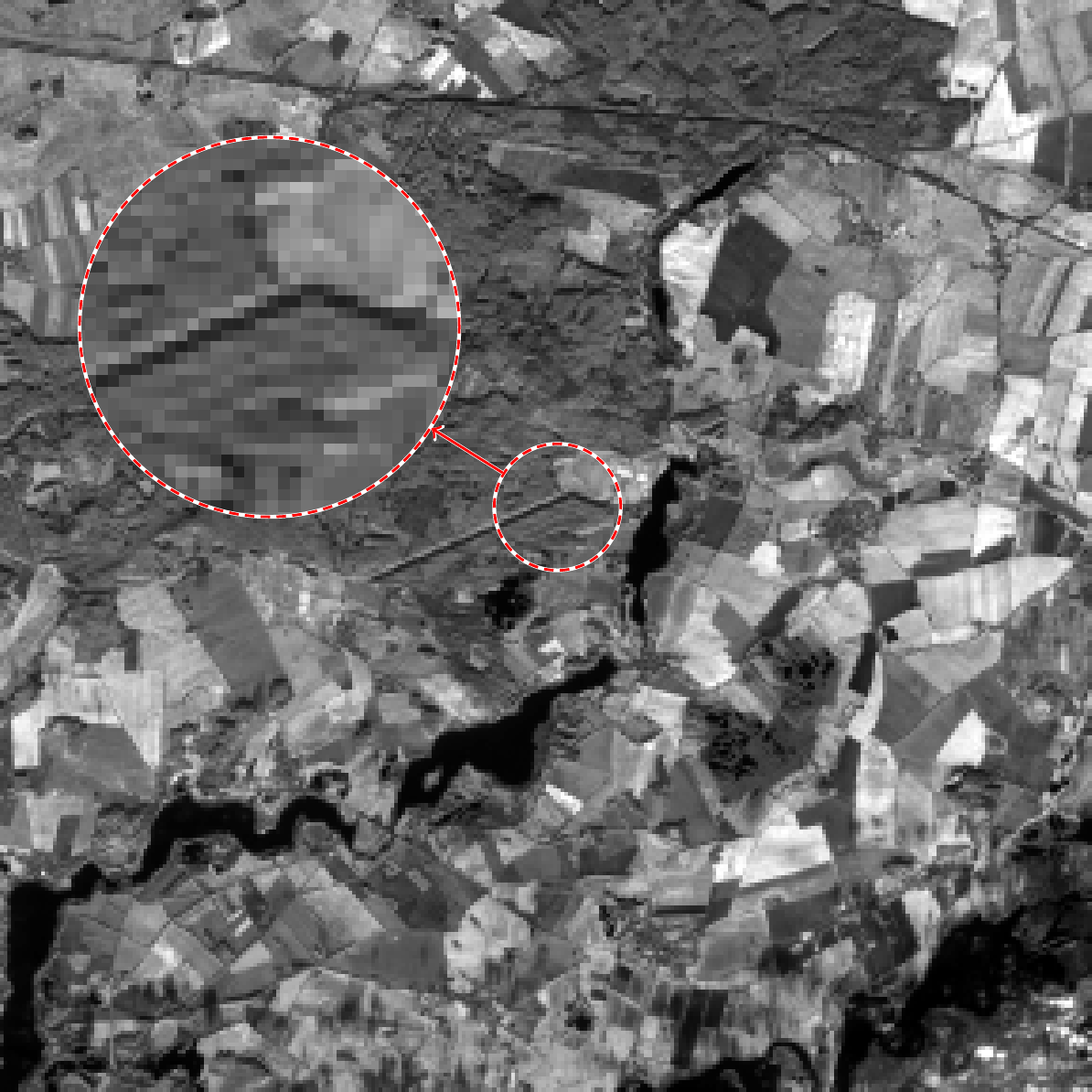}  \\

    \raisebox{13.5mm}{c)} &
    \includegraphics[width=\mywidth\textwidth]{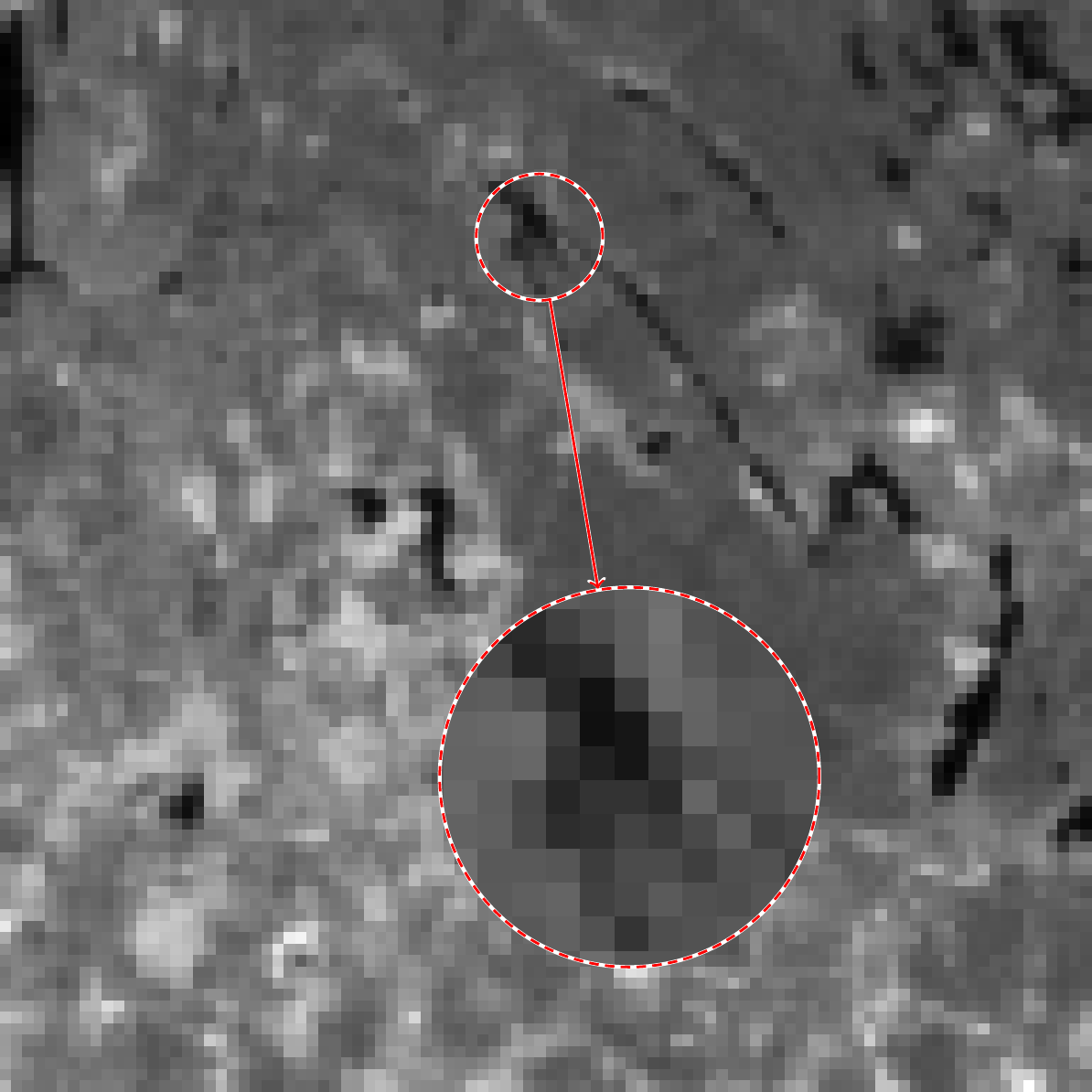} &
    \includegraphics[width=\mywidth\textwidth]{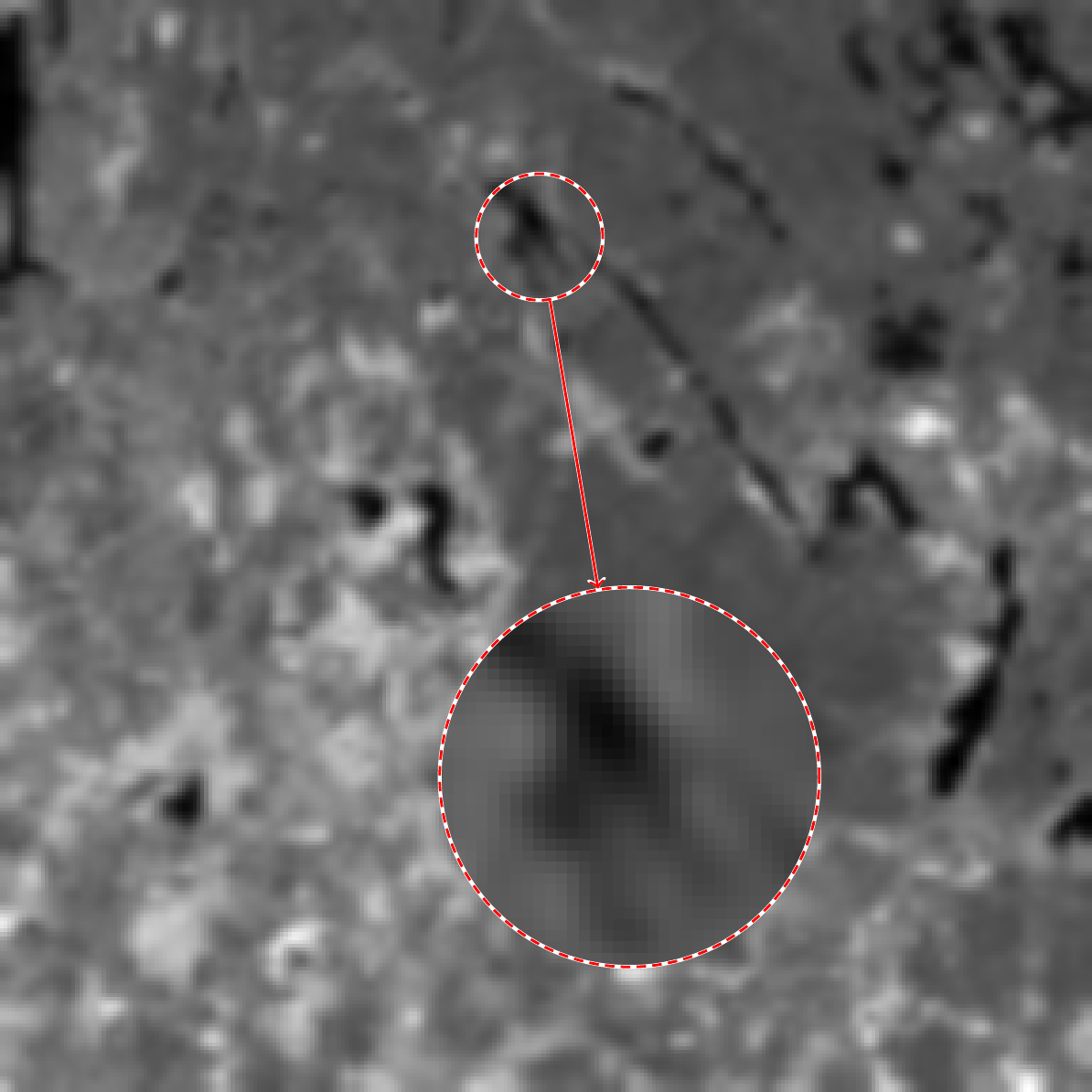} &
    \includegraphics[width=\mywidth\textwidth]{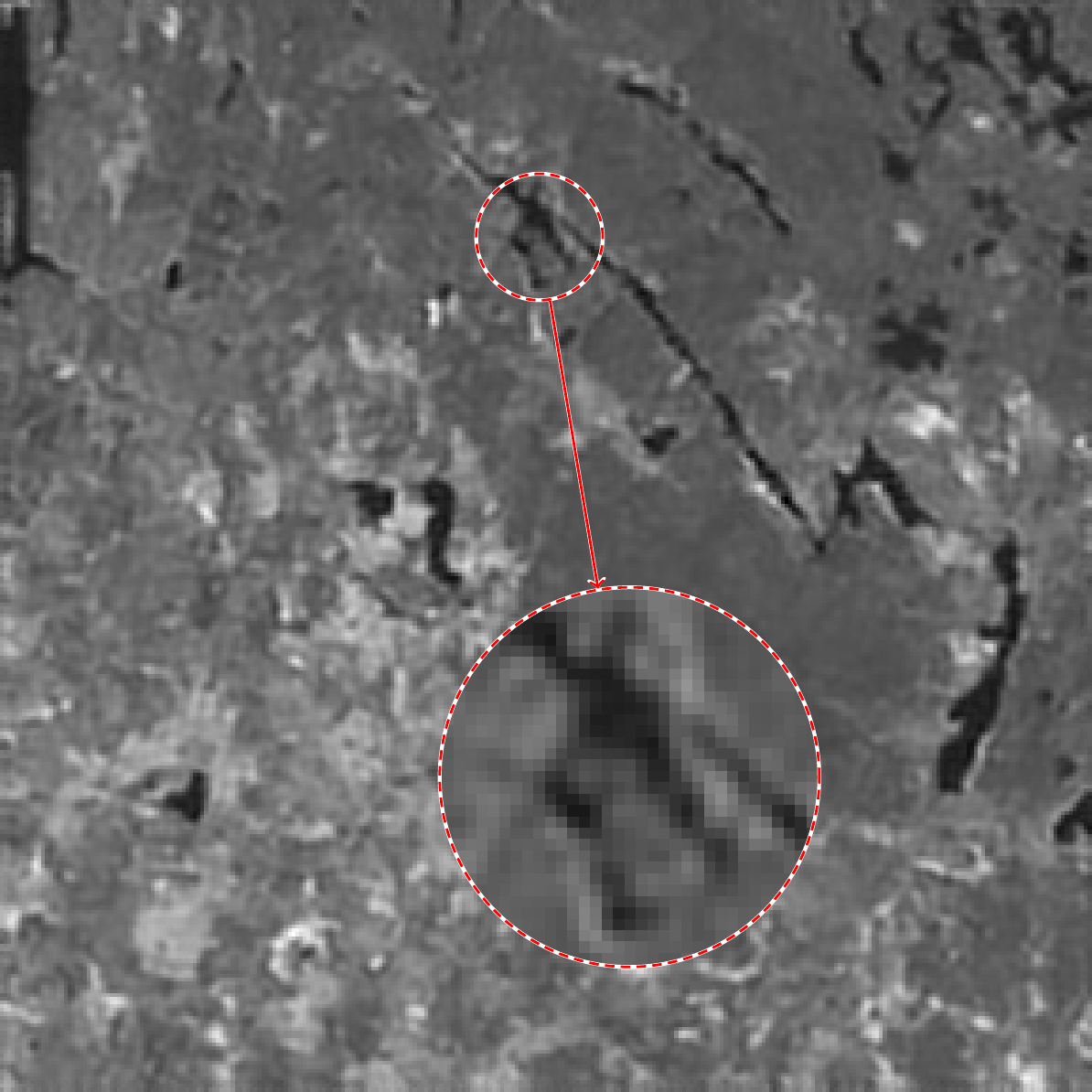} &
    \includegraphics[width=\mywidth\textwidth]{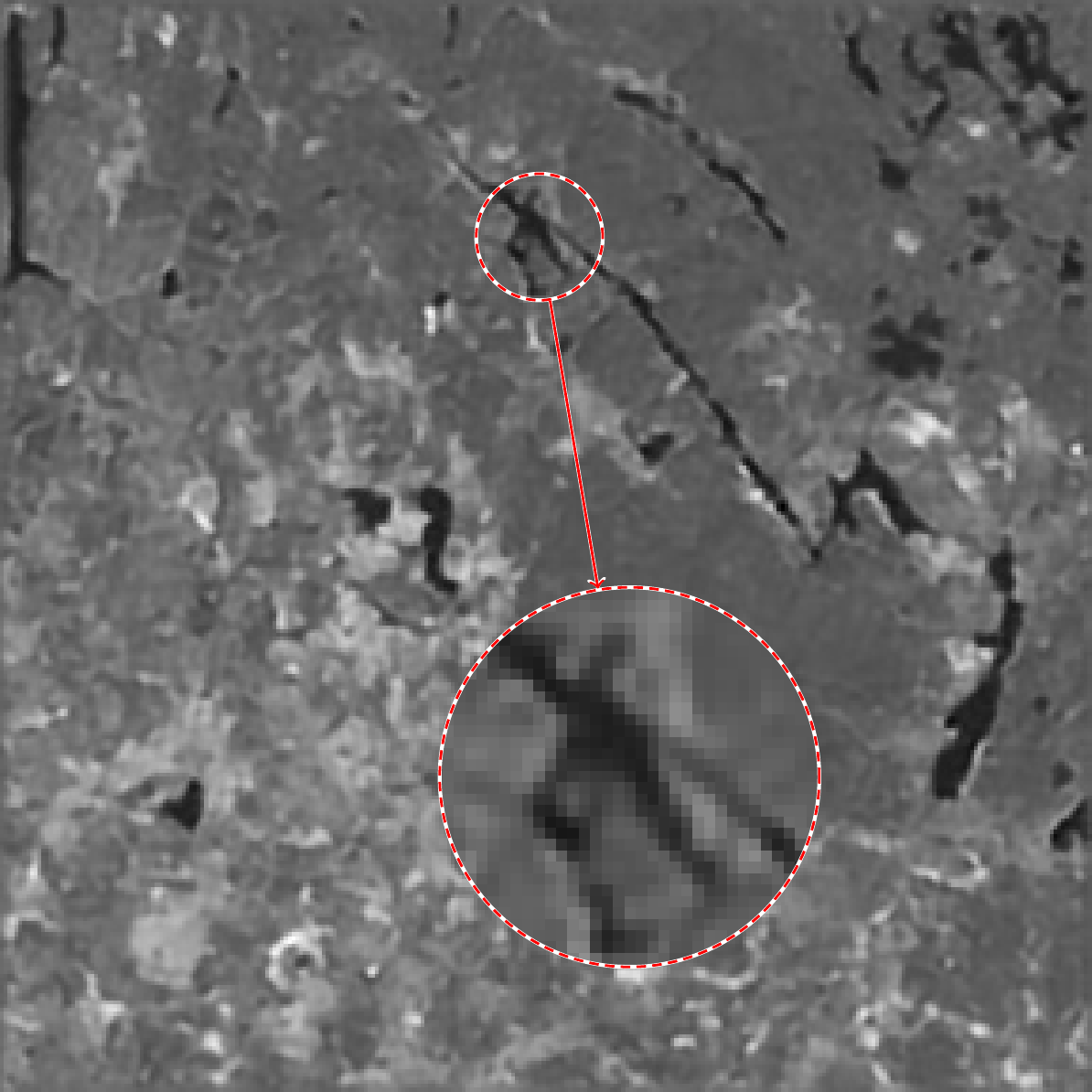} &
    \includegraphics[width=\mywidth\textwidth]{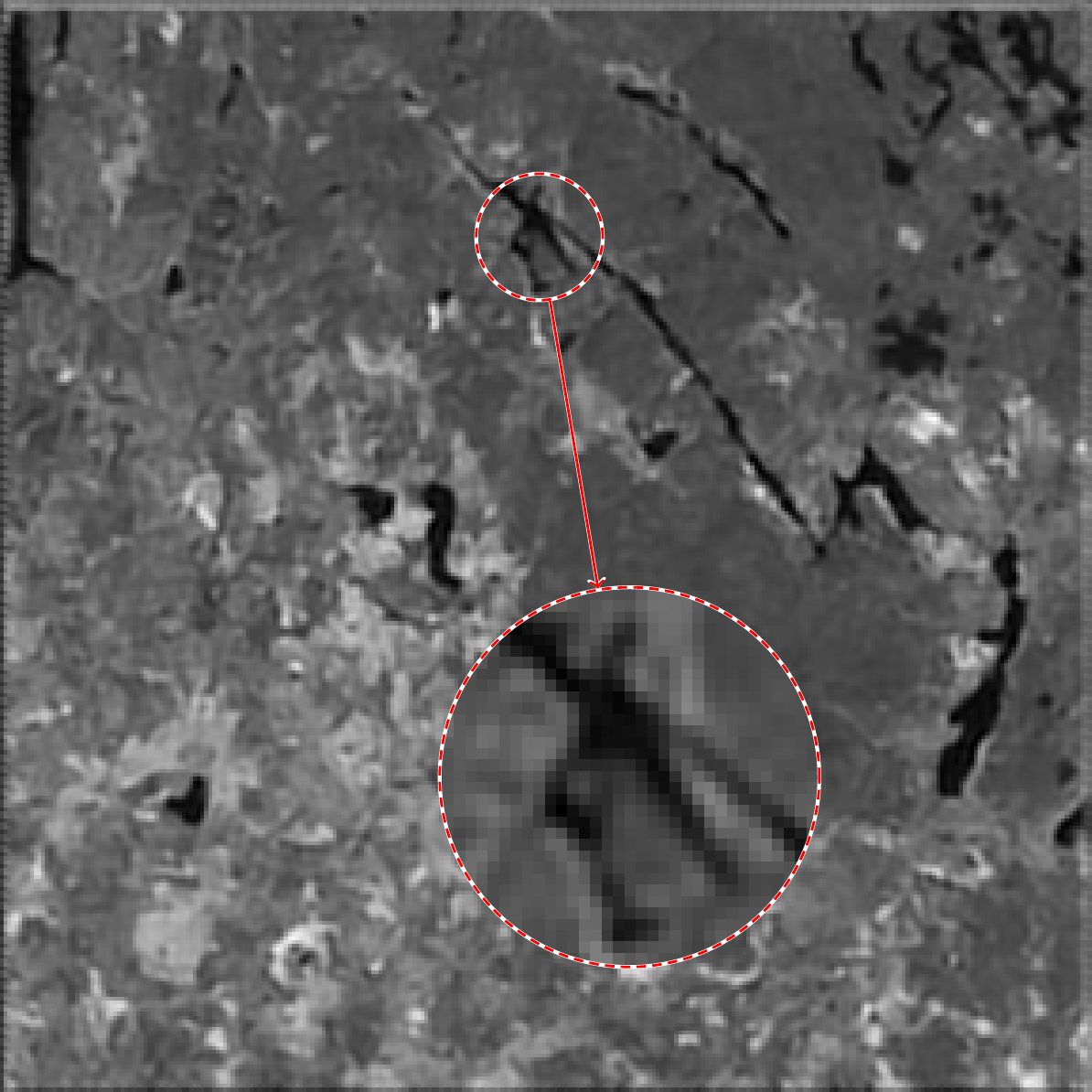} &
    \includegraphics[width=\mywidth\textwidth]{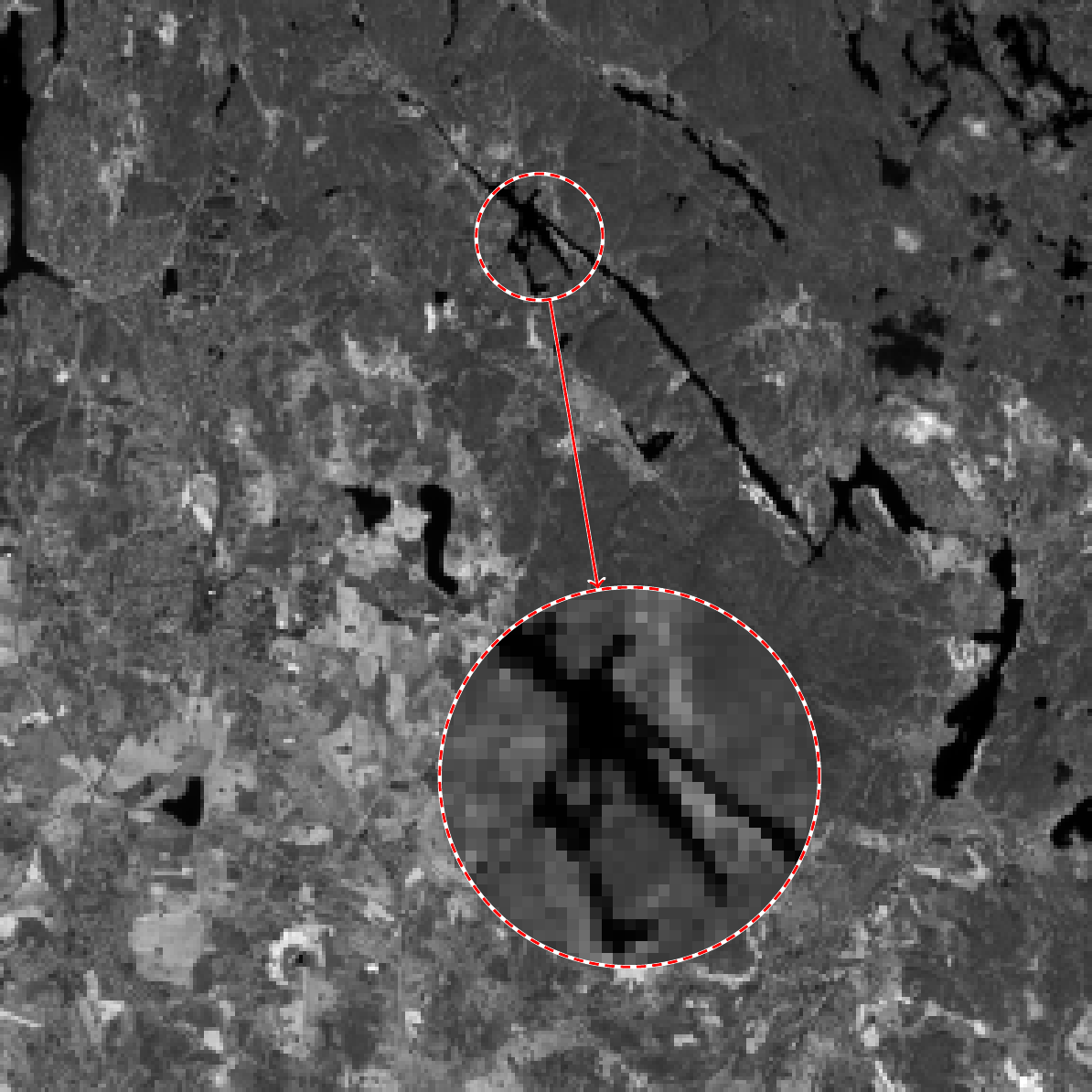}  \\


    \raisebox{13.5mm}{d)} &
    \includegraphics[width=\mywidth\textwidth]{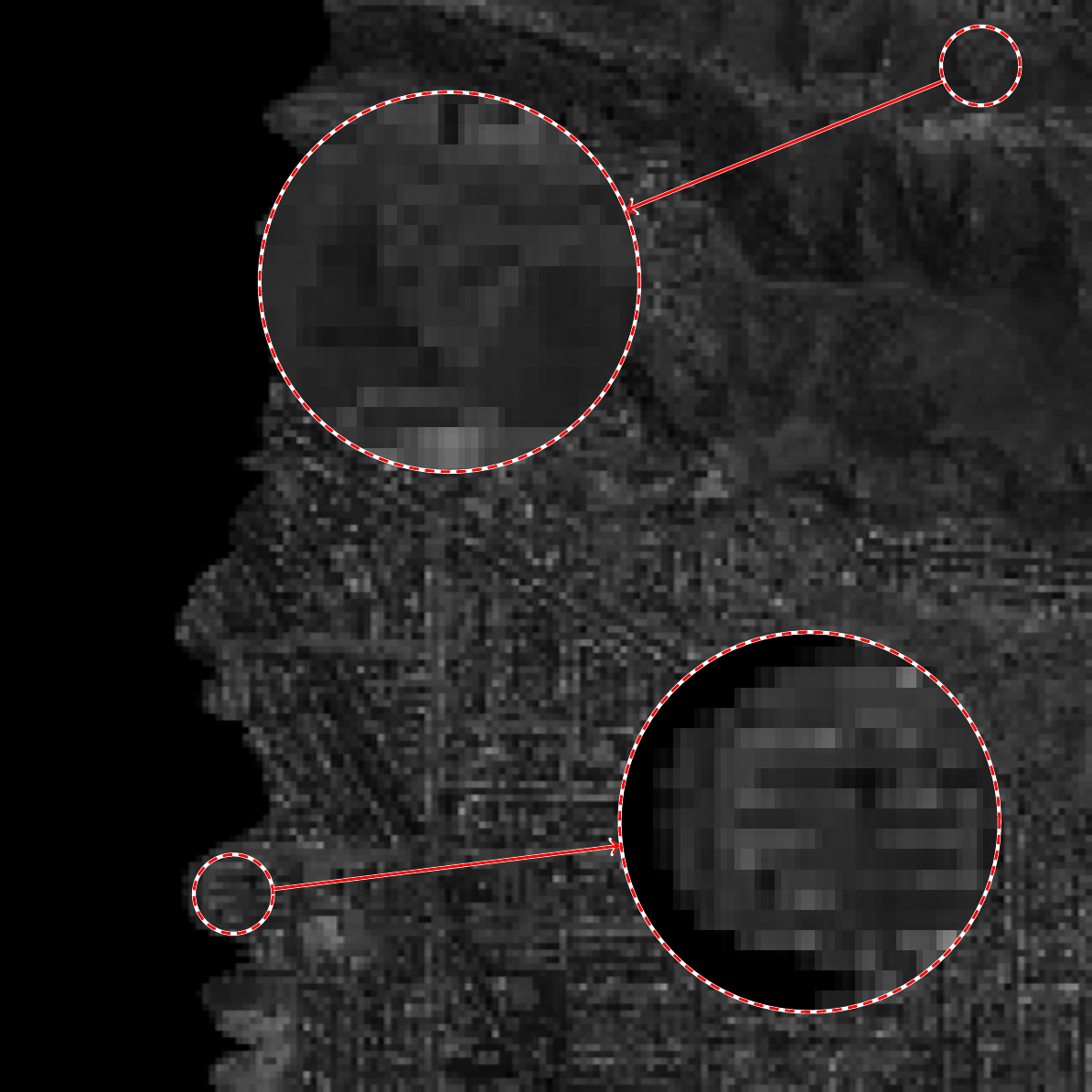} &
    \includegraphics[width=\mywidth\textwidth]{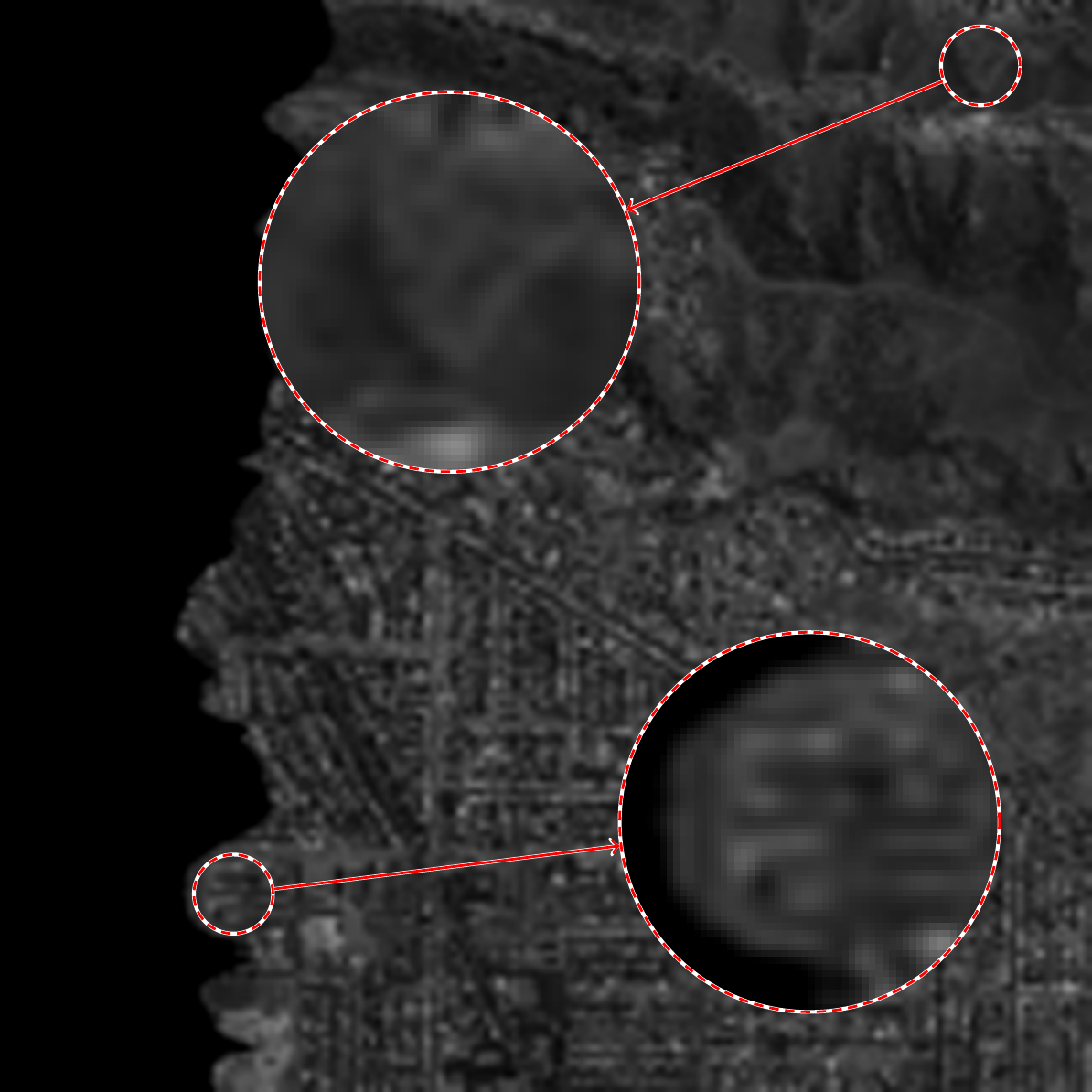} &
    \includegraphics[width=\mywidth\textwidth]{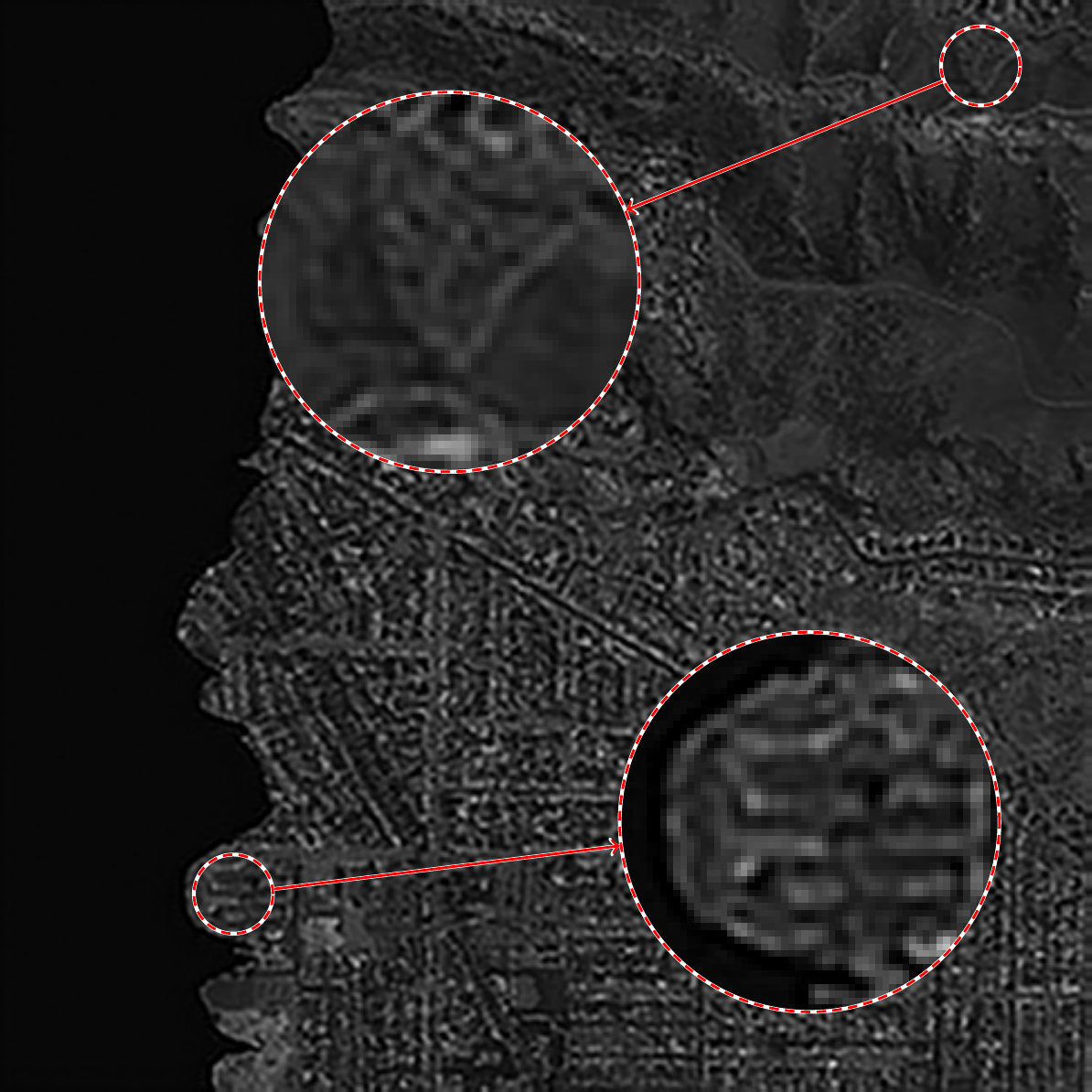} &
    \includegraphics[width=\mywidth\textwidth]{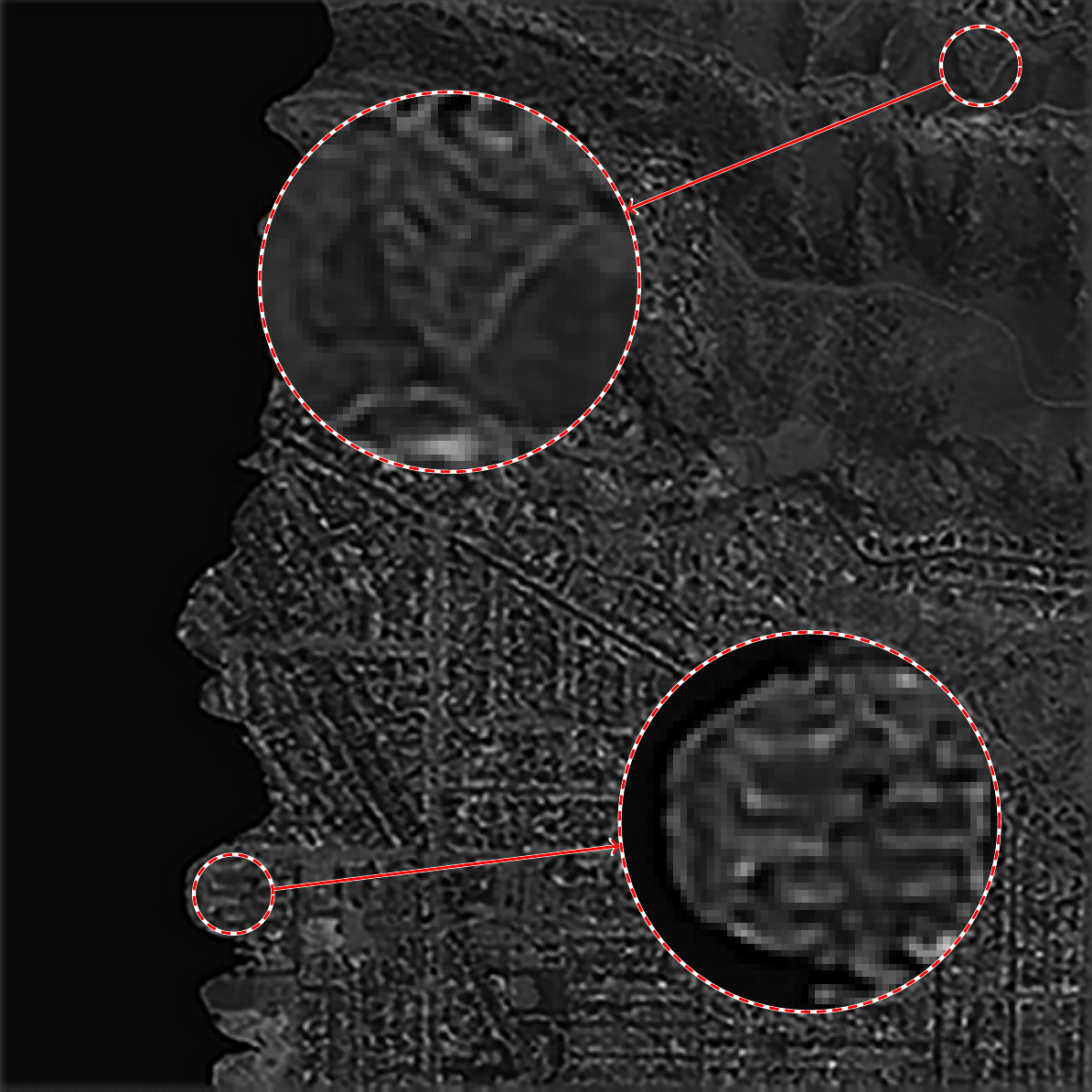} &
    \includegraphics[width=\mywidth\textwidth]{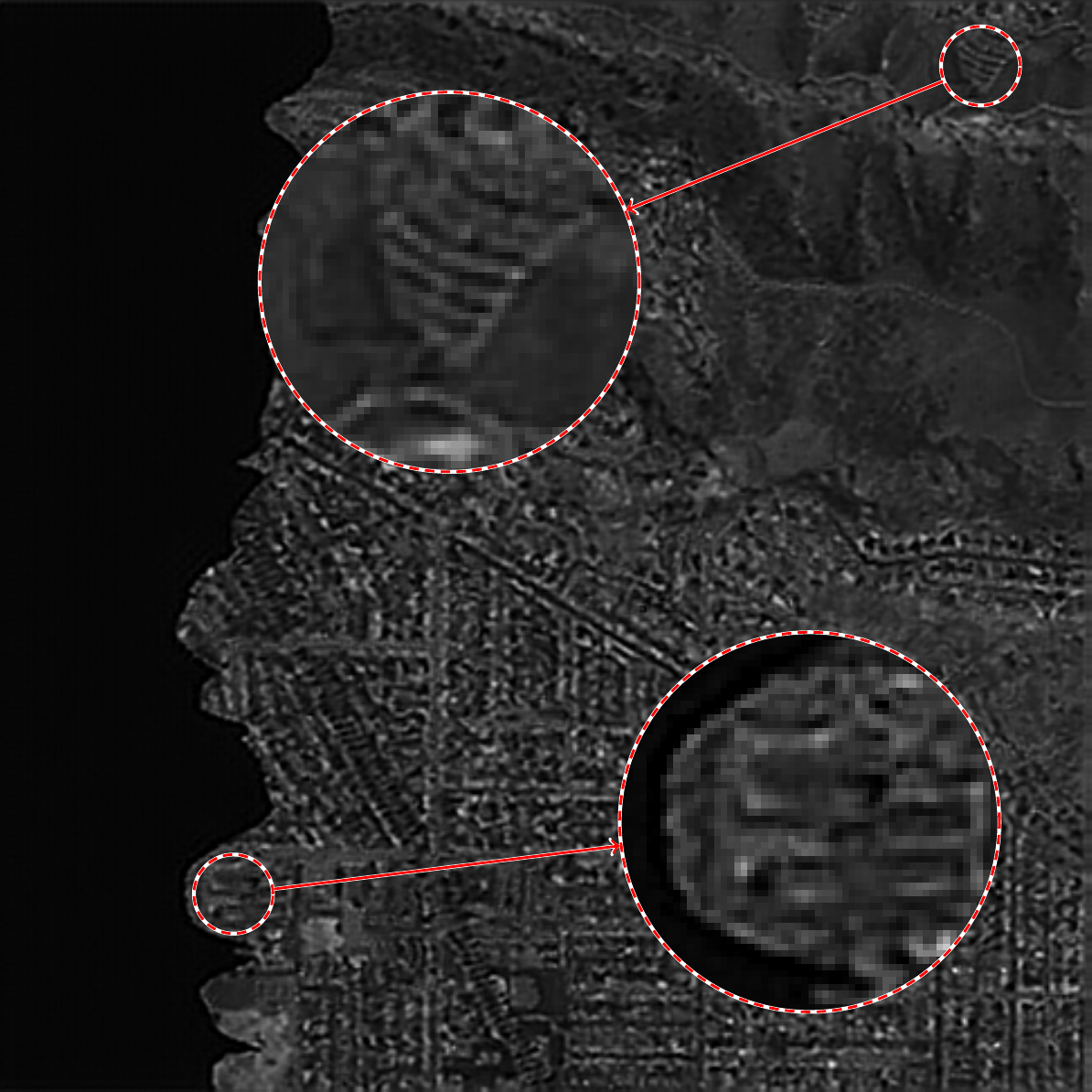} &
    \includegraphics[width=\mywidth\textwidth]{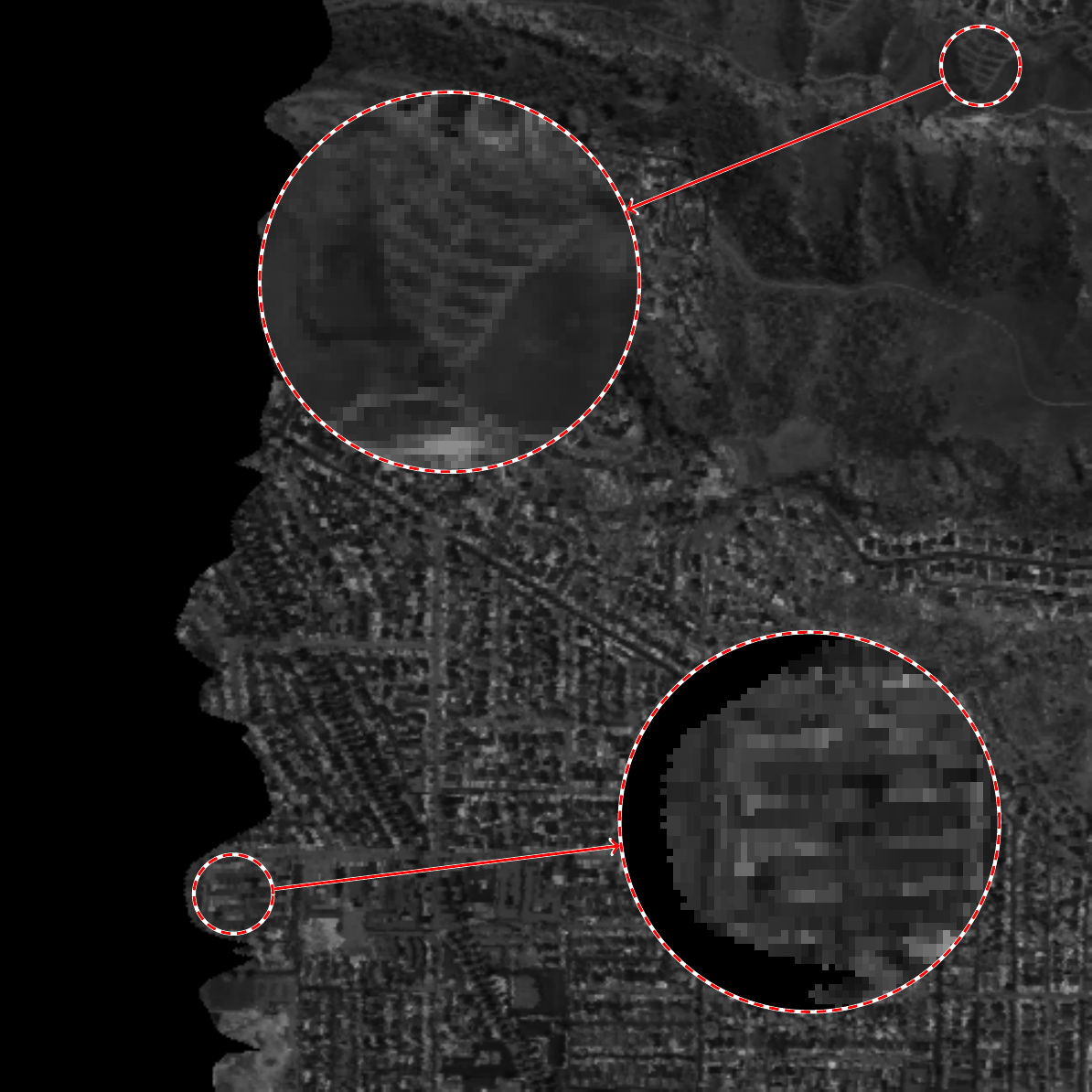}  \\

\end{tabular}
\caption{Super-resolved band B08 images (from \sims\, (a--c) and \simhsi\, (d) datasets) obtained using different techniques. The reconstruction was performed from multiple simulated LR images---one of them is shown for each region along with the HR reference. For \sims\, images (a--c), the HR images are of 60\,m GSD, and for the \simhsi\, images, the HR reference is of 3.4\,m GSD. The results for the remaining bands are included in the Supplementary Material.}
\label{fig:10m_sim_results}
\end{figure*}

\begin{figure*}[ht!]
\centering
\footnotesize
\renewcommand{\tabcolsep}{0.3mm}
\renewcommand{\arraystretch}{0.66}
\newcommand{\mywidth}{0.16}
\begin{tabular}{ccccccc}
    ~~~& LR image & Bicubic interpolation & RAMS & HighRes-net  & DeepSent & HR image\\

&    \includegraphics[width=\mywidth\textwidth]{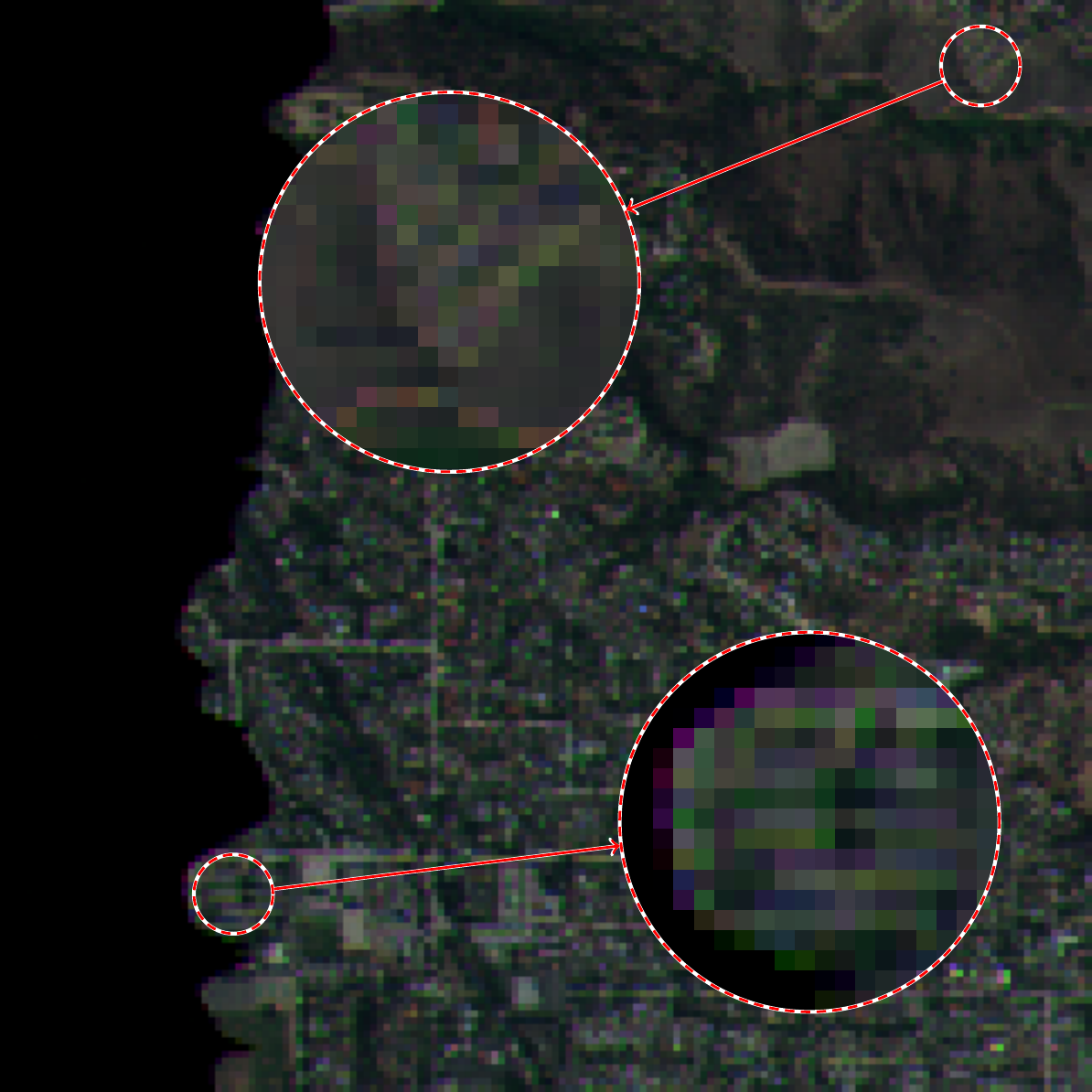} &
    \includegraphics[width=\mywidth\textwidth]{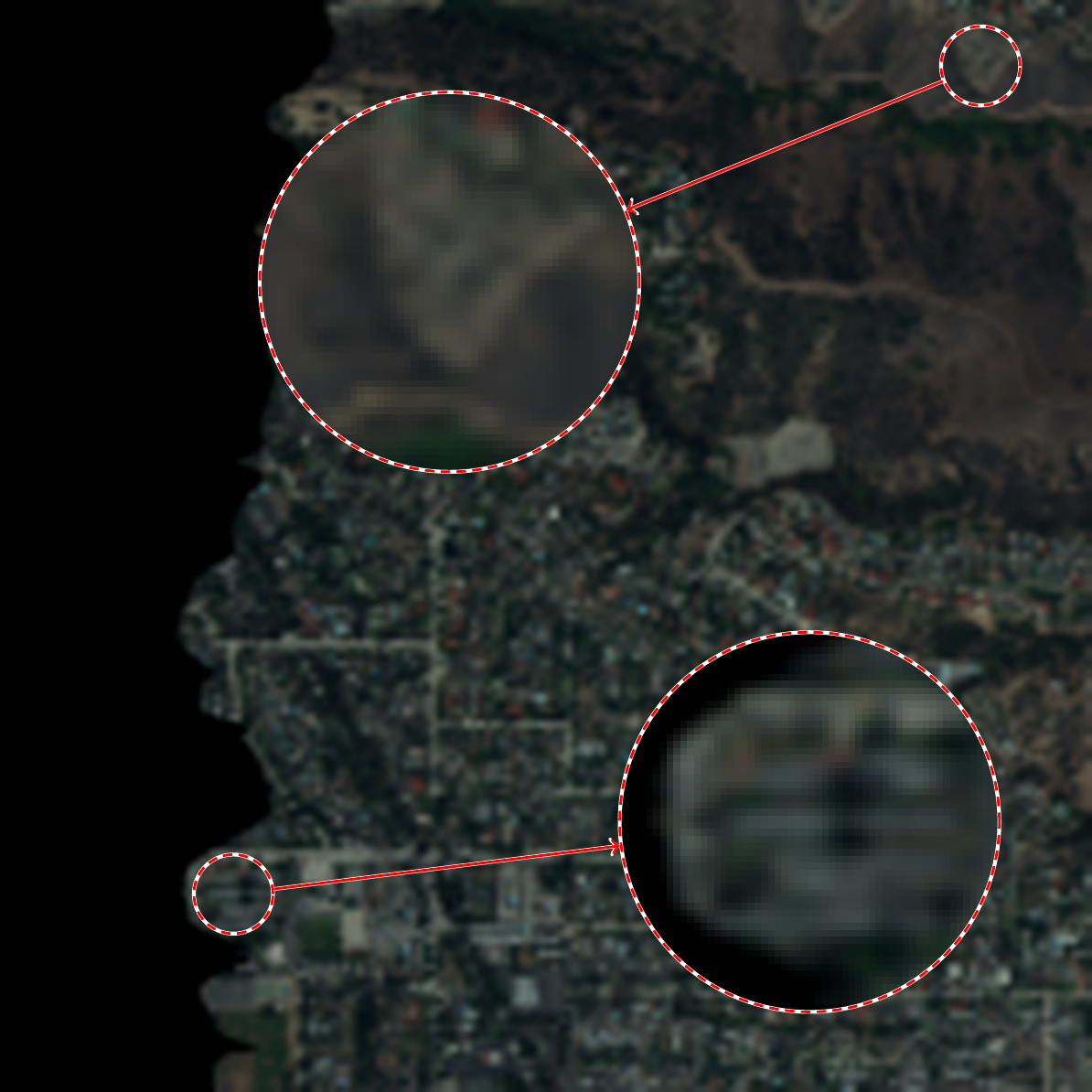} &
    \includegraphics[width=\mywidth\textwidth]{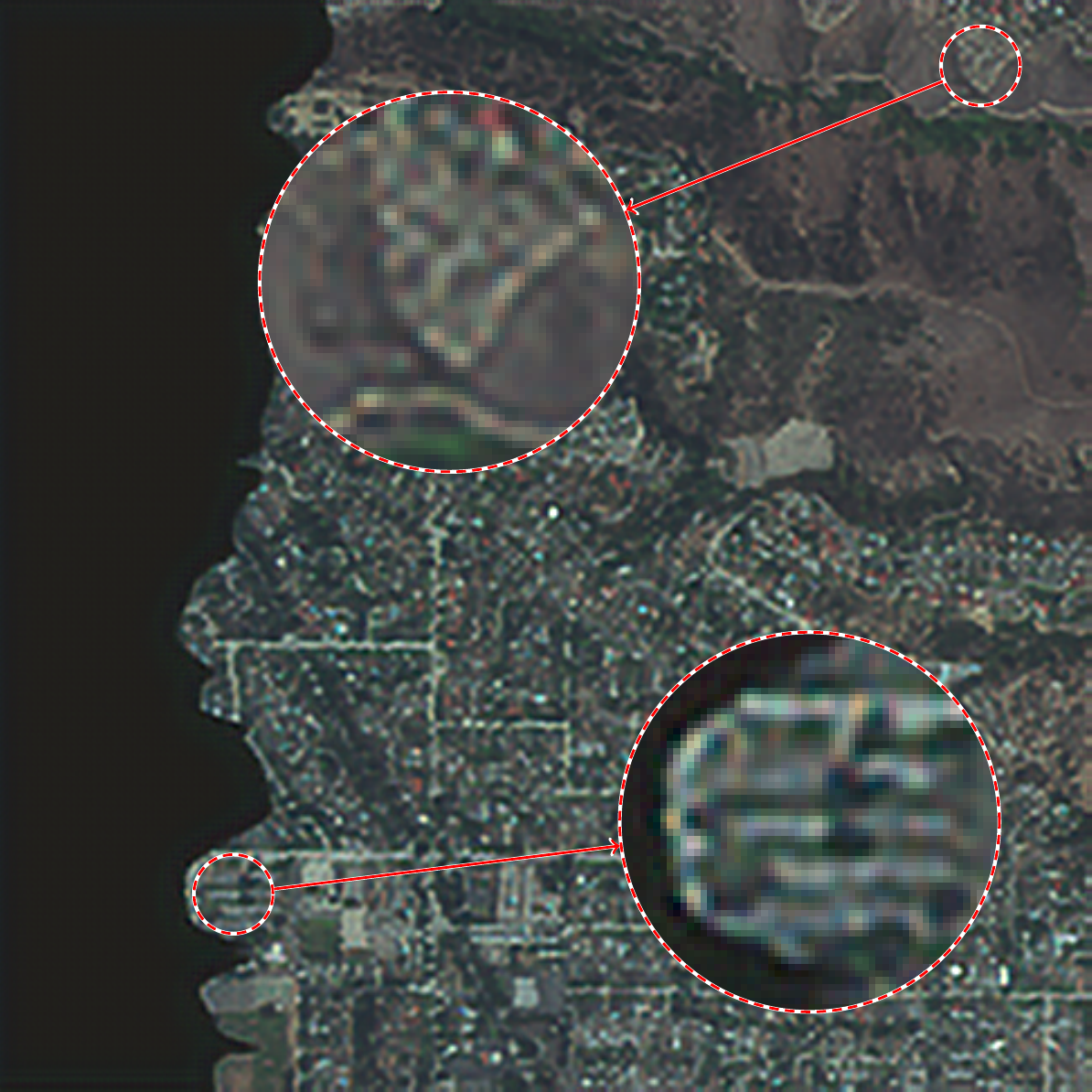} &
    \includegraphics[width=\mywidth\textwidth]{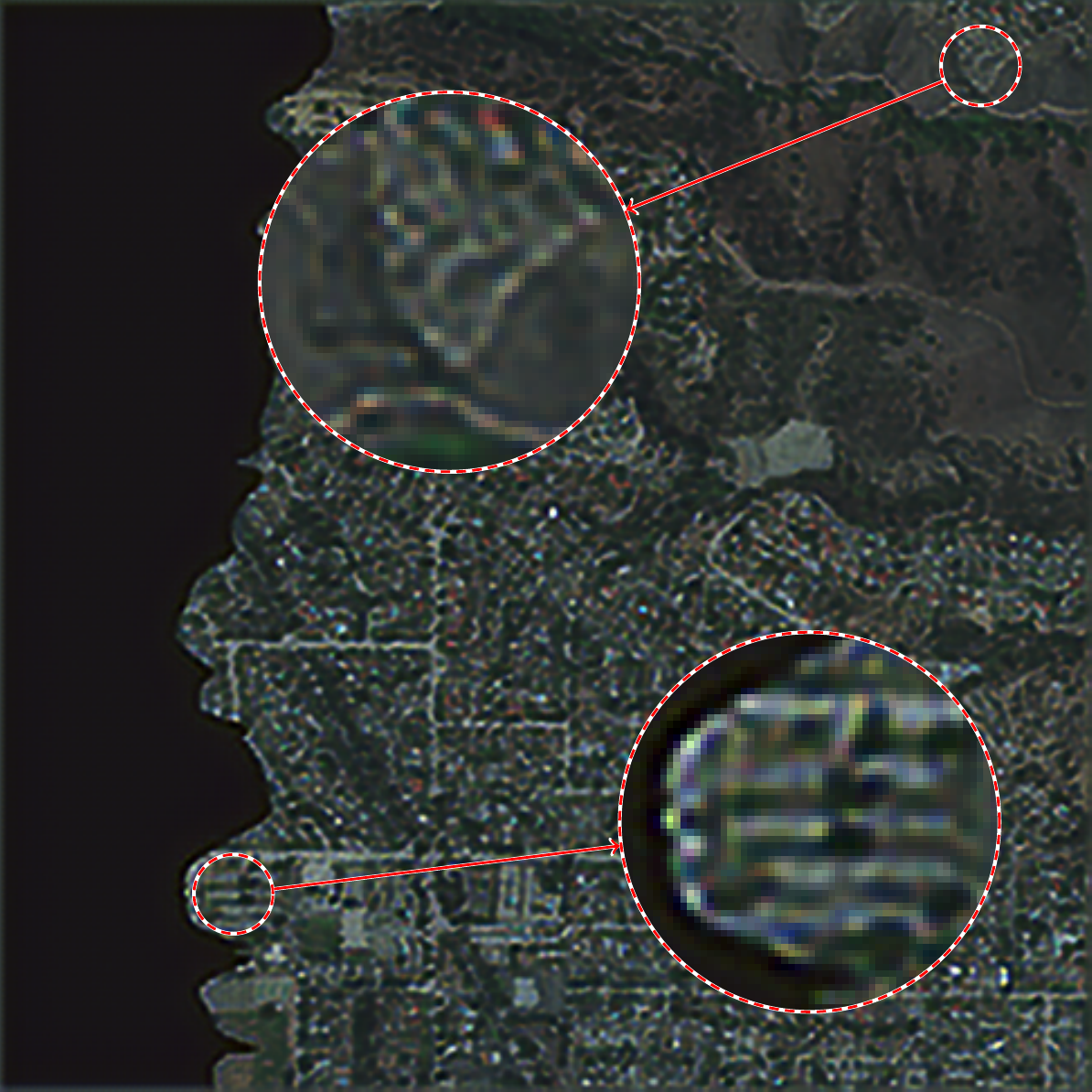} &
    \includegraphics[width=\mywidth\textwidth]{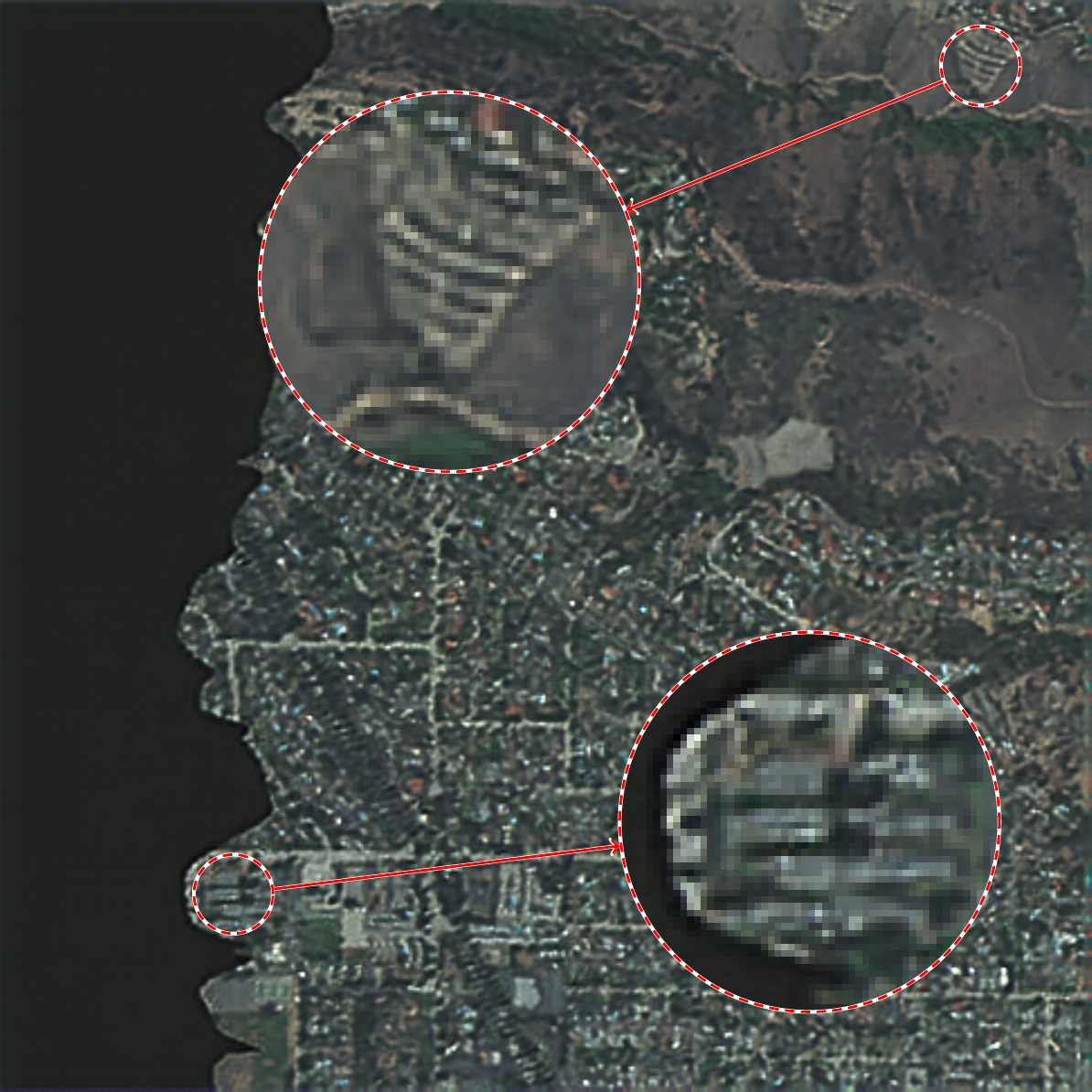} &
    \includegraphics[width=\mywidth\textwidth]{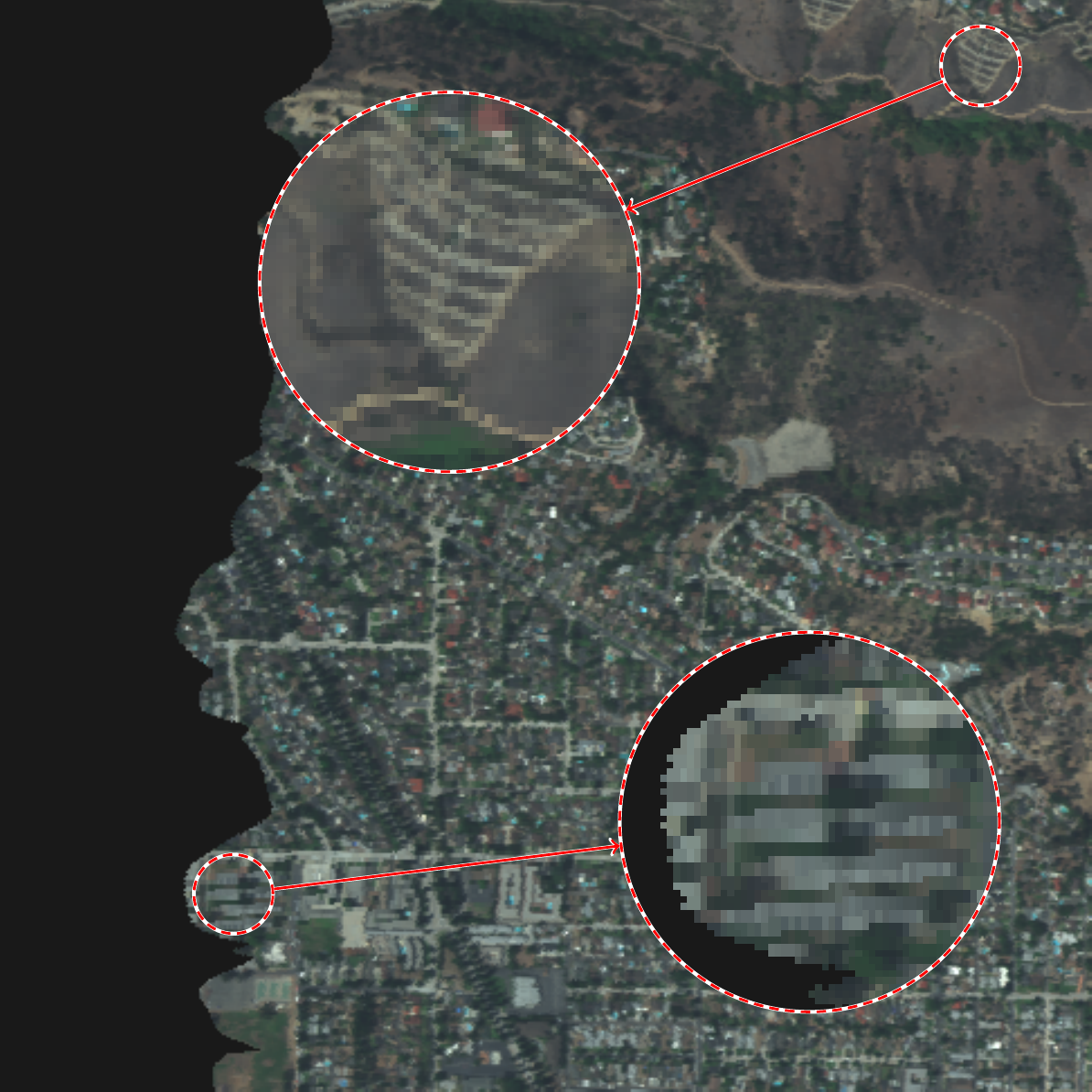}  \\

\end{tabular}
\caption{A color image composed from 10\,m B02, B03, and B04 bands  for the image from the \simhsi\, dataset, reconstructed using different techniques. The reconstruction was performed from multiple simulated LR images---one of them is shown for each region along with the HR reference. The results for the remaining bands are included in the Supplementary Material.}
\label{fig:hsi_rgb}
\end{figure*}

\begin{figure}[ht!]
\centering
\footnotesize
\renewcommand{\tabcolsep}{0.3mm}
\renewcommand{\arraystretch}{0.66}
\newcommand{\mywidth}{0.141}
\newcommand{\rowsep}{26mm}
\begin{tabular}{p{8mm}c}
    \footnotesize B06 & \begin{tabular}{ccc}
    \includegraphics[width=\mywidth\textwidth]{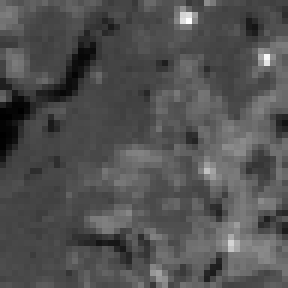} &
    \includegraphics[width=\mywidth\textwidth]{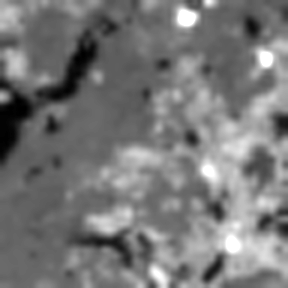} &
    \includegraphics[width=\mywidth\textwidth]{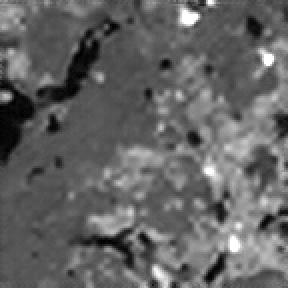} \\
    LR image & Bicubic interp. & RAMS \\
    \includegraphics[width=\mywidth\textwidth]{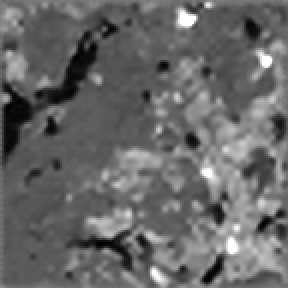} &
    \includegraphics[width=\mywidth\textwidth]{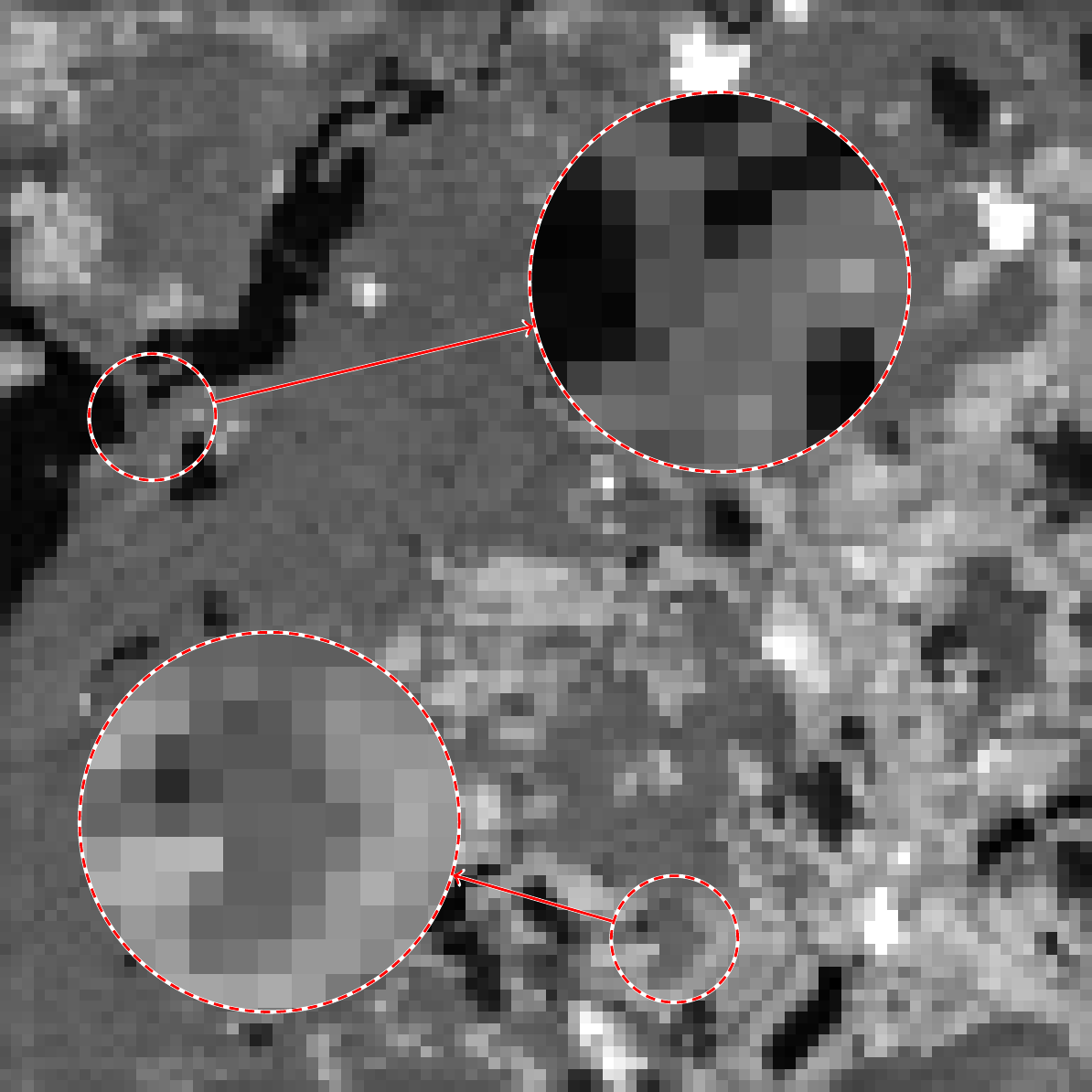} &
    \includegraphics[width=\mywidth\textwidth]{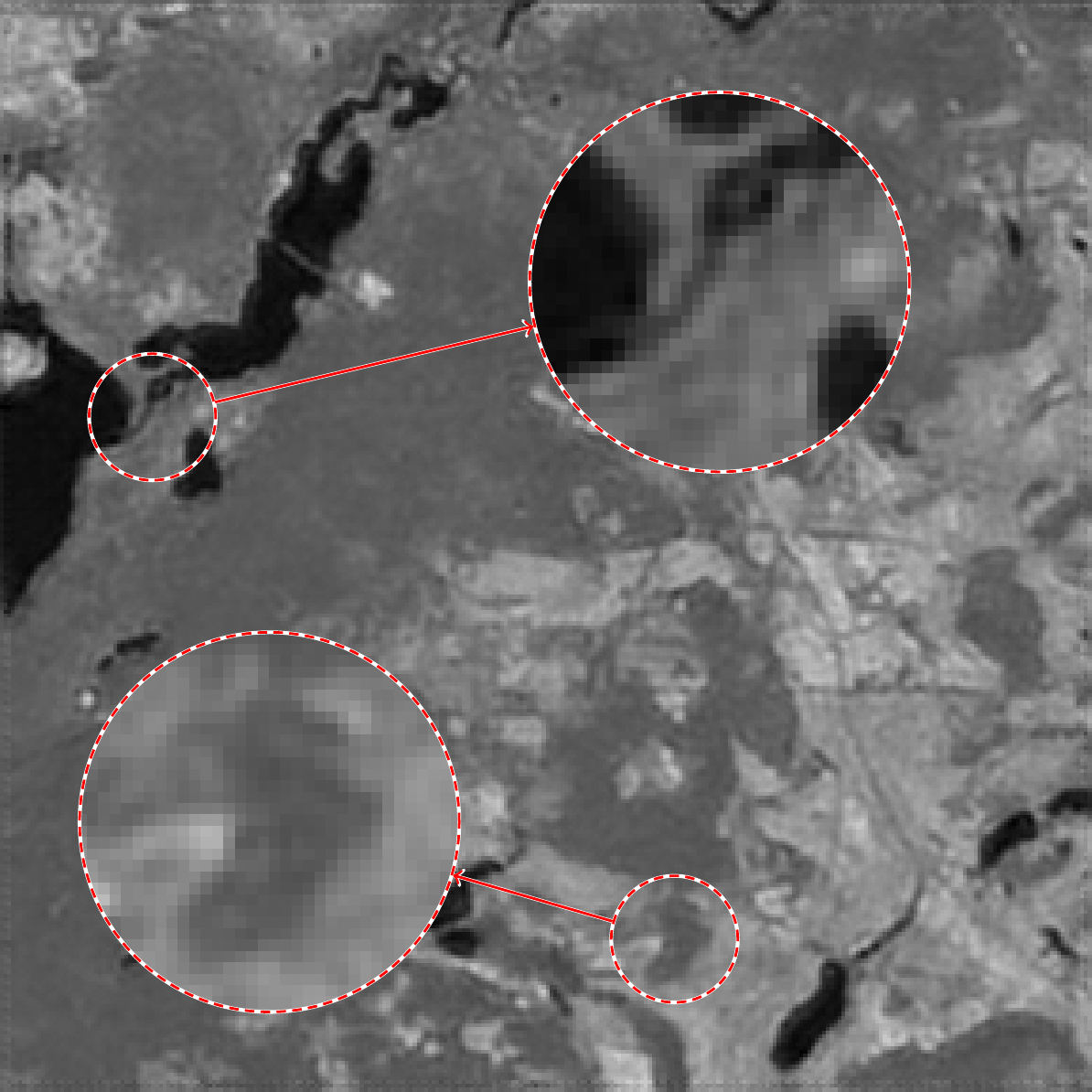} \rule{0pt}{\rowsep}\\
    HighRes-net & DSen2 & \DSenR\\
    \includegraphics[width=\mywidth\textwidth]{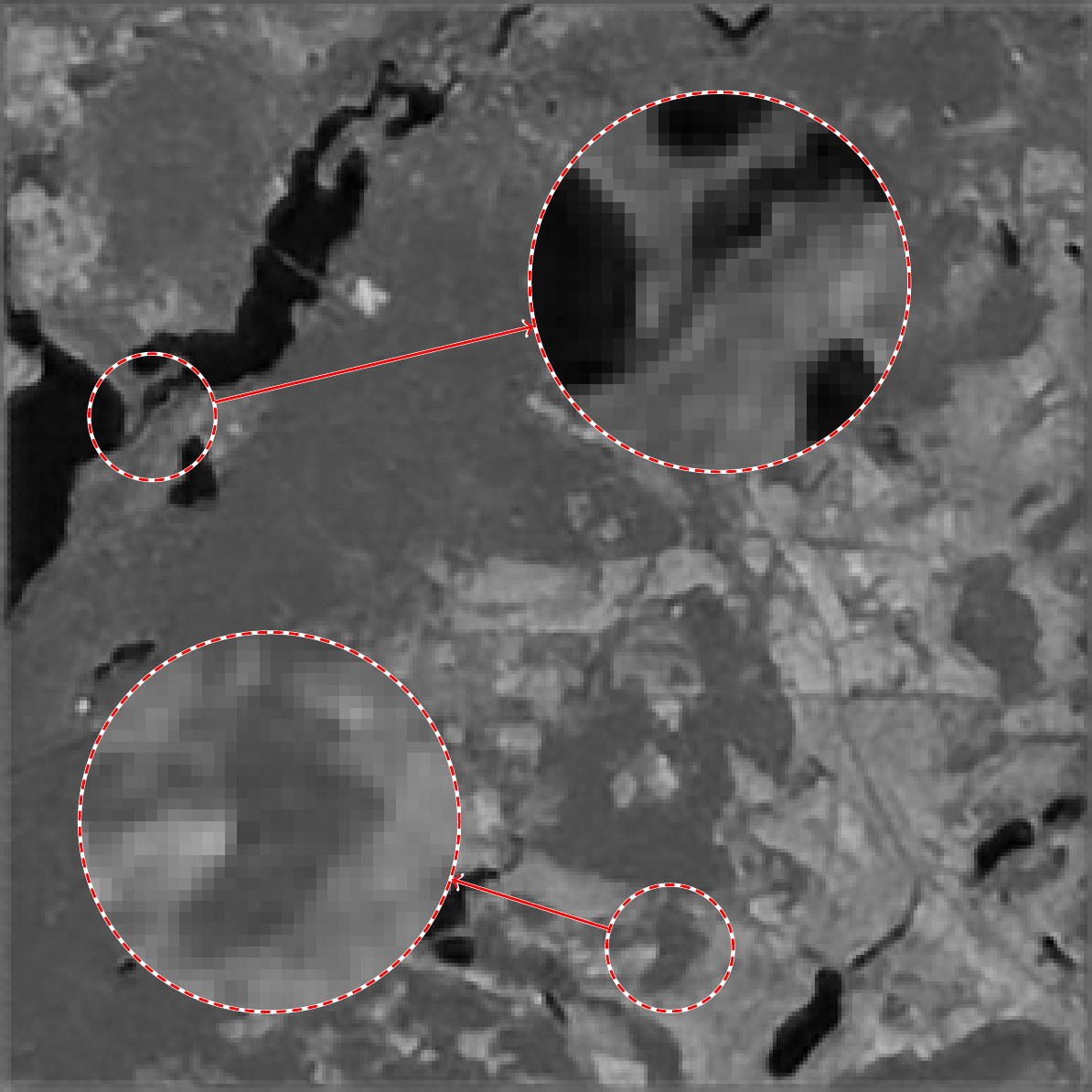} &
    \includegraphics[width=\mywidth\textwidth]{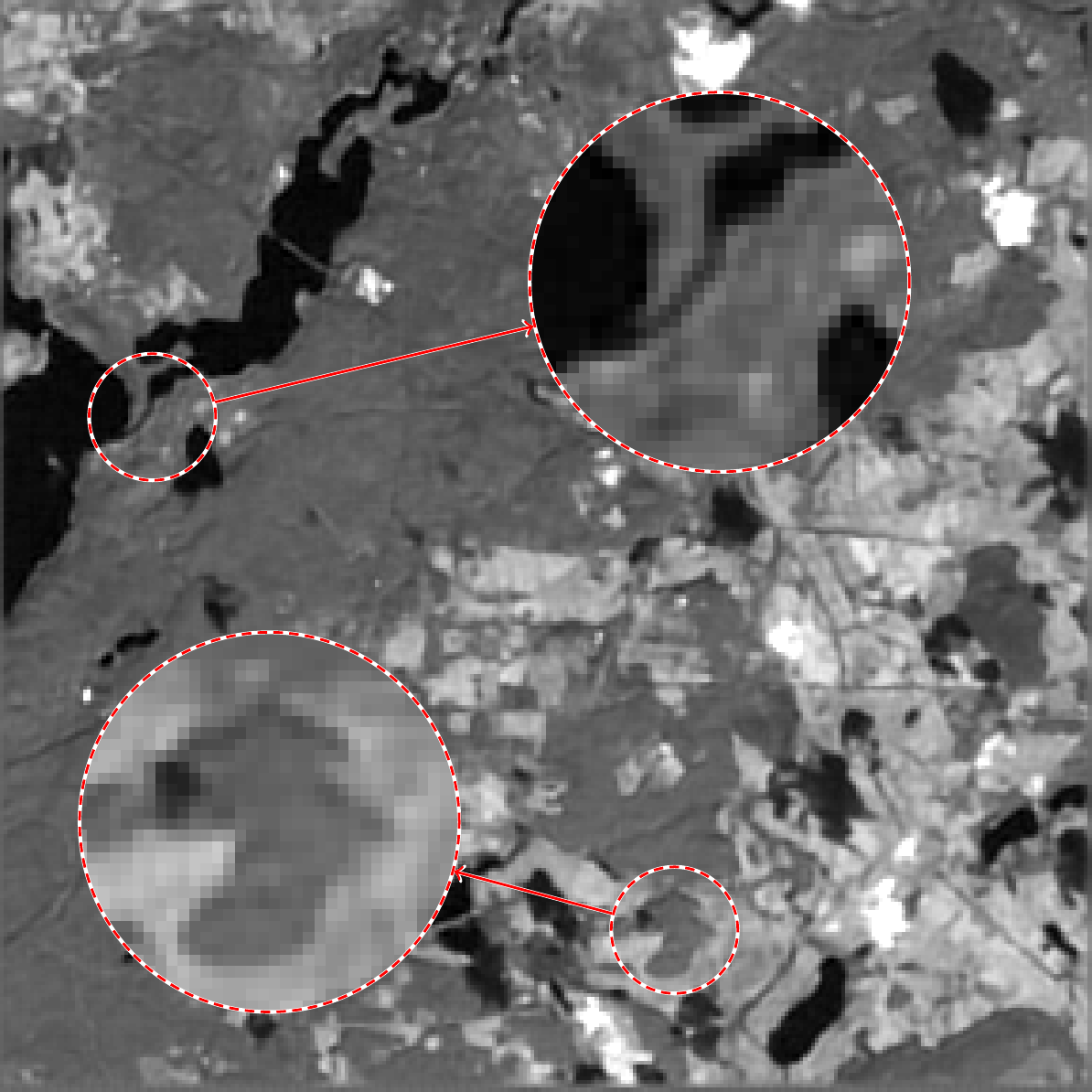} &
    \includegraphics[width=\mywidth\textwidth]{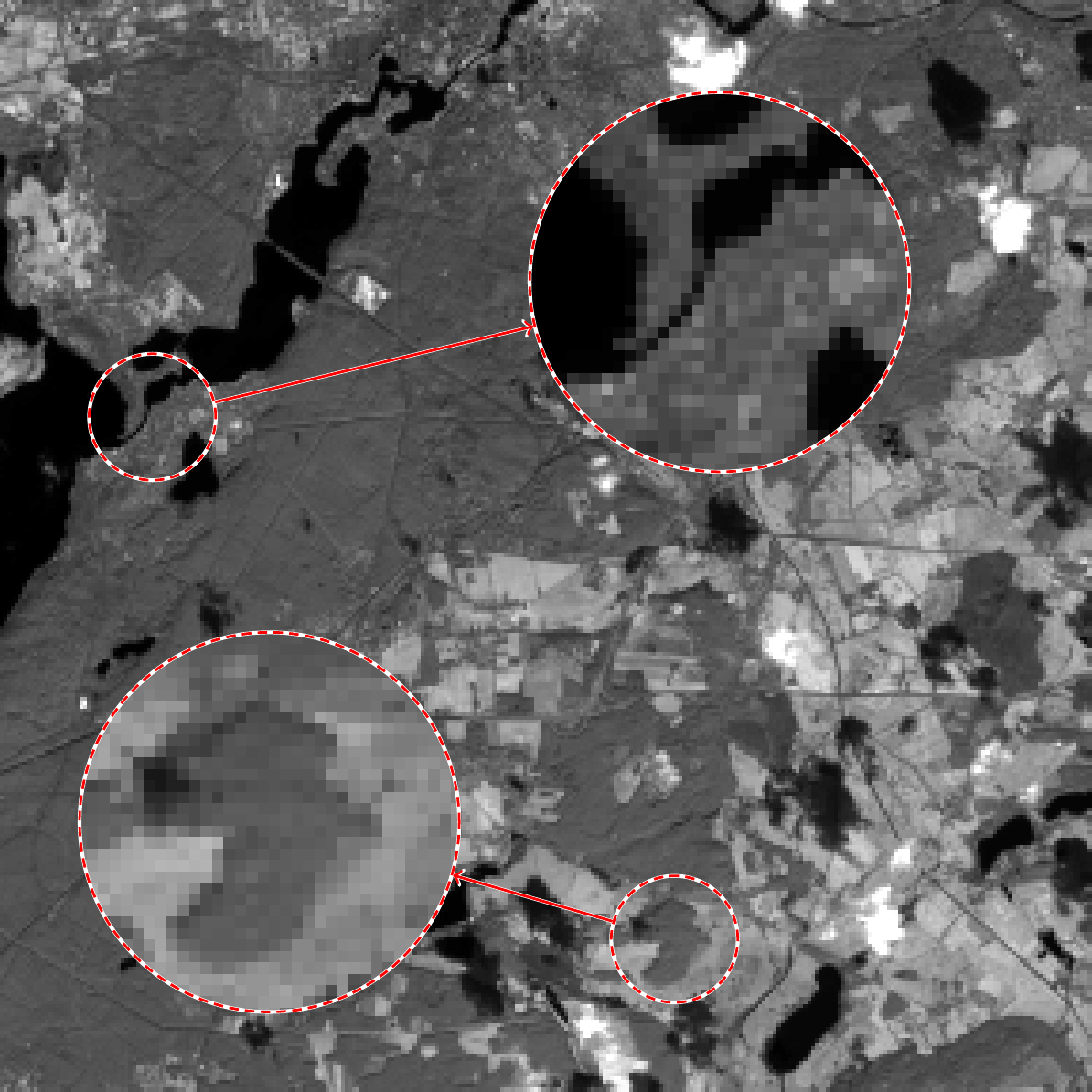} \rule{0pt}{\rowsep}\\
    \DSenHRn & \mymethod & HR image\\
    \end{tabular} \\
    & \\
    \hline
    & \rule{0pt}{3mm}\\

    \footnotesize B09 & \begin{tabular}{ccc}
    \includegraphics[width=\mywidth\textwidth]{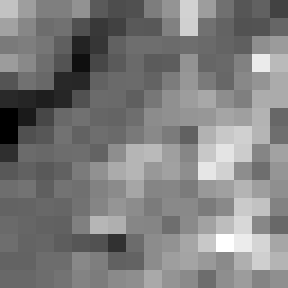} &
    \includegraphics[width=\mywidth\textwidth]{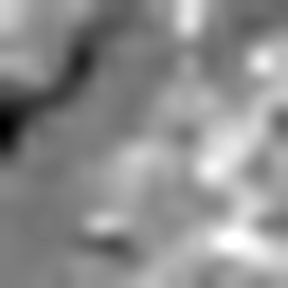} &
    \includegraphics[width=\mywidth\textwidth]{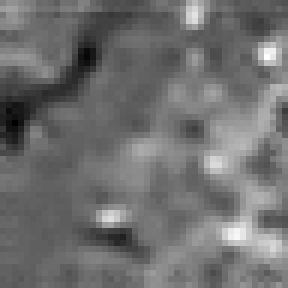} \\
    LR image & Bicubic interp. & RAMS \\
    \includegraphics[width=\mywidth\textwidth]{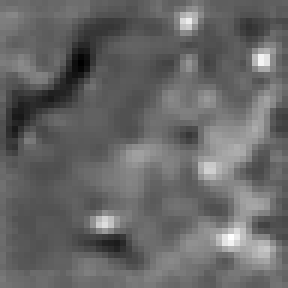} &
    \includegraphics[width=\mywidth\textwidth]{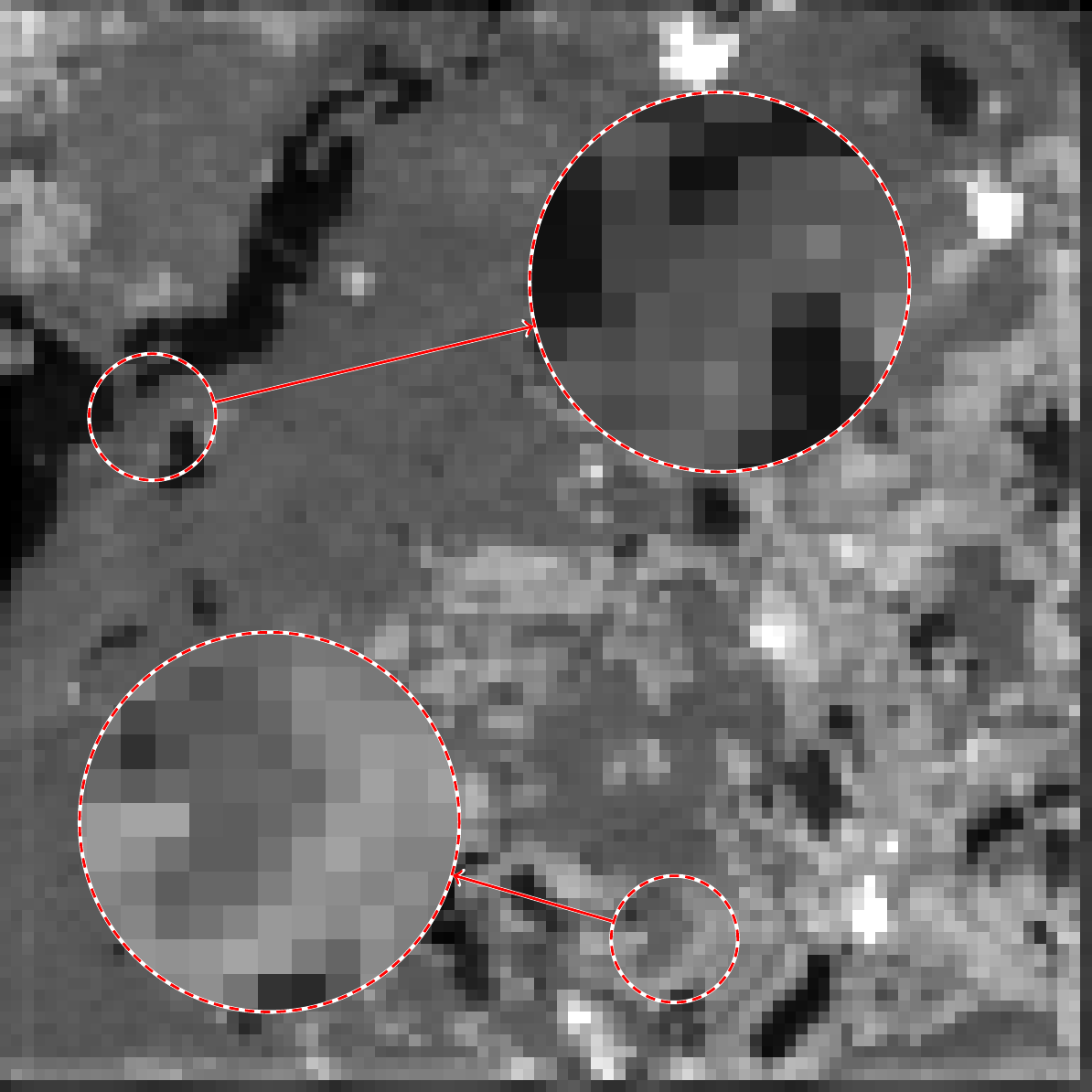} &
    \includegraphics[width=\mywidth\textwidth]{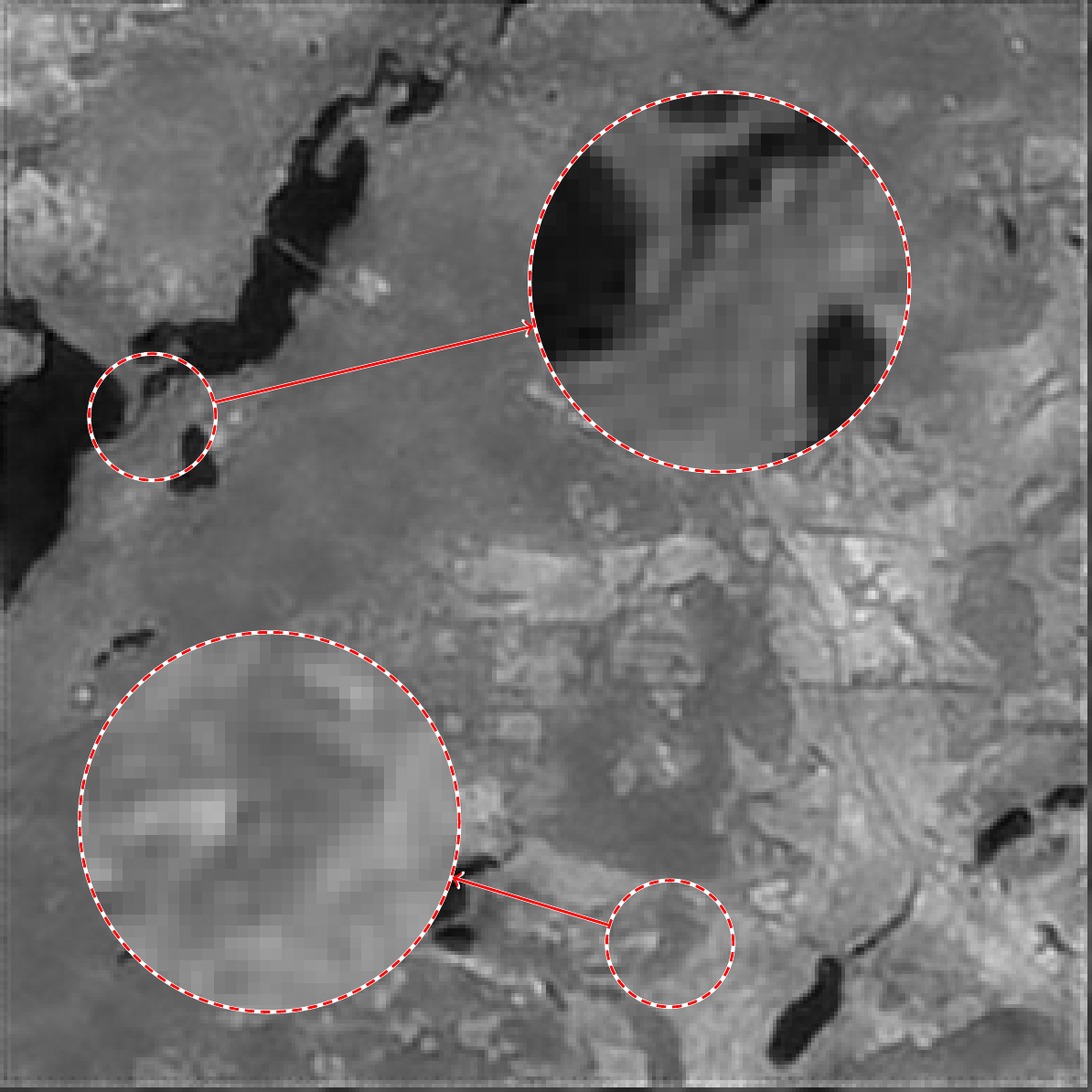} \rule{0pt}{\rowsep}\\
    HighRes-net & DSen2 & \DSenR\\
    \includegraphics[width=\mywidth\textwidth]{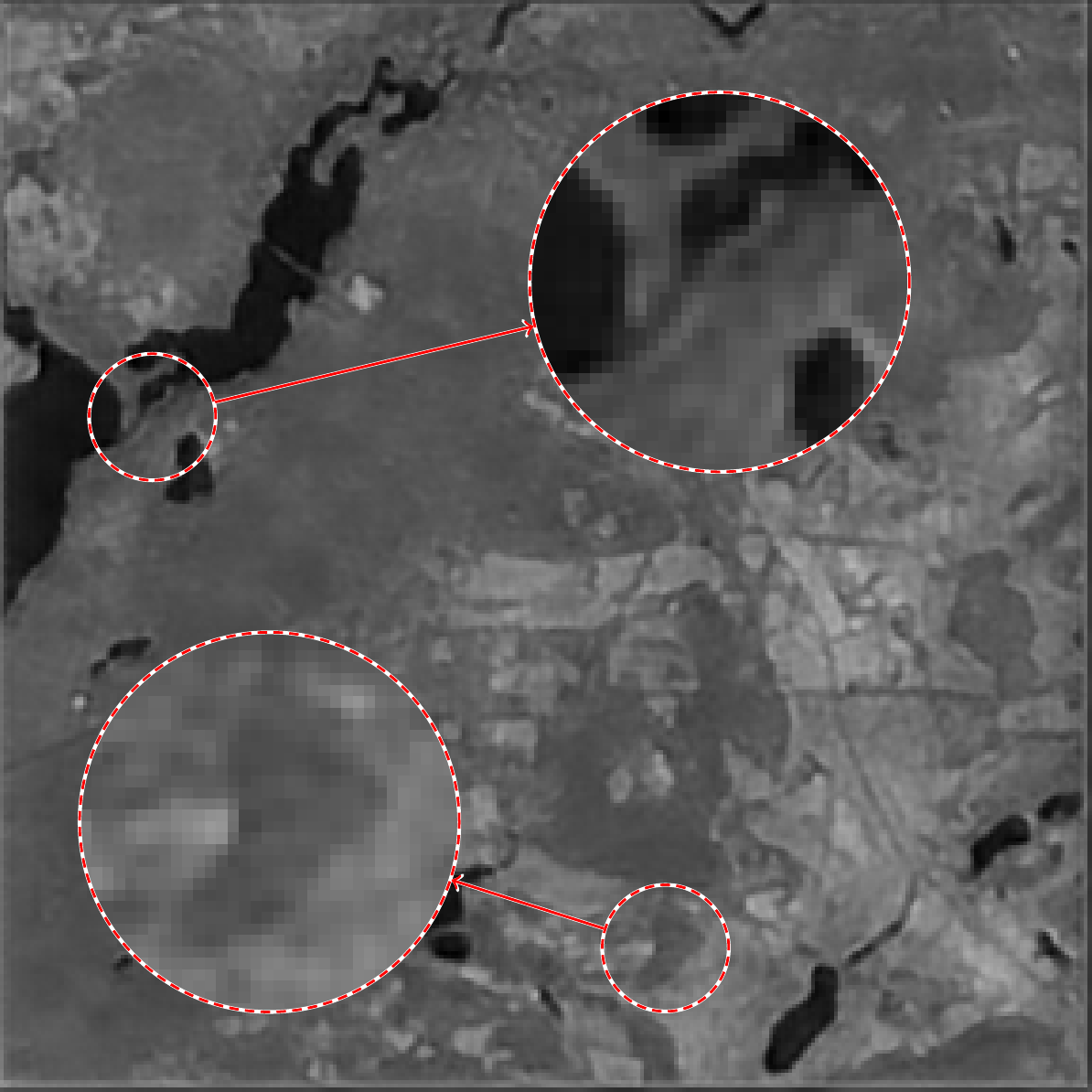} &
    \includegraphics[width=\mywidth\textwidth]{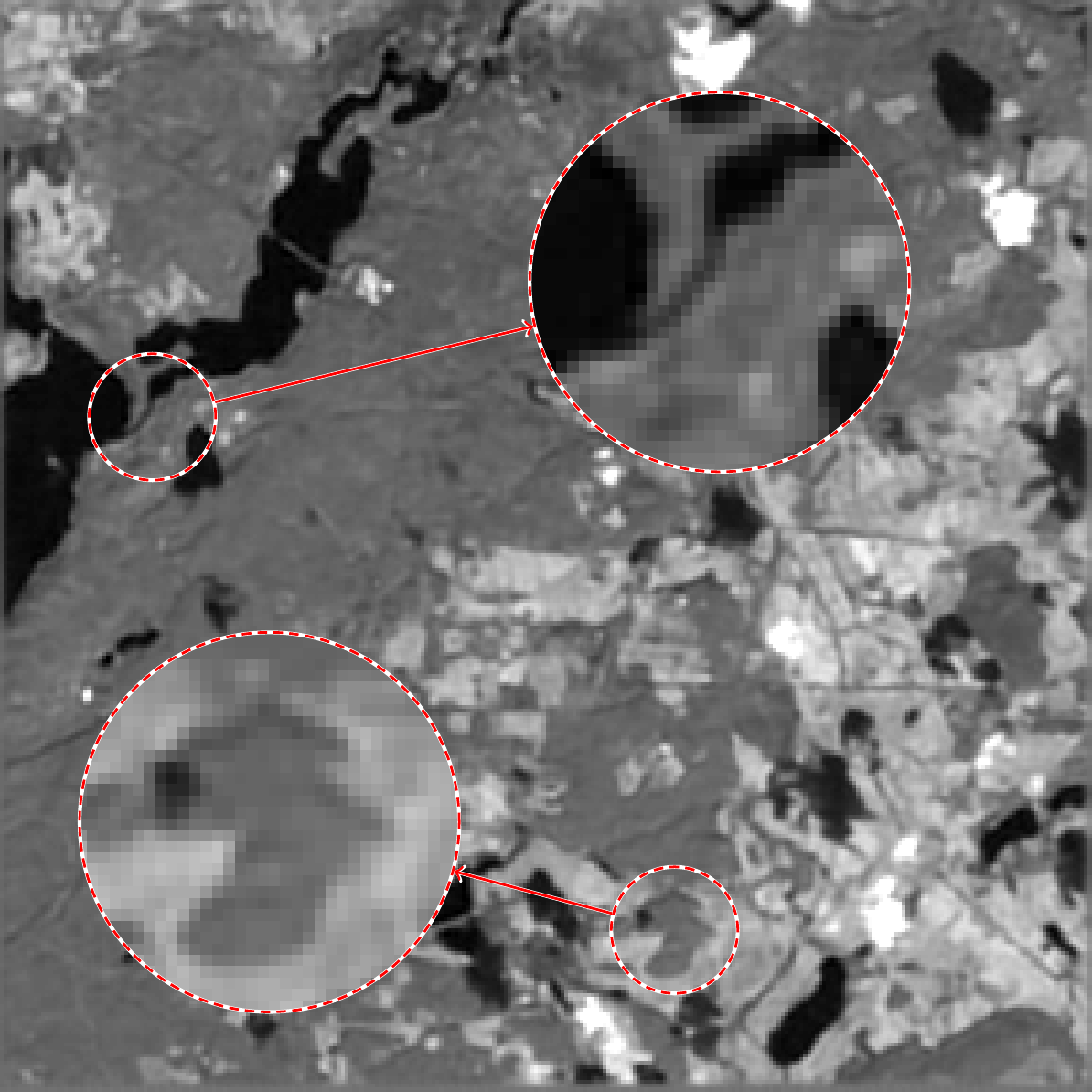} &
    \includegraphics[width=\mywidth\textwidth]{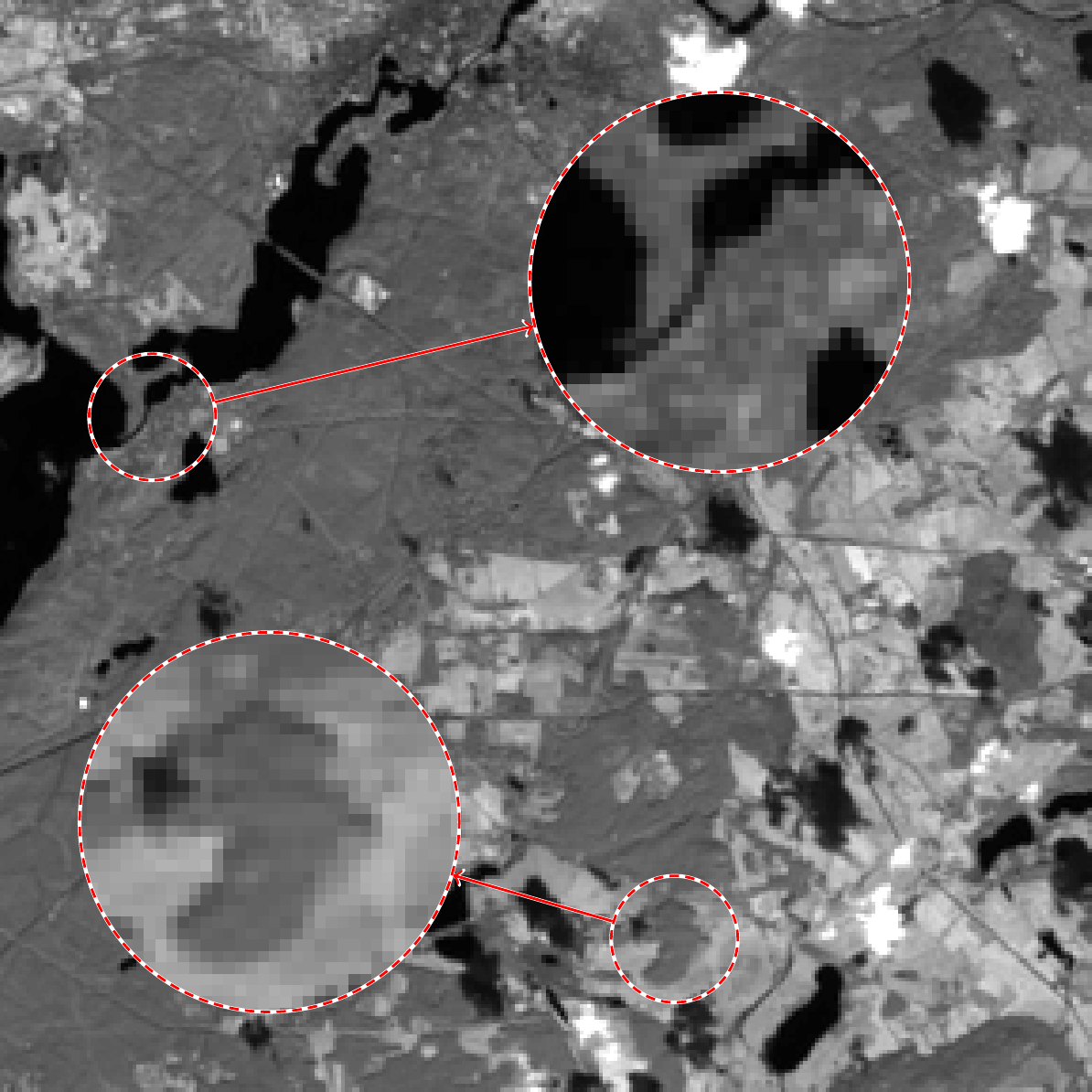} \rule{0pt}{\rowsep}\\
    \DSenHRn & \mymethod & HR image \\
    \end{tabular}
\end{tabular}
\caption{Super-resolved images obtained by different methods for the B06 (20\,m GSD) and B09 (60\,m GSD) bands. The reconstruction was performed from multiple simulated LR images---one of them is shown along with the HR reference. The results for the remaining bands are included in the Supplementary Material.}
\label{fig:20_60_sim_results}
\end{figure}

\begin{figure}[ht!]
\centering
\footnotesize
\renewcommand{\tabcolsep}{0.2mm}
\renewcommand{\arraystretch}{0.4}
\newcommand{\mywidth}{0.1}
\newcommand{\shiftdown}[1]{\smash{\raisebox{-.3\normalbaselineskip}{#1}}}
\newcolumntype{C}{>{\collectcell\shiftdown}l<{\endcollectcell}}
\begin{adjustbox}{max width=\textwidth}
\begin{tabular}{ccccccC}
    & B02 & B03 & B04 & B08 & & 1000 \\
    \raisebox{4.5mm}{\begin{turn}{90} Bicubic \end{turn}} &
    \includegraphics[width=\mywidth\textwidth]{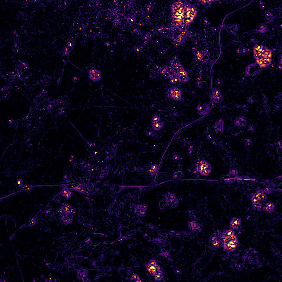} &
    \includegraphics[width=\mywidth\textwidth]{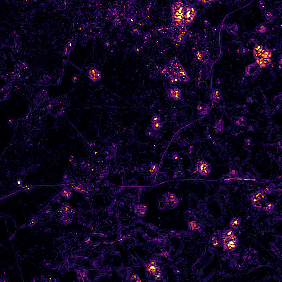} &
    \includegraphics[width=\mywidth\textwidth]{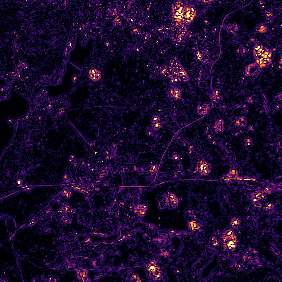} &
    \includegraphics[width=\mywidth\textwidth]{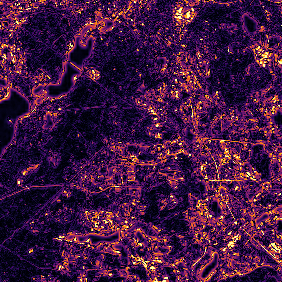} & & 750 \\
    \raisebox{2mm}{\begin{turn}{90} HighRes-net \end{turn}} &
    \includegraphics[width=\mywidth\textwidth]{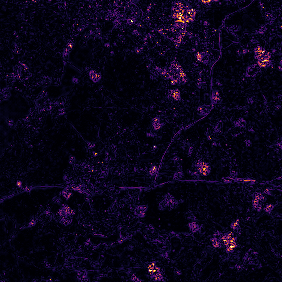} &
    \includegraphics[width=\mywidth\textwidth]{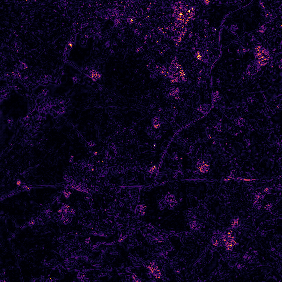} &
    \includegraphics[width=\mywidth\textwidth]{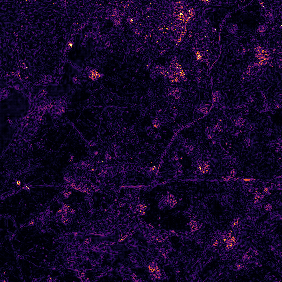} &
    \includegraphics[width=\mywidth\textwidth]{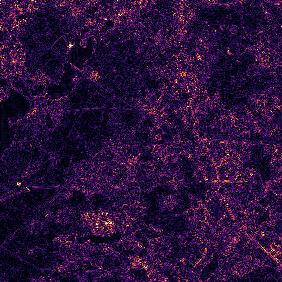} & & 500 \\
    \raisebox{5.5mm}{\begin{turn}{90} RAMS \end{turn}} &
    \includegraphics[width=\mywidth\textwidth]{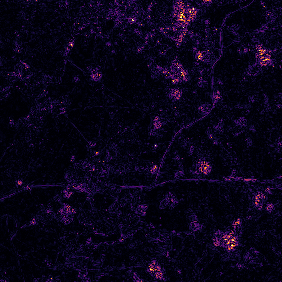} &
    \includegraphics[width=\mywidth\textwidth]{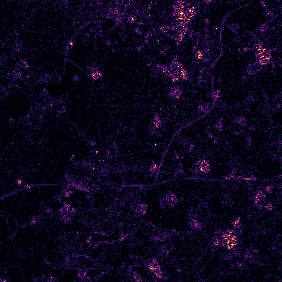} &
    \includegraphics[width=\mywidth\textwidth]{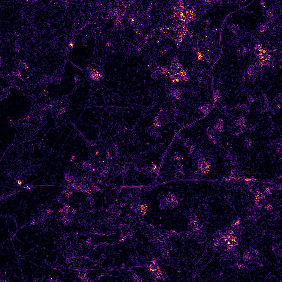} &
    \includegraphics[width=\mywidth\textwidth]{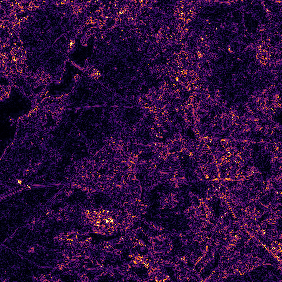} & & 250 \\
    \raisebox{4mm}{\begin{turn}{90} DeepSent \end{turn}} &
    \includegraphics[width=\mywidth\textwidth]{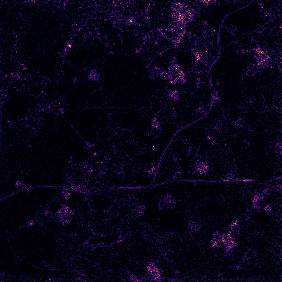} &
    \includegraphics[width=\mywidth\textwidth]{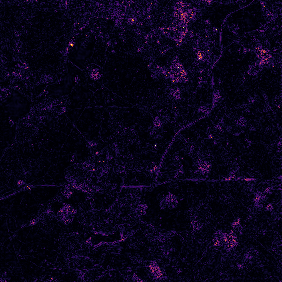} &
    \includegraphics[width=\mywidth\textwidth]{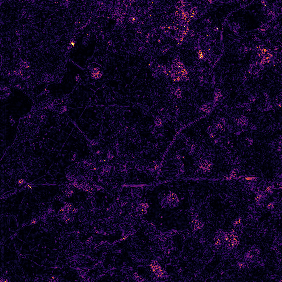} &
    \includegraphics[width=\mywidth\textwidth]{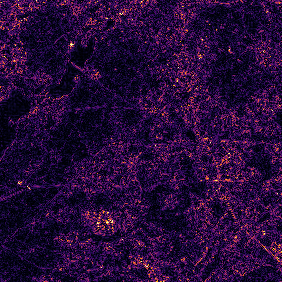} &
    \raisebox{-0.2mm}[0pt][0pt]{\includegraphics[height=0.409\textwidth, width=0.015\textwidth]{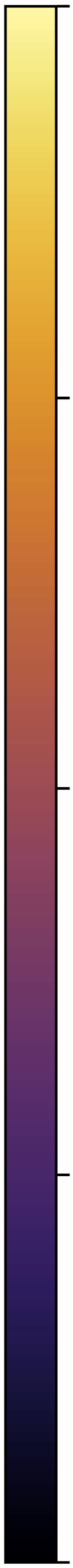}} & 0\\
\end{tabular}
\end{adjustbox}
\caption{Absolute differences between the super-resolved images (for the same scene as in Fig.~\ref{fig:10m_sim_results}a and Fig.~\ref{fig:20_60_sim_results}) and their HR counterparts for 10\,m bands at the 14-bit depth.}
\label{fig:sim_diffs_10m}
\end{figure}

\begin{figure*}[ht!]
\centering
\footnotesize
\renewcommand{\tabcolsep}{0.2mm}
\renewcommand{\arraystretch}{0.4}
\newcommand{\mywidth}{0.1}
\newcommand{\shiftdown}[1]{\smash{\raisebox{-.3\normalbaselineskip}{#1}}}
\newcolumntype{C}{>{\collectcell\shiftdown}l<{\endcollectcell}}
\begin{adjustbox}{max width=\textwidth}
\begin{tabular}{ccccccccccC}
    & B05 & B06 & B07 & B08a & B11 & B12 & B01 & B09 & & 1000\\
    \raisebox{4.5mm}{\begin{turn}{90} Bicubic \end{turn}} &
    \includegraphics[width=\mywidth\textwidth]{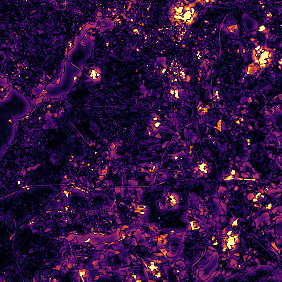} &
    \includegraphics[width=\mywidth\textwidth]{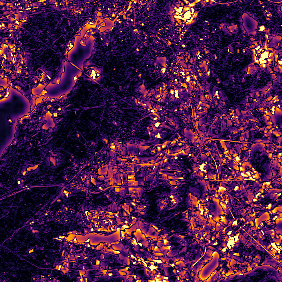} &
    \includegraphics[width=\mywidth\textwidth]{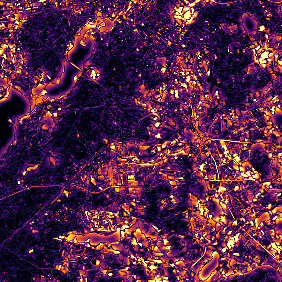} &
    \includegraphics[width=\mywidth\textwidth]{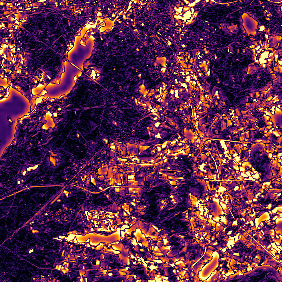} &
    \includegraphics[width=\mywidth\textwidth]{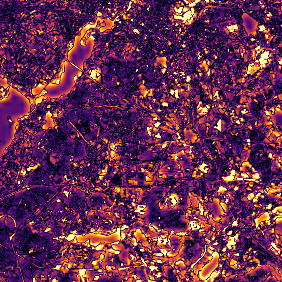} &
    \includegraphics[width=\mywidth\textwidth]{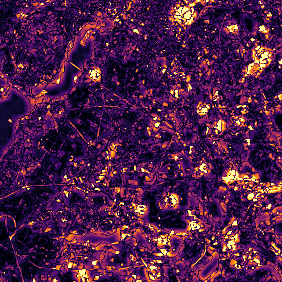} &
    \includegraphics[width=\mywidth\textwidth]{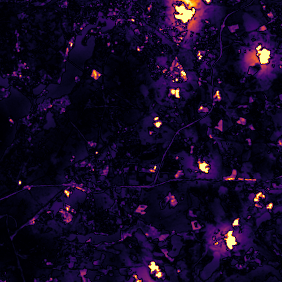} &
    \includegraphics[width=\mywidth\textwidth]{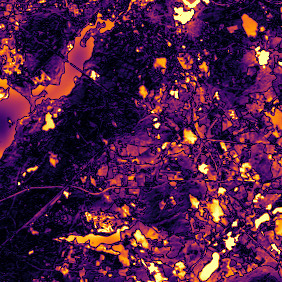} & & 750 \\
    \raisebox{2mm}{\begin{turn}{90} HighRes-net \end{turn}} &
    \includegraphics[width=\mywidth\textwidth]{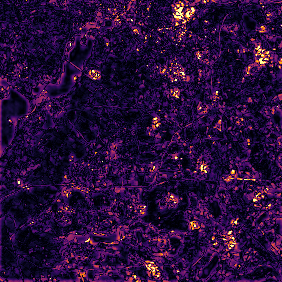} &
    \includegraphics[width=\mywidth\textwidth]{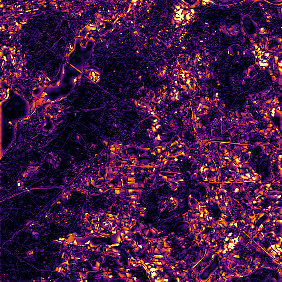} &
    \includegraphics[width=\mywidth\textwidth]{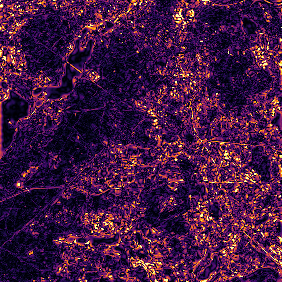} &
    \includegraphics[width=\mywidth\textwidth]{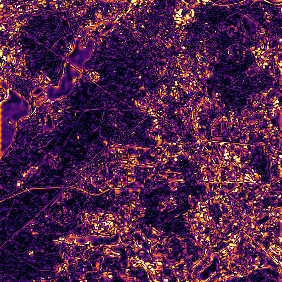} &
    \includegraphics[width=\mywidth\textwidth]{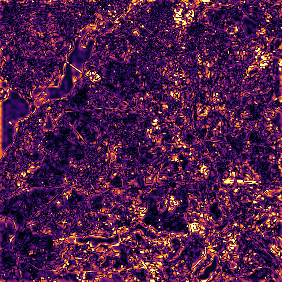} &
    \includegraphics[width=\mywidth\textwidth]{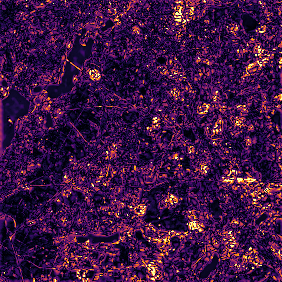} &
    \includegraphics[width=\mywidth\textwidth]{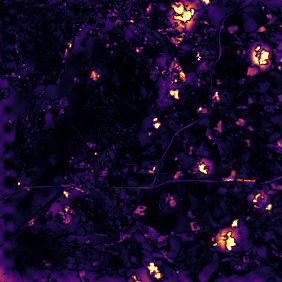} &
    \includegraphics[width=\mywidth\textwidth]{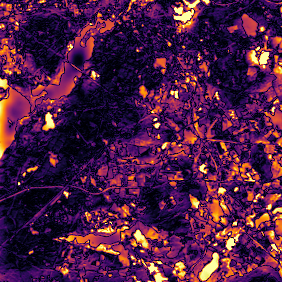} & & 500 \\
    \raisebox{5.5mm}{\begin{turn}{90} RAMS \end{turn}} &
    \includegraphics[width=\mywidth\textwidth]{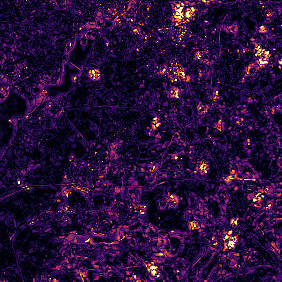} &
    \includegraphics[width=\mywidth\textwidth]{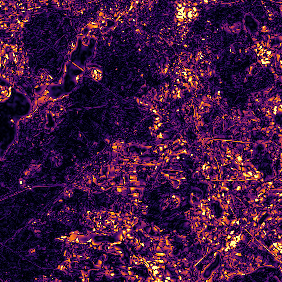} &
    \includegraphics[width=\mywidth\textwidth]{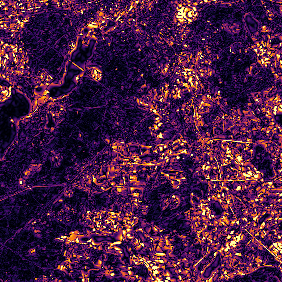} &
    \includegraphics[width=\mywidth\textwidth]{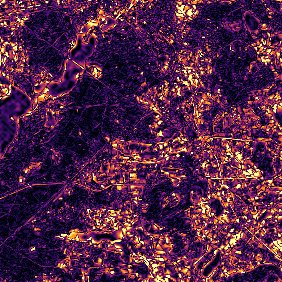} &
    \includegraphics[width=\mywidth\textwidth]{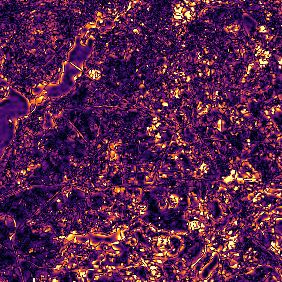} &
    \includegraphics[width=\mywidth\textwidth]{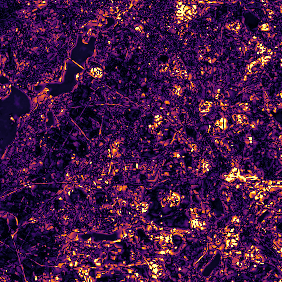} &
    \includegraphics[width=\mywidth\textwidth]{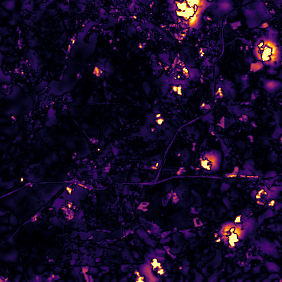} &
    \includegraphics[width=\mywidth\textwidth]{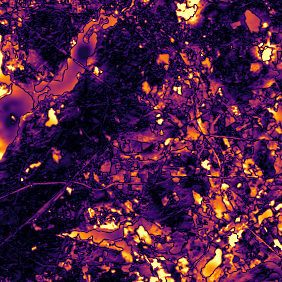} & & 250 \\
    \raisebox{4mm}{\begin{turn}{90} DeepSent \end{turn}} &
    \includegraphics[width=\mywidth\textwidth]{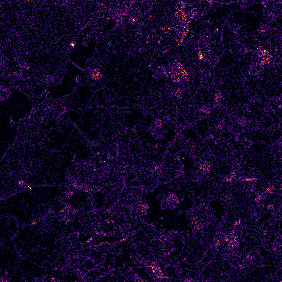} &
    \includegraphics[width=\mywidth\textwidth]{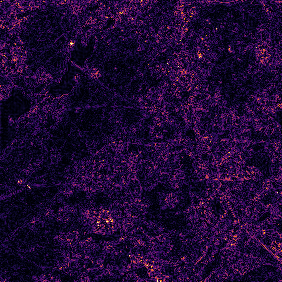} &
    \includegraphics[width=\mywidth\textwidth]{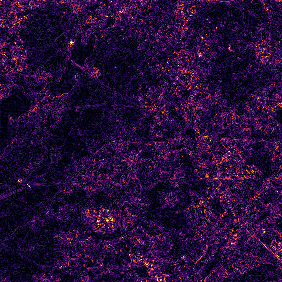} &
    \includegraphics[width=\mywidth\textwidth]{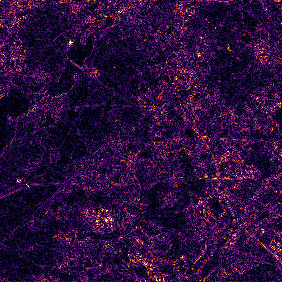} &
    \includegraphics[width=\mywidth\textwidth]{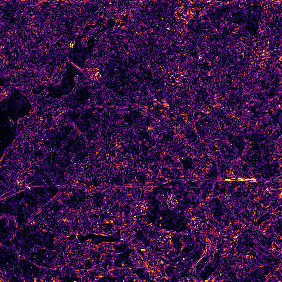} &
    \includegraphics[width=\mywidth\textwidth]{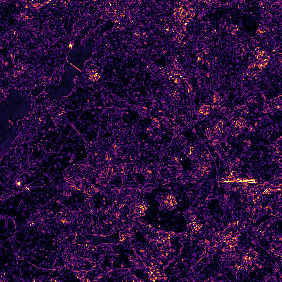} &
    \includegraphics[width=\mywidth\textwidth]{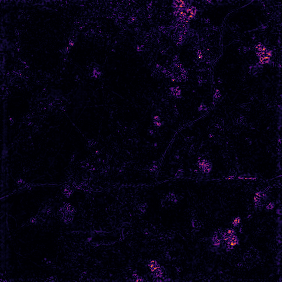} &
    \includegraphics[width=\mywidth\textwidth]{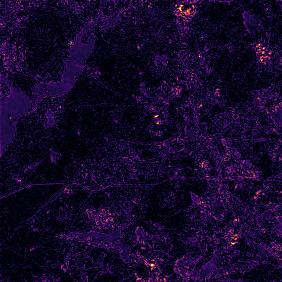} &
    \raisebox{-0.2mm}[0pt][0pt]{\includegraphics[height=0.409\textwidth, width=0.015\textwidth]{figures/sim_diff_maps/plasma_legend.png}} & 0\\
\end{tabular}
\end{adjustbox}
\caption{The absolute differences between the super-resolved images  (for the same scene as in Fig.~\ref{fig:10m_sim_results}a and Fig.~\ref{fig:20_60_sim_results}) and their HR counterparts for 20\,m and 60\,m bands at the 14-bit depth.}
\label{fig:sim_diffs_20m_60m}
\end{figure*}

In Fig.~\ref{fig:10m_sim_results}, we present an example of the super-resolved B08 band\footnote{The remaining bands for these scenes are presented in our Supplementary Material.} images (by a factor of $3\times$) for four different \sims\, and \simhsi\, scenes. We can appreciate that the results produced by \mymethod\, contain more high-frequency details and they are most similar to the HR references. This can also be seen in Fig.~\ref{fig:hsi_rgb}, where we show a color image composed from the B02, B03, and B04 bands (for the same scene as in Fig.~\ref{fig:10m_sim_results}d). It is worth noting that the colors are well reconstructed with \mymethod\, which is coherent with the low SAM values reported in Table~\ref{table:sim_results_table}. In Fig.~\ref{fig:20_60_sim_results}, we present the outcome of reconstructing the B06 (20\,m GSD) and B09 (60\,m GSD) bands (for the same scene as in Fig.~\ref{fig:10m_sim_results}a). Here, our model renders images $6\times$ and $18\times$ larger than the LR inputs which is also achieved with the \DSenHRn\, and \DSenR\, methods, however their outcome is of substantially lower quality than \mymethod. The remaining techniques operate with lower magnification ratios which results in the fine-grained image details being much worse reconstructed. Finally, Figs.~\ref{fig:sim_diffs_10m} and~\ref{fig:sim_diffs_20m_60m} depict the absolute difference maps with respect to the HR reference image. Black pixels indicate the zero error, whereas the yellow ones annotate the maximal difference. The maps associated with \mymethod~are significantly darker, which indicates lower reconstruction errors.

\subsection{Validation results for real-world S-2 data}\label{sec:real_world_data}

\begin{table}[ht!]
\centering
\caption{Reconstruction accuracy scores obtained for the real-world images from the MuS2 benchmark (10\,m bands). The best results in each group are boldfaced and the second best are underlined.}
\renewcommand{\tabcolsep}{1.6mm}
\begin{tabular}{lcccc}
    \Xhline{2\arrayrulewidth}
    & cPSNR & cSSIM & LPIPS & $\BalancedScore$ \\
    \hline
    \multicolumn{1}{l}{Bicubic}     & \textbf{30.30} & \textbf{0.8168} & 0.4220 & 1\\
    \multicolumn{1}{l}{RAMS}        & 29.91 & 0.7892 & \underline{0.3186} & \underline{0.9411}\\
    \multicolumn{1}{l}{HighRes-Net} & 29.45 & 0.7673 & 0.3212 & 0.9707\\
    \multicolumn{1}{l}{DeepSent}    &
    \underline{30.15} & \underline{0.7903} & \textbf{0.2967} & \textbf{0.9179}\\

\Xhline{2\arrayrulewidth}

\end{tabular}
\label{table:sim_results_table_mus2}
\end{table}

\begin{figure}[ht!]
\centering
\renewcommand{\tabcolsep}{0mm}
\begin{tabular}{c}
\includegraphics[width=0.48\textwidth]{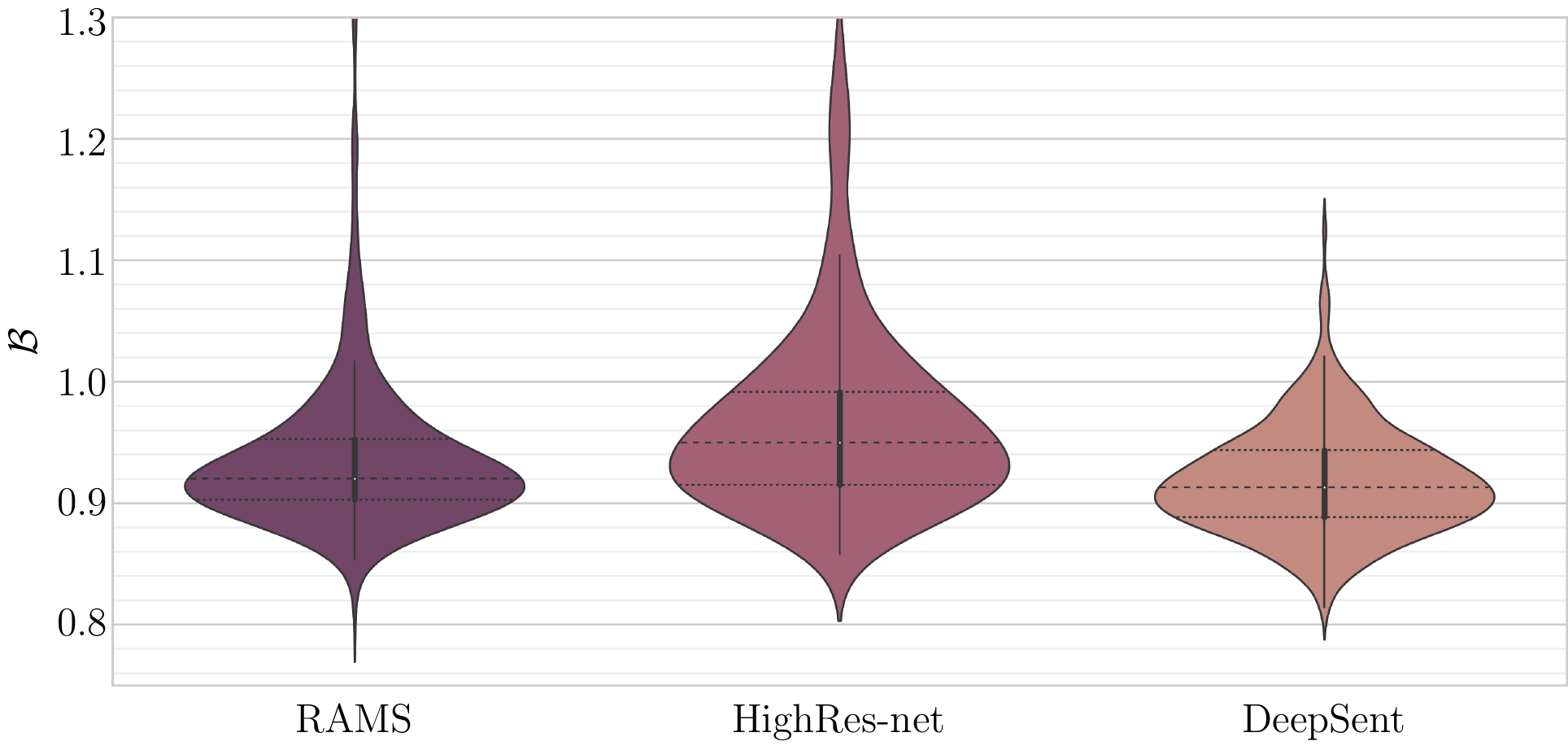} \\
\end{tabular}

\caption{Distribution of the balanced score ($\BalancedScore$) obtained for the MuS2 benchmark dataset. The dashed lines indicate the median and quartile values.}
\label{fig:mus2_beta_plot}
\end{figure}

\begin{figure*}[ht!]
\centering
\footnotesize
\renewcommand{\tabcolsep}{0.3mm}
\renewcommand{\arraystretch}{0.66}
\newcommand{\mywidth}{0.158}
\newcommand{\halfwidth}{0.0765}
\begin{tabular}{ccccccc}
    & LR image(s) & Bicubic interpolation & RAMS & HighRes-net  & DeepSent & HR image\\
    \raisebox{14.5mm}{a)} &
    \includegraphics[width=\mywidth\textwidth, trim={0 150 0 0}, clip]{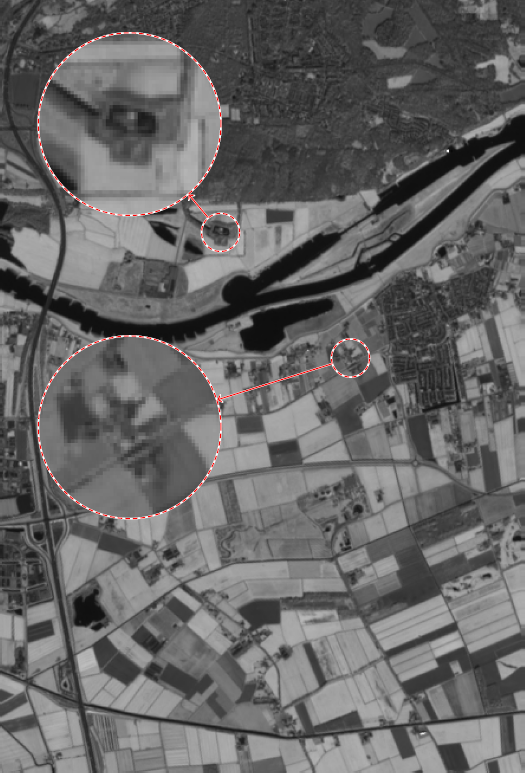} &
    \includegraphics[width=\mywidth\textwidth, trim={0 150 0 0}, clip]{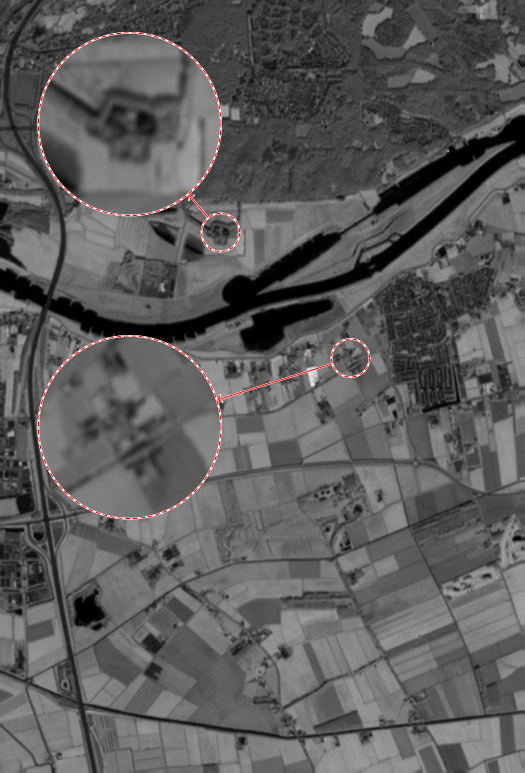} &
    \includegraphics[width=\mywidth\textwidth, trim={0 150 0 0}, clip]{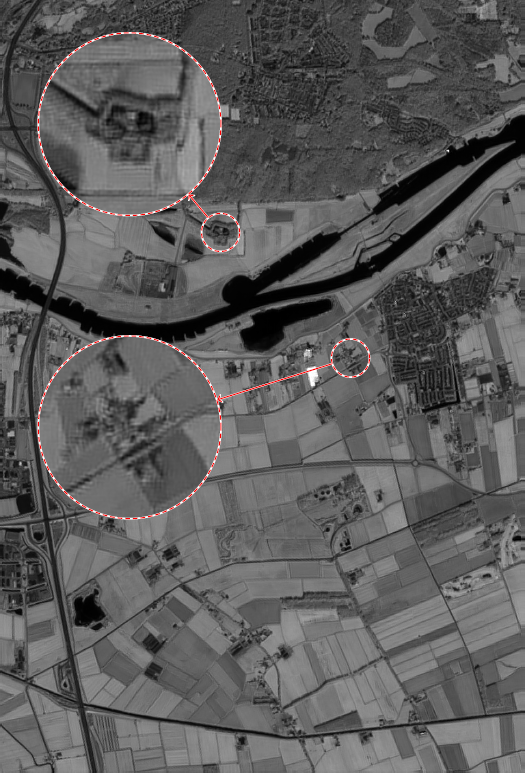} &
    \includegraphics[width=\mywidth\textwidth, trim={0 150 0 0}, clip]{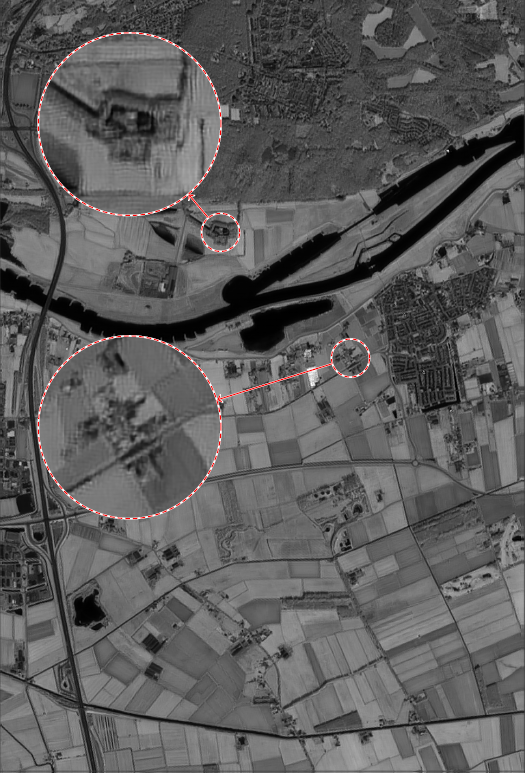} &
    \includegraphics[width=\mywidth\textwidth, trim={0 150 0 0}, clip]{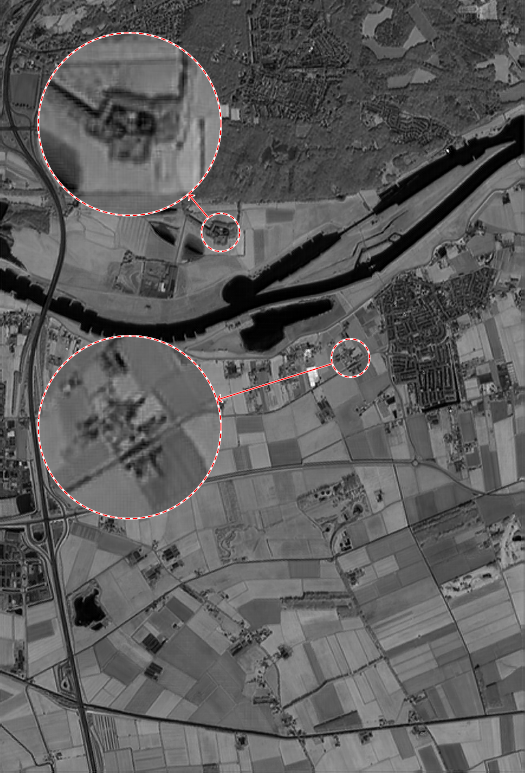} &
    \includegraphics[width=\mywidth\textwidth, trim={0 150 0 0}, clip]{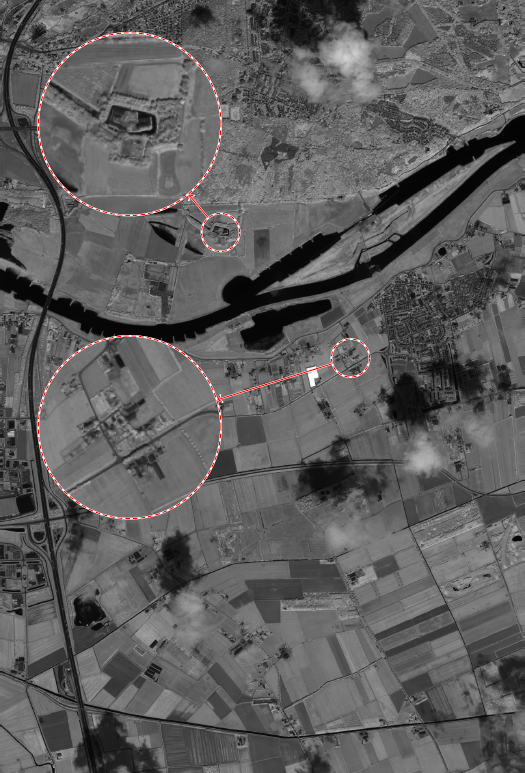}  \\

     \\

    \rule{0pt}{2mm}\\

    \raisebox{13.0mm}{b)} &
    \raisebox{12.0mm}{
    \begin{tabular}{cc}
       \includegraphics[width=\halfwidth\textwidth]{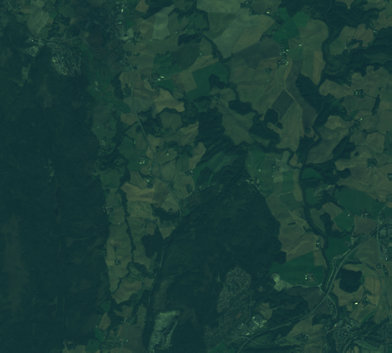}  &  \includegraphics[width=\halfwidth\textwidth]{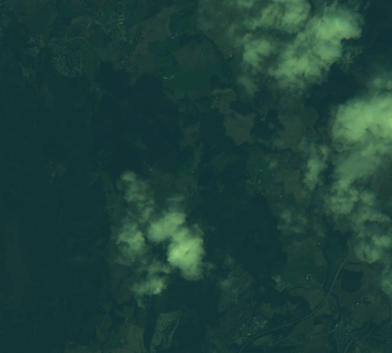}\\
       \includegraphics[width=\halfwidth\textwidth]{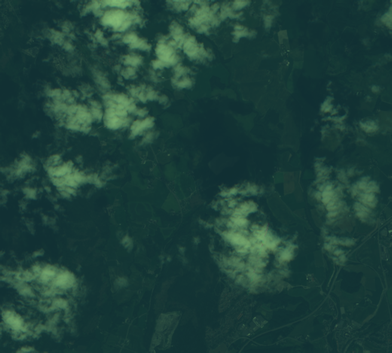}  &  \includegraphics[width=\halfwidth\textwidth]{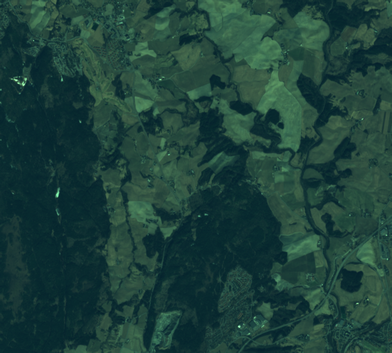}\\
    \end{tabular}
    }

    &

    \includegraphics[width=\mywidth\textwidth]{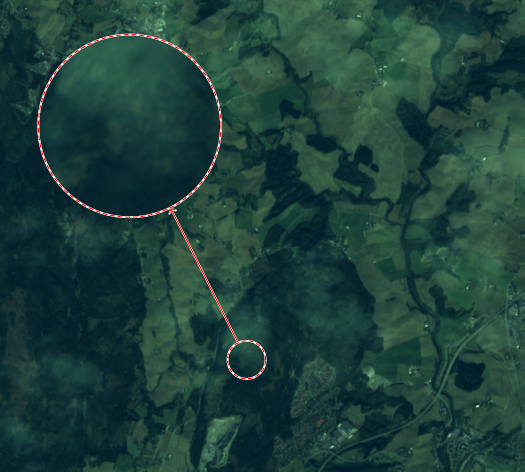} &
    \includegraphics[width=\mywidth\textwidth]{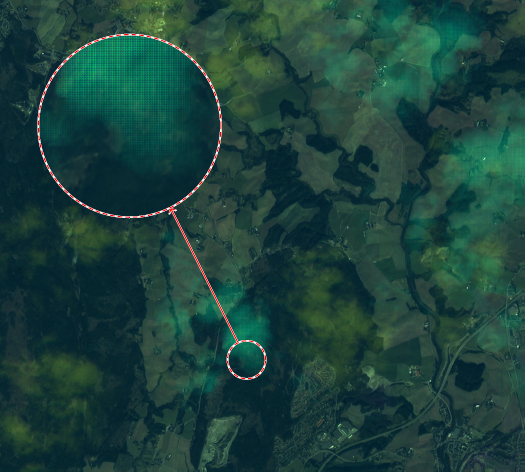} &
    \includegraphics[width=\mywidth\textwidth]{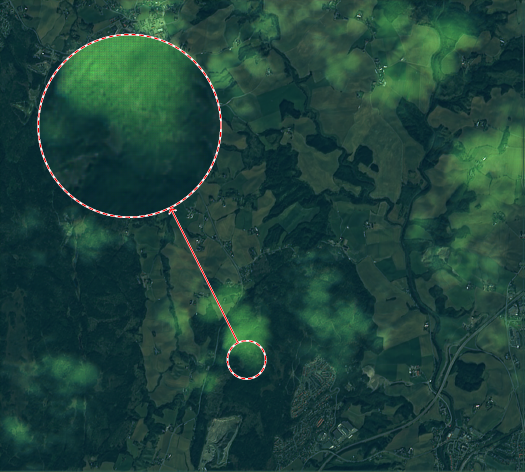} &
    \includegraphics[width=\mywidth\textwidth]{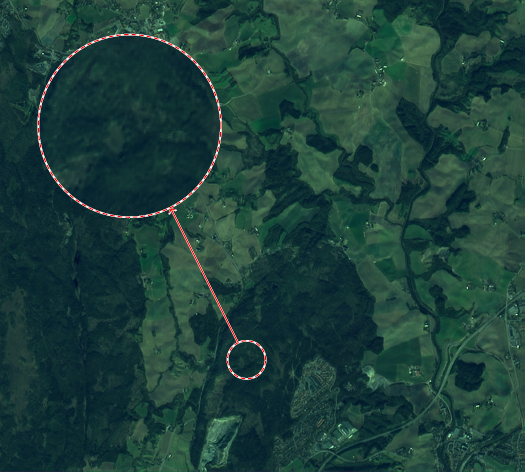} &
    \includegraphics[width=\mywidth\textwidth]{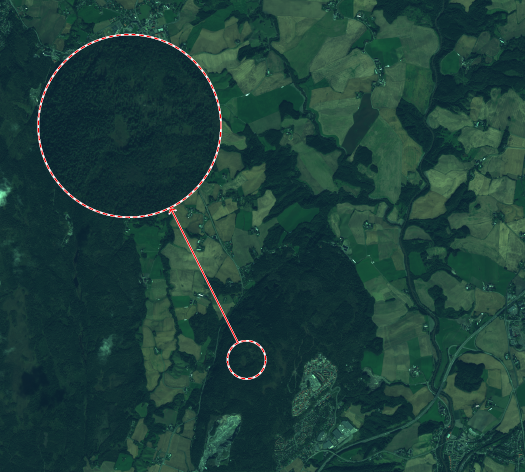}  \\

\end{tabular}
\caption{An example of super-resolving images from the MuS2 dataset using different techniques: B08 band~(a) and a color image composed of the B02, B03, and B04 bands~(b).  The reconstruction was performed from multiple simulated LR images---for (a), one of them is shown and for (b)~four examples are presented to show the differences between them. Also, the HR reference acquired with WorldView-2 is provided. The results for the remaining bands are included in the Supplementary Material.}
\label{fig:mus2_rgb}
\end{figure*}

The capabilities of \mymethod\, and other techniques for super-resolving real-world images were validated with \reals\, and MuS2 images. The former are not coupled with any HR reference, and for the latter, HR references are provided for the 10\,m S-2 bands. In Table~\ref{table:sim_results_table_mus2}, we present the quantitative scores obtained for the MuS2 dataset. As argued in~\cite{Kowaleczko2022}, cPSNR and cSSIM are not sufficiently robust in this case due to substantial differences between LR and HR images resulting from the fact that they were acquired using different satellites. LPIPS is much more informative here, along with the balanced score ($\BalancedScore$), specified in the MuS2 benchmark protocol. $\BalancedScore$ normalizes each score (per each band) with the result obtained using bicubic interpolation (the smaller, the better), as originally proposed in~\cite{Martens2019}. Distribution of these scores is presented in Fig.~\ref{fig:mus2_beta_plot}---DeepSent is slightly better than RAMS, and the difference is significant in the statistical sense (though the differences are not significant for band B04 between \mymethod\, and HighRes-net, and for band B08 between \mymethod\, and RAMS). Qualitative results are presented in Fig.~\ref{fig:mus2_rgb}. For the B08 band (Fig.~\ref{fig:mus2_rgb}a), it can be seen that \mymethod\, reconstructs the image details that are quite close to the HR reference (it is apparent that there is substantial information gain compared with the input LR image and the outcome of bicubic interpolation). Although we can observe some high-frequency artifacts on the edges, they are much smaller than for RAMS and HighRes-net. An interesting observation can also be made from the color image (Fig.~\ref{fig:mus2_rgb}b) composed of B02, B03, and B04 bands. In this case, the Earth surface in some of LR images is partially occluded with clouds, which affects the result of RAMS and HighRes-net---the clouds are not filtered out and heavy artifacts can be noticed in these regions. Clearly, \mymethod\, demonstrates much higher robustness in this case, which may result from combining the multitemporal and multispectral data fusion, and the reconstructed HR information is quite close to the provided reference.

\begin{figure*}[ht!]
\centering
\footnotesize
\renewcommand{\tabcolsep}{0.2mm}
\renewcommand{\arraystretch}{0.4}
\newcommand{\mywidth}{0.14}
\begin{tabular}{rc
    >{\columncolor[HTML]{99c1de}}p{1mm}
    >{\columncolor[HTML]{99c1de}}c
    >{\columncolor[HTML]{99c1de}}c
    >{\columncolor[HTML]{99c1de}}p{1mm}
    p{1mm}
    >{\columncolor[HTML]{bcd4e6}}p{1mm}
    >{\columncolor[HTML]{bcd4e6}}c
    >{\columncolor[HTML]{bcd4e6}}c
    >{\columncolor[HTML]{bcd4e6}}p{1mm}
    p{1mm}
    >{\columncolor[HTML]{d6e2e9}}p{1mm}
    >{\columncolor[HTML]{d6e2e9}}c
    >{\columncolor[HTML]{d6e2e9}}c
    >{\columncolor[HTML]{d6e2e9}}p{1mm}}
    & & \multicolumn{4}{c}{\cellcolor[HTML]{99c1de}60\,m} & &
    \multicolumn{4}{c}{\cellcolor[HTML]{bcd4e6}20\,m} & &
    \multicolumn{4}{c}{\cellcolor[HTML]{d6e2e9}10\,m} \rule{0pt}{3mm} \\[1mm]

    & & & B01 & B09 & & & & B05 & B08a & & & & B04 & B08 &\\[1mm]
    \raisebox{11mm}{\begin{turn}{90}LR image\end{turn}} & &
    & \includegraphics[width=\mywidth\textwidth]{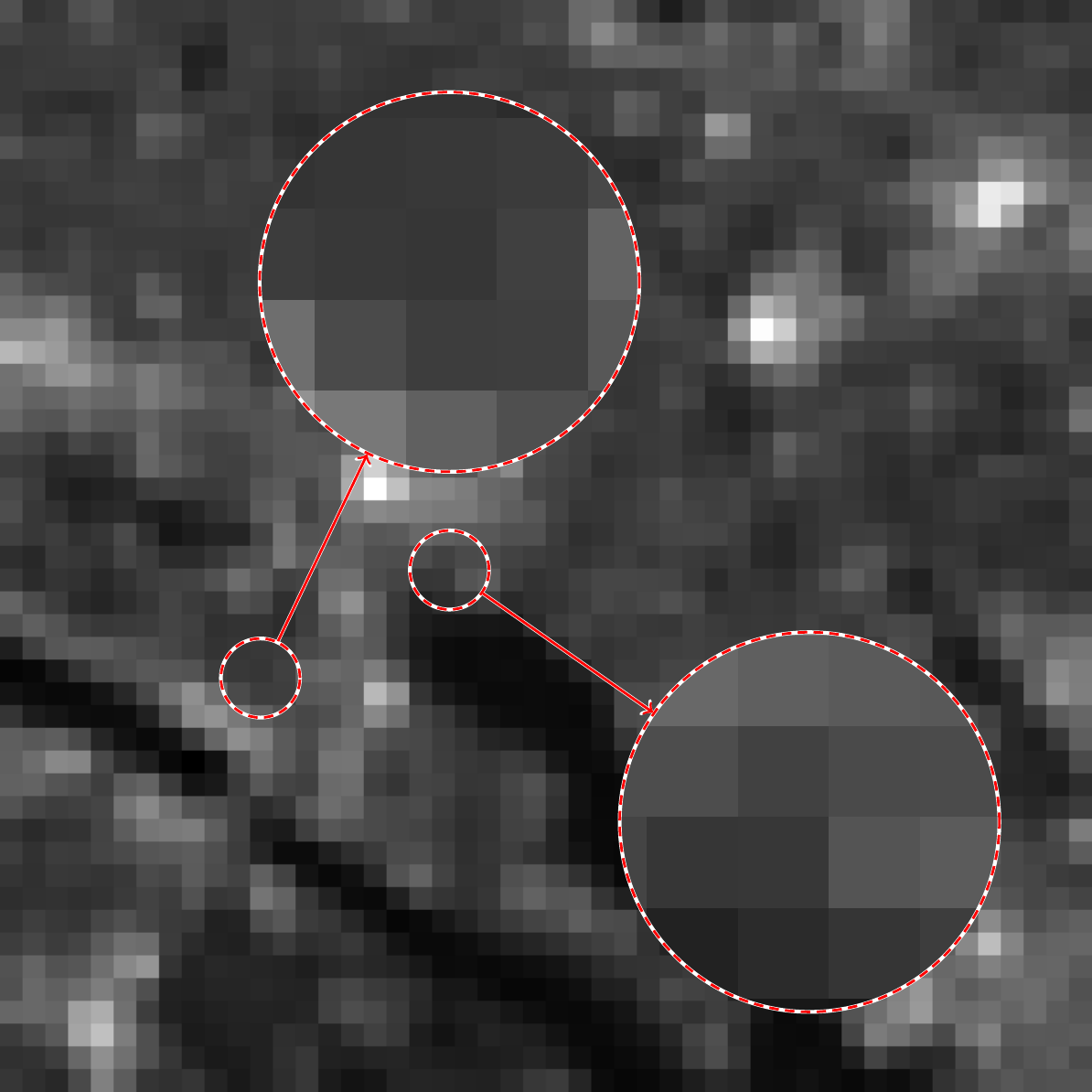} &
    \includegraphics[width=\mywidth\textwidth]{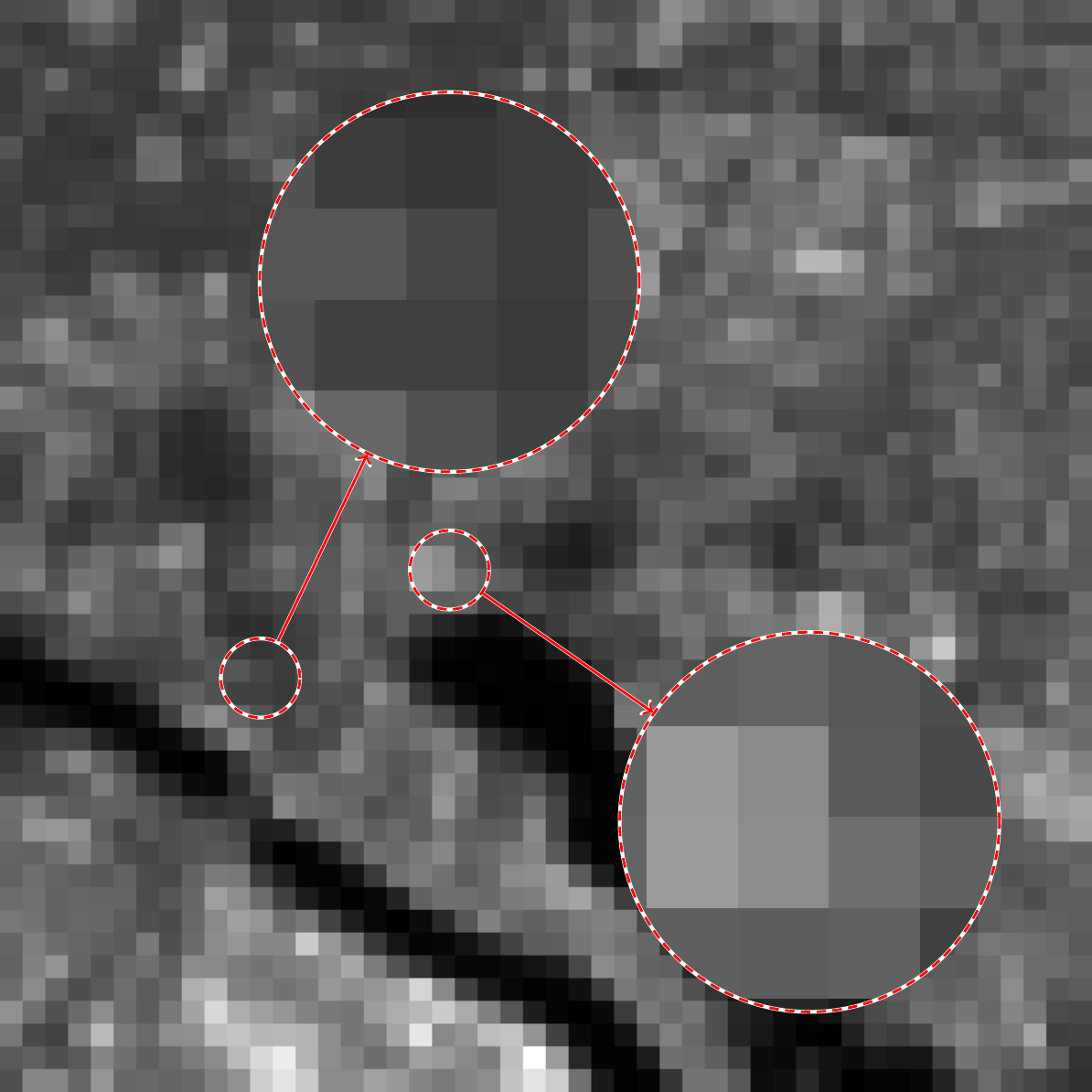} & & &
    & \includegraphics[width=\mywidth\textwidth]{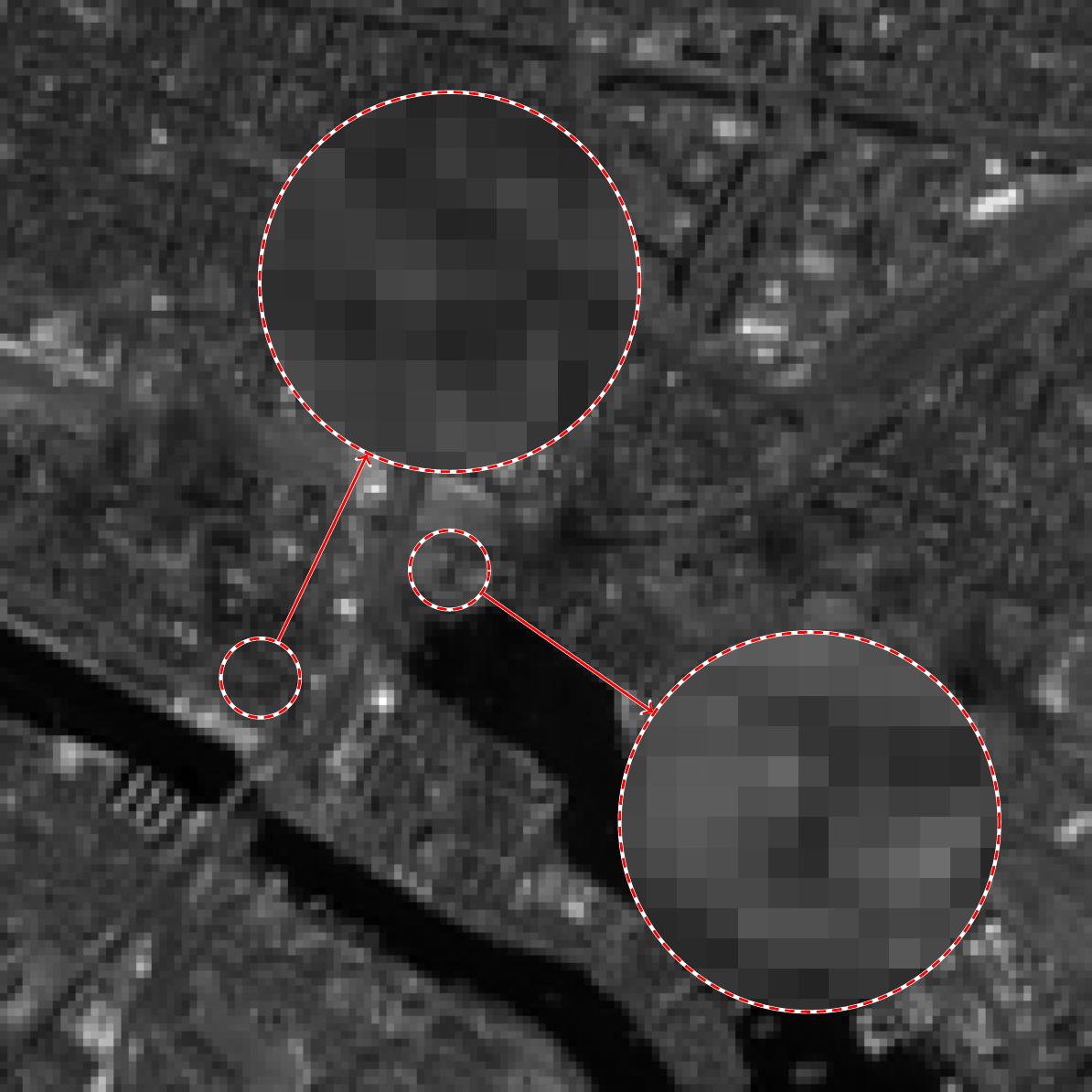} &
    \includegraphics[width=\mywidth\textwidth]{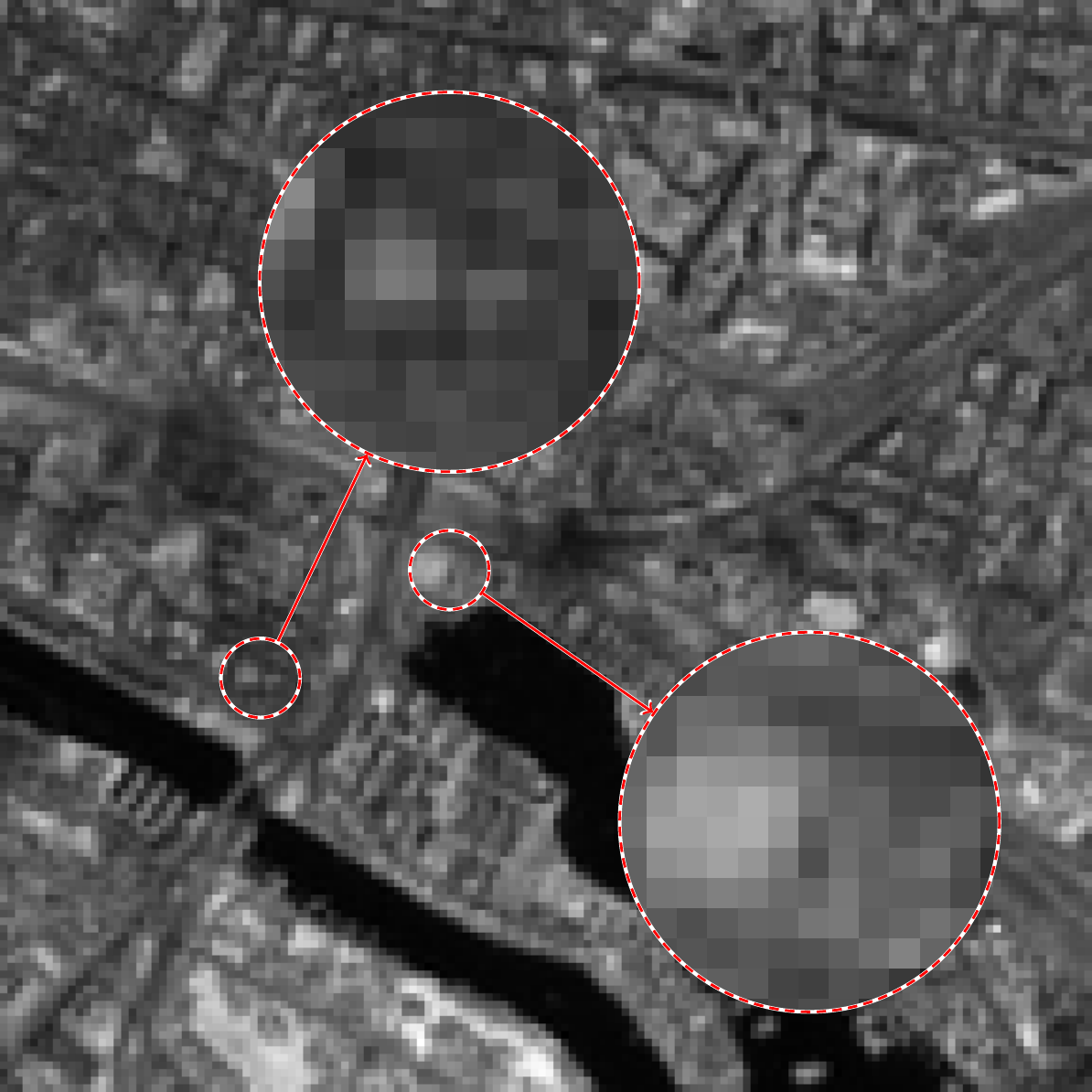} & & &
    & \includegraphics[width=\mywidth\textwidth]{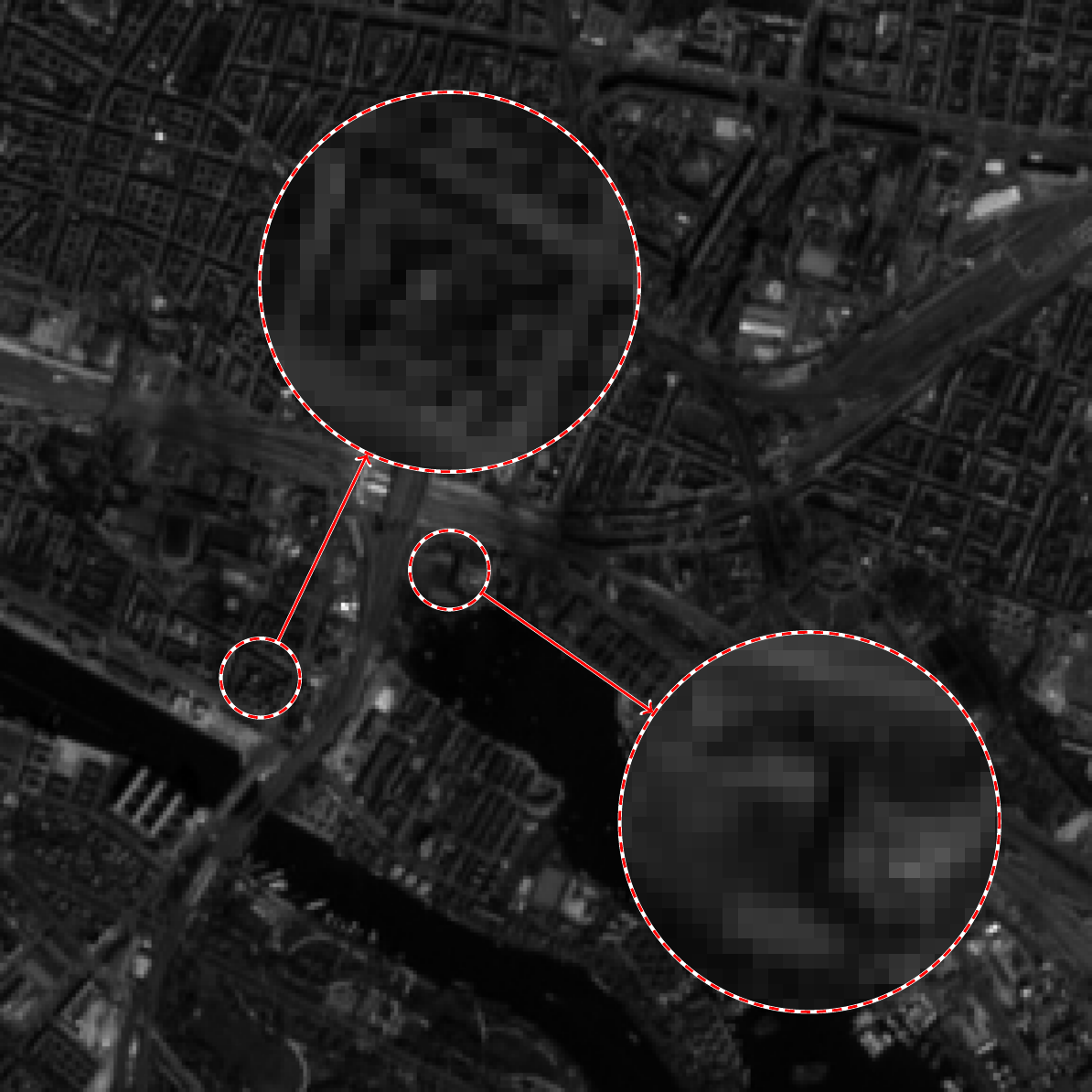} &
    \includegraphics[width=\mywidth\textwidth]{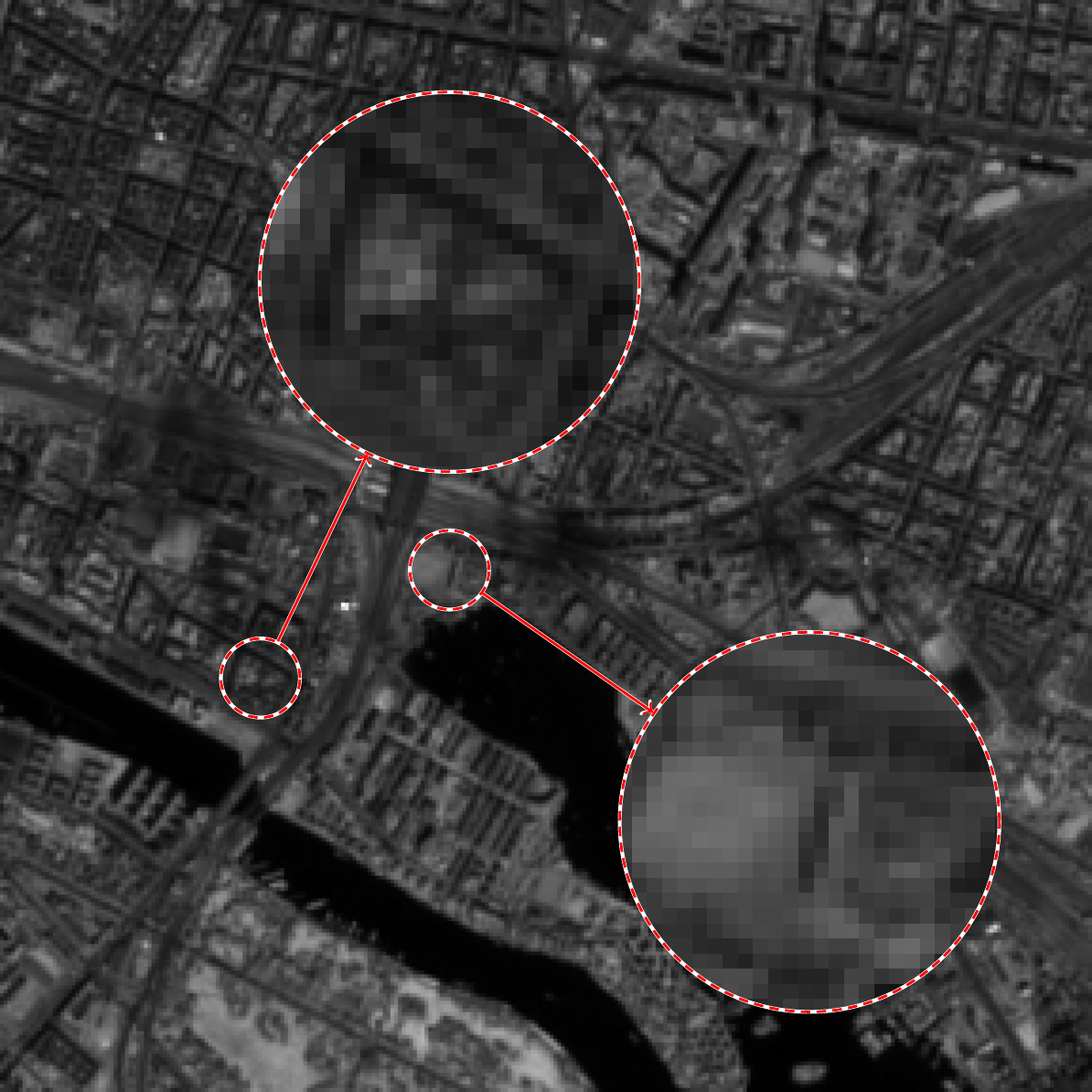} & \\

    \raisebox{9.5mm}{\begin{turn}{90}RAMS\end{turn}} & &
    & \includegraphics[width=\mywidth\textwidth]{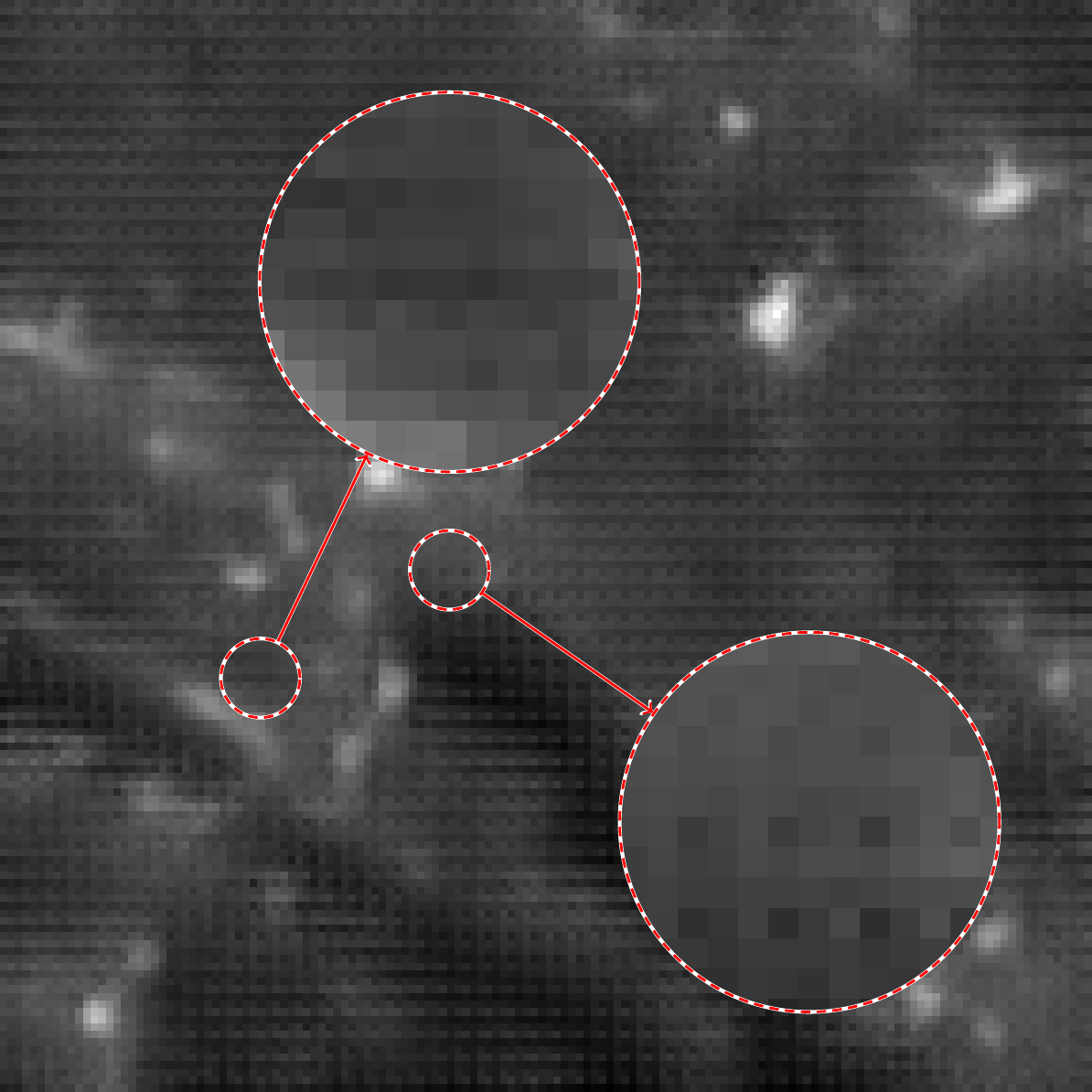} &
    \includegraphics[width=\mywidth\textwidth]{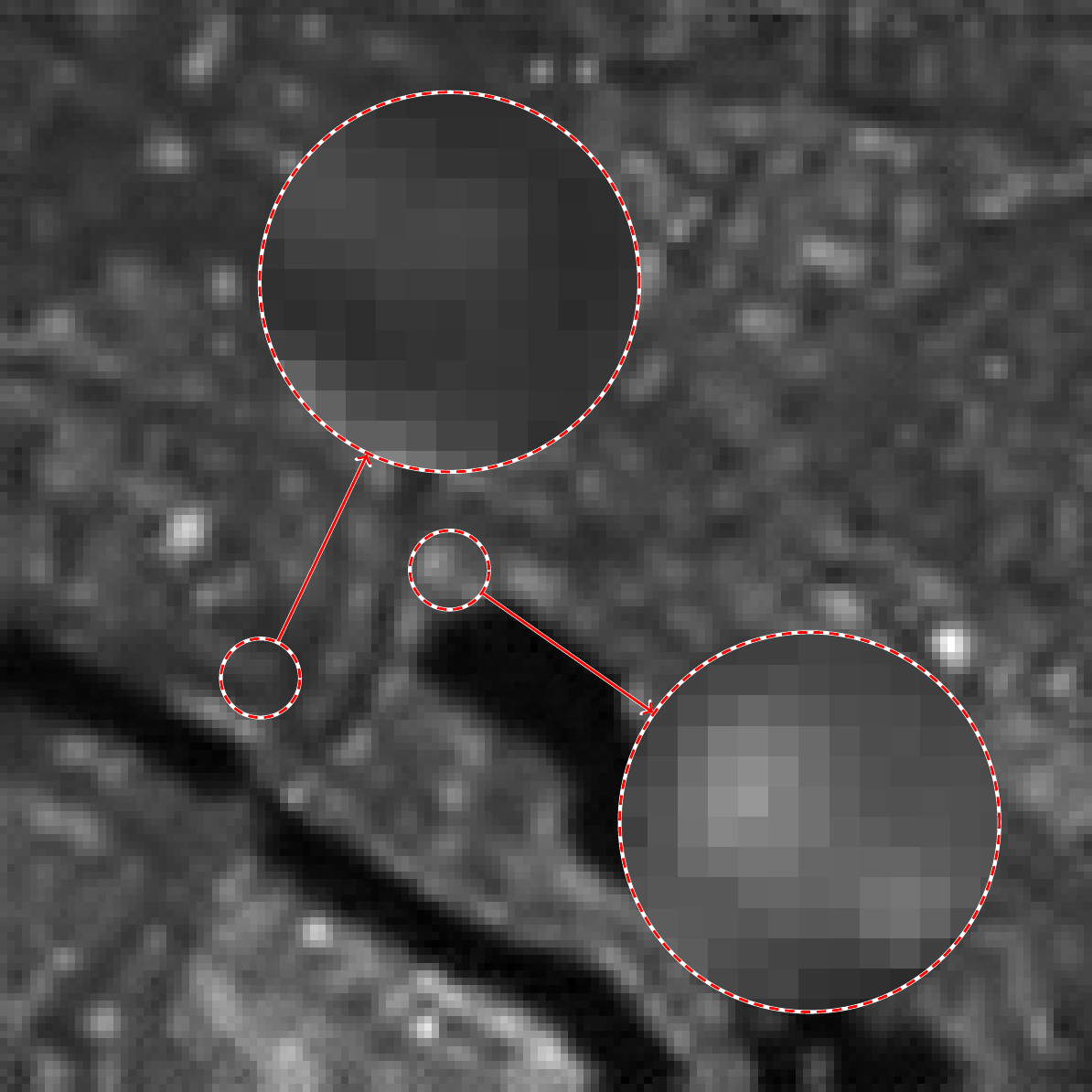} & & &
    & \includegraphics[width=\mywidth\textwidth]{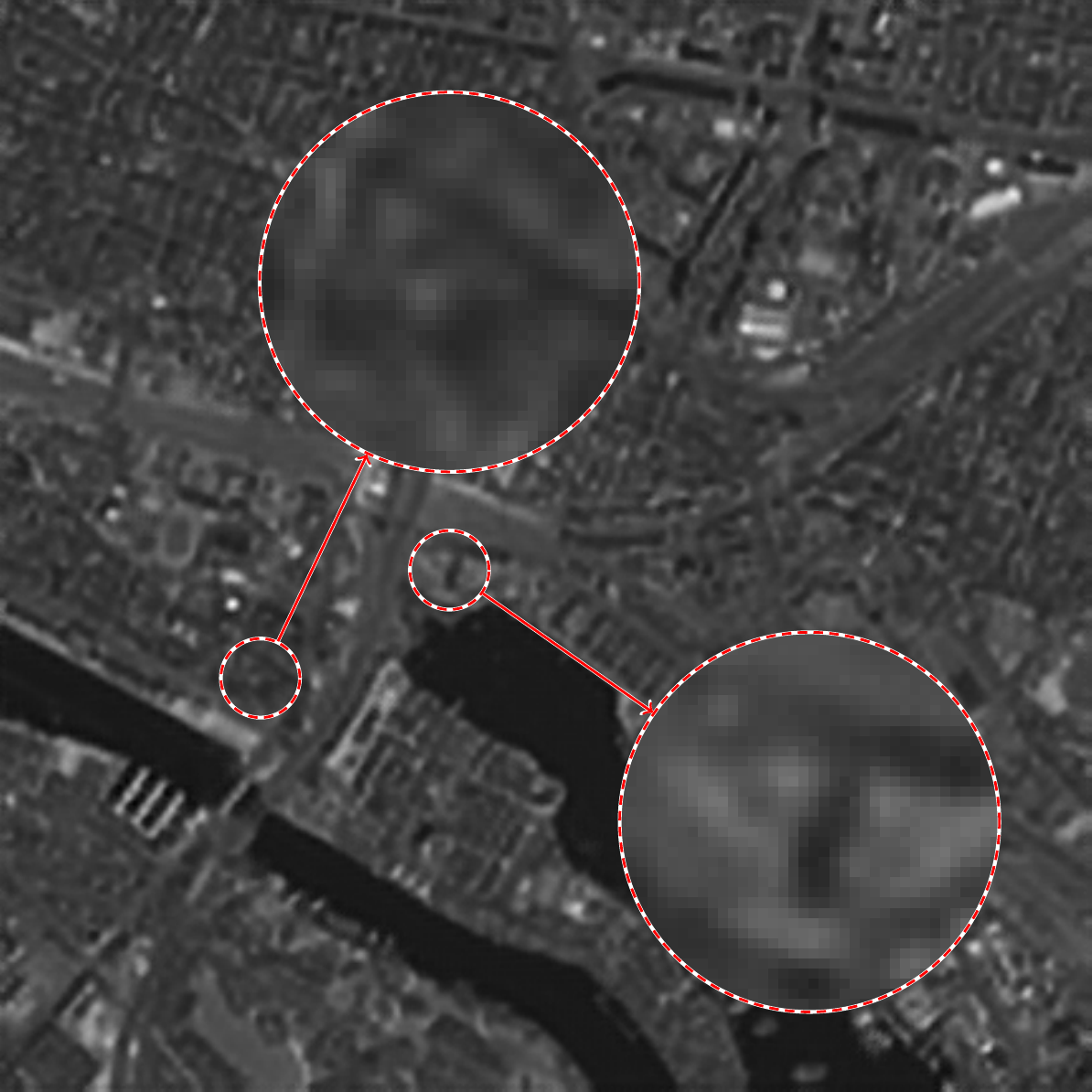} &
    \includegraphics[width=\mywidth\textwidth]{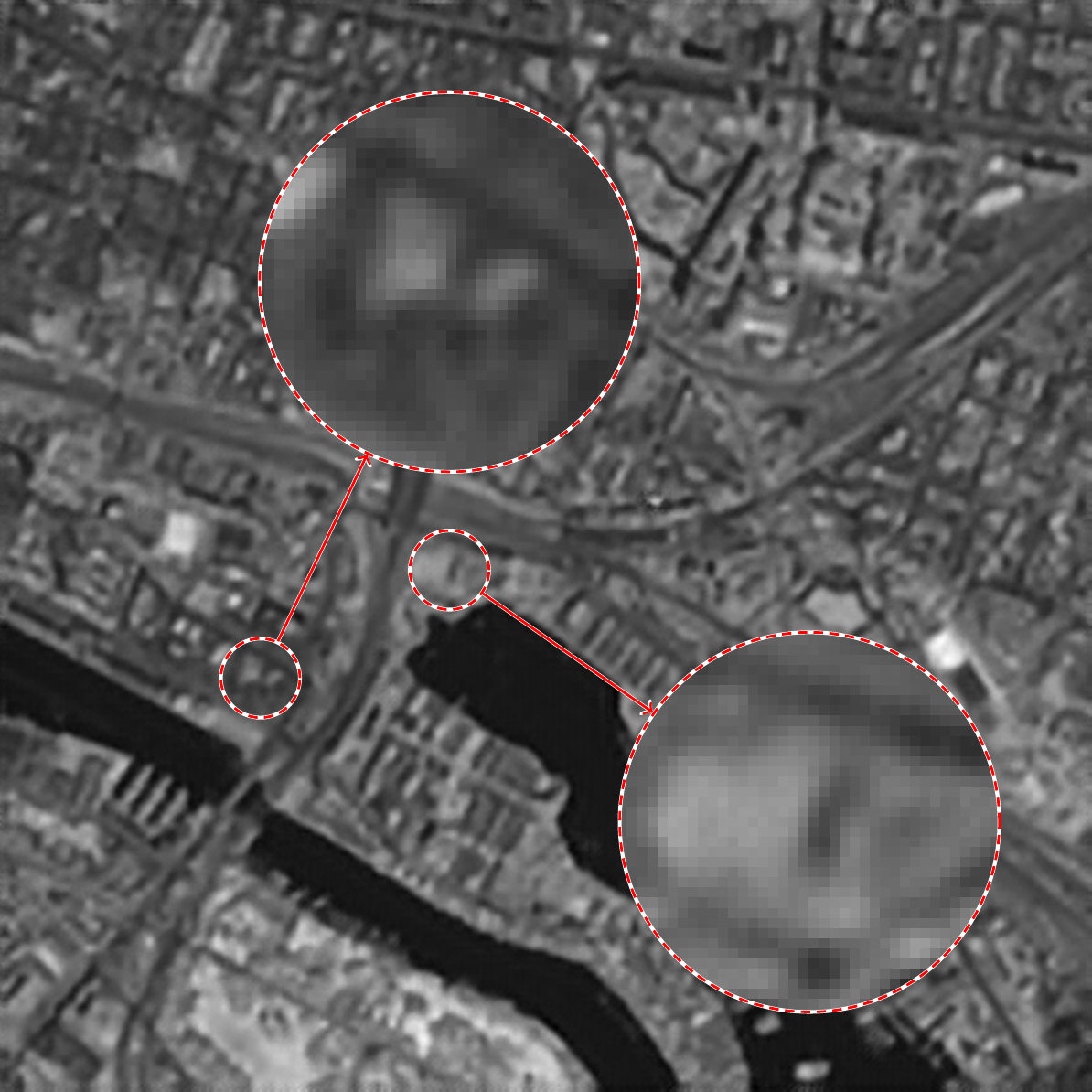} & & &
    & \includegraphics[width=\mywidth\textwidth]{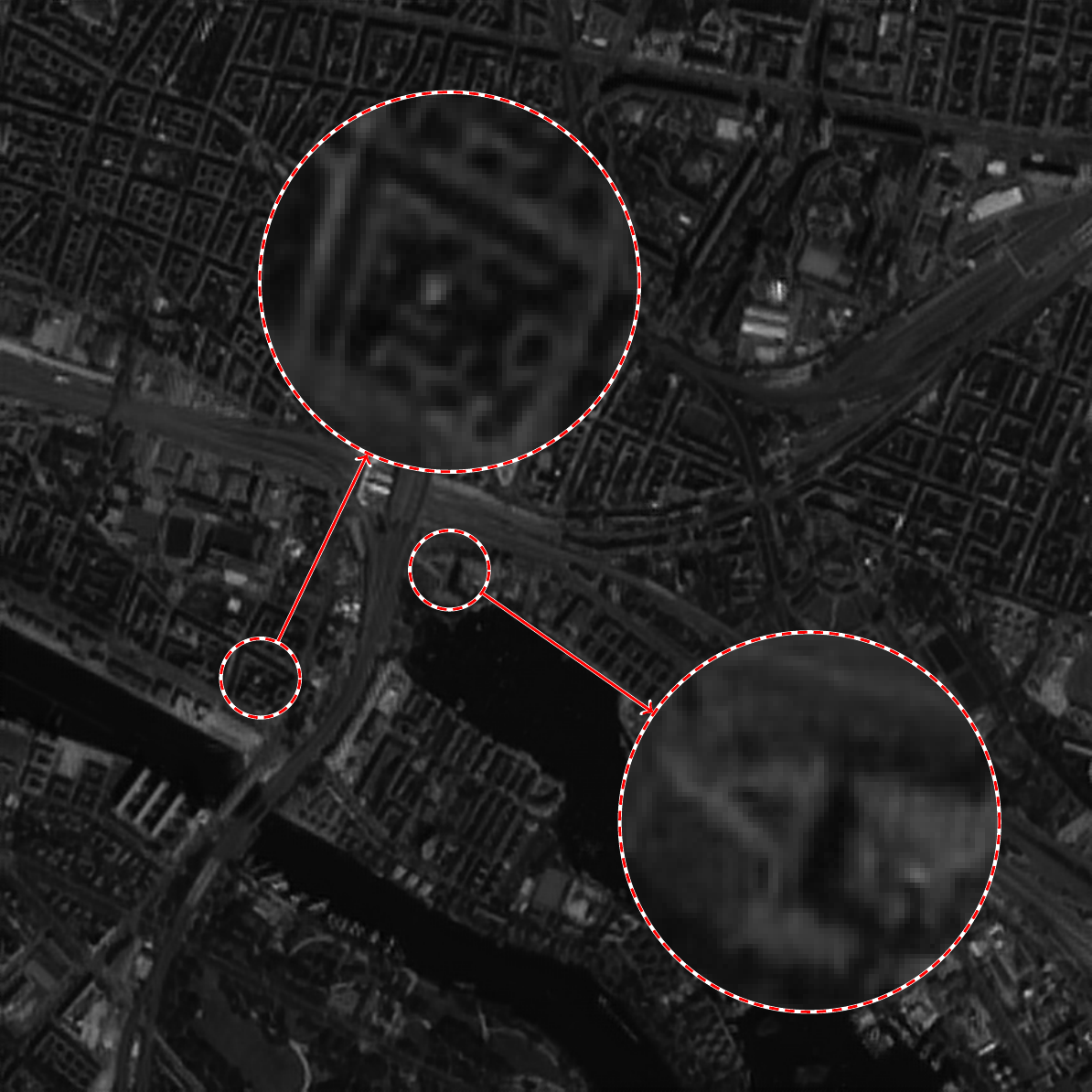} &
    \includegraphics[width=\mywidth\textwidth]{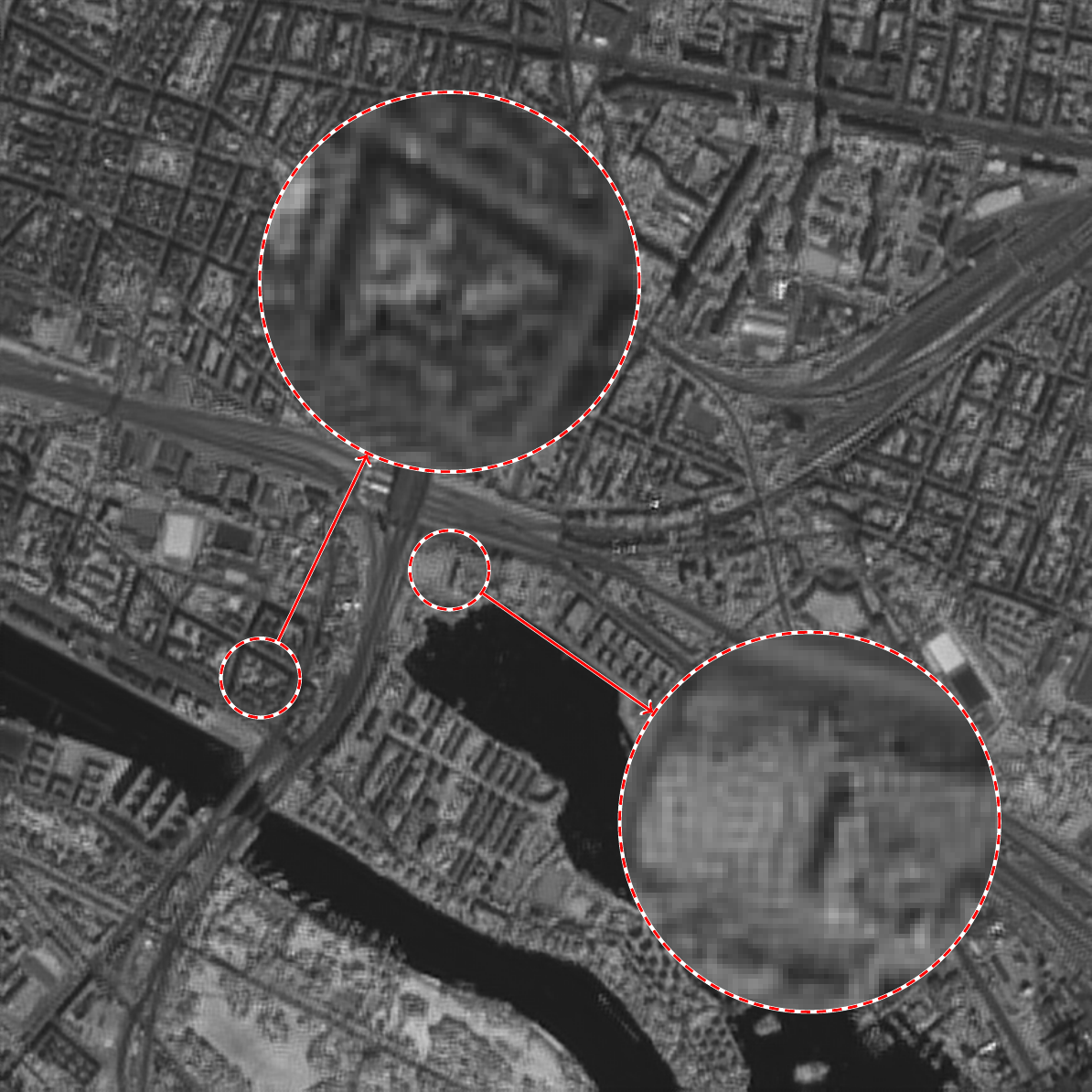} & \\

    \raisebox{6mm}{\begin{turn}{90}HighRes-net\end{turn}} & &
    & \includegraphics[width=\mywidth\textwidth]{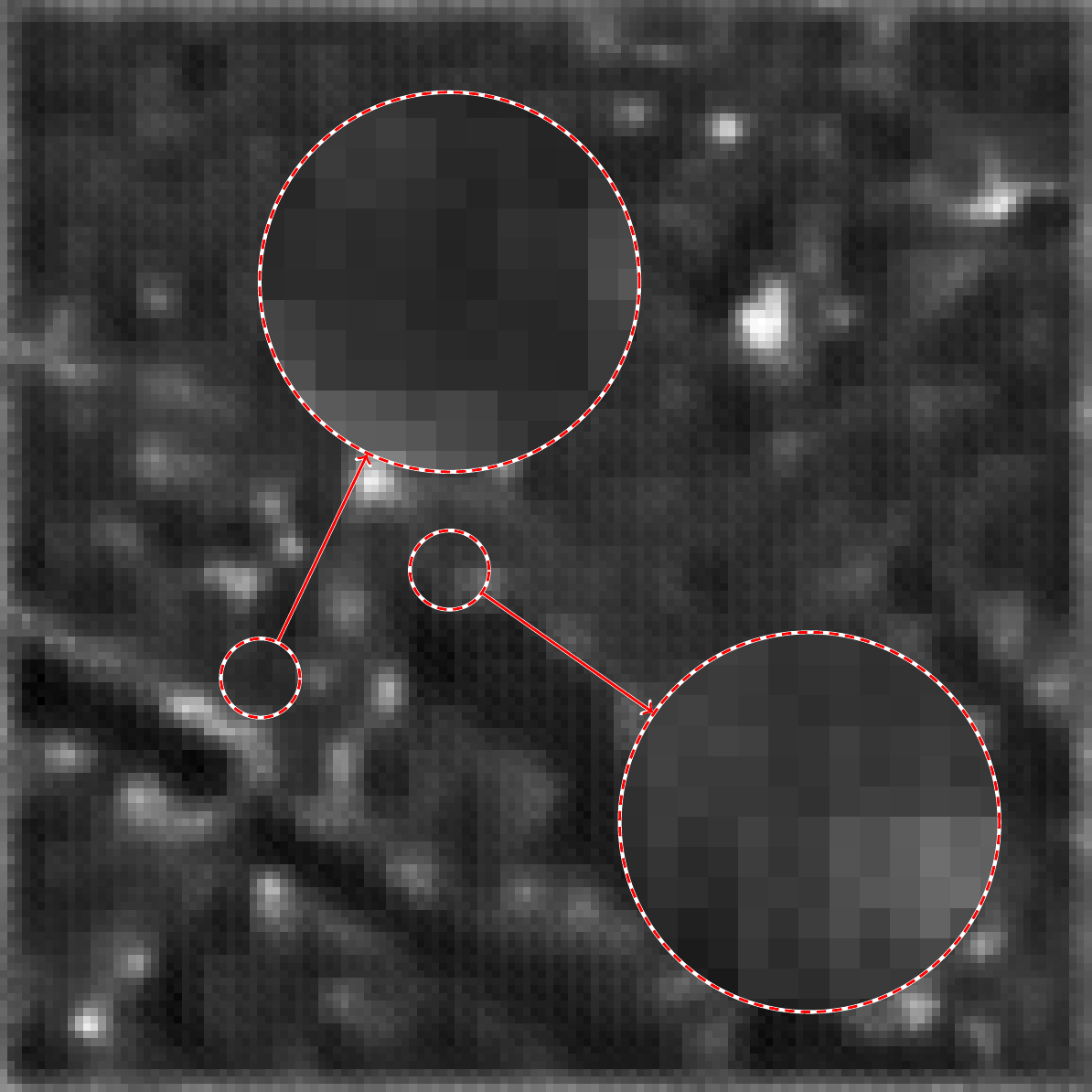} &
    \includegraphics[width=\mywidth\textwidth]{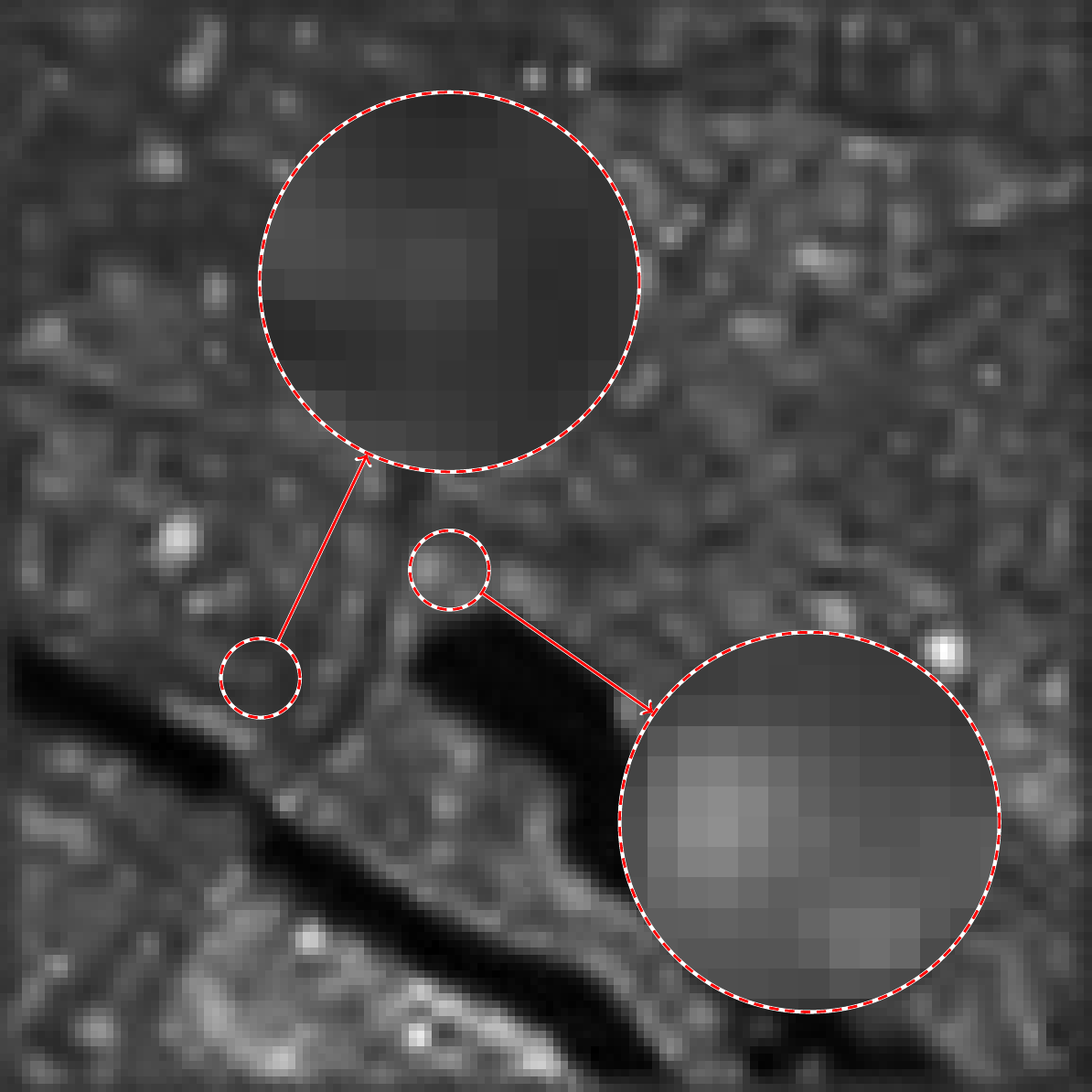} & & &
    & \includegraphics[width=\mywidth\textwidth]{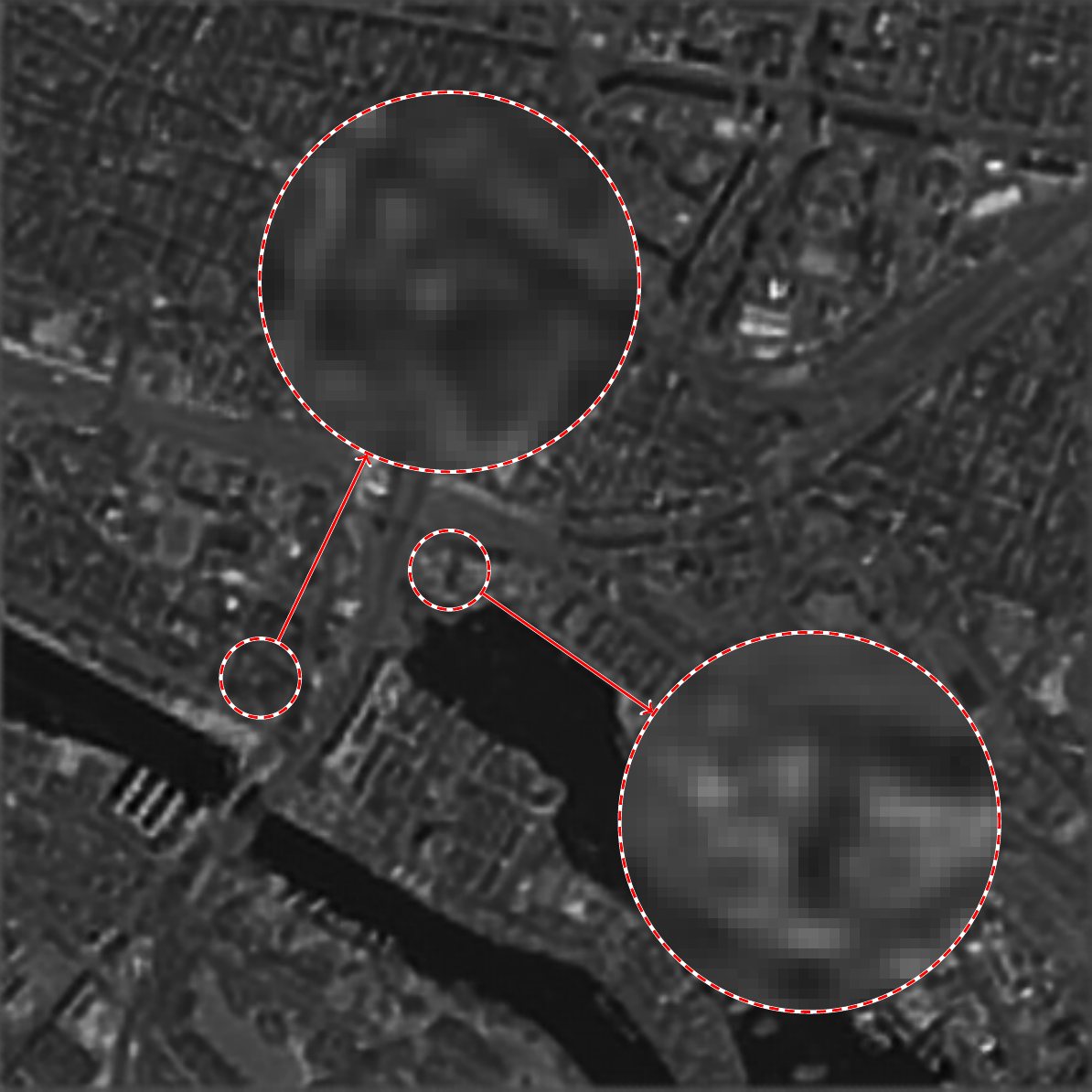} &
    \includegraphics[width=\mywidth\textwidth]{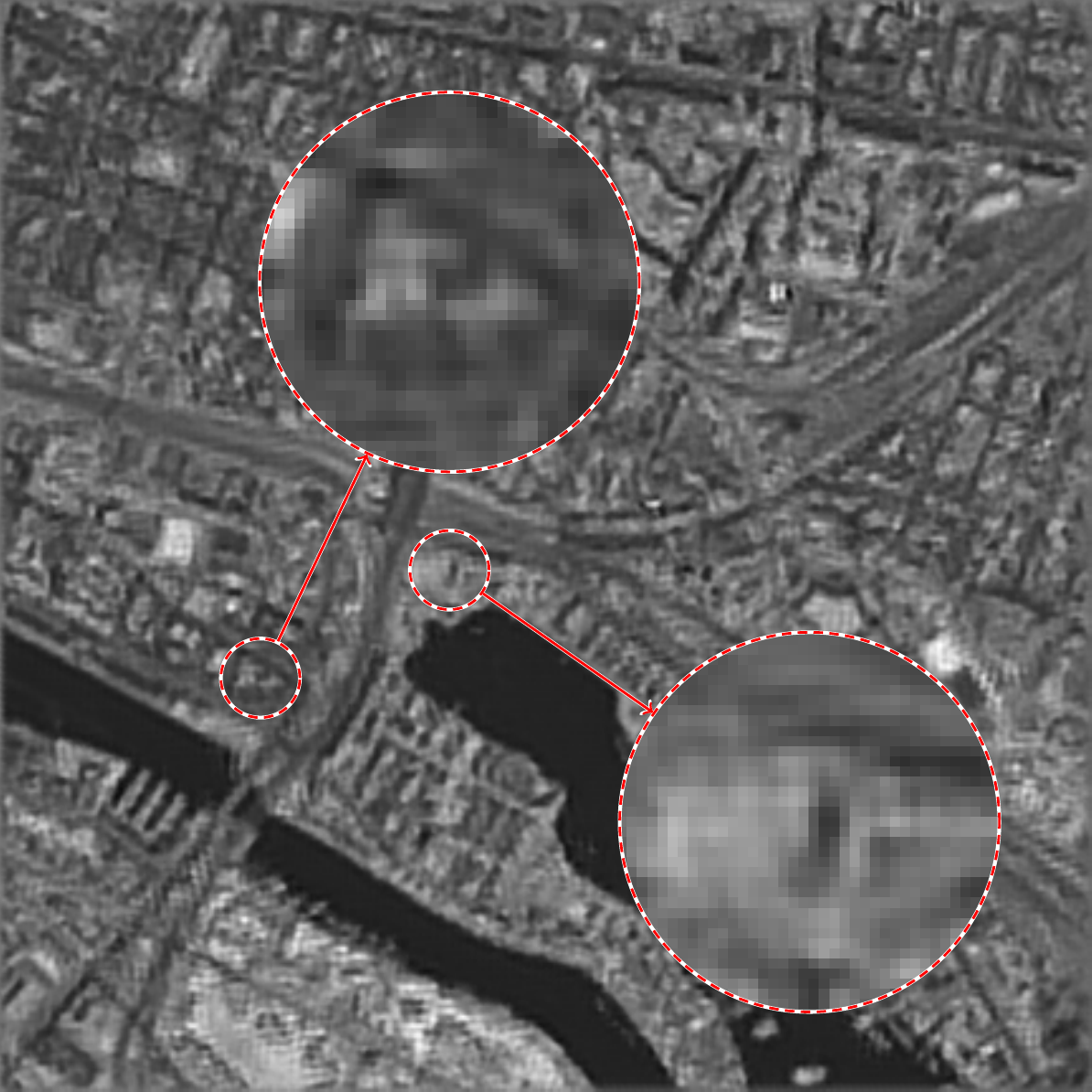} & & &
    & \includegraphics[width=\mywidth\textwidth]{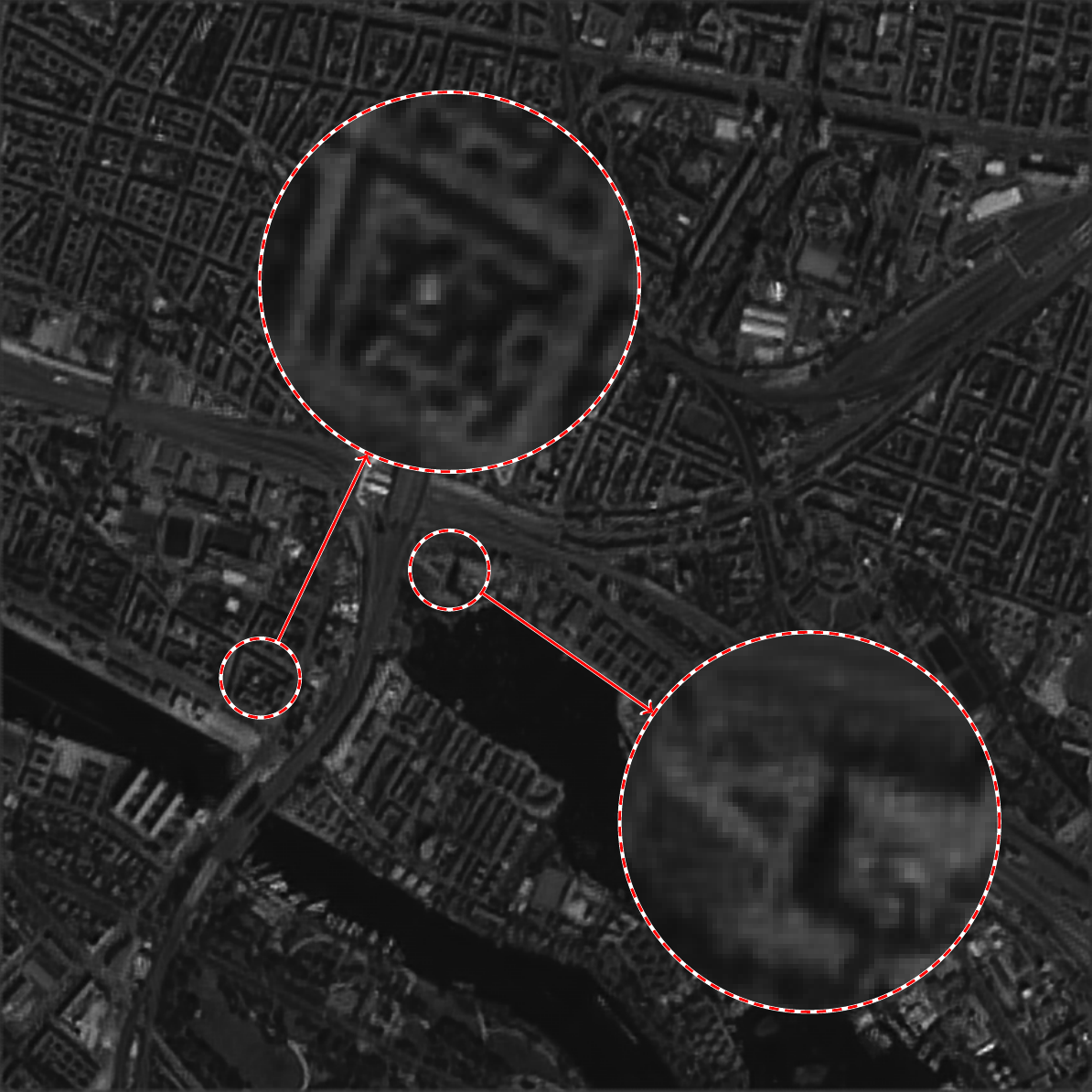} &
    \includegraphics[width=\mywidth\textwidth]{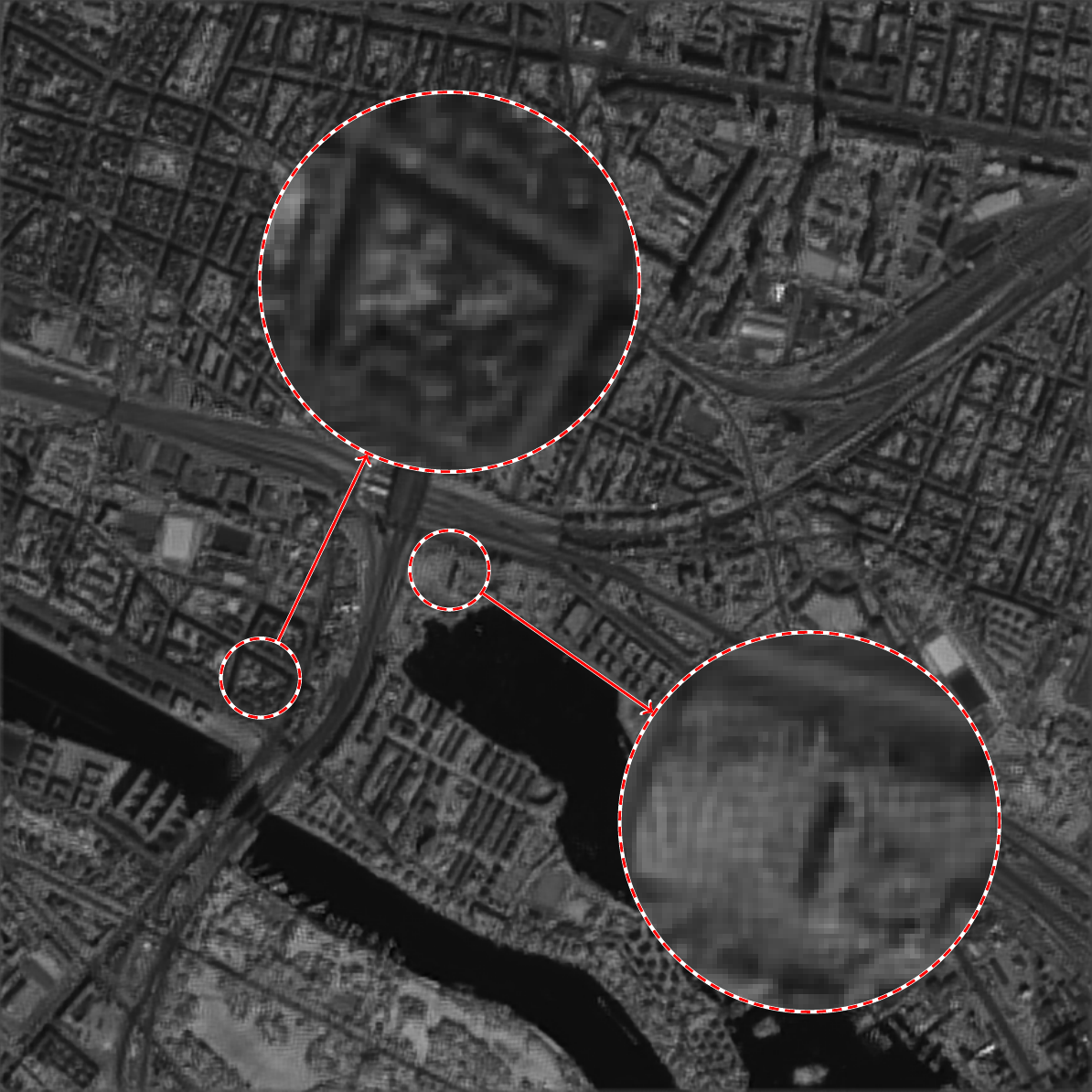} & \\

    \raisebox{10mm}{\begin{turn}{90}DSen2\end{turn}} & &
    & \includegraphics[width=\mywidth\textwidth]{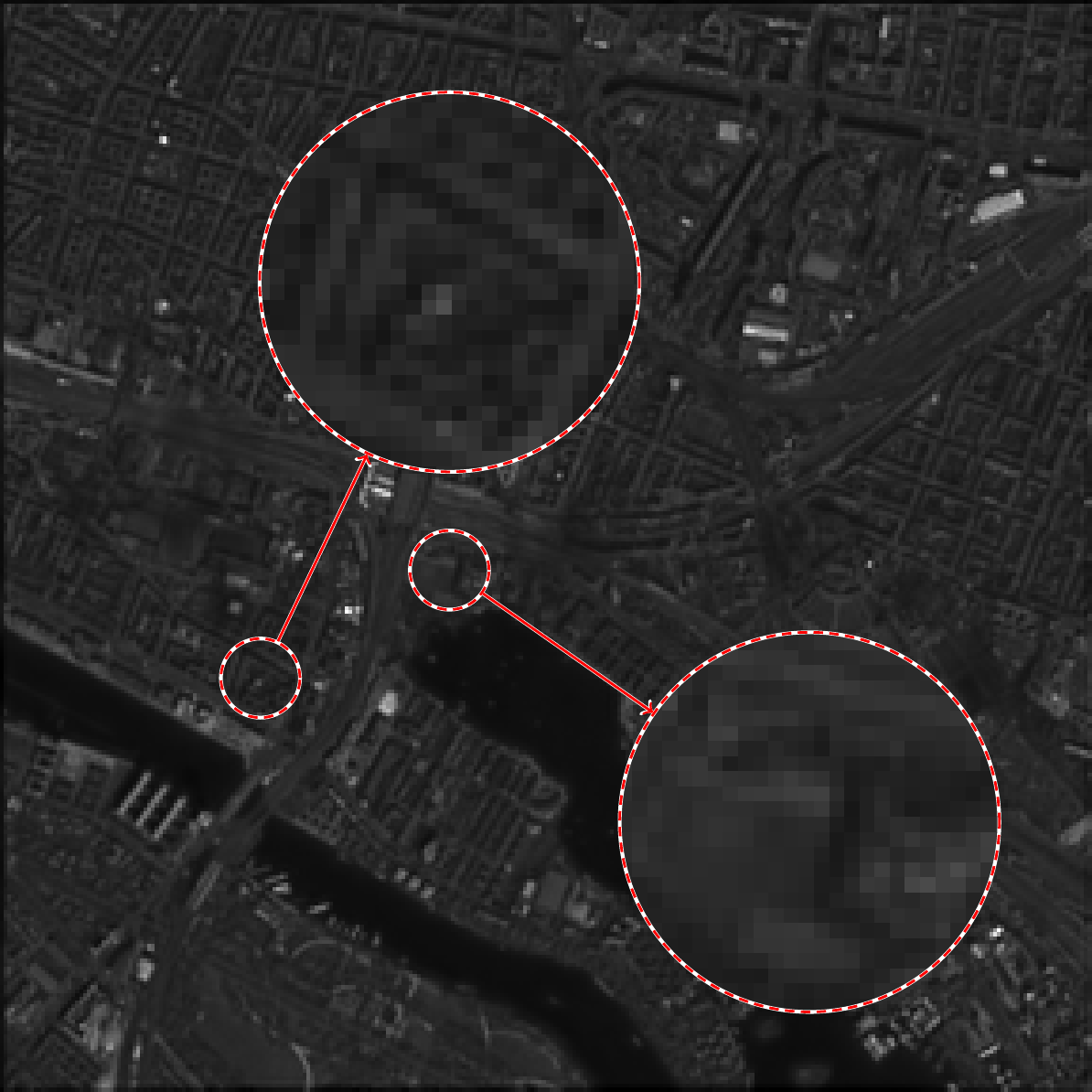} &
    \includegraphics[width=\mywidth\textwidth]{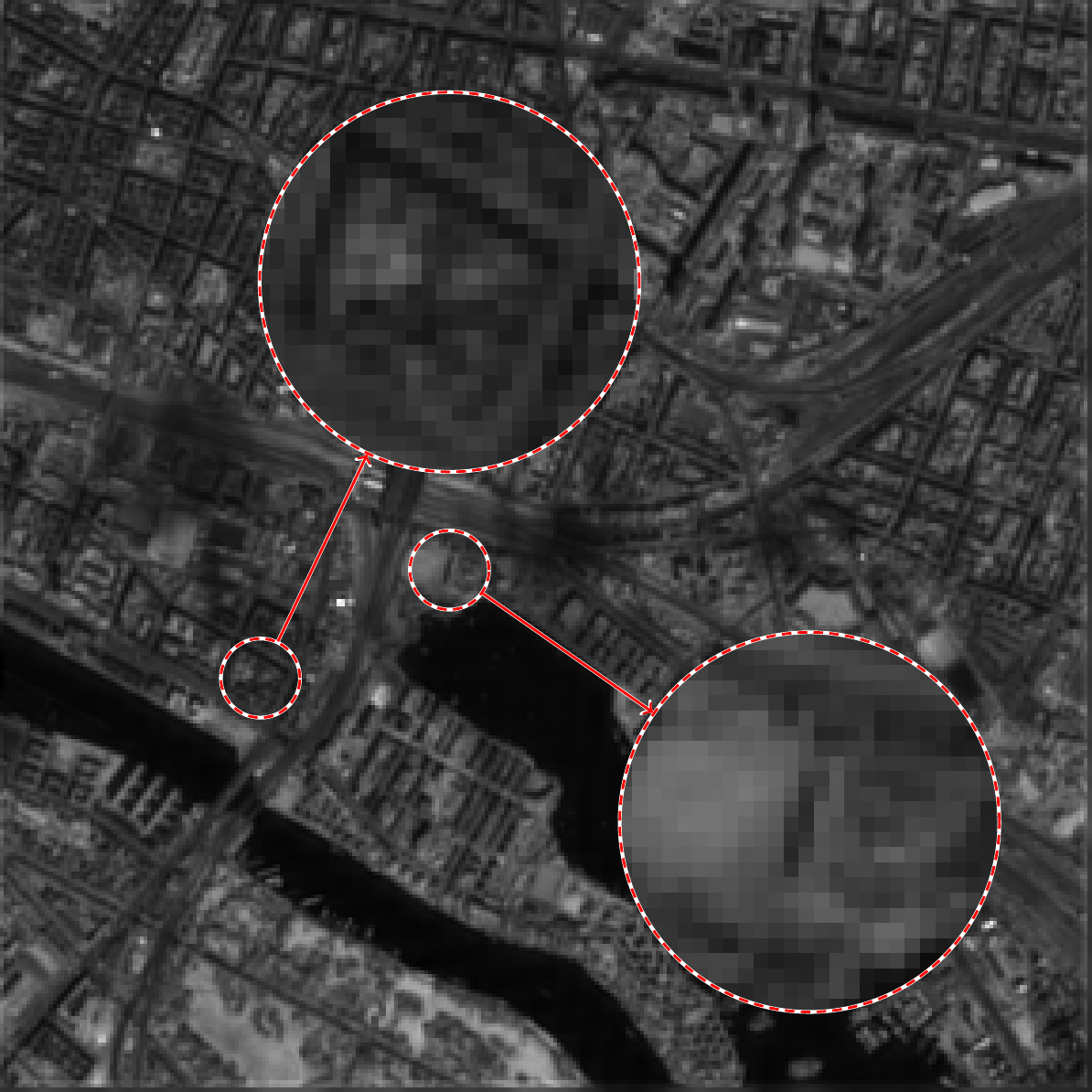} & & &
    & \includegraphics[width=\mywidth\textwidth]{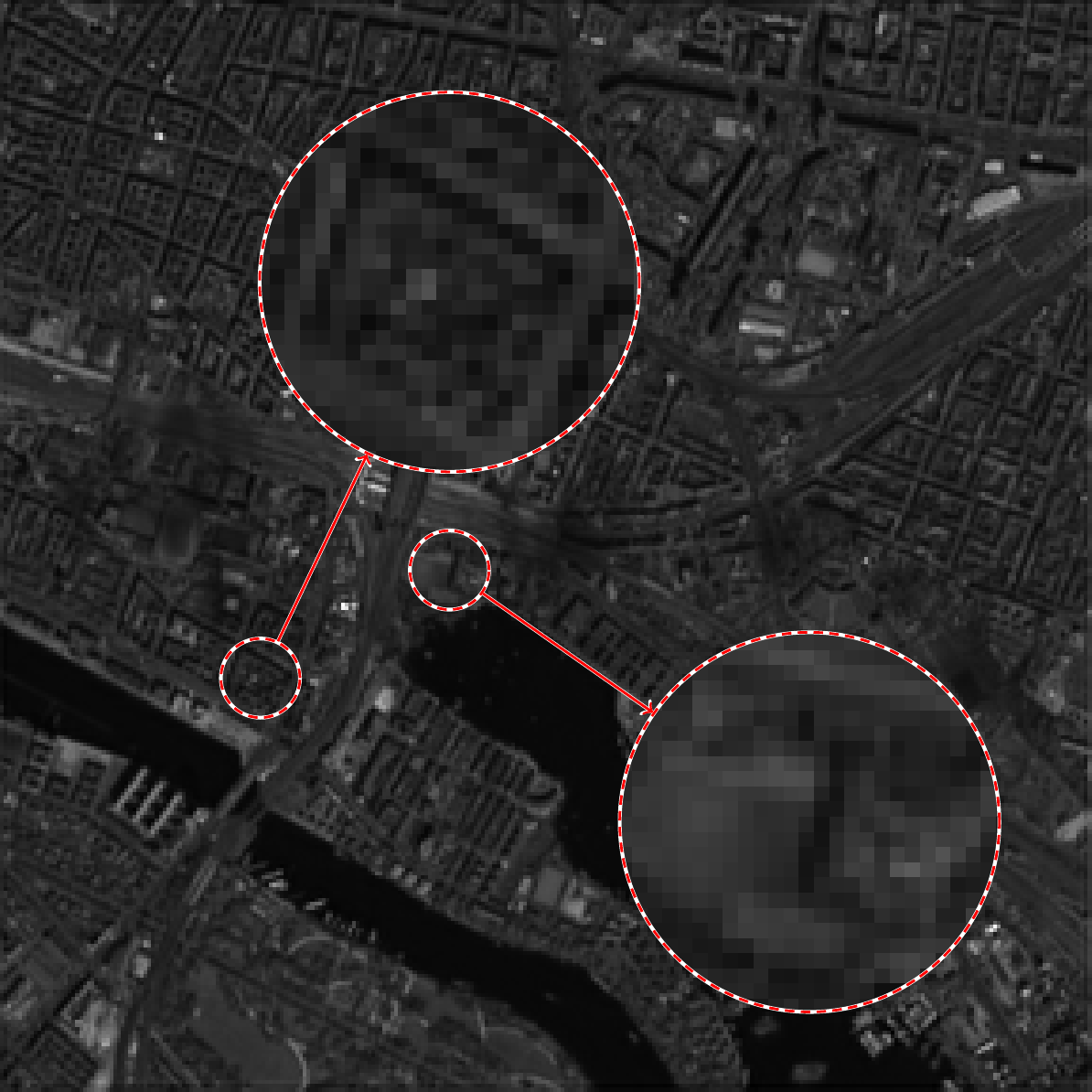} &
    \includegraphics[width=\mywidth\textwidth]{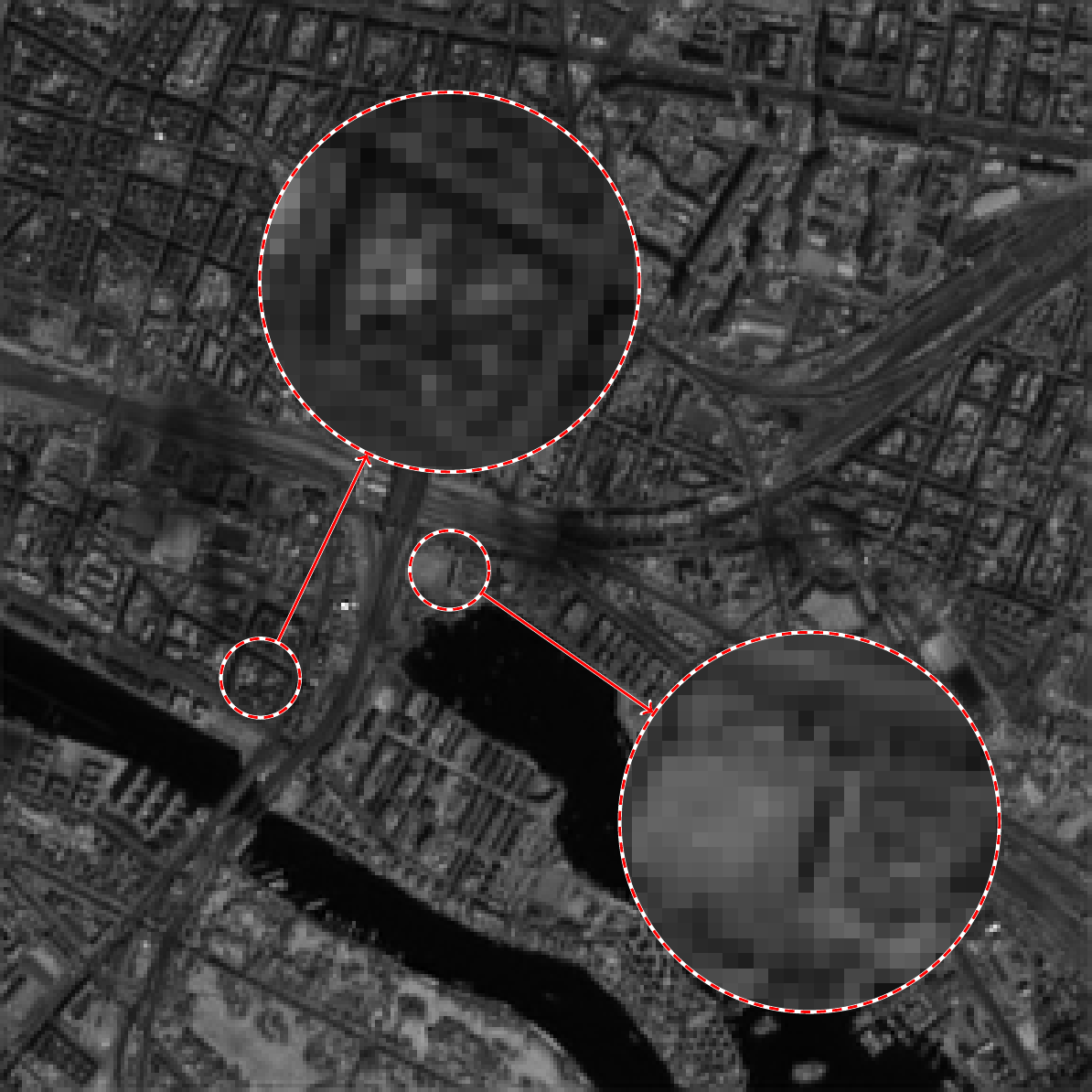} & & &
    \multicolumn{4}{c}{\cellcolor[HTML]{d6e2e9}} \\

    \raisebox{7mm}{\begin{turn}{90}DeepSent\end{turn}} & &
    & \includegraphics[width=\mywidth\textwidth]{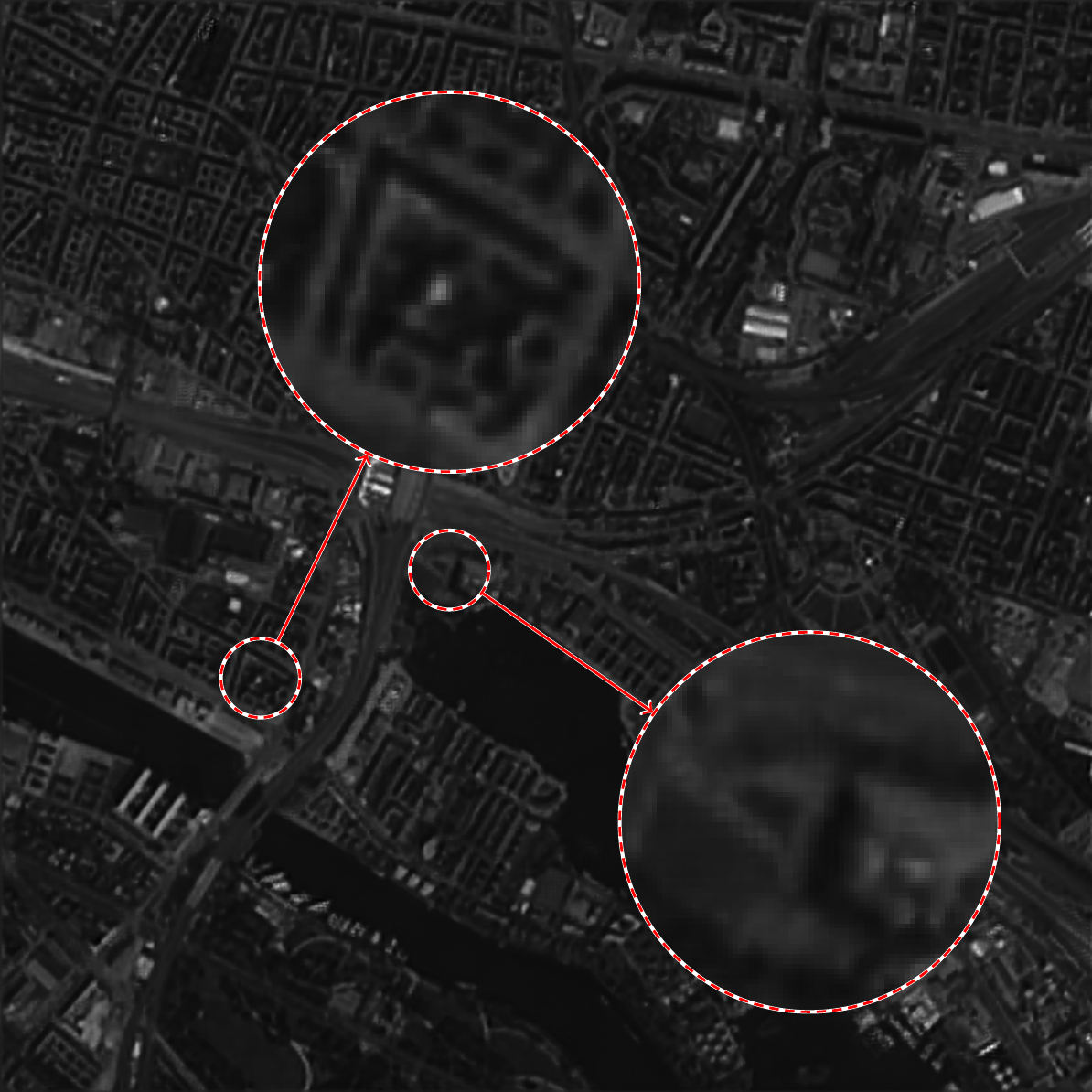} &
    \includegraphics[width=\mywidth\textwidth]{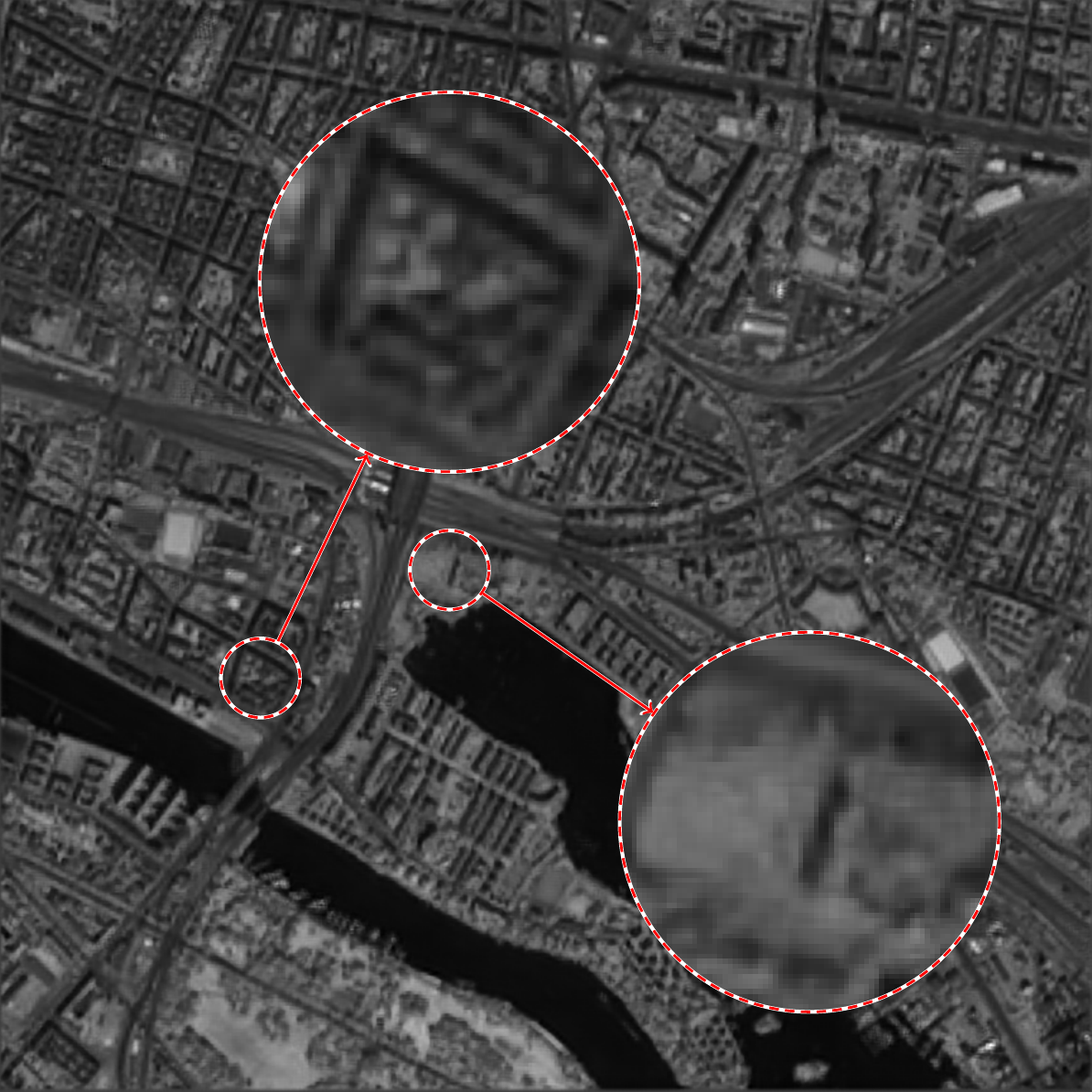} & & &
    & \includegraphics[width=\mywidth\textwidth]{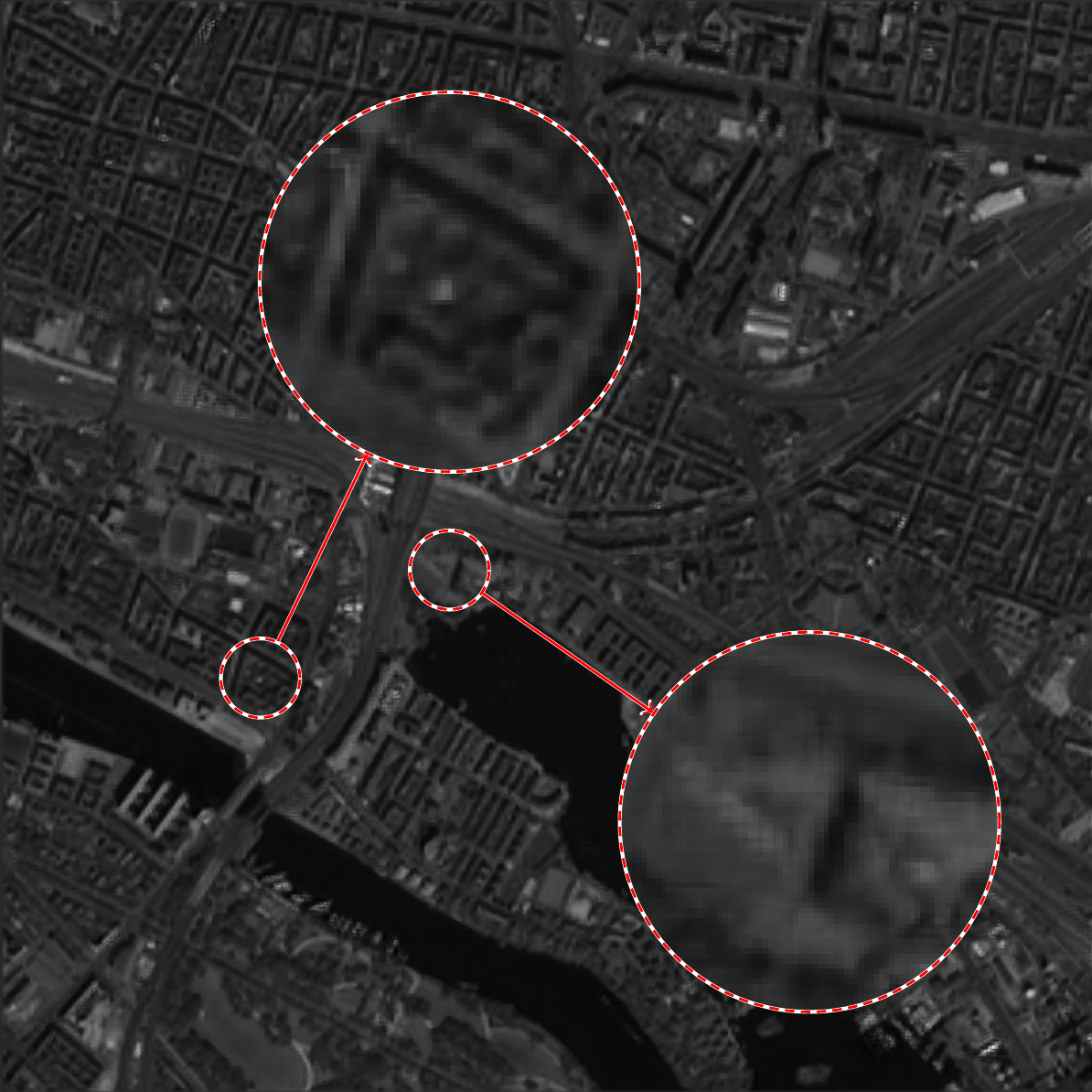} &
    \includegraphics[width=\mywidth\textwidth]{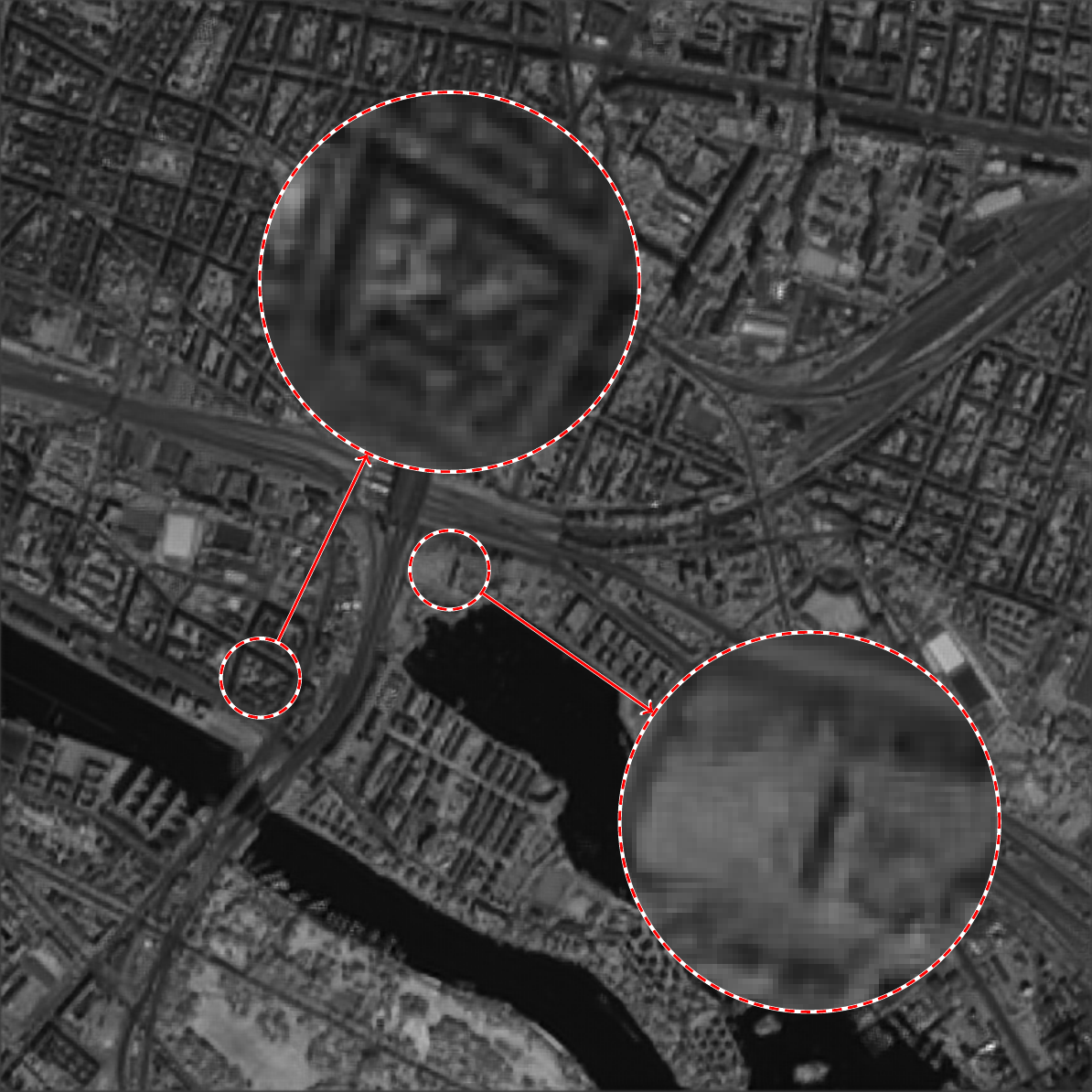} & & &
    & \includegraphics[width=\mywidth\textwidth]{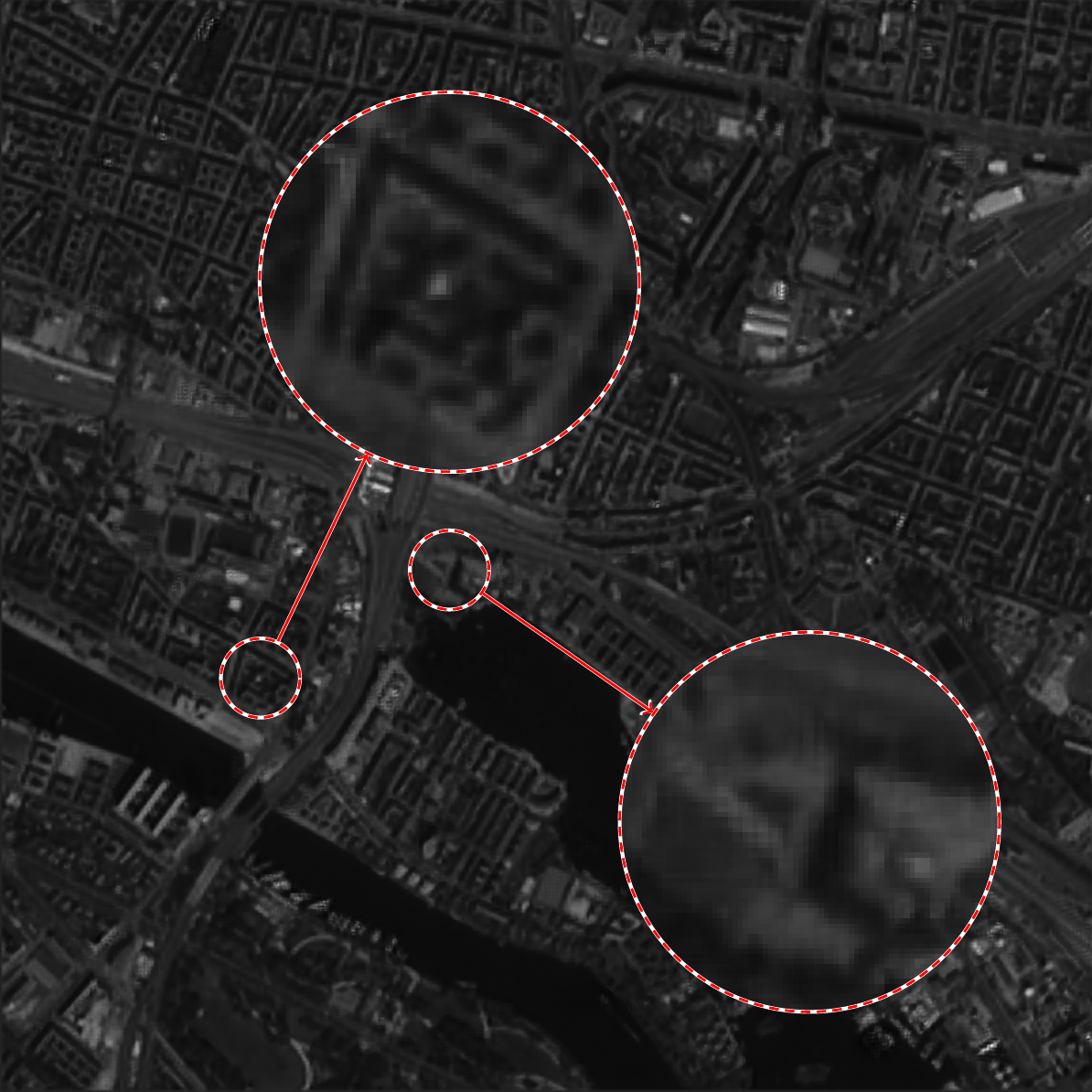} &
    \includegraphics[width=\mywidth\textwidth]{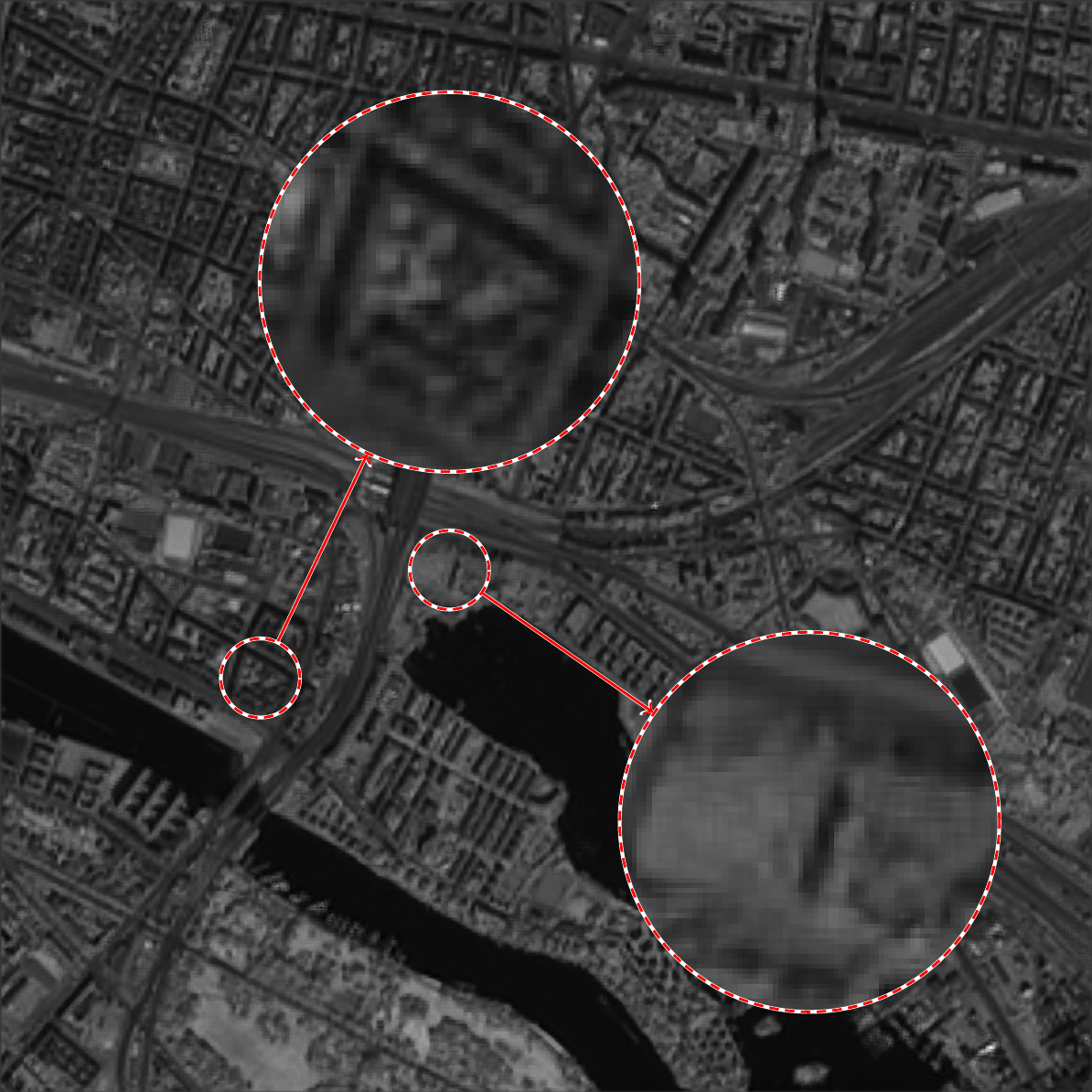} & \\[1mm]

\end{tabular}
\caption{Super-resolved images obtained from the real-world S-2 image stacks (from the \reals\, dataset). The input LR and reconstructed images are rendered for six selected bands of the same scene. DSen2 processes only 20\,m and 60\,m bands, so there is no outcome shown for the 10\,m bands.}
\label{fig:real_single_scene}
\end{figure*}

An example of the reconstruction outcome for an image from the \reals\, dataset is presented in Fig.~\ref{fig:real_single_scene}. Considering the 20\,m and 60\,m bands, \mymethod\, generates images with much more high-frequency details compared to other models, primarily due to its higher upsampling capabilities for these bands. For the 10\,m bands, the results generated by \mymethod\, are comparable with those produced by RAMS and HighRes-net models in terms of the visible details (like building shapes or streets), however the latter are heavily contaminated with high-frequency artifacts which affect the visual quality. This observation is confirmed with the naturalness image quality evaluator (NIQE)~\cite{Mittal2013} reported in Table~\ref{table:tab_niqe_real} (the lower, the better). We focus on the problem of the aforementioned artifacts later in this section. Importantly, \mymethod\, super-resolves the images for all the bands to the resolution of 3.3\,m (nominal) GSD, and it can be seen that their detail level is consistent due to the fusion in the spectral dimension, and visibly higher than that observed in the input LR images---the gain compared with the original 10\,m bands is attributed to the temporal fusion.

\begin{figure}[ht!]
    \centering
    \includegraphics[width=\columnwidth]{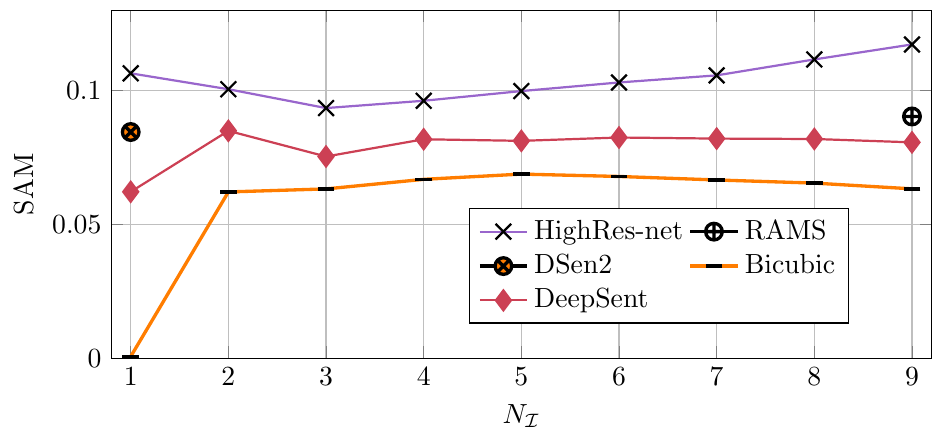}
    \caption{SAM scores depending on the number of input LR images for reconstruction performed from original S-2 images.}
    \label{fig:sam_var_lrs}
\end{figure}

Although for the \reals\, dataset we do not have any HR reference, we decided to verify whether the reconstruction preserves the radiometric characteristics. This is particularly important for the spectral fusion techniques (i.e., \mymethod\, and DSen2) to reassure that they do not copy the spectral properties of 10\,m bands while exploiting them to enlarge the bands of lower spatial resolution. To do so, we measure SAM between each set of output images and their corresponding LR counterparts. As the spectral characteristics vary between the input images captured at different times (mainly due to the vegetation), we select the input image, whose spectral distance to the reconstruction outcome is the smallest. Since SAM is calculated in a pixel-wise manner, it requires a constant size of all images within a stack. Thus, we downsample both input and output images to the resolution of the input 60\,m bands. This downsampling operation also serves as a naive registration by minimizing the sub-pixel shifts between super-resolved images of different bands. In Fig.~\ref{fig:sam_var_lrs}, we report how the number of input images affects the SAM scores for all investigated algorithms. We present the results obtained for the bicubic interpolation which can be treated as the SAM lower bound, as it does not modify the spectral information---the increase in SAM when more LR images are used results from the differences among them (the plot shows the difference between the averaged bicubically-upsampled image and the most similar input). It can be seen that the results retrieved with \mymethod\, are slightly worse than this lower-bound baseline, and it outperforms other models in terms of maintaining the spectral relations. Also, it preserves almost a constant value when more LR images are included into the input entry. 

\begin{figure}[ht!]
    \centering
    \footnotesize
    \renewcommand{\tabcolsep}{0.2mm}
    \renewcommand{\arraystretch}{0.5}
    \newcommand{\mywidth}{0.15}
    \begin{tabular}{cp{1.5mm}ccc}
    & & 2 LR images & 5 LR images & 9 LR images \\
    \multirow{2}{*}{a)} & &
    \includegraphics[width=\mywidth\textwidth]{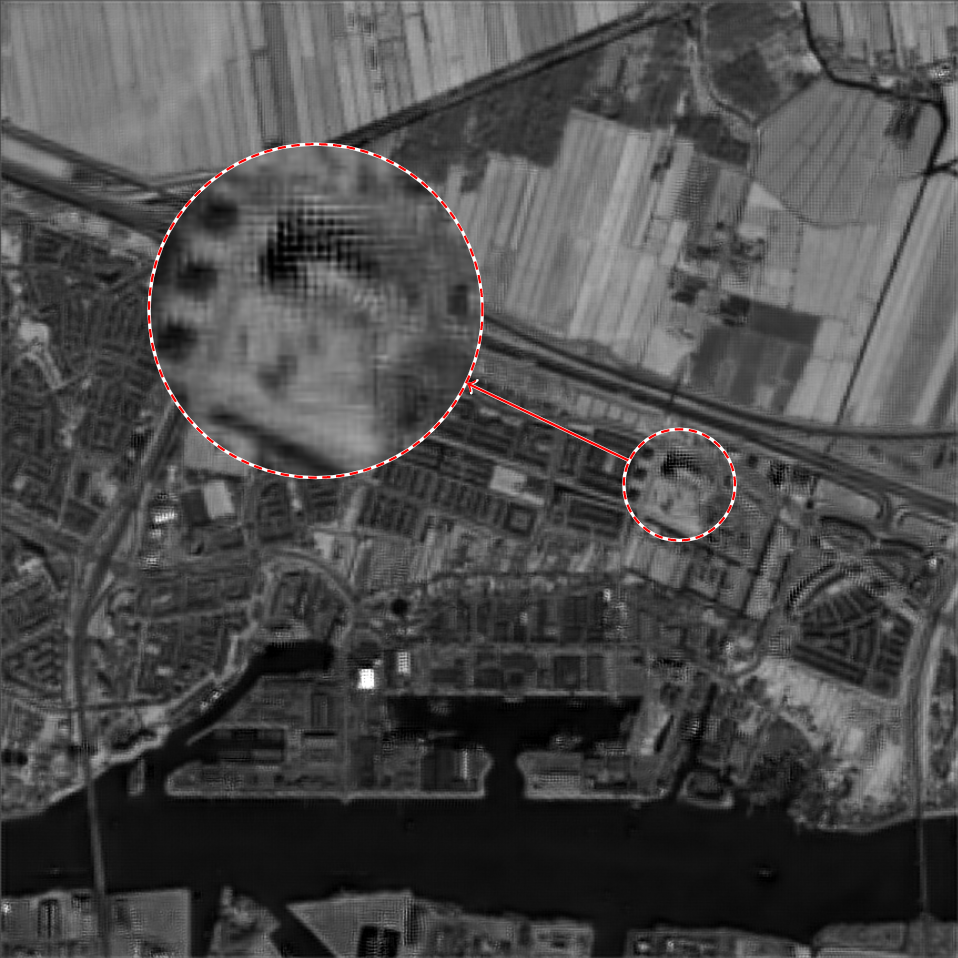} &
    \includegraphics[width=\mywidth\textwidth]{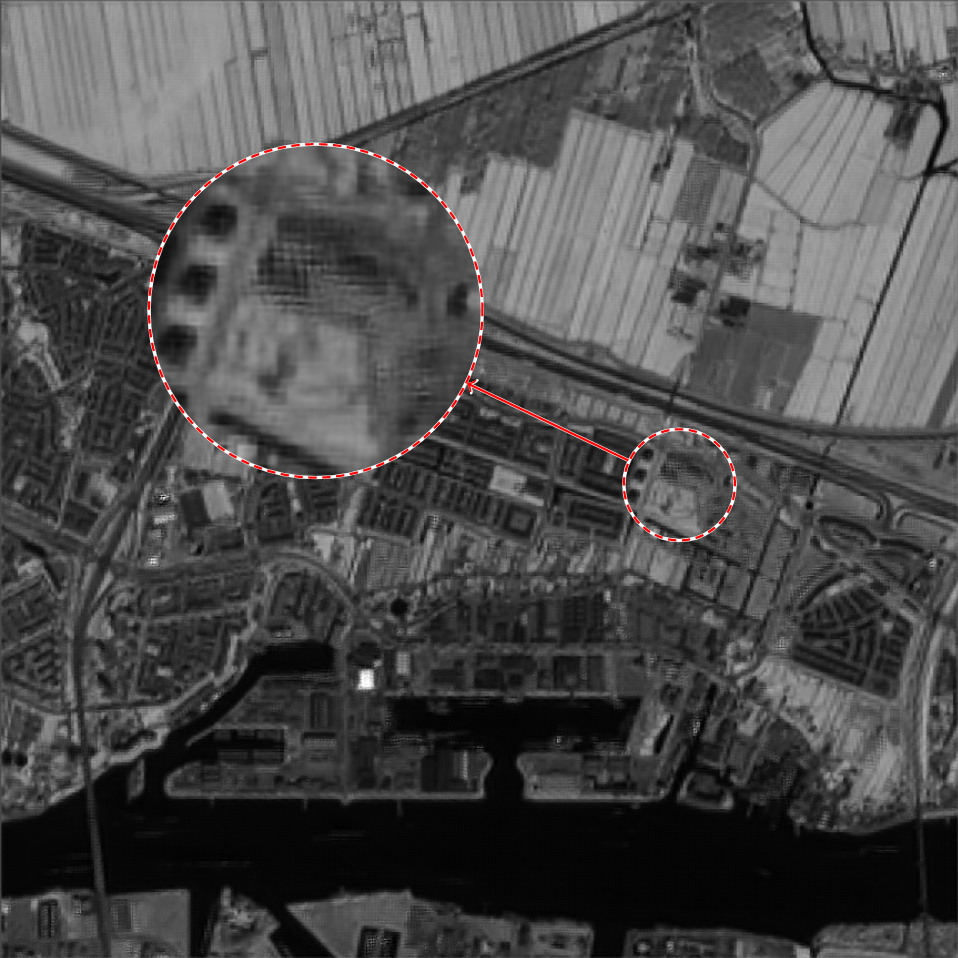} &
    \includegraphics[width=\mywidth\textwidth]{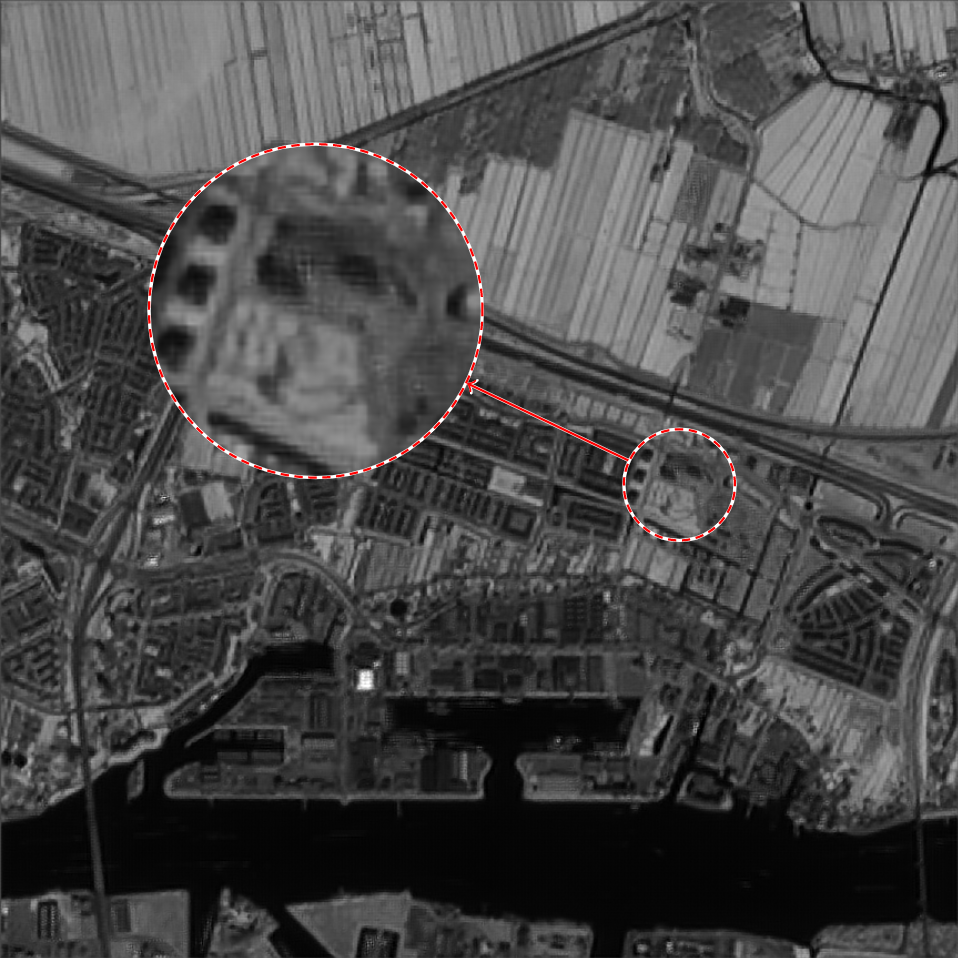} \\
    & & \includegraphics[width=\mywidth\textwidth]{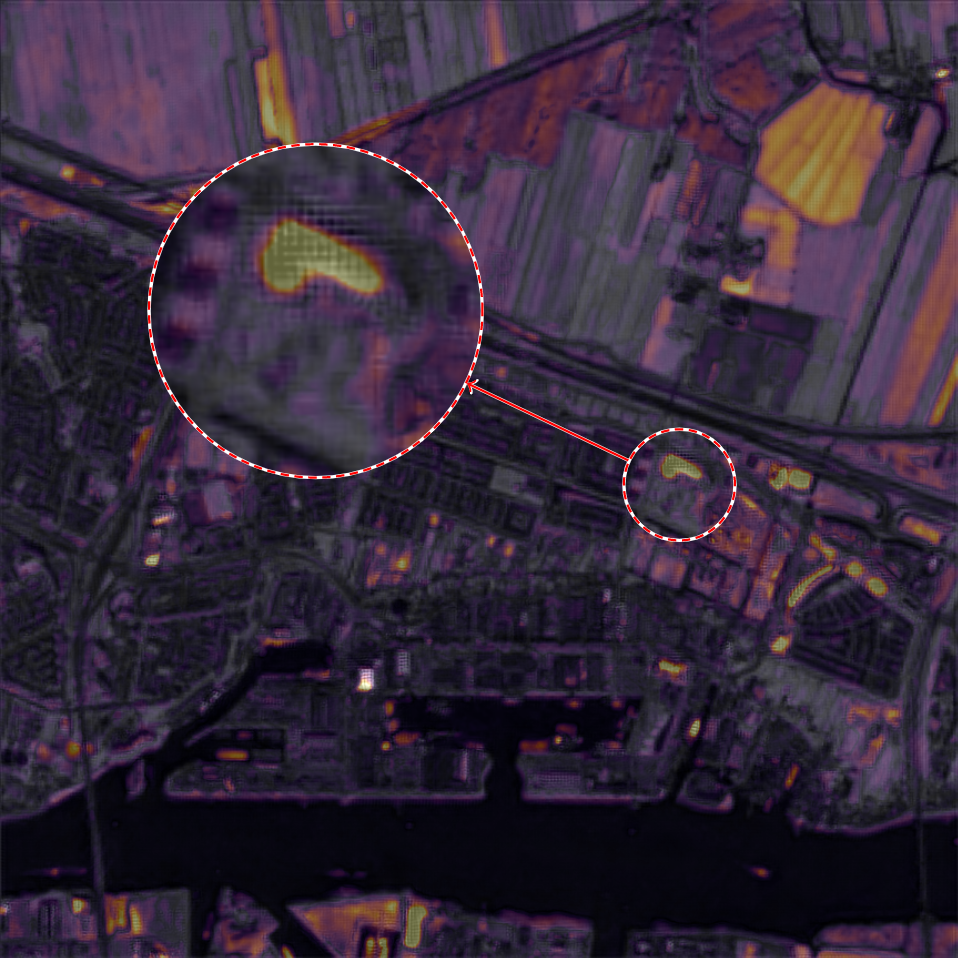} &
    \includegraphics[width=\mywidth\textwidth]{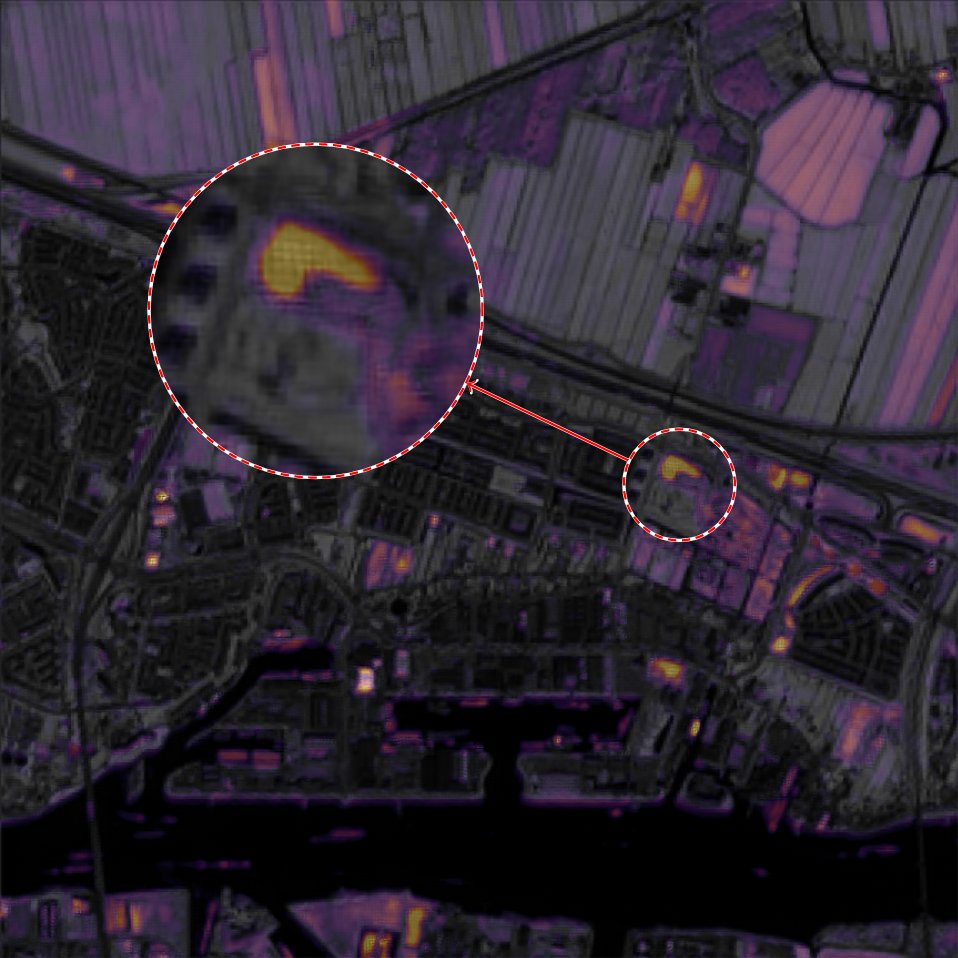} &
    \includegraphics[width=\mywidth\textwidth]{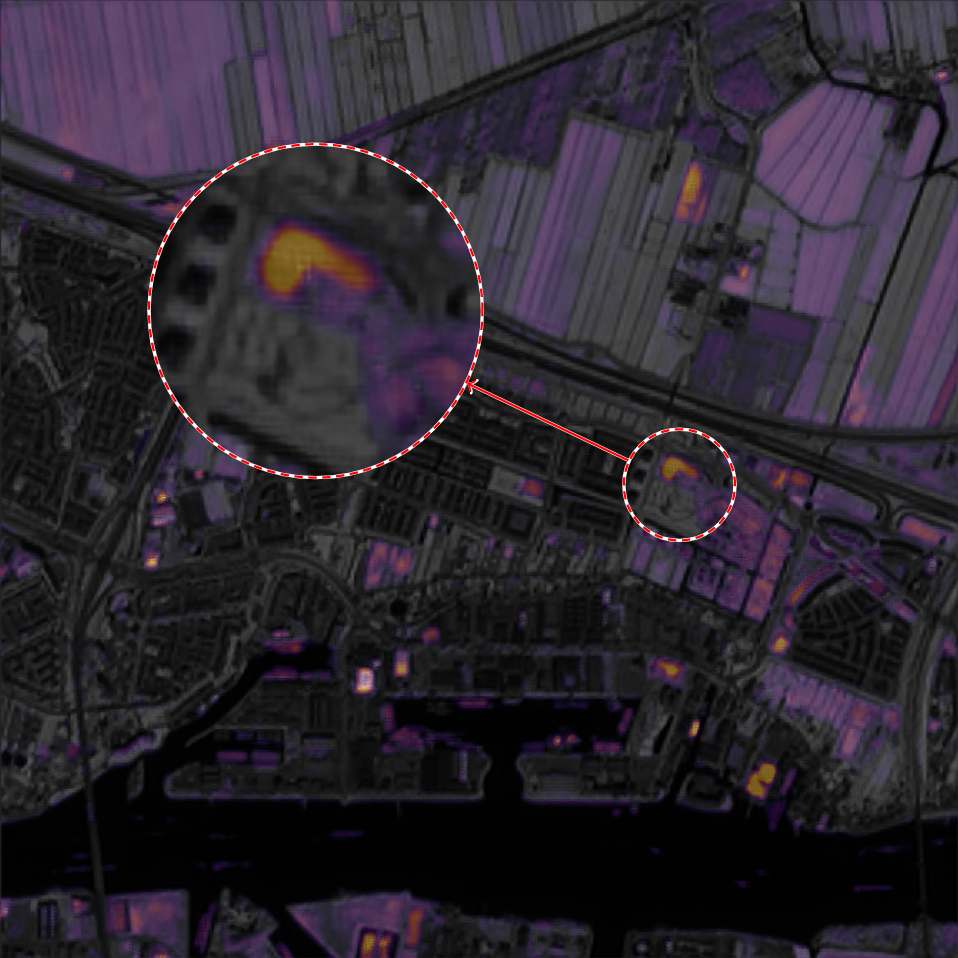} \\
    & & & & \rule{0pt}{0mm} \\
    \multirow{2}{*}{b)} & &
    \includegraphics[width=\mywidth\textwidth]{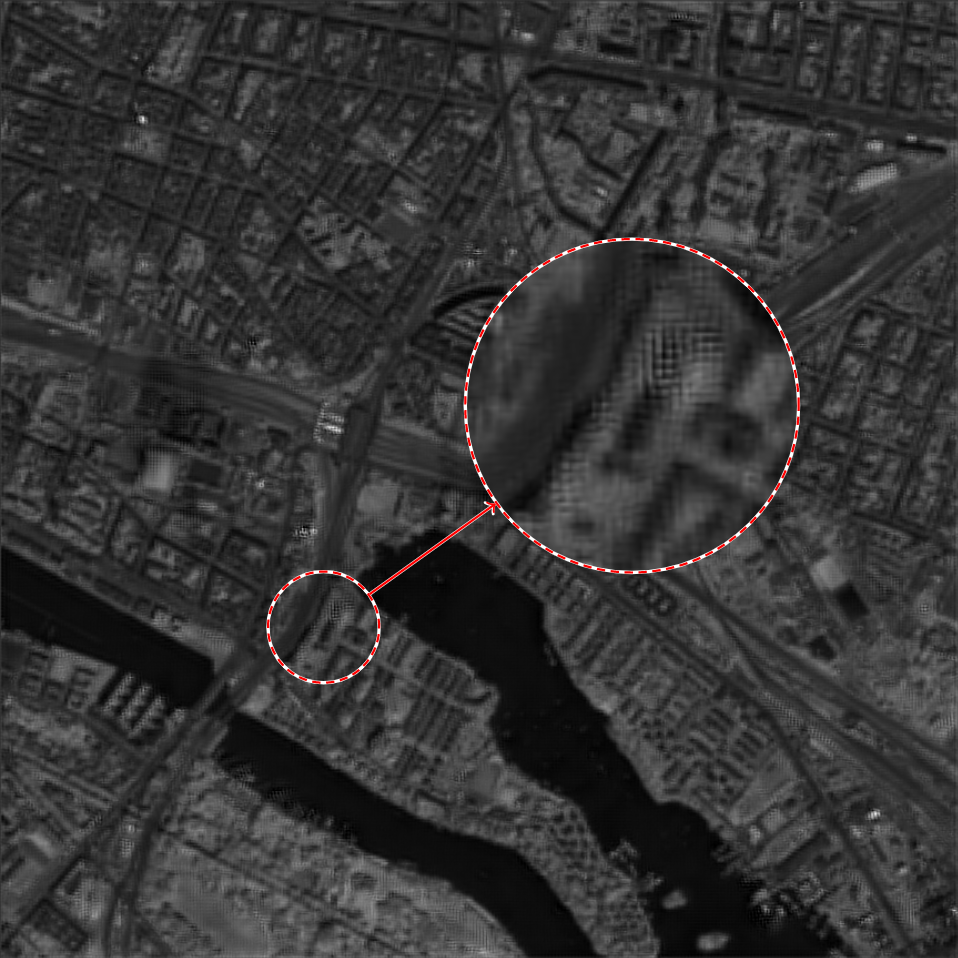} &
    \includegraphics[width=\mywidth\textwidth]{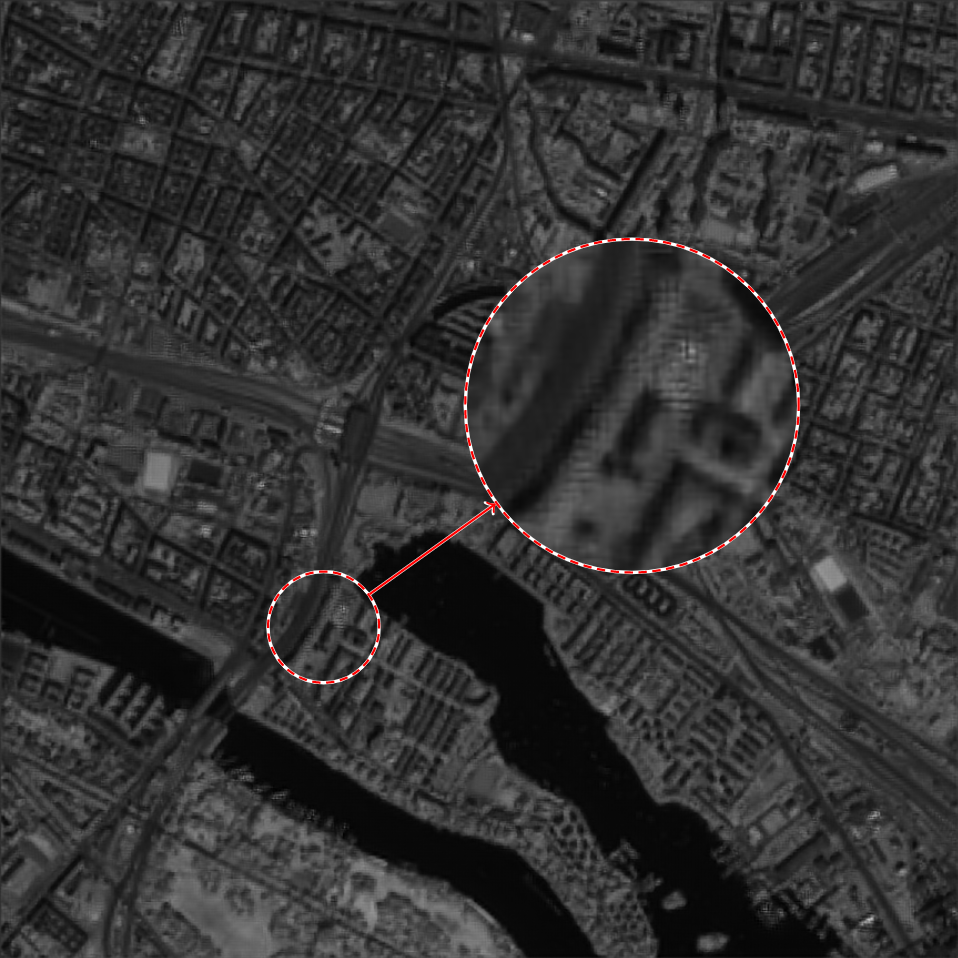} &
    \includegraphics[width=\mywidth\textwidth]{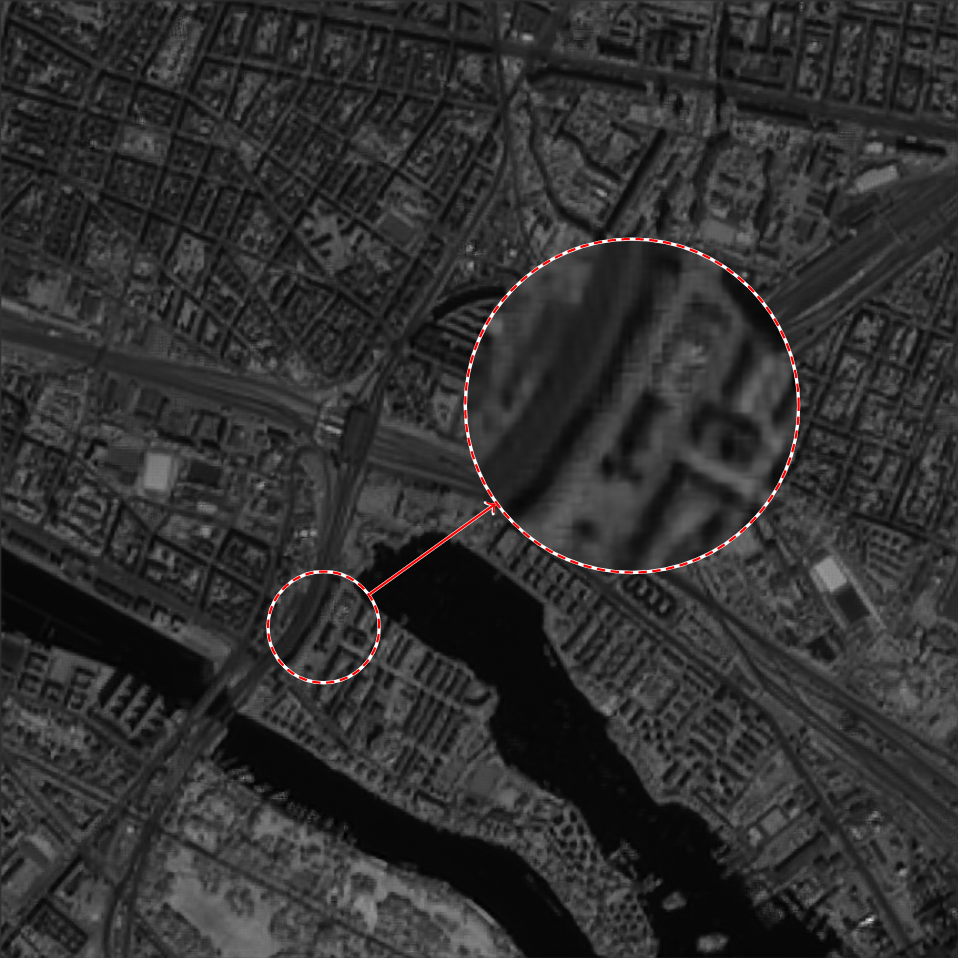} \\
    & & \includegraphics[width=\mywidth\textwidth]{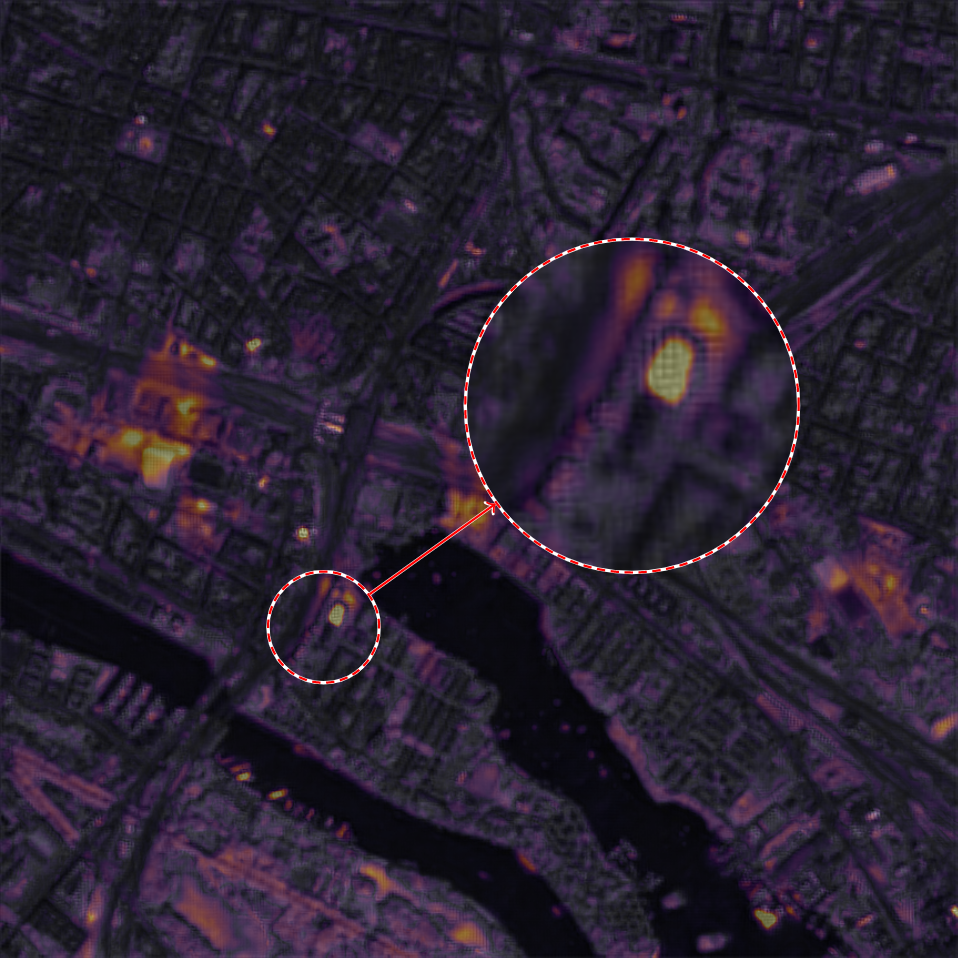} &
    \includegraphics[width=\mywidth\textwidth]{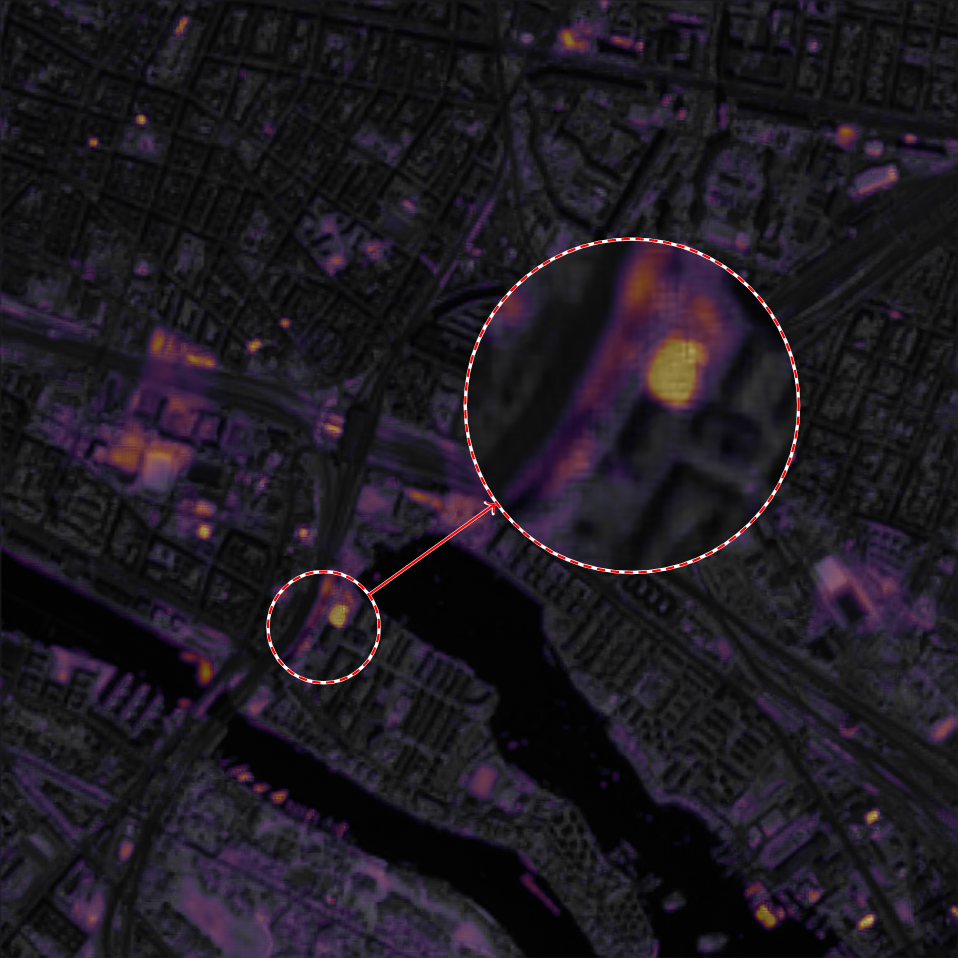} &
    \includegraphics[width=\mywidth\textwidth]{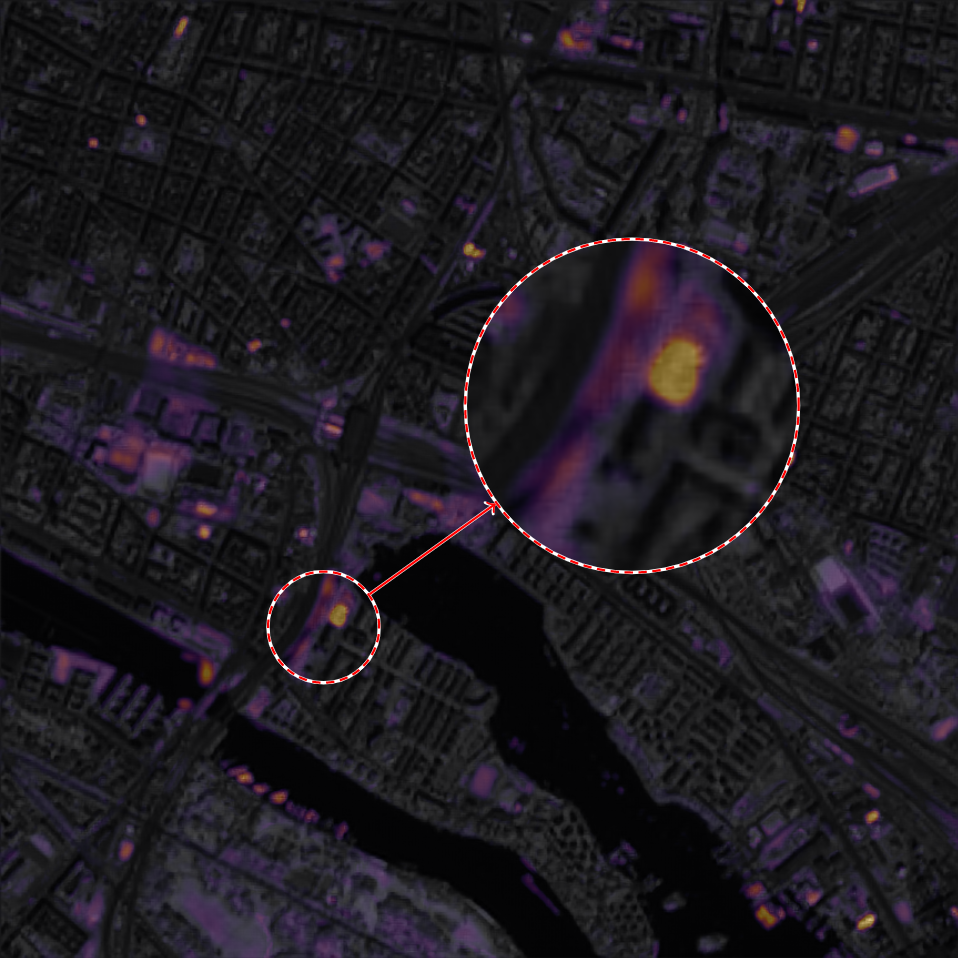} \\
    \end{tabular}
    \caption{Two scenes (a, b) super-resolved by \mymethod~from different number of real-world LR images (from \reals). The top rows for each scene show the reconstruction outcome, and the bottom rows depict the artifact heat-maps superimposed on the outcomes---the brighter (yellowish) pixels are supposed to indicate higher probability of the artifacts. It can be seen that the lower values in the heat-maps are observed for using more LR images, and they correspond well with the artifacts intensiveness observed in the top rows.}
    \label{fig:real_var_lrs}
\end{figure}

An important problem for real-world MISR lies in the high-frequency artifacts which are usually not observed for the simulated data. In Fig.~\ref{fig:real_single_scene}, we can observe significant differences in the brightness between the LR images and the corresponding reconstructed ones. In the case of MISR, they result from fusing information provided by multiple input images acquired under inherently different conditions, hence varying in terms of their brightness. Moreover, the time difference between each acquisition implies deviations in local information contained in multiple regions of the very same scene. As we noticed, such content-related variations have a significant negative impact on the reconstruction quality. The regions described by inconsistent local information across the time domain have visible artifacts after the SR reconstruction, mainly in the areas containing high-frequency details. To verify our observations, we reconstructed each real-world scene for a variable number of LR images, and qualitatively compared the results. Additionally, we created the heat-maps of estimated probability that some artifacts will occur, by calculating the standard deviation for each pixel in the stack of band-specific temporal images, and averaged them across all bands. From Fig.~\ref{fig:real_var_lrs}, depicting the SR results and superimposed heat-maps, we can see that such artifacts mainly appear in the areas of significant variance and propagate around them. However, as we add more images to the stack, the standard deviation of the local regions diminishes, and the artifacts become less visible. This shows the importance of how the input data are selected for reconstruction, and it explains less intensive artifacts observed for \mymethod\, than for other techniques, as they exploit less data for the reconstruction. 

\begin{table}[ht!]
\centering
\caption{The NIQE scores (the lower, the better) averaged across all real-world predictions and aggregated by resolution of the input bands. The best results are boldfaced and the second best are underlined.}
\renewcommand{\tabcolsep}{3mm}
\begin{tabular}{lrrr}
    \Xhline{2\arrayrulewidth}
    \multicolumn{1}{r}{Bands $\rightarrow$} &
      \multicolumn{1}{c}{60\,m} &
      \multicolumn{1}{c}{20\,m} &
      \multicolumn{1}{c}{10\,m}\\ \hline
    \multicolumn{1}{l}{LR images}             & 10.71  & 8.96  & 7.52 \\
    \multicolumn{1}{l}{RAMS}            & 10.64  & \underline{6.67}  & \underline{6.58} \\
    \multicolumn{1}{l}{HighRes-net}     & 9.23  & 7.38  & 6.92 \\
    \multicolumn{1}{l}{DSen2}           & \underline{7.50}  & 6.87  & --- \\
    \multicolumn{1}{l}{DeepSent} &
      \textbf{6.98} &
      \textbf{6.32} &
      \textbf{6.53} \\
\Xhline{2\arrayrulewidth}
\end{tabular}
\label{table:tab_niqe_real}
\end{table}

\subsection{Ablation study}
\label{ablation_study}

Our ablation study is primarily aimed at investigating the influence of the temporal fusion and spectral fusion on the final outcome.
\mymethod~architecture (Fig.~\ref{fig:architecture}) allows for selecting its forward paths dynamically based on the bands received at its input, as our recursive fusion blocks (at all fusion levels) do not produce any output, if no input is provided. Therefore, the forward pass may be effectively executed with specific parts of the network being switched on or off. Additionally, such recursive blocks support feeding the model with different numbers of LR images for each band independently. Given these flexibility features of \mymethod, the ablation study is focused on understanding the abilities of our architecture to exploit different fusion types and the range of the input data. 

\begin{figure}[ht!]
    \centering
    \includegraphics[width=\columnwidth]{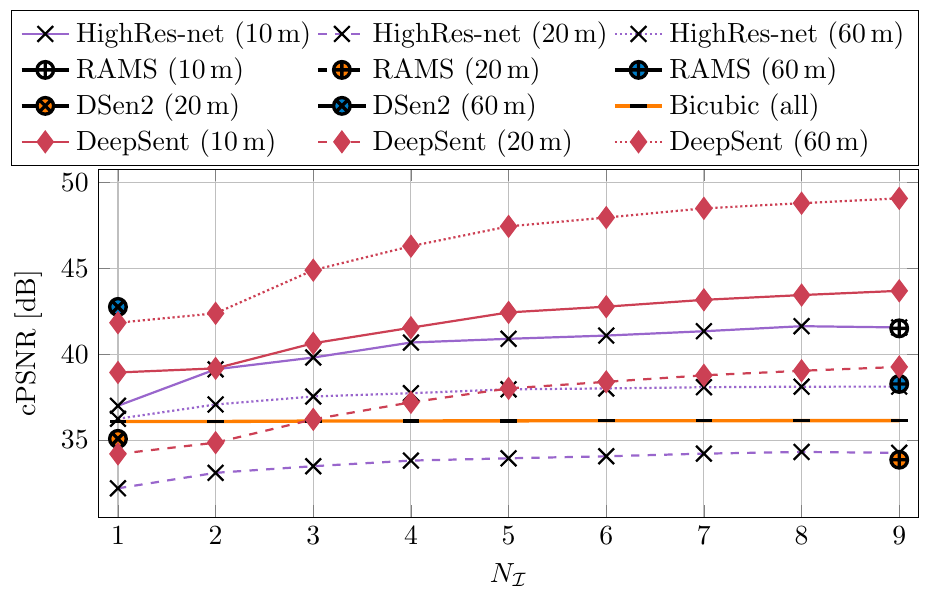}
    \includegraphics[width=\columnwidth]{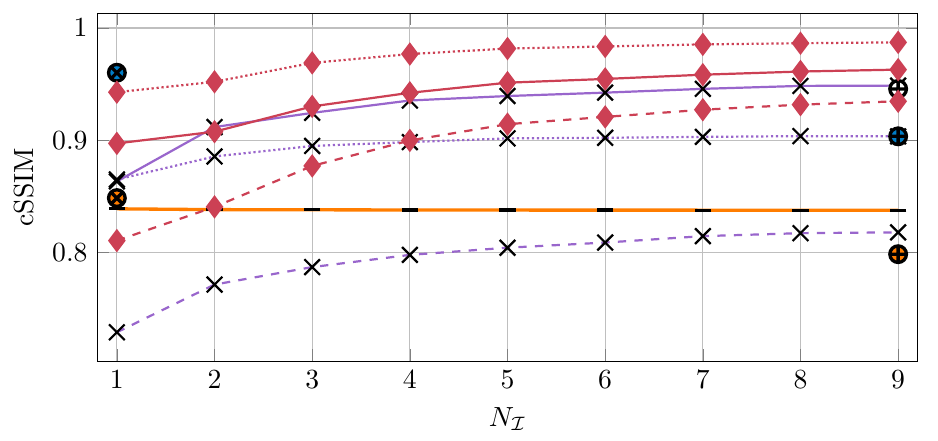}
    \caption{Reconstruction quality depending on the number of input LR images, obtained with different techniques for 10\,m, 20\,m, and 60\,m bands (for bicubic interpolation, we show an averaged score for all the bands, as it little depends on $\NumberOfLRs$ for all the bands). 
    }
    \label{fig:sim_var_lrs}
\end{figure}

To test the capabilities of our network to fuse multitemporal information contained in a stack of LR images, 
we calculated the quality metrics for the outcomes of SR performed from different number of LR images (from one to nine), for the \sims\, dataset. In Fig.~\ref{fig:sim_var_lrs}, we compare the results to other methods tested with the same approach---DSen2 and RAMS are visualized as single points, because they utilize the constant input stack size, of one and nine LR images, respectively. When SR is performed from a single image (hence without multitemporal data fusion), DSen2 retrieves higher scores (for 20\,m and 60\,m bands) than \mymethod, but for $\NumberOfLRs>2$, the temporal fusion performed with our method allows for better reconstruction---this is not the case for RAMS and HighRes-net which do not benefit from spectral fusion. We can appreciate that the metric values improve as we increase the number of input images for both \mymethod~and HighRes-net. However, our model benefits from it the most, as each additional LR image provides significantly more information contained in its spectral relations, which translates into a greater increase in the quality metrics.  In all cases, the improvements in both metrics (and all bands) are statistically significant for \mymethod.

To analyze the spectral fusion capabilities of \mymethod, we compared the reconstruction quality for the 10\,m bands executed based on five different groups of bands: (\textit{i})~a single 10\,m band that is to be reconstructed (thus without any spectral fusion), (\textit{ii})~all four of the 10\,m bands (spectral fusion of bands having the same spatial resolution), (\textit{iii})~10\,m bands with the 60\,m ones, (\textit{iv})~10\,m bands with the 20\,m ones, and finally (v)~all of the 12 S-2 bands.
The results are reported in Table~\ref{table:results_various_input_bands} and the scores distributions are shown in Fig.~\ref{fig:sim_results_ablation_plot}. They clearly demonstrate that the increase in the amount of spectral information enhances the scores (the improvements are statistically significant for all pairs of investigated configurations and for all bands). Interestingly, the information from the lower-resolution bands (20\,m and 60\,m) is beneficial for reconstructing the 10\,m bands, even though they contain less details than the four 10\,m bands.

\begin{table}[ht!]
\newcommand{\VarInputBands}{\bm{B}_{in}}
\centering
\caption{Reconstruction quality for the 10\,m bands obtained using \mymethod\, from various sets of input bands ($\VarInputBands$) and a constant number of nine temporal images per scene. The number of bands ($|\VarInputBands|$) is provided in brackets for each group. The best results are boldfaced and the second best are underlined.}\label{table:results_various_input_bands}
\resizebox{\columnwidth}{!}{
\renewcommand{\tabcolsep}{1mm}
\begin{tabular}{llllllllll}
\Xhline{2\arrayrulewidth}
                \multicolumn{1}{r}{Metric$\rightarrow$} & \multicolumn{4}{c}{cPSNR} & & \multicolumn{4}{c}{cSSIM}\\
               \cline{2-5} \cline{7-10}
\multicolumn{1}{l}{Input $\VarInputBands$ } $|\VarInputBands|$~&
\multicolumn{1}{c}{B02} &
\multicolumn{1}{c}{B03} &
\multicolumn{1}{c}{B04} &
\multicolumn{1}{c}{B08} &~&
\multicolumn{1}{c}{B02} &
\multicolumn{1}{c}{B03} &
\multicolumn{1}{c}{B04} &
\multicolumn{1}{c}{B08} \\
\Xhline{2\arrayrulewidth}
\multicolumn{1}{l}{Single band} (1)   & 42.83 & 42.00 & 40.30 & 35.88 & & 0.9412 & 0.9308 & 0.9293 & 0.8854 \\
\multicolumn{1}{l}{$\BandGroup_{10}$} (4)      & 44.90 & 43.39 & 41.02 & 36.44 & & 0.9574 & 0.9435 & 0.9433 & 0.9049 \\
\multicolumn{1}{l}{$\BandGroup_{10}, \BandGroup_{60}$} (6)  & 45.67 & 44.52 & 41.64 & 37.10 & & 0.9745 & 0.9705 & 0.9540 & \underline{0.9141} \\
\multicolumn{1}{l}{$\BandGroup_{10}, \BandGroup_{20}$} (10) & \underline{46.06} & \underline{44.99} & \underline{42.11} & \underline{37.20} & & \underline{0.9760} & \underline{0.9727} & \underline{0.9577} & {0.9138} \\
\multicolumn{1}{l}{$\BandGroup_{10}, \BandGroup_{20}, \BandGroup_{60}$} (12) &
  \textbf{47.11} &
  \textbf{46.02} &
  \textbf{43.06} &
  \textbf{37.98} & &
  \textbf{0.9810} &
  \textbf{0.9782} &
  \textbf{0.9650} &
  \textbf{0.9264} \\
\Xhline{2\arrayrulewidth}
\end{tabular}
}
\end{table}

\begin{figure}[ht!]
\centering
\renewcommand{\tabcolsep}{0mm}
\begin{tabular}{c}
\scriptsize cPSNR \\
\includegraphics[width=0.48\textwidth]{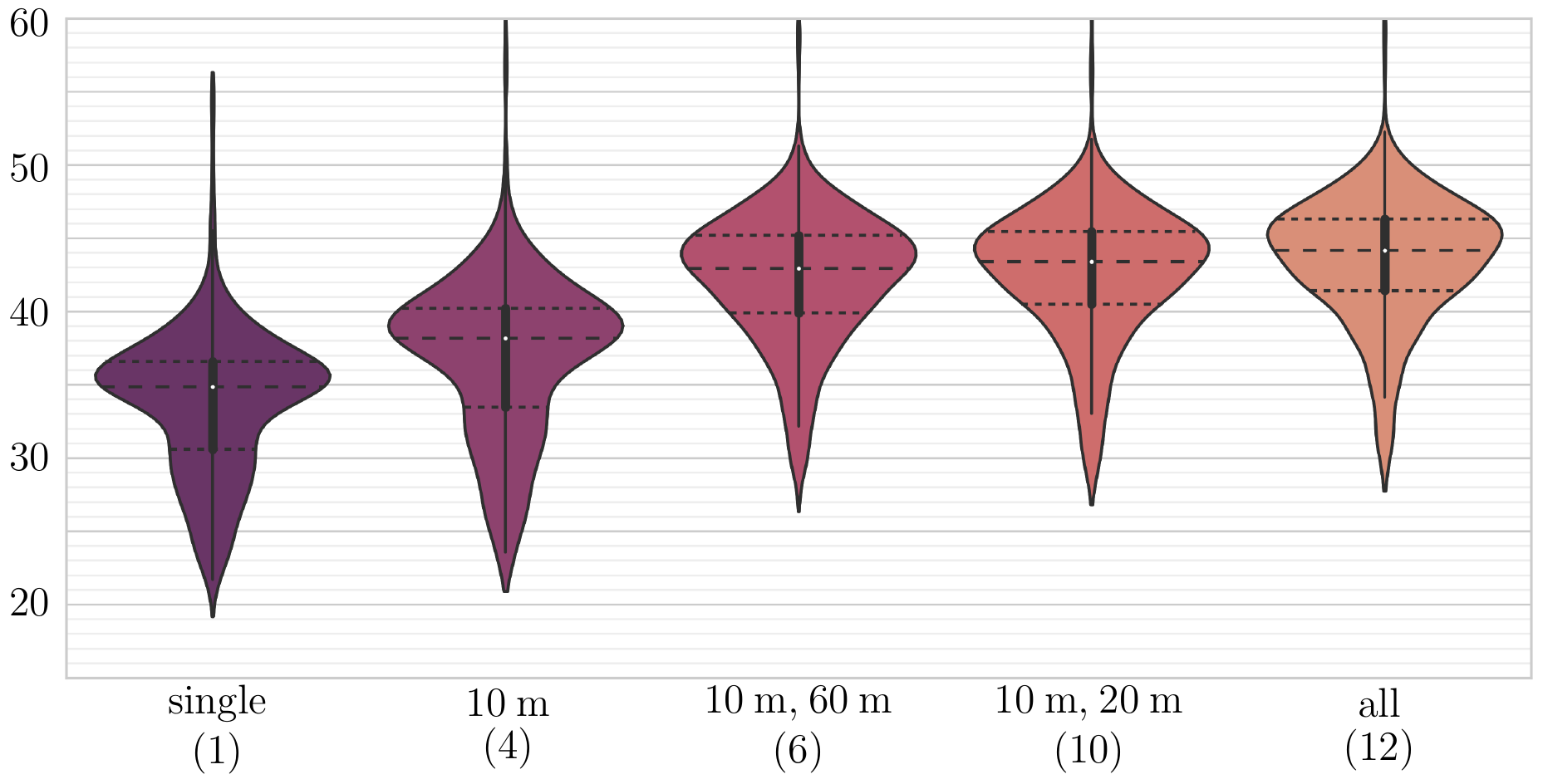} \\
\rule{0pt}{5mm}
\scriptsize cSSIM\\
\includegraphics[width=0.48\textwidth]{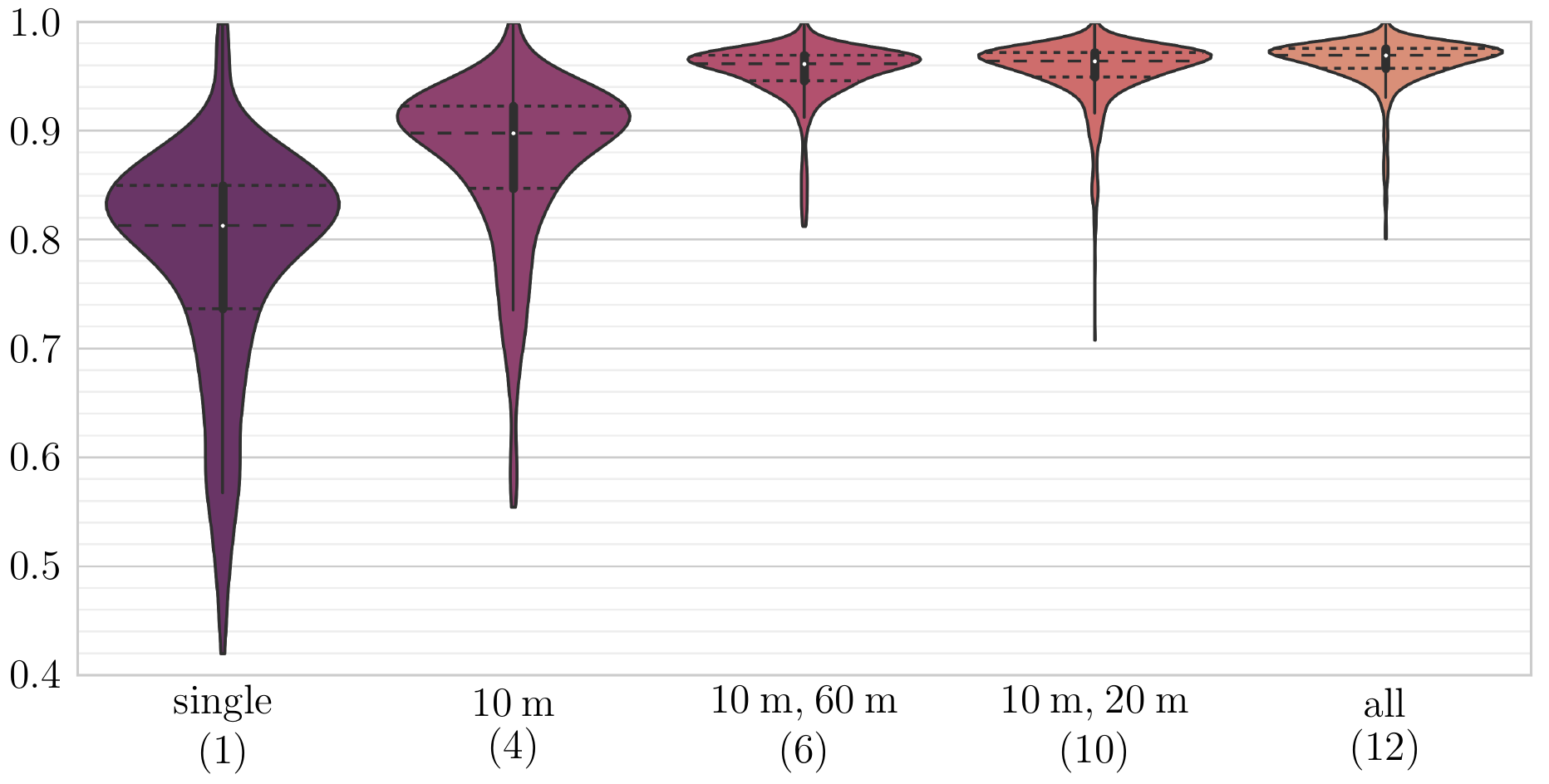} \\
\end{tabular}

\caption{The distribution of the quality metrics obtained over the test set for 10\,m bands with different combinations of input bands. The values reported in parentheses indicate the number of provided bands. The dashed lines
indicate the median and quantile values.}
\label{fig:sim_results_ablation_plot}
\end{figure}

\section{Conclusions}
In this paper, we introduced \mymethod---a new network designed for super-resolving  multispectral S-2 images that allows for information fusion in the spectral and temporal dimensions. Our main contribution consists in combining the information provided by multiple images captured at different times along with multiple spectral bands. To the best of our knowledge, such collation of temporal and spectral information fusion has not been reported in the field of SR so far. A unique feature of \mymethod\, is that a single model is capable of super-resolving all of the S-2 bands to the uniform resolution of 3.3\,m nominal GSD, while preserving the spectral relations between the bands. 
Quantitative and qualitative results for both the simulated data, as well as for the original Level-2A S-2 images, clearly indicate that \mymethod\, outperforms other state-of-the-art solutions which incorporate only one type of fusion (i.e., RAMS and HighRes-net for temporal, and DSen2 for spectral fusion). The reconstruction quality is also higher than when the single-fusion techniques are combined in a sequential manner (i.e., DSen2 followed by RAMS or HighRes-net).  In the case of super-resolving real-world data, we have demonstrated that the outcome retrieved by \mymethod\, is characterized with much less artifacts than the outcomes obtained with other state-of-the-art methods. We have also provided evidence that the artifacts result from inconsistent local information contained in an image time series, which may be a helpful hint for other researchers working in this field.

The results reported in this paper constitute an exciting departure point for further research. First, even though \mymethod\, allows for reconstructing all the spectral bands using the same earlier-trained model, they have to be retrieved one after another by passing appropriate input band through the residual connection---reconstructing the whole MSI in one go would be much more convenient, however it is challenging due to memory issues. Furthermore, the residual connection transmits an averaged image of the whole temporal series, making it impossible to select a particular image in the series that should be reconstructed. Even though the super-resolved images preserve the spectral characteristics, the ability to pick a single image in the series that should be reconstructed using the remaining images in that series would be an extremely useful feature of high practical importance. Finally, although \mymethod\, offers a lower level of artifacts than other methods when fed with original S-2 images, this remains an open research problem worthwhile being addressed in the future. One of the possibilities is to focus on developing more advanced methods of generating simulated datasets for training that would better reflect the real-world data characteristics. Another possibility would be to enhance the recursive fusion blocks with attention mechanism to select most valuable features that do not lead to visual artifacts. Last, but not least, the proposed framework could be exploited for solving specific image analysis tasks which require higher spatial resolution than that of the original S-2 products. This would also help in determining future research pathways that could lead to novel practical applications of SR.

\section*{Acknowledgment}

This research was supported by the National Science Centre, Poland, under Research Grant 2019/35/B/ST6/03006 (TT, JN, MK). The work was partially funded by European Space Agency (DeepSent project).

This work was supported within Project WARM-EO financed by the Malta Council for Science \& Technology, for and on behalf of the Foundation for Science and Technology, through the Space Research Fund.

TT benefits from the European Union scholarship through the European Social Fund (grant POWR.03.05.00-00-Z305). MK was supported by the SUT funds through the Rector’s Research and Development Grant 02/080/RGJ22/0024.

\ifCLASSOPTIONcaptionsoff
  \newpage
\fi

\begin{IEEEbiography}[{\includegraphics[width=1in,height=1.25in,clip,keepaspectratio]{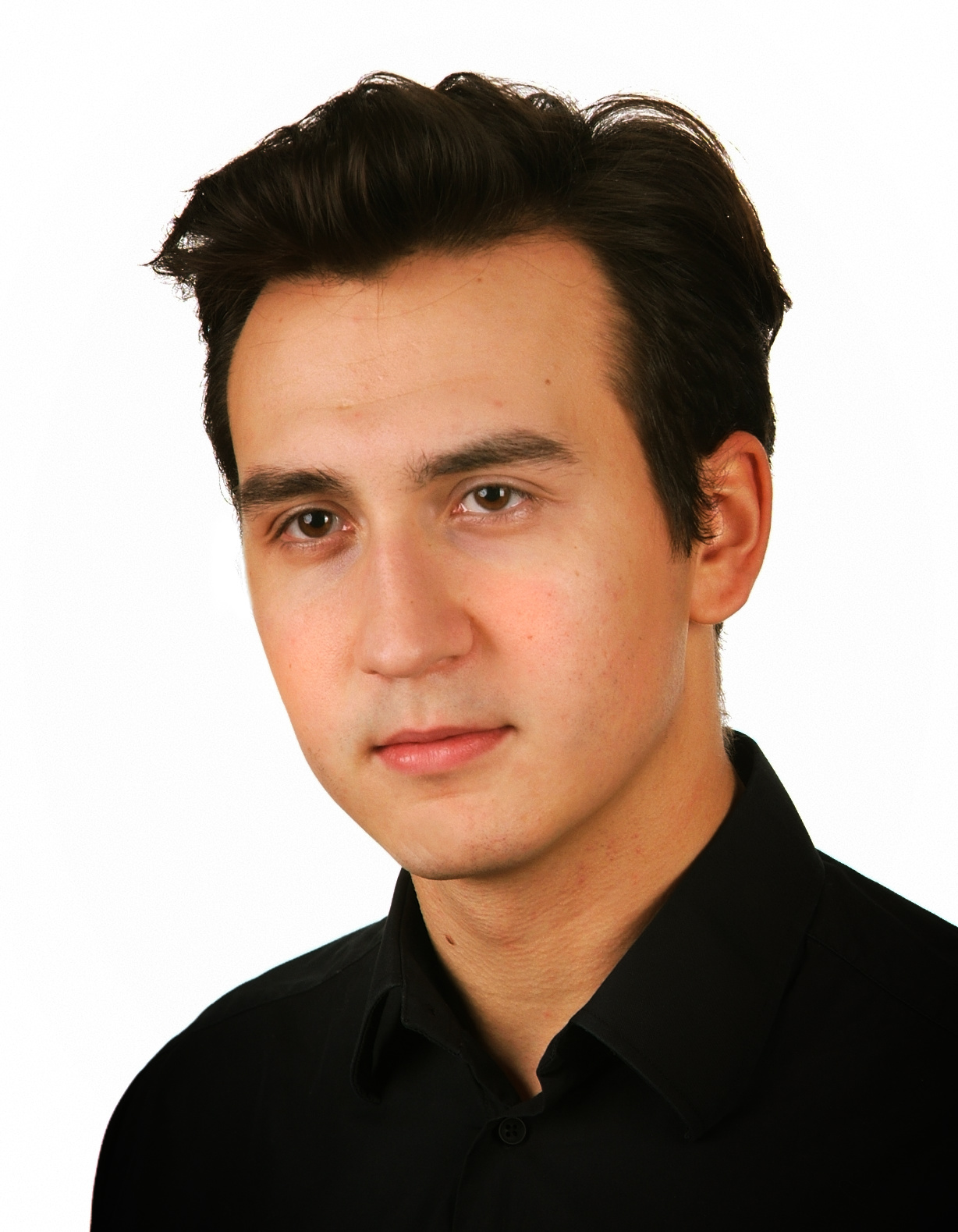}}]{Tomasz Tarasiewicz} is a PhD student at the Silesian University of Technology, where he earned his MSc (2019). He is involved in research on deep learning solutions for problems such as human skin segmentation, and super-resolution reconstruction with an emphasis on remote sensing applications. To date, the results of his research have been published in five peer-reviewed conference proceedings.
\end{IEEEbiography}


\begin{IEEEbiography}[{\includegraphics[width=1in,height=1.25in,clip,keepaspectratio]{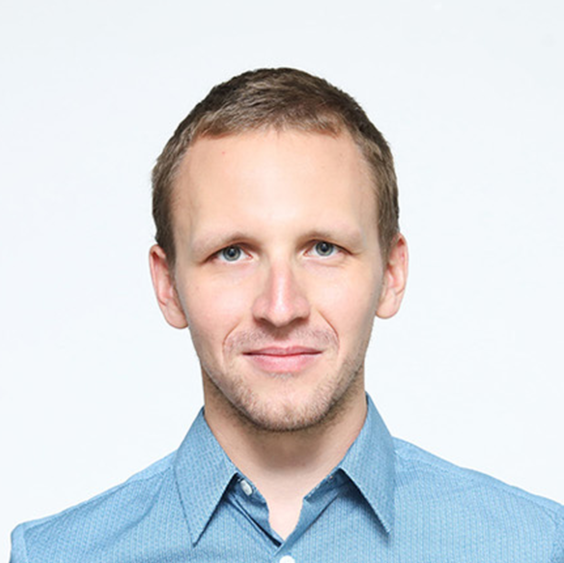}}]{Jakub Nalepa}
received his MSc (2011), PhD (2016), and DSc (2021) in Computer Science from Silesian University of Technology, Gliwice, Poland, where he is currently an Associate Professor. Jakub is the Head of AI at KP Labs where he shapes the scientific and industrial AI objectives of the company related to, among others, Earth observation, on-board and on-the-ground satellite data analysis, and anomaly detection from the satellite telemetry data. He has been pivotal in designing the on-board deep learning capabilities of Intuition-1, and has contributed to other missions, including CHIME and OPS-SAT. His research interests focus on (deep) machine learning, hyperspectral data analysis, signal processing, remote sensing, and tackling practical challenges which arise in Earth observation to deploy scalable EO solutions.

\end{IEEEbiography}


\begin{IEEEbiography}[{\includegraphics[width=1in,height=1.25in,clip,keepaspectratio]{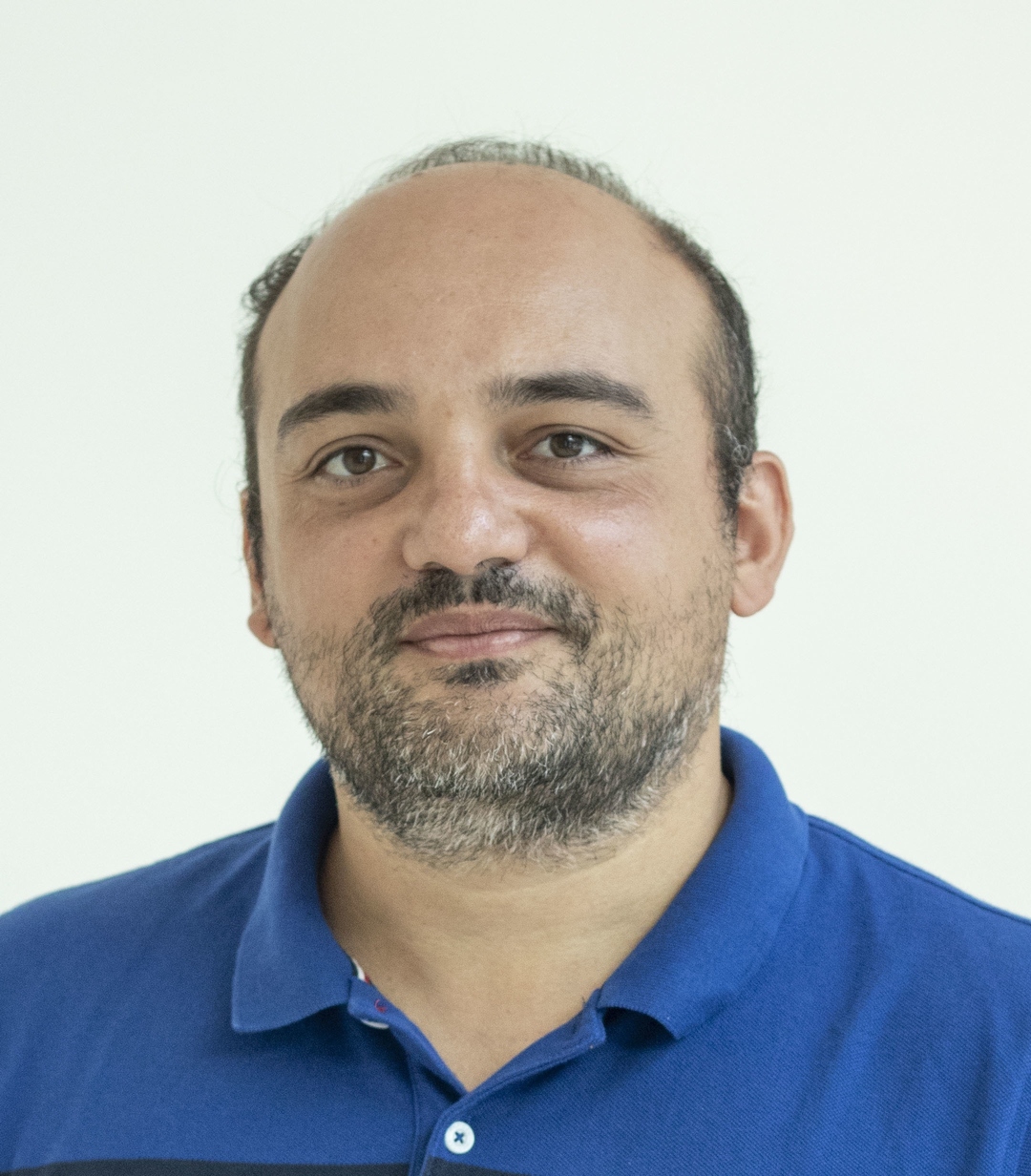}}]{Reuben A. Farrugia}
(Senior Member, IEEE) is an Associate Professor at the University of Malta and received his Ph.D. degree in 2009. He has been working in the fields of image processing and pattern recognition with the University of Malta since 2004 with a specific focus on image processing and computer vision. He has published more than 80 international peer-reviewed publications, and has been involved in technical and organizational committees of several national and international conferences. Prof. Farrugia is an Area Editor of the Signal Processing: Image Communications journal (Elsevier). In September 2013, he was appointed as the National Contact Point of the European Association of Biometrics (EAB).
\end{IEEEbiography}

\begin{IEEEbiography}[{\includegraphics[width=1in,height=1.25in,clip,keepaspectratio]{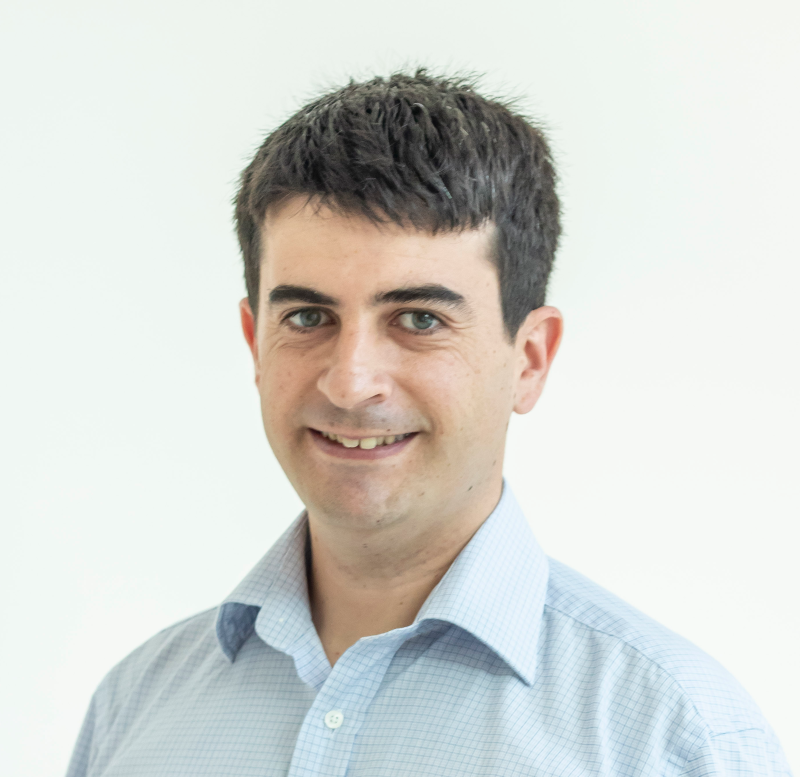}}]{Gianluca Valentino}
Gianluca Valentino received his BSc (2010) and PhD (2013) in Computer Engineering from the Faculty of ICT of the University of Malta, where he is currently a Senior Lecturer. He is also a Visiting Scientist at CERN and a Senior Member of the IEEE. To date, he has published over 100 peer-reviewed papers in journals and conferences. His research interests include machine learning, computer vision as well as their applications to remote sensing and high-energy physics.
\end{IEEEbiography}

\begin{IEEEbiography}[{\includegraphics[width=1in,height=1.25in,clip,keepaspectratio]{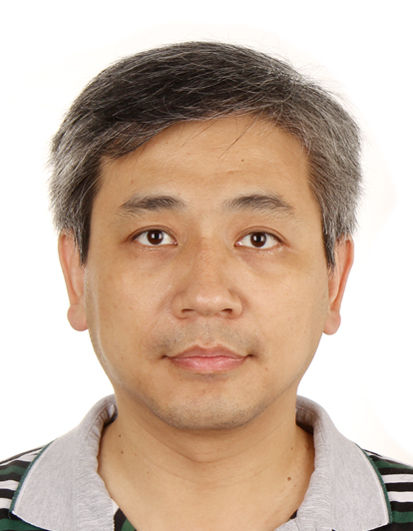}}]{Mang Chen} received his PhD degree in image processing and pattern recognition from the Shanghai Jiaotong University (SJTU), China, in 2006. Since then, he had joined Sharp Labs China (SLC) and had participated in the research and development of OCR, font-CAD, attention statistics and augmented reality. Since 2020, as a member of Data Science Research Group at the University of Malta, he has been working in the fields of computer vision and image restoration for Earth Observation.
\end{IEEEbiography}

\begin{IEEEbiography}[{\includegraphics[width=1in,height=1.25in,clip,keepaspectratio]{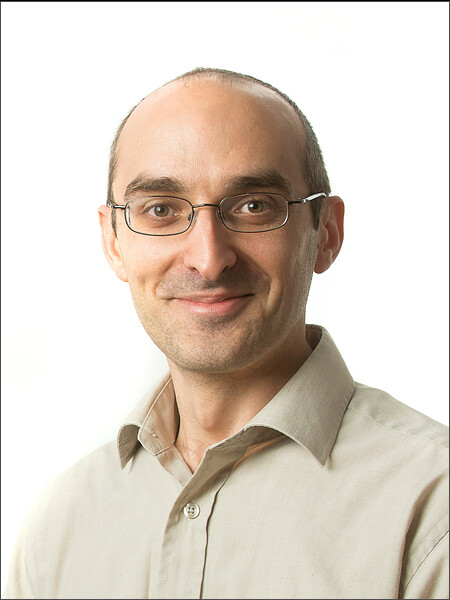}}]{Johann A. Briffa} is an associate professor and head of the Department of
Communications and Computer Engineering at the University of Malta. He obtained
his PhD in Systems Engineering from Oakland University, Rochester MI in December
2003. His research interests include information theory and image processing,
most recently applied to quantum key distribution and remote sensing
respectively.
\end{IEEEbiography}

\begin{IEEEbiography}[{\includegraphics[width=1in,height=1.25in,clip,keepaspectratio]{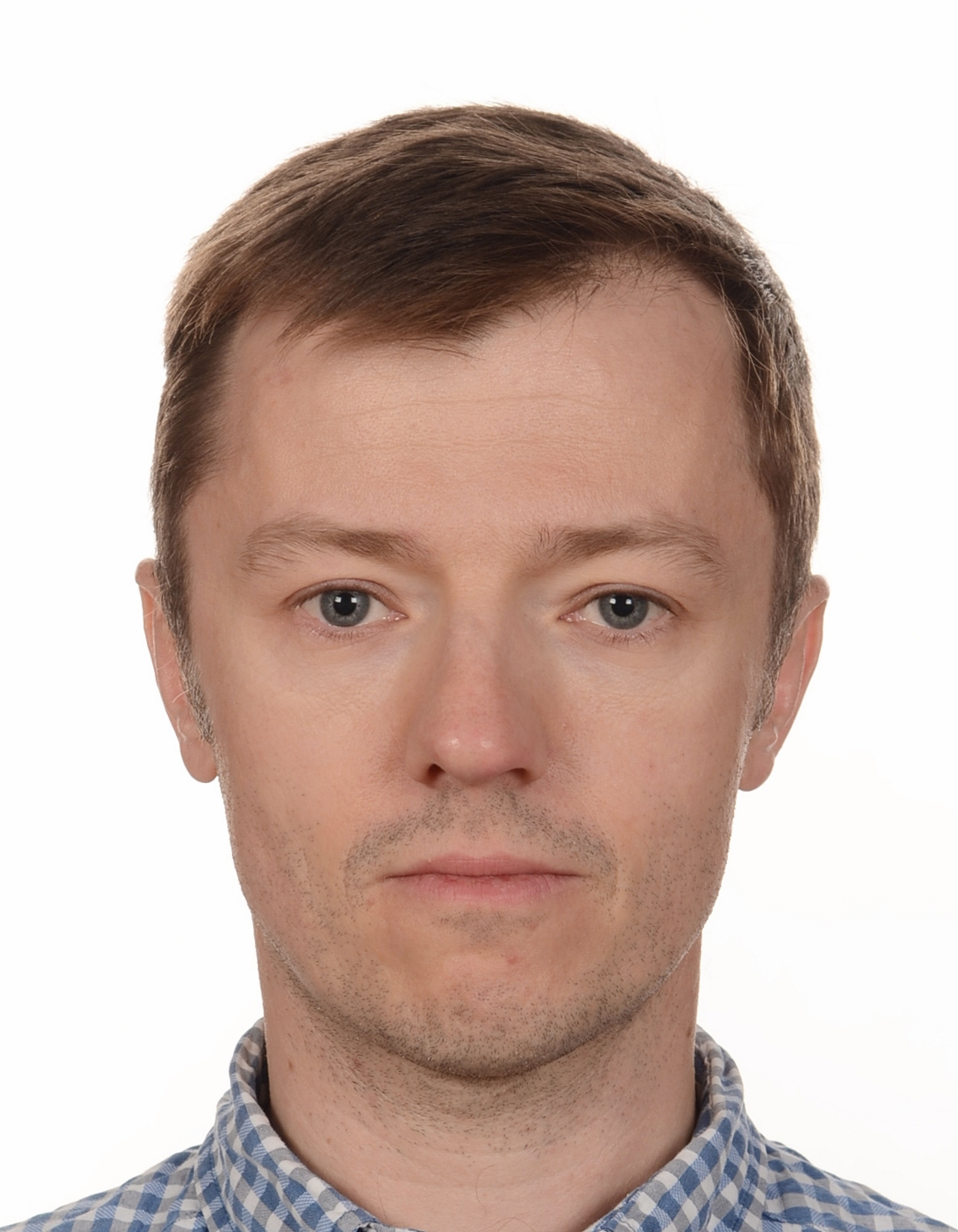}}]{Michal Kawulok,} MSc (2003), PhD (2007), DSc (2015), IEEE member, is an Associate Professor at the Silesian University of Technology (Gliwice, Poland). He has published over 100 peer-reviewed papers in journals and conference proceedings. His research interests are concerned with image processing, evolutionary computation, pattern recognition, and machine learning, with particular attention given to super-resolution reconstruction, face and gesture recognition, dimensionality reduction techniques, and support vector machines.

\end{IEEEbiography}





\begin{thebibliography}{10}
\providecommand{\url}[1]{#1}
\csname url@samestyle\endcsname
\providecommand{\newblock}{\relax}
\providecommand{\bibinfo}[2]{#2}
\providecommand{\BIBentrySTDinterwordspacing}{\spaceskip=0pt\relax}
\providecommand{\BIBentryALTinterwordstretchfactor}{4}
\providecommand{\BIBentryALTinterwordspacing}{\spaceskip=\fontdimen2\font plus
\BIBentryALTinterwordstretchfactor\fontdimen3\font minus
  \fontdimen4\font\relax}
\providecommand{\BIBforeignlanguage}[2]{{%
\expandafter\ifx\csname l@#1\endcsname\relax
\typeout{** WARNING: IEEEtran.bst: No hyphenation pattern has been}%
\typeout{** loaded for the language `#1'. Using the pattern for}%
\typeout{** the default language instead.}%
\else
\language=\csname l@#1\endcsname
\fi
#2}}
\providecommand{\BIBdecl}{\relax}
\BIBdecl

\bibitem{Drusch2012}
M.~Drusch, U.~Del~Bello, S.~Carlier, O.~Colin, V.~Fernandez, F.~Gascon,
  B.~Hoersch, C.~Isola, P.~Laberinti, P.~Martimort \emph{et~al.}, ``Sentinel-2:
  {ESA}'s optical high-resolution mission for {GMES} operational services,''
  \emph{Remote Sensing of Environment}, vol. 120, pp. 25--36, 2012.

\bibitem{Phiri2020}
D.~Phiri, M.~Simwanda, S.~Salekin, V.~R. Nyirenda, Y.~Murayama, and
  M.~Ranagalage, ``Sentinel-2 data for land cover/use mapping: a review,''
  \emph{Remote Sensing}, vol.~12, no.~14, p. 2291, 2020.

\bibitem{Segarra2020}
J.~Segarra, M.~L. Buchaillot, J.~L. Araus, and S.~C. Kefauver, ``Remote sensing
  for precision agriculture: Sentinel-2 improved features and applications,''
  \emph{Agronomy}, vol.~10, no.~5, p. 641, 2020.

\bibitem{Sekertekin2018}
A.~Sekertekin, S.~Y. Cicekli, and N.~Arslan, ``Index-based identification of
  surface water resources using {Sentinel-2} satellite imagery,'' in
  \emph{Proc. International Symposium on Multidisciplinary Studies and
  Innovative Technologies (ISMSIT)}.\hskip 1em plus 0.5em minus 0.4em\relax
  IEEE, 2018, pp. 1--5.

\bibitem{Kaplan2017}
G.~Kaplan and U.~Avdan, ``Object-based water body extraction model using
  {Sentinel-2} satellite imagery,'' \emph{European Journal of Remote Sensing},
  vol.~50, no.~1, pp. 137--143, 2017.

\bibitem{Gibson2020}
R.~Gibson, T.~Danaher, W.~Hehir, and L.~Collins, ``A remote sensing approach to
  mapping fire severity in south-eastern {Australia} using {Sentinel-2} and
  random forest,'' \emph{Remote Sensing of Environment}, vol. 240, p. 111702,
  2020.

\bibitem{DelFrate2014}
F.~Del~Frate, D.~Latini, M.~Picchiani, G.~Schiavon, and C.~Vittucci, ``A neural
  network architecture combining {VHR SAR} and multispectral data for precision
  farming in viticulture,'' in \emph{Proc. IEEE International Geoscience and
  Remote Sensing Symposium (IGARSS)}, 2014, pp. 1508--1511.

\bibitem{LiuTang2022}
Z.~Liu, H.~Tang, and W.~Huang, ``Building outline delineation from {VHR} remote
  sensing images using the convolutional recurrent neural network embedded with
  line segment information,'' \emph{IEEE Transactions on Geoscience and Remote
  Sensing}, vol.~60, pp. 1--13, 2022.

\bibitem{Razzak2021}
M.~T. Razzak, G.~Mateo-Garc{\'\i}a, G.~Lecuyer, L.~G{\'o}mez-Chova, Y.~Gal, and
  F.~Kalaitzis, ``Multi-spectral multi-image super-resolution of {Sentinel-2}
  with radiometric consistency losses and its effect on building delineation,''
  \emph{ISPRS Journal of Photogrammetry and Remote Sensing}, vol. 195, pp.
  1--13, 2023.

\bibitem{Liebel2016}
L.~Liebel and M.~K{\"o}rner, ``Single-image super resolution for multispectral
  remote sensing data using {CNNs},'' in \emph{Proc. ISPRSC}, 2016, pp.
  883--890.

\bibitem{Lanaras2018}
C.~Lanaras, J.~Bioucas-Dias, S.~Galliani, E.~Baltsavias, and K.~Schindler,
  ``Super-resolution of {Sentinel-2} images: Learning a globally applicable
  deep neural network,'' \emph{ISPRS Journal of Photogrammetry and Remote
  Sensing}, vol. 146, pp. 305--319, 2018.

\bibitem{Paris2018}
C.~Paris, J.~Bioucas-Dias, and L.~Bruzzone, ``A novel sharpening approach for
  superresolving multiresolution optical images,'' \emph{IEEE Transactions on
  Geoscience and Remote Sensing}, vol.~57, no.~3, pp. 1545--1560, 2018.

\bibitem{Gargiulo2019}
M.~Gargiulo, A.~Mazza, R.~Gaetano, G.~Ruello, and G.~Scarpa, ``Fast
  super-resolution of 20 m {Sentinel-2} bands using convolutional neural
  networks,'' \emph{Remote Sensing}, vol.~11, no.~22, p. 2635, 2019.

\bibitem{Ulfarsson2019}
M.~O. Ulfarsson, F.~Palsson, M.~Dalla~Mura, and J.~R. Sveinsson, ``Sentinel-2
  sharpening using a reduced-rank method,'' \emph{IEEE Transactions on
  Geoscience and Remote Sensing}, vol.~57, no.~9, pp. 6408--6420, 2019.

\bibitem{TaoXiong2021}
Y.~Tao, S.~Xiong, R.~Song, and J.-P. Muller, ``Towards streamlined single-image
  super-resolution: Demonstration with 10 m {Sentinel-2} colour and 10--60 m
  multi-spectral {VNIR and SWIR} bands,'' \emph{Remote Sensing}, vol.~13,
  no.~13, p. 2614, 2021.

\bibitem{QianJiang2023}
X.~Qian, T.-X. Jiang, and X.-L. Zhao, ``{SelfS2}: Self-supervised transfer
  learning for {Sentinel-2} multispectral image super-resolution,'' \emph{IEEE
  Journal of Selected Topics in Applied Earth Observations and Remote Sensing},
  vol.~16, pp. 215--227, 2023.

\bibitem{WangChen2021}
Z.~Wang, J.~Chen, and S.~C.~H. Hoi, ``Deep learning for image super-resolution:
  A survey,'' \emph{IEEE Transactions on Pattern Analysis and Machine
  Intelligence}, vol.~43, no.~10, pp. 3365--3387, 2021.

\bibitem{Romero2020}
L.~Salgueiro~Romero, J.~Marcello, and V.~Vilaplana, ``Super-resolution of
  {Sentinel-2} imagery using generative adversarial networks,'' \emph{Remote
  Sensing}, vol.~12, no.~15, p. 2424, 2020.

\bibitem{Galar2020}
M.~Galar, R.~Sesma, C.~Ayala, L.~Albizua, and C.~Aranda, ``Learning
  super-resolution for {Sentinel-2} images with real ground truth data from a
  reference satellite,'' \emph{ISPRS Annals of the Photogrammetry, Remote
  Sensing and Spatial Information Sciences}, vol.~1, pp. 9--16, 2020.

\bibitem{Yue2016}
L.~Yue, H.~Shen, J.~Li, Q.~Yuan, H.~Zhang, and L.~Zhang, ``Image
  super-resolution: The techniques, applications, and future,'' \emph{Signal
  Processing}, vol. 128, pp. 389--408, 2016.

\bibitem{Molini2020}
A.~B. Molini, D.~Valsesia, G.~Fracastoro, and E.~Magli, ``{DeepSUM}: Deep
  neural network for super-resolution of unregistered multitemporal images,''
  \emph{IEEE Transactions on Geoscience and Remote Sensing}, vol.~58, no.~5,
  pp. 3644--3656, 2020.

\bibitem{Kawulok2021IGARSS}
M.~Kawulok, T.~Tarasiewicz, J.~Nalepa, D.~Tyrna, and D.~Kostrzewa, ``Deep
  learning for multiple-image super-resolution of {Sentinel-2} data,'' in
  \emph{Proc. IEEE International Geoscience and Remote Sensing Symposium
  (IGARSS)}, 2021, pp. 3885--3888.

\bibitem{Vaqueiro2021}
M.~Vaqueiro, J.~M. Fonseca, H.~Oliveira, and A.~Mora, ``Multi-image
  super-resolution algorithm supported on {Sentinel-2} satellite images
  geolocation error,'' in \emph{Proc. International Young Engineers Forum
  (YEF-ECE)}.\hskip 1em plus 0.5em minus 0.4em\relax IEEE, 2021, pp. 50--57.

\bibitem{HuHuang2021}
J.-F. Hu, T.-Z. Huang, L.-J. Deng, T.-X. Jiang, G.~Vivone, and J.~Chanussot,
  ``Hyperspectral image super-resolution via deep spatiospectral attention
  convolutional neural networks,'' \emph{IEEE Transactions on Neural Networks
  and Learning Systems}, pp. 1--15, 2021.

\bibitem{Lin2020}
C.-H. Lin and J.~M. Bioucas-Dias, ``An explicit and scene-adapted definition of
  convex self-similarity prior with application to unsupervised {Sentinel-2}
  super-resolution,'' \emph{IEEE Transactions on Geoscience and Remote
  Sensing}, vol.~58, no.~5, pp. 3352--3365, 2020.

\bibitem{Armannsson2021}
S.~E. Armannsson, M.~O. Ulfarsson, J.~Sigurdsson, H.~V. Nguyen, and J.~R.
  Sveinsson, ``A comparison of optimized {Sentinel-2} super-resolution methods
  using {Wald’s} protocol and {Bayesian} optimization,'' \emph{Remote
  Sensing}, vol.~13, no.~11, p. 2192, 2021.

\bibitem{deudon2020highresnet}
M.~Deudon, A.~Kalaitzis, I.~Goytom, M.~R. Arefin, Z.~Lin, K.~Sankaran,
  V.~Michalski, S.~E. Kahou, J.~Cornebise, and Y.~Bengio, ``{HighRes}-net:
  Recursive fusion for multi-frame super-resolution of satellite imagery,''
  \emph{arXiv preprint arXiv:2002.06460}, 2020.

\bibitem{Salvetti2020}
F.~Salvetti, V.~Mazzia, A.~Khaliq, and M.~Chiaberge, ``Multi-image super
  resolution of remotely sensed images using residual attention deep neural
  networks,'' \emph{Remote Sensing}, vol.~12, no.~14, p. 2207, 2020.

\bibitem{Topan2009}
H.~Topan, D.~Maktav, K.~Jacobsen, and G.~Buyuksalih, ``Information content of
  optical satellite images for topographic mapping,'' \emph{International
  Journal of Remote Sensing}, vol.~30, no.~7, pp. 1819--1827, 2009.

\bibitem{Meissner2020}
H.~Mei{\ss}ner, M.~Cramer, and R.~Reulke, ``Evaluation of structures and
  methods for resolution determination of remote sensing sensors,'' in
  \emph{Pacific-Rim Symposium on Image and Video Technology}.\hskip 1em plus
  0.5em minus 0.4em\relax Springer, 2020, pp. 59--69.

\bibitem{Kowaleczko2022}
P.~Kowaleczko, T.~Tarasiewicz, M.~Ziaja, D.~Kostrzewa, J.~Nalepa, P.~Rokita,
  and M.~Kawulok, ``{MuS2}: A benchmark for {Sentinel-2} multi-image
  super-resolution,'' \emph{arXiv preprint arXiv:2210.02745}, 2022.

\bibitem{HuangLi2021}
Y.~Huang, J.~Li, X.~Gao, Y.~Hu, and W.~Lu, ``Interpretable detail-fidelity
  attention network for single image super-resolution,'' \emph{IEEE
  Transactions on Image Processing}, vol.~30, pp. 2325--2339, 2021.

\bibitem{YuanZheng2017}
Y.~Yuan, X.~Zheng, and X.~Lu, ``Hyperspectral image superresolution by transfer
  learning,'' \emph{IEEE Journal of Selected Topics in Applied Earth
  Observations and Remote Sensing}, vol.~10, no.~5, pp. 1963--1974, 2017.

\bibitem{Dong2014}
C.~Dong, C.~C. Loy, K.~He, and X.~Tang, ``Learning a deep convolutional network
  for image super-resolution,'' in \emph{Proc. European Conference on Computer
  Vision (ECCV)}.\hskip 1em plus 0.5em minus 0.4em\relax Springer, 2014, pp.
  184--199.

\bibitem{LimSon2017}
B.~Lim, S.~Son, H.~Kim, S.~Nah, and K.~Mu~Lee, ``Enhanced deep residual
  networks for single image super-resolution,'' in \emph{Proc. IEEE/CVF
  Conference on Computer Vision and Pattern Recognition Workshops}, 2017, pp.
  136--144.

\bibitem{LuLi2022}
Z.~Lu, J.~Li, H.~Liu, C.~Huang, L.~Zhang, and T.~Zeng, ``Transformer for single
  image super-resolution,'' in \emph{Proc. IEEE/CVF Conference on Computer
  Vision and Pattern Recognition}, 2022, pp. 457--466.

\bibitem{Ledig2017}
C.~Ledig, L.~Theis, F.~Husz{\'a}r \emph{et~al.}, ``Photo-realistic single image
  super-resolution using a generative adversarial network.'' in \emph{Proc.
  IEEE Conference on Computer Vision and Pattern Recognition (CVPR)}, vol.~2,
  no.~3, 2017, p.~4.

\bibitem{WangJiang2020}
Z.~Wang, K.~Jiang, P.~Yi, Z.~Han, and Z.~He, ``Ultra-dense {GAN} for satellite
  imagery super-resolution,'' \emph{Neurocomputing}, vol. 398, pp. 328--337,
  2020.

\bibitem{KimChung2019}
D.-W. Kim, J.-R. Chung, J.~Kim, D.~Y. Lee, S.~Y. Jeong, and S.-W. Jung,
  ``Constrained adversarial loss for generative adversarial network-based
  faithful image restoration,'' \emph{ETRI Journal}, vol.~41, no.~4, pp.
  415--425, 2019.

\bibitem{Nasrollahi2014}
K.~Nasrollahi and T.~B. Moeslund, ``Super-resolution: a comprehensive survey,''
  \emph{Machine Vision and Applications}, vol.~25, no.~6, pp. 1423--1468, 2014.

\bibitem{Farsiu2004}
S.~Farsiu, M.~D. Robinson, M.~Elad, and P.~Milanfar, ``Fast and robust
  multiframe super resolution,'' \emph{IEEE Transactions on Image Processing},
  vol.~13, no.~10, pp. 1327--1344, 2004.

\bibitem{Zhu2016}
H.~Zhu, W.~Song, H.~Tan, J.~Wang, and D.~Jia, ``Super resolution reconstruction
  based on adaptive detail enhancement for {ZY-3} satellite images,''
  \emph{ISPRS Annals of Photogrammetry, Remote Sensing and Spatial Information
  Sciences}, pp. 213--217, 2016.

\bibitem{Kawulok2020GRSL}
M.~Kawulok, P.~Benecki, S.~Piechaczek, K.~Hrynczenko, D.~Kostrzewa, and
  J.~Nalepa, ``Deep learning for multiple-image super-resolution,'' \emph{IEEE
  Geoscience and Remote Sensing Letters}, vol.~17, no.~6, pp. 1062--1066, 2020.

\bibitem{Kawulok2018Gecco}
M.~Kawulok, P.~Benecki, D.~Kostrzewa, and L.~Skonieczny, ``Evolving imaging
  model for super-resolution reconstruction,'' in \emph{Proc GECCO}.\hskip 1em
  plus 0.5em minus 0.4em\relax New York, NY, USA: ACM, 2018, pp. 284--285.

\bibitem{Molini2020IGARSS}
A.~B. Molini, D.~Valsesia, G.~Fracastoro, and E.~Magli, ``{DeepSUM++}:
  Non-local deep neural network for super-resolution of unregistered
  multitemporal images,'' in \emph{Proc. IEEE International Geoscience and
  Remote Sensing Symposium (IGARSS)}, 2020, pp. 609--612.

\bibitem{Martens2019}
M.~M{\"a}rtens, D.~Izzo, A.~Krzic, and D.~Cox, ``Super-resolution of {PROBA-V}
  images using convolutional neural networks,'' \emph{Astrodynamics}, vol.~3,
  no.~4, pp. 387--402, 2019.

\bibitem{Arefin2020}
M.~Rifat~Arefin, V.~Michalski, P.-L. St-Charles, A.~Kalaitzis, S.~Kim, S.~E.
  Kahou, and Y.~Bengio, ``Multi-image super-resolution for remote sensing using
  deep recurrent networks,'' in \emph{Proc. IEEE Conference on Computer Vision
  and Pattern Recognition (CVPR) Workshops}, 2020, pp. 206--207.

\bibitem{AnZhang2022}
T.~An, X.~Zhang, C.~Huo, B.~Xue, L.~Wang, and C.~Pan, ``{TR-MISR}: Multiimage
  super-resolution based on feature fusion with transformers,'' \emph{IEEE
  Journal of Selected Topics in Applied Earth Observations and Remote Sensing},
  vol.~15, pp. 1373--1388, 2022.

\bibitem{Tarasiewicz2021}
T.~Tarasiewicz, J.~Nalepa, and M.~Kawulok, ``A graph neural network for
  multiple-image super-resolution,'' in \emph{Proc. IEEE International
  Conference on Image Processing (ICIP)}, 2021, pp. 1824--1828.

\bibitem{WuLin2022}
J.~Wu, L.~Lin, T.~Li, Q.~Cheng, C.~Zhang, and H.~Shen, ``Fusing {Landsat 8 and
  Sentinel-2} data for 10-m dense time-series imagery using a degradation-term
  constrained deep network,'' \emph{International Journal of Applied Earth
  Observation and Geoinformation}, vol. 108, p. 102738, 2022.

\bibitem{Saunier2022}
S.~Saunier, B.~Pflug, I.~M. Lobos, B.~Franch, J.~Louis, R.~De~Los~Reyes,
  V.~Debaecker, E.~G. Cadau, V.~Boccia, F.~Gascon \emph{et~al.}, ``{Sen2Like}:
  Paving the way towards harmonization and fusion of optical data,''
  \emph{Remote Sensing}, vol.~14, no.~16, p. 3855, 2022.

\bibitem{HuangXiao2015}
W.~Huang, L.~Xiao, Z.~Wei, H.~Liu, and S.~Tang, ``A new pan-sharpening method
  with deep neural networks,'' \emph{IEEE Geoscience and Remote Sensing
  Letters}, vol.~12, no.~5, pp. 1037--1041, 2015.

\bibitem{Sara2021}
D.~Sara, A.~K. Mandava, A.~Kumar, S.~Duela, and A.~Jude, ``Hyperspectral and
  multispectral image fusion techniques for high resolution applications: A
  review,'' \emph{Earth Science Informatics}, vol.~14, no.~4, pp. 1685--1705,
  2021.

\bibitem{YangXiao2022}
J.~Yang, L.~Xiao, Y.-Q. Zhao, and J.~C.-W. Chan, ``Variational regularization
  network with attentive deep prior for hyperspectral–multispectral image
  fusion,'' \emph{IEEE Transactions on Geoscience and Remote Sensing}, vol.~60,
  pp. 1--17, 2022.

\bibitem{LiWang2021}
Q.~Li, Q.~Wang, and X.~Li, ``Exploring the relationship between {2D/3D}
  convolution for hyperspectral image super-resolution,'' \emph{IEEE
  Transactions on Geoscience and Remote Sensing}, vol.~59, no.~10, pp.
  8693--8703, 2021.

\bibitem{XueZhao2022TC}
J.~Xue, Y.~Zhao, Y.~Bu, J.~C.-W. Chan, and S.~G. Kong, ``When {Laplacian} scale
  mixture meets three-layer transform: A parametric tensor sparsity for tensor
  completion,'' \emph{IEEE Transactions on Cybernetics}, vol.~52, no.~12, pp.
  13\,887--13\,901, 2022.

\bibitem{XueZhao2022TNNLS}
J.~Xue, Y.~Zhao, S.~Huang, W.~Liao, J.~C.-W. Chan, and S.~G. Kong, ``Multilayer
  sparsity-based tensor decomposition for low-rank tensor completion,''
  \emph{IEEE Transactions on Neural Networks and Learning Systems}, vol.~33,
  no.~11, pp. 6916--6930, 2022.

\bibitem{XueZhao2019}
J.~Xue, Y.~Zhao, W.~Liao, and J.~C.-W. Chan, ``Nonlocal low-rank regularized
  tensor decomposition for hyperspectral image denoising,'' \emph{IEEE
  Transactions on Geoscience and Remote Sensing}, vol.~57, no.~7, pp.
  5174--5189, 2019.

\bibitem{DianLi2019}
R.~Dian, S.~Li, and L.~Fang, ``Learning a low tensor-train rank representation
  for hyperspectral image super-resolution,'' \emph{IEEE Transactions on Neural
  Networks and Learning Systems}, vol.~30, no.~9, pp. 2672--2683, 2019.

\bibitem{BuZhao2021}
Y.~Bu, Y.~Zhao, J.~Xue, J.~C.-W. Chan, S.~G. Kong, C.~Yi, J.~Wen, and B.~Wang,
  ``Hyperspectral and multispectral image fusion via graph laplacian-guided
  coupled tensor decomposition,'' \emph{IEEE Transactions on Geoscience and
  Remote Sensing}, vol.~59, no.~1, pp. 648--662, 2021.

\bibitem{YangXiao2021}
J.~Yang, L.~Xiao, Y.-Q. Zhao, and J.~C.-W. Chan, ``Hybrid local and nonlocal
  {3-D} attentive {CNN} for hyperspectral image super-resolution,'' \emph{IEEE
  Geoscience and Remote Sensing Letters}, vol.~18, no.~7, pp. 1274--1278, 2021.

\bibitem{KwanChoi2018}
C.~Kwan, J.~H. Choi, S.~H. Chan, J.~Zhou, and B.~Budavari, ``A super-resolution
  and fusion approach to enhancing hyperspectral images,'' \emph{Remote
  Sensing}, vol.~10, no.~9, p. 1416, 2018.

\bibitem{LiZhang2018}
Y.~Li, L.~Zhang, C.~Dingl, W.~Wei, and Y.~Zhang, ``Single hyperspectral image
  super-resolution with grouped deep recursive residual network,'' in
  \emph{Proc. IEEE Fourth International Conference on Multimedia Big Data
  (BigMM)}, 2018, pp. 1--4.

\bibitem{Kawulok2018ACIIDS}
M.~Kawulok, P.~Benecki, J.~Nalepa, D.~Kostrzewa, and {\L}.~Skonieczny,
  ``Towards robust evaluation of super-resolution satellite image
  reconstruction,'' in \emph{Proc. Asian Conference on Intelligent Information
  and Database Systems}.\hskip 1em plus 0.5em minus 0.4em\relax Springer, 2018,
  pp. 476--486.

\bibitem{Kohler2019}
T.~K{\"o}hler, M.~B{\"a}tz, F.~Naderi, A.~Kaup, A.~Maier, and C.~Riess,
  ``Toward bridging the simulated-to-real gap: Benchmarking super-resolution on
  real data,'' \emph{IEEE Transactions on Pattern Analysis and Machine
  Intelligence}, vol.~42, no.~11, pp. 2944--2959, 2019.

\bibitem{Lugmayr2020}
A.~Lugmayr, M.~Danelljan, and R.~Timofte, ``{NTIRE} 2020 challenge on
  real-world image super-resolution: Methods and results,'' in \emph{Proc.
  IEEE/CVF Conference on Computer Vision and Pattern Recognition (CVPR)
  Workshops}, 2020, pp. 494--495.

\bibitem{ChenHe2022}
H.~Chen, X.~He, L.~Qing, Y.~Wu, C.~Ren, R.~E. Sheriff, and C.~Zhu, ``Real-world
  single image super-resolution: A brief review,'' \emph{Information Fusion},
  vol.~79, pp. 124--145, 2022.

\bibitem{Cornebise2022}
J.~Cornebise, I.~Orsolic, and F.~Kalaitzis, ``Open high-resolution satellite
  imagery: The {WorldStrat} dataset {\textendash} with application to
  super-resolution,'' in \emph{Proc. NeurIPS}, 2022.

\bibitem{NguyenAnger2021Proba}
N.~L. Nguyen, J.~Anger, A.~Davy, P.~Arias, and G.~Facciolo, ``{PROBA-V-REF:
  Repurposing the PROBA-V Challenge for Reference-Aware Super Resolution},'' in
  \emph{Proc. IEEE International Geoscience and Remote Sensing Symposium
  (IGARSS)}.\hskip 1em plus 0.5em minus 0.4em\relax IEEE, 2021, pp. 3881--3884.

\bibitem{TaoMuller2021}
Y.~Tao and J.-P. Muller, ``Super-resolution restoration of spaceborne
  ultra-high-resolution images using the {UCL OpTiGAN} system,'' \emph{Remote
  Sensing}, vol.~13, no.~12, p. 2269, 2021.

\bibitem{XuTang2021}
P.~Xu, H.~Tang, J.~Ge, and L.~Feng, ``{ESPC\_NASUnet}: An end-to-end
  super-resolution semantic segmentation network for mapping buildings from
  remote sensing images,'' \emph{IEEE Journal of Selected Topics in Applied
  Earth Observations and Remote Sensing}, vol.~14, pp. 5421--5435, 2021.

\bibitem{Latif2022}
H.~Latif, S.~Ghuffar, and H.~M. Ahmad, ``Super-resolution of {Sentinel-2}
  images using wasserstein gan,'' \emph{Remote Sensing Letters}, vol.~13,
  no.~12, pp. 1194--1202, 2022.

\bibitem{Beaulieu2018}
M.~Beaulieu, S.~Foucher, D.~Haberman, and C.~Stewart, ``Deep image-to-image
  transfer applied to resolution enhancement of {Sentinel-2} images,'' in
  \emph{Proc. IEEE International Geoscience and Remote Sensing Symposium
  (IGARSS)}.\hskip 1em plus 0.5em minus 0.4em\relax IEEE, 2018, pp. 2611--2614.

\bibitem{WangYu2018}
X.~Wang, K.~Yu, S.~Wu, J.~Gu, Y.~Liu, C.~Dong, Y.~Qiao, and C.~Change~Loy,
  ``{ESRGAN}: Enhanced super-resolution generative adversarial networks,'' in
  \emph{Proc. European Conference on Computer Vision (ECCV) Workshops}, 2018,
  pp. 0--0.

\bibitem{Zabalza2022}
M.~Zabalza and A.~Bernardini, ``Super-resolution of {Sentinel-2} images using a
  spectral attention mechanism,'' \emph{Remote Sensing}, vol.~14, no.~12, 2022.

\bibitem{Valsesia2022permutation}
D.~Valsesia and E.~Magli, ``Permutation invariance and uncertainty in
  multitemporal image super-resolution,'' \emph{IEEE Transactions on Geoscience
  and Remote Sensing}, vol.~60, pp. 1--12, 2022.

\bibitem{WangShi2016}
Q.~Wang, W.~Shi, Z.~Li, and P.~M. Atkinson, ``Fusion of {Sentinel-2} images,''
  \emph{Remote Sensing of Environment}, vol. 187, pp. 241--252, 2016.

\bibitem{Brodu2017}
N.~Brodu, ``Super-resolving multiresolution images with band-independent
  geometry of multispectral pixels,'' \emph{IEEE Transactions on Geoscience and
  Remote Sensing}, vol.~55, no.~8, pp. 4610--4617, 2017.

\bibitem{Lanaras2017}
C.~Lanaras, J.~Bioucas-Dias, E.~Baltsavias, and K.~Schindler,
  ``Super-resolution of multispectral multiresolution images from a single
  sensor,'' in \emph{Proc. IEEE Conference on Computer Vision and Pattern
  Recognition (CVPR) Workshops}, 2017, pp. 20--28.

\bibitem{Gargiulo2018}
M.~Gargiulo, A.~Mazza, R.~Gaetano, G.~Ruello, and G.~Scarpa, ``A {CNN-based}
  fusion method for super-resolution of {Sentinel-2} data,'' in \emph{Proc.
  IEEE International Geoscience and Remote Sensing Symposium (IGARSS)}.\hskip
  1em plus 0.5em minus 0.4em\relax IEEE, 2018, pp. 4713--4716.

\bibitem{Latte2020}
N.~Latte and P.~Lejeune, ``Planetscope radiometric normalization and
  {Sentinel-2} super-resolution (2.5 m): A straightforward spectral-spatial
  fusion of multi-satellite multi-sensor images using residual convolutional
  neural networks,'' \emph{Remote Sensing}, vol.~12, no.~15, p. 2366, 2020.

\bibitem{Lloyd2021}
D.~T. Lloyd, A.~Abela, R.~A. Farrugia, A.~Galea, and G.~Valentino, ``Optically
  enhanced super-resolution of sea surface temperature using deep learning,''
  \emph{IEEE Transactions on Geoscience and Remote Sensing}, vol.~60, pp.
  1--14, 2021.

\bibitem{Raiyani2021}
K.~Raiyani, T.~Gonçalves, L.~Rato, P.~Salgueiro, and J.~R. Marques~da Silva,
  ``{Sentinel-2 Image Scene Classification: A Comparison between Sen2Cor and a
  Machine Learning Approach},'' \emph{Remote Sensing}, vol.~13, no.~2, 2021.

\bibitem{ZhangIsola2018}
R.~Zhang, P.~Isola, A.~A. Efros, E.~Shechtman, and O.~Wang, ``The unreasonable
  effectiveness of deep features as a perceptual metric,'' in \emph{Proc.
  IEEE/CVF Conference on Computer Vision and Pattern Recognition}, 2018.

\bibitem{Mittal2013}
A.~Mittal, R.~Soundararajan, and A.~C. Bovik, ``Making a “completely blind”
  image quality analyzer,'' \emph{IEEE Signal Processing Letters}, vol.~20,
  no.~3, pp. 209--212, 2013.

\bibitem{Marcinkiewicz2019}
M.~Marcinkiewicz, M.~Kawulok, and J.~Nalepa, ``Segmentation of multispectral
  data simulated from hyperspectral imagery,'' in \emph{Proc. IEEE
  International Geoscience and Remote Sensing Symposium (IGARSS)}.\hskip 1em
  plus 0.5em minus 0.4em\relax IEEE, 2019, pp. 3336--3339.

\bibitem{Blonski2003}
S.~Blonski, G.~Glasser, J.~Russell, R.~Ryan, G.~Terrie, and V.~Zanoni,
  ``Synthesis of multispectral bands from hyperspectral data: Validation based
  on images acquired by {AVIRIS, Hyperion, ALI, and ETM+},'' Tech. Rep., 2003.

\bibitem{Dorr2020}
F.~Dorr, ``Satellite image multi-frame super resolution using {3D}
  wide-activation neural networks,'' \emph{Remote Sensing}, vol.~12, no.~22,
  2020.

\end{thebibliography}
\end{document}